\documentclass[lettersize,journal]{IEEEtran}
\usepackage{amsmath,amsfonts}
\usepackage{algorithmic}
\usepackage{algorithm}
\usepackage{array}
\usepackage[caption=false,font=normalsize,labelfont=sf,textfont=sf]{subfig}
\usepackage{textcomp}
\usepackage{stfloats}
\usepackage{url}
\usepackage{verbatim}
\usepackage{graphicx}
\usepackage{cite}
\usepackage{bm} 
\usepackage{microtype}
\usepackage[table,xcdraw]{xcolor} 
\usepackage{colortbl}
\usepackage{amsmath}
\usepackage{subfig}
\usepackage{multirow}
\usepackage{caption}
\usepackage{booktabs} 
\usepackage{multirow}
\usepackage{graphicx}
\usepackage[table]{xcolor} 
\usepackage[colorlinks=false, pdfborder={0 0 0}]{hyperref} 
\usepackage{cleveref} 

\begin{document}

\title{A Decoupled Basis-Vector-Driven Generative Framework for Dynamic Multi-Objective Optimization}

\author{Yaoming Yang†, Shuai Wang†, Bingdong Li*, Peng Yang,~\IEEEmembership{Senior Member,~IEEE,} and Ke Tang,~\IEEEmembership{Fellow,~IEEE}

}
        

\markboth{Journal of \LaTeX\ Class Files,~Vol.~14, No.~8, August~2021}%
{Shell \MakeLowercase{\textit{et al.}}: A Sample Article Using IEEEtran.cls for IEEE Journals}


\maketitle

\begin{abstract}
Dynamic multi-objective optimization requires continuous tracking of moving Pareto fronts. Existing methods struggle with irregular mutations and data sparsity, primarily facing three challenges: the non-linear coupling of dynamic modes, negative transfer from outdated historical data, and the cold-start problem during environmental switches. To address these issues, this paper proposes a decoupled basis-vector-driven generative framework (DB-GEN). First, to resolve non-linear coupling, the framework employs the discrete wavelet transform to separate evolutionary trajectories into low-frequency trends and high-frequency details. Second, to mitigate negative transfer, it learns transferable basis vectors via sparse dictionary learning rather than directly memorizing historical instances. Recomposing these bases under a topology-aware contrastive constraint constructs a structured latent manifold. Finally, to overcome the cold-start problem, a surrogate-assisted search paradigm samples initial populations from this manifold. Pre-trained on 120 million solutions, DB-GEN performs direct online inference without retraining or fine-tuning. This zero-shot generation process executes in milliseconds, requiring approximately 0.2 seconds per environmental change. Experimental results demonstrate that DB-GEN improves tracking accuracy across various dynamic benchmarks compared to existing algorithms.
\end{abstract}

\section{Introduction}  
\IEEEPARstart{I}{n} real-world scientific research and engineering applications, many optimization problems involve multiple conflicting objectives. Furthermore, their environmental parameters and objective functions often change dynamically over time \cite{mukhopadhyay2013survey}. These problems are collectively referred to as Dynamic Multi-Objective Optimization Problems (DMOPs) \cite{farina2004dynamic}. Unlike static multi-objective optimization, which seeks a fixed Pareto Set (PS), the core task in solving DMOPs lies in continuously and accurately tracking the time-varying PS and its corresponding image in the objective space, the Pareto Front (PF). 

To address this challenge, population-based Evolutionary Algorithms (EAs) have been widely adopted \cite{mukhopadhyay2013survey, jiang2022evolutionary, qian2022result}. Due to their inherent parallel search mechanisms, EAs have recently demonstrated exceptional capabilities in solving complex real-world multi-objective tasks, ranging from large language model merging \cite{li2025s} and adversarial textual attacks \cite{liu2024effective} to online load balancing in cloud services \cite{yang2024reducing}. Building upon these successes, researchers are continuously adapting evolutionary frameworks to tackle the temporal complexities of DMOPs.

However, the intrinsic contradiction between environmental dynamics and population evolution poses a rigorous test for algorithm design \cite{helbig2014benchmarks}. Specifically, environmental changes rarely follow simple linear or periodic patterns; instead, they often manifest as random jumps or complex non-linear correlations \cite{jiang2018benchmark}. Concurrently, algorithms must strike a delicate balance between rapid convergence to the current PF and maintaining diversity to accommodate future mutations \cite{helbig2014benchmarks, zhou2013population}. If this balance is disrupted, a highly converged population may become trapped in the optimal region of the previous environment during a sudden change, leading to tracking latency or even the loss of the global optimum.

To overcome these difficulties, researchers have proposed various Dynamic Multi-Objective Evolutionary Algorithms (DMOEAs).
Early approaches primarily relied on reactive evolutionary mechanisms to address environmental changes passively, including diversity-based methods, multi-population-based methods, and memory-based methods.

Diversity-based methods inject new individuals through random initialization and mutation to compensate for diversity loss.
Liu \textit{et al.} \cite{liu2021dynamic} adaptively introduced diversity by employing either objective inverse modeling or random initialization based on the severity of changes. 
Ma \textit{et al.} \cite{ma2021multiregional} mitigated diversity loss by partitioning the search space into subregions and randomly generating new individuals within their respective boundaries. 
Jiang and Yang \cite{jiang2016steady} maintained population diversity by constructing an elite archive, selecting parents from both the current population and the archive during the cross-mutation process.

Multi-population strategies maintain parallel sub-swarms to collaboratively balance exploration and exploitation. 
Aboud \textit{et al.} \cite{aboud2022dpb} proposed a bi-level parallel framework that utilizes an upper-level single population for global search and environmental monitoring, while triggering multiple lower-level sub-populations to collaboratively track the Pareto front upon detecting changes. 
Similarly, Jo{\'c}ko \textit{et al.} \cite{jocko2022dynamic} partitioned the swarm into multiple dedicated sub-swarms, assigning each to optimize a specific objective function to balance exploration and exploitation. 
Furthermore, Yang \textit{et al.} \cite{yang2024dynamic} divided the population into non-dominated and dominated sub-populations during the steady state, applying distinct genetic engineering operators to each group to simultaneously accelerate convergence and preserve diversity.

Memory-based approaches archive historical elite solutions and retrieve them upon detecting similar changes. For instance, Hu \textit{et al.} \cite{hu2023mahalanobis} established an environment pool and utilized the Mahalanobis distance to retrieve similar past environments, guiding the initialization of the new population. Sahmoud and Topcuoglu \cite{sahmoud2016memory} developed a memory-based NSGA-II that reuses archived non-dominated solutions for population reinitialization to maintain diversity. Similarly, Jiang and Yang \cite{jiang2016steady} maintained a reservoir of discovered non-dominated solutions to supply high-quality parents for genetic operations during environmental changes.

However, these reactive mechanisms search without actively learning the underlying dynamics. Overemphasizing diversity degrades convergence efficiency \cite{liu2021dynamic, ma2021multiregional}, multi-population schemes incur heavy computational overhead \cite{aboud2022dpb, jocko2022dynamic, yang2024dynamic}, and memory strategies are susceptible to negative transfer when confronting non-periodic or stochastic environments \cite{hu2023mahalanobis, jiang2016steady}.
To break the limitations of passive responses, proactive data-driven strategies have emerged, hypothesizing that environmental variations follow learnable patterns. 

Temporal and trajectory prediction methods attempt to estimate the coordinates of the new PS using historical sequences. 
To predict the evolutionary trajectory, mathematical and statistical models have been widely adopted. For instance, Wang \textit{et al.} \cite{wang2024adaptive} adaptively selected prediction strategies based on the extent of environmental changes, while Lei \textit{et al.} \cite{lei2025adaptive} utilized first- and second-order derivatives within clustered sub-regions to estimate future locations. Similarly, Jiang \textit{et al.} \cite{jiang2025dynamic} modeled the evolutionary trajectory using a vector autoregressive framework, and Wang \textit{et al.} \cite{wang2024moea} captured complex dynamics through spatial-temporal topological tensor prediction. 
To handle highly non-linear patterns, neural networks have also been introduced. Ye \textit{et al.} \cite{ye2024learning} trained a feedforward neural network to learn directional improvements between consecutive environments, and Hu \textit{et al.} \cite{hu2025dynamic} employed an incremental Long Short-Term Memory network comed with an inverse model to predict new Pareto sets.
Despite their applicability in tracking continuous dynamics, these prediction-based methods share a common limitation regarding online data dependency.

Recently, motivated by the recent development of generative models, manifold and distribution learning models have been introduced into DMOPs. These approaches assume that the PS resides on a low-dimensional manifold and attempt to learn the global probability distribution of historical elite solutions to sample high-quality, diverse populations. 
 Specifically, Li \textit{et al.} \cite{li2024adversarial} employed an adversarial autoencoder (AAE) to learn the prior distribution of historical data and generate diverse initial populations, utilizing a Markov chain to constrain the prediction scope. To explicitly capture the movement of the Pareto set, Wang \textit{et al.} \cite{wang2025generative} trained a Generative Adversarial Network (GAN) on up-sampled historical elite solutions to generate new candidates. Furthermore, Wang \textit{et al.} \cite{wang2025new} combined a variational autoencoder (VAE) with a diffusion model, mapping the Pareto set into a low-dimensional latent space for robust data generation. 
Despite their capability to generate well-distributed solutions, these generative models commonly rely on a preliminary anchor prediction.

Another paradigm circumventing exact coordinate prediction is domain knowledge transfer, which seeks shared latent spaces or mapping functions to reuse historical data. For instance, Zou \textit{et al.} \cite{zou2025knowledge} represented environmental knowledge using Gaussian Mixture Models for adaptive weight allocation. Song \textit{et al.} \cite{song2025multisource} clustered historical Pareto sets to identify source domains and guide the search via classification. Furthermore, Jiang \textit{et al.} \cite{jiang2017transfer} and Hu \textit{et al.} \cite{hu2025learning} mitigated negative transfer by projecting historical and current data into shared spaces to minimize their distribution discrepancies. While these methods primarily transfer knowledge within a single isolated problem, very recent studies have extended this paradigm across problems. For instance, Xie \textit{et al.} \cite{xie2025evolutionary} proposed a centralized model to leverage knowledge across different DMOPs through dynamic feature extraction and task-specific classification. 
Despite broadening the learning scope, the effectiveness of these transfer-based methods relies heavily on the apparent similarity between domains. Directly transferring raw experiences or distributions without fundamentally disentangling these dynamic modes makes algorithms highly susceptible to severe negative transfer.

While effective under smooth environmental changes, current methods exhibit limitations when addressing irregular variations, data sparsity, and complex topological transformations. Specifically, the coupling of multiple dynamic modes challenges traditional linear prediction models, particularly during non-linear topological shifts. Furthermore, the cold-start problem restricts online adaptation: immediately after an environmental change, the scarcity of evaluated solutions provides insufficient data for timely model training and calibration. Moreover, transfer learning strategies that rely on basic distance metrics or direct memory inheritance are prone to negative transfer. 

To overcome these challenges, we propose a Decoupled basis-vector-driven Generative framework that shifts the dynamic optimization paradigm from online autoregressive prediction to zero-shot latent manifold generation. By leveraging a pre-trained offline model, DB-GEN bypasses the need for online data collection, while employing frequency decoupling and interpretable basis learning to extract transferable physical primitives from historical trajectories. 

In summary, the major contributions of this paper are highlighted as follows:

\begin{itemize}
\item \textbf{A Generative Framework with Frequency Decoupling:} To address the non-linear coupling of dynamic modes, we introduce a Discrete Wavelet Transform (DWT)\cite{daubechies1992ten} module into the generative framework. This mechanism explicitly disentangles complex evolutionary trajectories into low-frequency global trends and high-frequency local details, reducing learning complexity and isolating predictable patterns from stochastic environmental noise.

\item \textbf{Cross-Problem Structure Learning via Transferable Bases:} To mitigate negative transfer caused by rigidly memorizing problem-specific historical data, we design a structure learning module based on sparse dictionary learning. By recomposing learnable basis vectors guided by an Inverted Generational Distance (IGD)-driven contrastive loss, the model extracts universal topological priors from diverse source problems. This constructs an interpretable latent manifold that enables robust \textit{cross-problem generalization} rather than traditional intra-problem tracking.

\item \textbf{Cross-Problem Zero-Shot Generative Search:} To overcome the absolute cold-start problem during environmental switches, DB-GEN operates as an offline-trained zero-shot generator for completely unseen tasks. Pre-trained on the largest DMOP dataset to our knowledge (120 million solutions), it employs joint centroid perturbation in the continuous latent space and Tchebycheff decomposition for precise candidate screening. This paradigm directly generates high-quality initial populations without requiring any target-specific historical accumulation or online model updates.
\end{itemize}

The remainder of this paper is organized as follows. Section~II details the components and the overall workflow of the proposed DB-GEN. Section~III presents the experimental configurations, performance comparisons, and corresponding analyses. Finally, Section~IV concludes this paper and outlines potential directions for future work.

\section{Proposed Method}
\begin{figure*}[t]  
    \centering
    \includegraphics[width=0.95\textwidth]{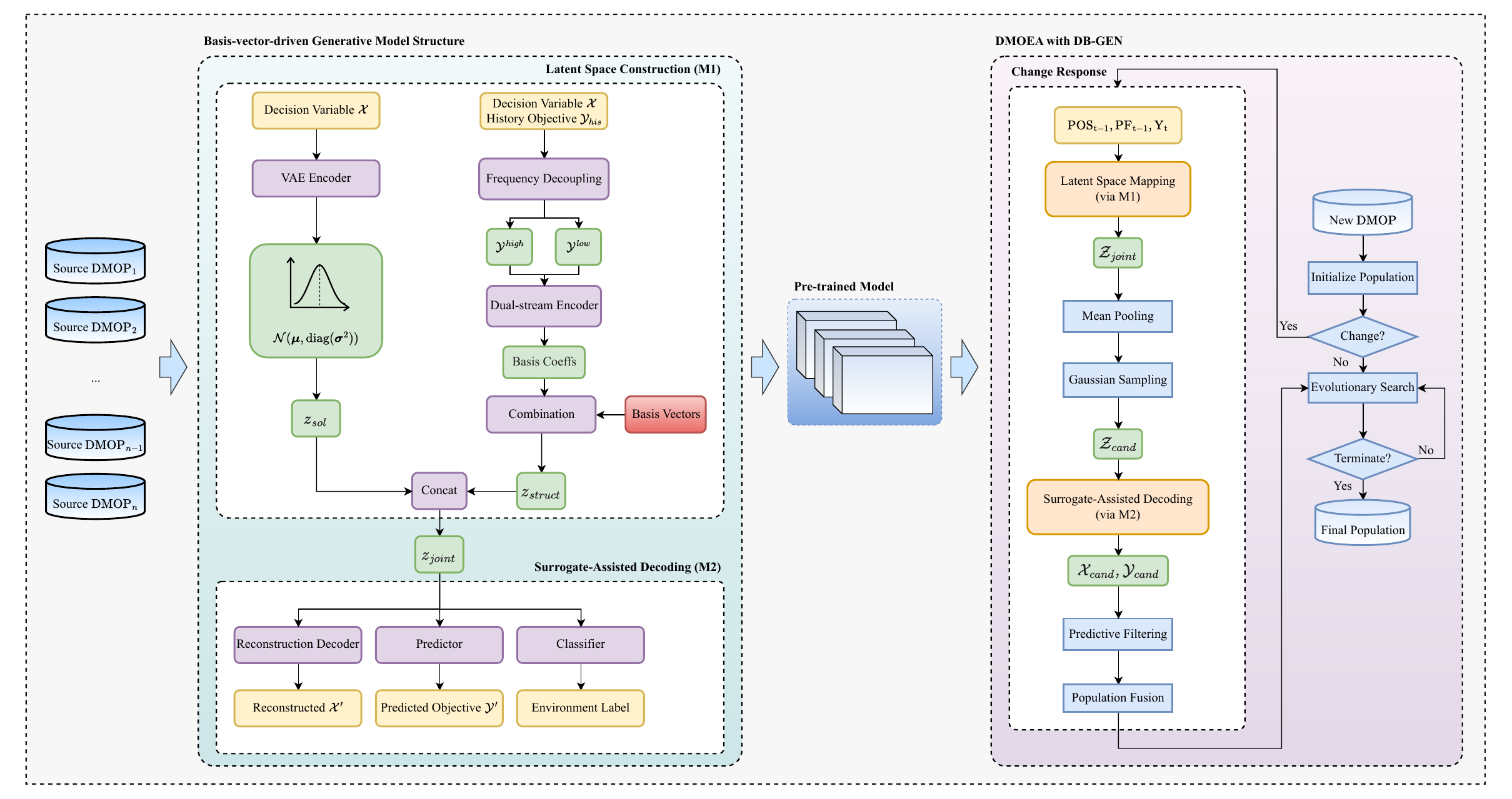} 
    
    \caption{The overall framework of the proposed DB-GEN. }
    \label{fig:framework}
\end{figure*}
\subsection{Overall Algorithm} 
\label{sec:overall}

To address the limitations of existing dynamic multi-objective evolutionary algorithms, we propose a novel generative structure learning framework for DMOPs. The core motivation of DB-GEN is to decouple the complex environmental dynamics into universal physical primitives and construct a topology-aware latent manifold for robust generative search. 

As illustrated in Fig.~\ref{fig:framework}, the proposed framework fundamentally operates in two primary phases: the \textit{Joint Latent Structure Learning} phase for end-to-end model training and the \textit{Multi-Task Generative Inference} phase for surrogate-assisted population initialization.

1) Offline Latent Space Pre-training Phase: Prior to online tracking, the model is jointly trained on historical evolutionary trajectories to construct a regularized latent space via three interconnected steps. First, \textit{Multi-scale Frequency Decoupling Encoding} (Sec.~\ref{sec:perception}) decomposes historical objective sequences into global trends and local details to extract a holistic perceptual embedding $\mathbf{z}_{perc}$. Subsequently, \textit{Structure Learning via Basis Vectors} (Sec.~\ref{sec:basis}) maps $\mathbf{z}_{perc}$ to a sparse coefficient vector $\mathbf{a}$, dynamically recomposing a learnable dictionary $\mathbf{W}$ into a structural latent variable $\mathbf{z}_{struct}$. A topology-aware contrastive loss is applied here to cluster geometrically similar Pareto Fronts. Finally, \textit{Joint Generative and Predictive Modeling} (Sec.~\ref{sec:joint_model}) extracts a solution latent $\mathbf{z}_{sol}$ from decision variables $\mathbf{x}$ using a VAE. Fusing $\mathbf{z}_{sol}$ and $\mathbf{z}_{struct}$ yields a unified embedding $\mathbf{z}_{joint}$, which is decoded via a multi-task strategy for simultaneous reconstruction, objective prediction, and semantic supervision.

2) Multi-Task Generative Inference Phase: When an environmental change is detected, the frozen trained model acts as an intelligent surrogate to generate a high-quality initial population for the new environment (Sec.~\ref{sec:search}). Rather than relying on individual noisy predictions, the model calculates a joint centroid $\bar{\mathbf{z}}_{joint}$ from the current population to capture the macroscopic topological trend. By applying Gaussian perturbation around this centroid, a pool of candidate latent codes is generated. These candidates are subsequently decoded into decision variables and objective values. Finally, a decomposition-based selection mechanism employing the augmented Tchebycheff scalarizing function generates an initial population with balanced convergence and diversity to accelerate the subsequent evolutionary search.

The complete online tracking execution process of the proposed DB-GEN is summarized in Algorithm~\ref{alg:db_gen}. The detailed mathematical formulations and specific mechanisms of each module are comprehensively elaborated in the following subsections.

\begin{algorithm}[htbp]
\caption{Online Tracking Phase of DB-GEN for DMOPs}
\label{alg:db_gen}
\begin{algorithmic}[1]
\REQUIRE Sequence of DMOPs $F_t$, Population size $N$, Weight vectors $\mathcal{W} = \{\mathbf{w}_1, \dots, \mathbf{w}_N\}$, Candidate pool size $N_{cand}$, Perturbation radius $\sigma$, Pre-trained DB-GEN model $\mathcal{M}$.
\ENSURE The tracked Pareto Optimal Sets $\mathcal{PS}$.

\STATE Initialize environment index $t \leftarrow 0$, generation index $gen \leftarrow 0$, $\mathcal{PS} \leftarrow \emptyset$
\STATE Initialize population $P_{gen}$ uniformly at random and evaluate on $F_0$
\WHILE{stopping criterion is not satisfied}
    \STATE Perform standard MOEA/D reproduction and selection on $P_{gen}$ to obtain $P_{gen+1}$
    \IF{environmental change is detected}
        \STATE $\mathcal{PS} \leftarrow \mathcal{PS} \cup P_{gen+1}$
        \STATE $t \leftarrow t + 1$
        \STATE Construct fitness trajectory matrix $\mathcal{T}$ by evaluating $P_{gen+1}$ on recent past environments
        \STATE {}$[\mathbf{Low}, \mathbf{High}] \leftarrow \text{DWT}(\mathcal{T})$
        \STATE Extract $\bar{\mathbf{z}} \leftarrow \text{Encoder}_{\mathcal{M}}(P_{gen+1}, \mathbf{Low}, \mathbf{High})$
        \STATE Generate $N_{cand}$ latent vectors $\mathcal{Z}_{cand} \leftarrow \{\bar{\mathbf{z}} + \epsilon_i \mid \epsilon_i \sim \mathcal{N}(0, \sigma_r)\}_{i=1}^{N_{cand}}$
        \STATE {}$[\mathcal{X}_{cand}, \mathcal{Y}_{cand}] \leftarrow \text{Decoder}_{\mathcal{M}}(\mathcal{Z}_{cand})$
        \STATE Calculate Tchebycheff scalarizing values for $\mathcal{Y}_{pred}$ using the predefined weight vectors
        \STATE Select $N$ solutions from $\mathcal{X}_{cand}$ with the minimum Tchebycheff values to form $P_{gen+1}$
        
        \STATE Re-evaluate $P_{gen+1}$ on the new environment $F_t$
        \STATE Update the ideal point $\mathbf{z}^*$ based on the exact evaluations
    \ENDIF
    \STATE $gen \leftarrow gen + 1$
\ENDWHILE

\STATE \textbf{return} $\mathcal{PS}$

\end{algorithmic}
\end{algorithm}

\subsection{Multi-scale Frequency Decoupling Encoding}
\label{sec:perception}
The dynamics of DMOPs typically exhibit complex multi-scale spatiotemporal characteristics, where rigid global movements are entangled with non-rigid local deformations\cite{farina2004dynamic, goh2008competitive, biswas2014evolutionary, jiang2017transfer}. 
Decomposing these mixed dynamics into distinct frequency components is a strategy for reducing learning complexity and isolating predictable trends from stochastic noise\cite{fang2023stwave, wu2021autoformer}.
To effectively disentangle these coupled variations while preserving the temporal resolution of the input sequence, we introduce a Multi-scale Frequency Decoupling Perception module based on Multiresolution Analysis\cite{mallat2002theory}. Our method employs an explicit \textit{Decomposition-Reconstruction} mechanism to separate the signal into trend and detail components of the original sequence length.

Formally, let $\mathcal{Y}_{his} \in \mathbb{R}^{L \times M}$ denote the historical objective sequence with length $L$ and $M$ objectives. The decoupling process consists of two stages:

1) Wavelet Decomposition: To decouple the temporal dynamics, we apply the DWT\cite{daubechies1992ten} along the time dimension of the historical sequence. Given the relatively short length of our sequences, the Haar wavelet is adopted as the basis \cite{torrence1998practical} due to its localization properties. The Haar basis consists of a scaling function $\phi(t)$ (low-pass filter) and a mother wavelet $\psi(t)$ (high-pass filter), explicitly defined for $t \in [0, 1)$ as:
\begin{equation}
    \phi(t) = 1, \quad
    \psi(t) = 
    \begin{cases} 
        1 & 0 \le t < 1/2 \\
        -1 & 1/2 \le t < 1 
    \end{cases}.
    \label{eq:haar_basis}
\end{equation}

In the discrete domain, this normalized continuous unit interval $[0, 1)$ conceptually maps to a pair of adjacent sequence elements. Thus, projecting a signal onto these basis functions is mathematically equivalent to passing the signal through a digital quadrature mirror filter bank. For a discrete historical sequence $\mathcal{Y}_{his} = \{y_1, y_2, \dots, y_L\}$ of length $L$, the inner products with $\phi(t)$ and $\psi(t)$ simplify to pairwise averaging and differencing. Specifically, the approximation coefficients $C_{low, k}$ and detail coefficients $C_{high, k}$ at index $k \in \{1, 2, \dots, L/2\}$ are computed as:
\begin{equation}
    \begin{aligned}
        C_{low, k} = \frac{y_{2k-1} + y_{2k}}{\sqrt{2}}, 
        C_{high, k} = \frac{y_{2k-1} - y_{2k}}{\sqrt{2}}.
    \end{aligned}
    \label{eq:dwt_discrete}
\end{equation}
where the $1/\sqrt{2}$ factor acts as the normalization constant to ensure the orthonormality of the transformation. 

2) Selective Reconstruction: To obtain independent feature streams in the original time domain, we utilize the Inverse Discrete Wavelet Transform (IDWT). By selectively zeroing out specific coefficient sets during reconstruction, we can isolate the macroscopic trends from the microscopic variations. 

Specifically, the low-frequency sequence $\mathcal{Y}^{low}$ is reconstructed using only $\mathbf{C}_{low}$ (with $\mathbf{C}_{high}$ zeroed out), while the high-frequency sequence $\mathcal{Y}^{high}$ is reconstructed using only $\mathbf{C}_{high}$ (with $\mathbf{C}_{low}$ zeroed out). Following the inverse filter properties of the Haar transform, the reconstructed time-domain elements at adjacent indices $2k-1$ and $2k$ are mathematically decoupled as:
\begin{equation}
    \begin{aligned}
        y^{low}_{2k-1}  &= \frac{C_{low, k} + C_{high, k}}{\sqrt{2}}  = \frac{C_{low, k}}{\sqrt{2}}, \\
        y^{low}_{2k}    &= \frac{C_{low, k} - C_{high, k}}{\sqrt{2}}  = \frac{C_{low, k}}{\sqrt{2}}, \\
        y^{high}_{2k-1} &= \frac{C_{low, k} + C_{high, k}}{\sqrt{2}} = \frac{C_{high, k}}{\sqrt{2}}, \\
        y^{high}_{2k}   &= \frac{C_{low, k} - C_{high, k}}{\sqrt{2}}  = -\frac{C_{high, k}}{\sqrt{2}}.
    \end{aligned}
    \label{eq:idwt_recon}
\end{equation}
This discrete formulation guarantees that $\mathcal{Y}^{low}$ captures the step-wise local mean of the historical trajectory, while $\mathcal{Y}^{high}$ extracts the local step-wise fluctuations. It satisfies the reconstruction property, ensuring no information loss during the process.

Subsequently, a dual-stream encoder utilizing Multilayer Perceptron (MLP) extracts features from these decoupled views:
\begin{equation}
    \begin{aligned}
        \mathbf{h}_{low} &= \text{MLP}_{L}(\text{Flatten}(\mathcal{Y}^{low}); \theta_{L}), \\
        \mathbf{h}_{high} &= \text{MLP}_{H}(\text{Flatten}(\mathcal{Y}^{high}); \theta_{H}),
    \end{aligned}
    \label{eq:encoder}
\end{equation}
where $\text{MLP}_{L}$ and $\text{MLP}_{H}$ are the low-frequency and high-frequency encoders parameterized by $\theta_{L}$ and $\theta_{H}$, respectively. $\mathbf{h}_{low}$ captures the global evolutionary trend of the Pareto Front, while $\mathbf{h}_{high}$ encodes the local geometric variations.

Finally, to construct a comprehensive representation of the current state, the features from both streams are fused. We employ a concatenation-based fusion strategy followed by a projection layer to integrate the multi-scale information:
\begin{equation}
    \mathbf{z}_{perc} = \text{MLP}_{F}([\mathbf{h}_{low} \oplus \mathbf{h}_{high}]; \theta_{F}),
    \label{eq:fusion}
\end{equation}
where $\oplus$ denotes the concatenation operation, and $\text{MLP}_{F}$ serves as the fusion network parameterized by $\theta_{F}$. The resulting vector $\mathbf{z}_{perc}$ serves as a holistic perceptual embedding of the environment, containing both trend information and detail features required for subsequent structure learning.

\subsection{Structure Learning via Basis Vectors}
\label{sec:basis}
A fundamental challenge in transfer-learning-based DMOP solvers is the risk of negative transfer {when reusing knowledge from} different historical problems, where models may retrieve misleading historical solutions due to superficial similarities in environmental parameters, failing to adapt to novel topological changes\cite{pan2009survey, jiang2017transfer}. 
To overcome this, we propose a Structure Learning mechanism based on Sparse Dictionary Learning\cite{olshausen1996emergence, mairal2009online}. Instead of memorizing specific historical solutions, our model learns to capture universal \textit{basis vectors} of environmental evolution. By dynamically recomposing these learned basis vectors, the model can generalize to unseen dynamics through compositional generation.

\subsubsection{Sparse Basis Representation}
To construct the basis vector set, we introduce a learnable latent dictionary matrix $\mathbf{W} \in \mathbb{R}^{K \times D}$, where $K$ denotes the dictionary size and $D$ represents the latent feature dimension.  After training, each row of  $\mathbf{W}$ represents a basis vector encoding a distinct mode of environmental change.

To reconstruct the latent structure of the current environment, the model generates a sparse coefficient vector $\mathbf{a} \in \mathbb{R}^{K}$ based on the perceptual embedding $\mathbf{z}_{perc}$ obtained from the previous stage.
Furthermore, we employ a Batch Normalization (BN) layer followed by a ReLU activation to adaptively induce sparsity in $\mathbf{a}$:
\begin{equation}
    \mathbf{a} = \text{ReLU}(\text{BN}(\text{MLP}_{struct}(\mathbf{z}_{perc}))),
    \label{eq:sparsity}
\end{equation}
where $\text{MLP}_{struct}$ maps the perceptual features to the coefficient space. The BN layer normalizes the pre-activation values to a standard distribution, and the ReLU function naturally truncates negative values to zero. It ensures that only a small subset of relevant basis vectors is activated for any given environment.

The final structural latent variable $\mathbf{z}_{struct}$ is then obtained by the linear combination of basis vectors:
\begin{equation}
    \mathbf{z}_{struct} = \mathbf{a} \cdot \mathbf{W} = \sum_{k=1}^{K} a_k \mathbf{w}_k.
    \label{eq:reconstruction}
\end{equation}
Through this decomposable representation, the model can flexibly describe complex, unseen environmental dynamics by recombining known primitives, thereby achieving robust zero-shot generalization.


\subsubsection{Topology-Aware Contrastive Regularization}
{To ensure that the learned basis vectors capture topological semantics rather than arbitrary statistical correlations, we introduce a Topology-Aware Contrastive Loss based on the IGD metric }.

Specifically, we quantify the topological difference between two environments, $E_i$ and $E_j$, by calculating the IGD value between their ground-truth Pareto Fronts $\text{PF}_i$ and $\text{PF}_j$. Prior to training, we compute a global distance matrix $\mathbf{D} \in \mathbb{R}^{N \times N}$ covering all historical environments. For each anchor environment $E_i$, we pre-select two candidate sets:
\begin{itemize}
    \item Positive Set $\mathcal{S}_i^{+}$: Comprising the top-$k$ environments with the smallest IGD distances to $E_i$ excluding $E_i$ itself.
    \item Negative Set $\mathcal{S}_i^{-}$: Comprising the top-$k$ environments with the largest IGD distances to $E_i$.
\end{itemize}

During the training phase, for a given anchor environment $E_i$, we randomly sample a positive instance $E_p \in \mathcal{S}_i^{+}$ and a negative instance $E_n \in \mathcal{S}_i^{-}$. Drawing inspiration from deep metric learning \cite{schroff2015facenet}, we use a triplet margin loss to explicitly shape the geometry of the latent manifold. Rather than relying solely on unsupervised reconstruction gradients, this contrastive objective directly enforces that the sparse coefficient vector $\mathbf{a}_i$ is closer to its topological neighbor $\mathbf{a}_p$ than to the distant instance $\mathbf{a}_n$:
\begin{equation}
    \mathcal{L}_{contrast} = \max(0, \|\mathbf{a}_i - \mathbf{a}_p\|^2 - \|\mathbf{a}_i - \mathbf{a}_n\|^2 + m),
    \label{eq:triplet}
\end{equation}
where $m=1.0$  defines the minimum separation boundary between non-similar environments. 
By injecting this physical topological prior into the optimization process, the contrastive loss prevents the dictionary matrix $\mathbf{W}$ from degenerating into a purely mathematical compression tool. Instead, it forces the latent space to cluster environments with similar Pareto Front geometries. This structural alignment is crucial for the subsequent evolutionary search, as it ensures that random perturbations or interpolations of the latent coefficients $\mathbf{a}$ will reliably map to continuous, physically valid environmental changes in the decision space.

\subsection{Joint Generative and Predictive Modeling}
\label{sec:joint_model}

{To bridge the gap between the learned environmental structure and the decision space, we design a framework based on latent variable fusion. }This module integrates the decision variables $\mathbf{x}$ and the environmental context $\mathbf{z}_{struct}$ to perform simultaneous solution reconstruction and objective prediction.
\subsubsection{Latent Variable Fusion}
The objective of this module is to construct a joint representation $\mathbf{z}_{joint}$ that encapsulates both the intrinsic manifold of the solution space and the physical laws of the current environment. To achieve this, we synthesize two complementary latent codes. 

First, following the VAE framework \cite{kingma2013auto}, the Solution Latent ($\mathbf{z}_{sol}$) is extracted from the decision variable $\mathbf{x}$ via a probabilistic encoder $E_{\phi}$, which maps the input to the mean vector $\boldsymbol{\mu}$ and variance vector $\boldsymbol{\sigma}^2$. Modeled as a diagonal Gaussian distribution $q_{\phi}(\mathbf{z}_{sol}|\mathbf{x}) = \mathcal{N}(\boldsymbol{\mu}, \text{diag}(\boldsymbol{\sigma}^2))$, this variable captures the diversity and distribution patterns of the population. To maintain the generative capability of the framework and ensure a continuous latent space, a Kullback-Leibler (KL) divergence loss $\mathcal{L}_{KL}$ is introduced here. This term regularizes the approximate posterior distribution $q_{\phi}(\mathbf{z}_{sol}|\mathbf{x})$ to closely match a prior distribution, defined as a standard multivariate normal distribution $\mathcal{N}(\mathbf{0}, \mathbf{I})$. This regularization prevents the latent codes from becoming disjointed or overfitting to discrete points, which would otherwise degrade the model's ability to sample diverse solutions. The KL divergence is computed analytically as:
\begin{equation}
    \begin{aligned}
        \mathcal{L}_{KL} &= D_{KL}(q_{\phi}(\mathbf{z}_{sol}|\mathbf{x}) \| \mathcal{N}(\mathbf{0}, \mathbf{I})) \\
        &= -\frac{1}{2} \sum_{j=1}^{d_z} \left(1 + \log(\sigma_j^2) - \mu_j^2 - \sigma_j^2 \right),
    \end{aligned}
    \label{eq:kl_loss}
\end{equation}
where $d_z$ is the dimensionality of the solution latent space $\mathbf{z}_{sol}$, and $\mu_j$ and $\sigma_j$ are the $j$-th components of the mean vector $\boldsymbol{\mu}$ and standard deviation vector $\boldsymbol{\sigma}$ produced by the encoder.

Second, the Structural Context ($\mathbf{z}_{struct}$), obtained from the Basis Module (Eq.~\ref{eq:reconstruction}), represents the sparse combination of physical primitives specific to the current environment.

These two distinct codes are concatenated to form the joint embedding:
\begin{equation}
    \mathbf{z}_{joint} = [\mathbf{z}_{sol} \oplus \mathbf{z}_{struct}],
\end{equation}
where $\oplus$ denotes concatenation. This design ensures that the subsequent generation is conditioned on the learned environmental dynamics while retaining the stochasticity required for robust exploration \cite{sohn2015learning}.

\subsubsection{Multi-Task Decoding}
The joint embedding $\mathbf{z}_{joint}$ serves as the unified representation that encapsulates both the solution characteristics and the environmental context. To utilize this representation, we employ a multi-head decoding strategy consisting of four parallel streams \cite{caruana1997multitask}.

The first branch focuses on reconstructing the decision variables. We employ an MLP as the backbone decoder to map the latent embedding back to the decision space. To satisfy the boundary constraints of the decision variables, a Sigmoid activation function is applied to the final output layer:
\begin{equation}
    \hat{\mathbf{x}} = \text{Sigmoid}(\mathbf{W}_{rec} \text{MLP}_{rec}(\mathbf{z}_{joint}) + \mathbf{b}_{rec}),
    \label{eq:recon_x}
\end{equation}
where $\mathbf{W}_{rec}$ and $\mathbf{b}_{rec}$ denote the weight matrix and bias vector of the output layer, respectively. The reconstruction loss $\mathcal{L}_{rec} = \|\mathbf{x} - \hat{\mathbf{x}}\|^2$ forces the latent space to capture the geometric features of the Pareto Set.

Simultaneously, the second branch acts as a surrogate predictor to estimate the objective values. The predictor aims to regress the unbounded objective values directly:
\begin{equation}
    \hat{\mathbf{y}} = \mathbf{W}_{pred} \text{MLP}_{pred}(\mathbf{z}_{joint}) + \mathbf{b}_{pred}.
    \label{eq:pred_y}
\end{equation}
The prediction loss $\mathcal{L}_{pred} = \|\mathbf{y} - \hat{\mathbf{y}}\|^2$ renders the latent space objective-aware, enabling efficient candidate screening.

To ensure the latent space retains distinct semantic structures of the problem environment, we introduce two auxiliary heads that also operate on $\mathbf{z}_{joint}$. A problem classifier predicts the problem identity logits $\mathbf{p}_{id}$ via a linear transformation to differentiate between optimization landscapes:
\begin{equation}
    \mathbf{p}_{id} = \mathbf{W}_{cls} \mathbf{z}_{joint} + \mathbf{b}_{cls}.
\end{equation}
Concurrently, an environmental regressor estimates the evolutionary status $t_{norm} \in [0, 1]$ using a Sigmoid-activated linear layer to capture continuous temporal dynamics:
\begin{equation}
    \hat{t}_{norm} = \text{Sigmoid}(\mathbf{W}_{reg} \mathbf{z}_{joint} + \mathbf{b}_{reg}).
\end{equation}
These auxiliary tasks implicitly regularize the structure learning module by minimizing the auxiliary loss $\mathcal{L}_{aux}$, composed of the Cross-Entropy loss for classification and the Mean Squared Error for time regression.

\subsubsection{Overall Objective}
The entire framework is trained end-to-end by minimizing the weighted sum of the reconstruction, prediction, topological, auxiliary, and KL-divergence losses:
\begin{equation}
    \begin{split}
        \mathcal{L}_{total} &= \mathcal{L}_{rec} +  \mathcal{L}_{pred} + \lambda_1 \mathcal{L}_{contrast} \\
        &\quad + \lambda_2\mathcal{L}_{aux} + \lambda_3 \mathcal{L}_{KL},
    \end{split}
    \label{eq:total_loss}
\end{equation}
where $\lambda_1, \lambda_2$, and $\lambda_3$ act as penalty coefficients to balance the contributions of the respective task components.

\subsection{Surrogate-Assisted Generative Search}
\label{sec:search}

{While the learned joint model provides the capability to reconstruct solutions, identifying the distribution for the \textit{next} moment requires an effective search strategy within the latent manifold.} We propose a Centroid-based Neighborhood Sampling strategy with a surrogated screening mechanism to generate high-quality, diverse initial populations.

\subsubsection{Centroid-based Latent Mapping}
Given the current population $\mathcal{P}_t$, we map individuals through the encoding modules to obtain a set of joint embeddings $\{\mathbf{z}_{joint}^{(i)}\}$. Geometrically, each embedding represents a localized semantic coordinate of a solution under the current environmental state. However, due to the stochasticity of evolutionary search, individual representations often suffer from approximation noise and may not perfectly align with the true Pareto manifold.

To mitigate this individual-level noise, we calculate the joint centroid of the population:
\begin{equation}
    \bar{\mathbf{z}}_{joint} = \frac{1}{|\mathcal{P}_t|} \sum_{\mathbf{x} \in \mathcal{P}_t} \mathbf{z}_{joint}(\mathbf{x}).
\end{equation}
By aggregating the individual embeddings, this centroid effectively acts as a stable anchor representing the dominant topological trend of the current environment. This mechanism filters out stochastic micro-fluctuations, providing a robust macroscopic descriptor of the population distribution.

Benefiting from the topological smoothness and continuity induced by our latent space construction mechanism, we assume that the optimal solutions for the \textit{next} moment reside in the local geometric neighborhood of this centroid. Furthermore, introducing stochasticity into population updates rather than relying solely on deterministic greedy strategies can significantly enhance search efficiency and prevent premature convergence in MOEAs \cite{bian2025stochastic}. Therefore, we inject controlled randomness into the generation phase to thoroughly explore the latent neighborhood. Specifically, we generate a candidate pool $\mathcal{Z}_{cand}$ consisting of $N_{cand}$ samples via Gaussian perturbation:
\begin{equation}
    \mathbf{z}_{cand}^{(j)} = \bar{\mathbf{z}}_{joint} + \sigma_r \cdot \boldsymbol{\epsilon}^{(j)}, 
\end{equation}
where $\boldsymbol{\epsilon}^{(j)} \sim \mathcal{N}(\mathbf{0}, \mathbf{I})$ is the standard multivariate Gaussian noise for $j = 1, 2, \dots, N_{cand}$, and $\sigma_r$ controls the exploration intensity. These perturbed joint codes are then fed into the frozen Decoder $D_{\psi}$ and the Surrogate Predictor $P_{\xi}$ to obtain the reconstructed candidate solutions $\mathcal{X}_{cand}$ and their predicted objective values $\mathcal{Y}_{cand}$, respectively.

\subsubsection{Decomposition-based Screening}
To ensure the generated population is not only convergent but also diverse and aligned with the sub-problems of the downstream optimizer, we employ a greedy matching strategy based on the Augmented Tchebycheff decomposition.

Let $\{\mathbf{w}_1, \dots, \mathbf{w}_N\}$ be the set of weight vectors defining the sub-problems. We identify the ideal point $\mathbf{z}^*$ from the predicted values $\mathcal{Y}_{cand}$. For each sub-problem $k$, we aim to select a unique candidate $\mathbf{x}^{(k)}$ from the pool that minimizes the Augmented Tchebycheff cost:
\begin{equation} 
\text{cost}(\mathbf{x}, \mathbf{w}_k) = \max_{j} \{w_{k,j} | \hat{f}_j(\mathbf{x}) - z^*_j| \} + \rho \sum_{j} | \hat{f}_j(\mathbf{x}) - z^*_j |,
\end{equation}
where $\hat{f}(\mathbf{x})$ is the predicted objective vector.

The selection process proceeds iteratively: for each weight vector, we assign the candidate with the minimum cost that has not yet been selected. This bijective mapping is designed to maintain the distribution of the generated initial population along the Pareto Front, providing an evenly distributed starting point for the subsequent evolutionary search.

\section{EXPERIMENTS}
\subsection{Benchmark Test Problems and Performance Indicators}
\subsubsection{Test Problems}
{To comprehensively evaluate the effectiveness of the proposed framework, we conduct experiments on two widely recognized DMOP benchmark suites}: the FDA test suite \cite{farina2004dynamic} and the CEC 2018 benchmark suite (DF1 through DF14)\cite{jiang2018benchmark}. In addition to these synthetic benchmarks, we also incorporate the real-world inspired Dynamic Resource Allocation (DRA) and Dynamic Path Planning (DPP) problems \cite{lavalle2006planning} to verify the practical applicability of the model.

\subsubsection{Performance Indicators}
The performance of the algorithms is quantified using two metrics: Mean Inverted Generational Distance (MIGD) \cite{zitzler2003performance} and Mean Hypervolume (MHV) \cite{while2006faster}. 
The MIGD metric calculates the average IGD across all environments over the entire evolutionary process. It serves as a comprehensive indicator of both the convergence of the solution set and its diversity along the true Pareto Front. A smaller MIGD value indicates superior performance. 
Conversely, the MHV metric measures the average Hypervolume (HV) covered by the solution set across all time steps. It assesses the ability of algorithm to maintain a wide and well-distributed coverage in the objective space. Unlike MIGD, a larger MHV value signifies better performance.

\subsection{Competitors and Parameter Settings}
\subsubsection{Competitors}
To demonstrate the performance of our proposed method, we compare it against four state-of-the-art algorithms that represent different mainstream categories in dynamic multi-objective optimization. 
Specifically, the selected competitors include STT-MOEA/D~\cite{wang2024moea}, which employs spatial-temporal topological tensor prediction to forecast the location of the PS in the subsequent environment; DIP-DMOEA~\cite{ye2024learning}, a method based on directional improvement prediction; the recently proposed VARE~\cite{jiang2025dynamic}, which utilizes vector autoregressive evolution; and SIKT-DMOEA~\cite{hu2025learning}, an algorithm driven by similarity identification and knowledge transfer.

\subsubsection{Parameter Settings}
To comprehensively evaluate the performance of the algorithm under different dynamic characteristics, we adopt three standard configurations for the change frequency ($\tau_t$) and change severity ($n_t$): $(10, 10)$ for standard changes, $(5, 10)$ for high-frequency changes, and $(10, 5)$ for severe changes. These settings are widely adopted in recent DMOP literature \cite{jiang2025dynamic, wang2024moea}. 

During the online tracking phase, the framework employs MOEA/D equipped with Differential Evolution (DE) operators for static optimization. For the offline preparation, the pre-training dataset is constructed by instantiating multiple historical problems under various $n_t$ configurations, yielding a collection of approximately $1.2 \times 10^8$ samples to ensure a robust approximation of the latent manifolds. All algorithms are executed for 20 independent runs on each test instance to mitigate statistical randomness. Due to space constraints, the detailed parameter configuration tables for the DMOP benchmarks, the DB-GEN architecture, and the static optimizer, along with the specific methodology for training data generation, are provided in Appendix A-A.

\subsection{Overall Performance Comparison}
As shown in Tables \ref{tab:igd_comparison} and \ref{tab:hv_comparison}, our method exhibits highly competitive performance across the majority of test instances. The evaluation employs the MIGD (smaller is better) and MHV (larger is better) metrics. To facilitate observation, the best and second-best mean values are highlighted with dark and light gray backgrounds, respectively. Furthermore, the symbols "$+$", "$-$", and "$\approx$" appended to the baseline results indicate that the compared algorithm performs significantly better than, worse than, or similarly to our proposed method, according to the Wilcoxon rank-sum test at a 5\% significance level.

Statistically, out of the 57 tested dynamic environment configurations, DB-GEN achieves the optimal MIGD (dark gray) on 49 instances and the sub-optimal (light gray) on 5 instances, yielding a remarkable Top-2 hit rate of 94.7\%. For the MHV metric, Table \ref{tab:hv_comparison} presents the comparison results on the FDA1 to FDA5 instances due to space limitations. On this subset, our method secures the optimal MHV results in 13 out of the 15 configurations. Furthermore, the non-parametric Friedman test at a 5\% significance level reveals that DB-GEN achieves an outstanding average rank of 1.29 for the MIGD metric (across all 57 instances) and 1.20 for the MHV metric (on the presented FDA subset). These statistically significant scores, closely approaching the theoretical optimum of 1.0, quantitatively demonstrate that DB-GEN consistently secures the top position across the test environments.

It is worth noting that DF6 and DF9 are generally considered challenging problems due to their complex and irregular Pareto Front changes. Consequently, baseline methods often exhibit poor IGD performance and struggle to track the shifting PF, particularly under severe environmental mutations ($n_t=10, \tau_t=5$). Under such severe settings, explicit numerical predictions frequently fail. In contrast, DB-GEN demonstrates remarkable robustness, yielding a substantial quantitative improvement. For instance, on the DF6 problem ($n_t=10, \tau_t=5$), DB-GEN achieves an IGD of 4.26E-01, reducing the tracking error by approximately 66.4\% compared to the best-performing baseline (DIP-DMOEA with 1.27E+00). Similarly, on the DF9 problem under the same severe configuration, DB-GEN lowers the IGD from the baseline best of 2.78E-01 to 8.00E-02, marking a 71.2\% performance gain. This indicates that the continuous latent space modeling enables reliable zero-shot solution generation, effectively handling environments where historical data is severely deceptive.

\begin{table*}[htbp]
\centering
\caption{The MIGD values and standard deviation in parentheses obtained by SOTAs and Ours.}
\label{tab:igd_comparison}

\definecolor{bestbg}{gray}{0.75} 
\definecolor{secondbg}{gray}{0.85} 
\newcommand{\hlbest}[1]{\cellcolor{bestbg}#1}
\newcommand{\hlsecond}[1]{\cellcolor{secondbg}#1}

\resizebox{\textwidth}{!}{
\begin{tabular}{c|c|c|c|c|c|c|c}
\toprule
Problem & $n_t$ & $\tau_t$ & STT-MOEA/D & DIP-DMOEA & VARE & SIKT-DMOEA & Ours \\
\midrule
\multirow{3}{*}{DF1} & 5 & 10 & 7.36e-02(8.78e-03)$-$ & \hlsecond{3.71e-02(3.15e-04)}$-$ & 4.22e-02(3.33e-03)$-$ & 5.64e-02(8.69e-03)$-$ & \hlbest{1.31e-02(4.08e-03)} \\
 & 10 & 5 & 6.75e-01(3.59e-03)$-$ & \hlsecond{6.48e-02(5.87e-03)}$-$ & 1.24e-01(1.69e-02)$-$ & 1.50e-01(2.38e-02)$-$ & \hlbest{2.10e-02(9.62e-04)} \\
 & 10 & 10 & 6.33e-01(7.77e-02)$-$ & \hlsecond{2.33e-02(4.03e-03)}$-$ & 2.85e-02(1.85e-03)$-$ & 5.35e-02(7.42e-03)$-$ & \hlbest{1.05e-02(2.22e-04)} \\
\midrule
\multirow{3}{*}{DF2} & 5 & 10 & \hlsecond{2.19e-02(1.38e-02)}$\approx$ & 4.97e-02(6.99e-03)$-$ & 6.51e-02(5.91e-03)$-$ & 4.21e-02(4.36e-03)$-$ & \hlbest{2.15e-02(5.04e-03)} \\
 & 10 & 5 & 4.02e-01(6.23e-03)$-$ & \hlsecond{9.55e-02(8.71e-03)}$-$ & 1.36e-01(9.32e-02)$-$ & 1.08e-01(7.67e-03)$-$ & \hlbest{4.69e-02(1.64e-03)} \\
 & 10 & 10 & 3.85e-01(3.81e-03)$-$ & 3.82e-02(8.73e-04)$-$ & 4.22e-02(3.67e-03)$-$ & \hlsecond{3.63e-02(3.91e-03)}$-$ & \hlbest{1.57e-02(1.06e-03)} \\
\midrule
\multirow{3}{*}{DF3} & 5 & 10 & 2.05e-01(1.91e-02)$-$ & \hlsecond{1.05e-01(7.91e-03)}$+$ & 1.32e-01(2.43e-02)$\approx$ & \hlbest{4.43e-02(3.74e-03)}$+$ & 1.41e-01(3.67e-03) \\
 & 10 & 5 & 6.48e-01(2.77e-02)$-$ & 1.51e-01(5.11e-04)$-$ & 3.37e-01(1.44e-01)$-$ & \hlbest{1.07e-01(6.85e-03)}$+$ & \hlsecond{1.37e-01(8.45e-03)} \\
 & 10 & 10 & 6.70e-01(1.24e-02)$-$ & \hlsecond{8.65e-02(6.30e-04)}$+$ & 1.41e-01(2.23e-02)$-$ & \hlbest{3.67e-02(1.93e-03)}$+$ & 9.68e-02(1.16e-02) \\
\midrule
\multirow{3}{*}{DF4} & 5 & 10 & 4.31e-01(6.49e-02)$-$ & 1.18e-01(5.28e-02)$-$ & \hlbest{4.07e-02(2.44e-03)}$+$ & \hlsecond{6.23e-02(4.12e-03)}$\approx$ & \hlsecond{6.23e-02(2.17e-03)} \\ 
 & 10 & 5 & 8.68e-01(5.39e-02)$-$ & 1.94e-01(6.51e-02)$-$ & \hlsecond{7.63e-02(1.29e-03)}$\approx$ & 1.23e-01(8.37e-03)$-$ & \hlbest{7.56e-02(2.14e-03)} \\
 & 10 & 10 & 8.34e-01(1.43e-02)$-$ & 1.23e-01(9.53e-02)$-$ & \hlbest{3.49e-02(9.83e-03)}$+$ & 5.67e-02(2.63e-03)$\approx$ & \hlsecond{5.59e-02(9.25e-04)} \\
\midrule
\multirow{3}{*}{DF5} & 5 & 10 & 2.77e-02(1.06e-02)$-$ & \hlsecond{2.10e-02(9.42e-04)}$-$ & 4.65e-02(4.04e-03)$-$ & 9.73e-02(2.13e-02)$-$ & \hlbest{1.38e-02(3.58e-04)} \\
 & 10 & 5 & 6.42e-01(5.29e-03)$-$ & \hlsecond{4.29e-02(7.37e-04)}$-$ & 1.30e-01(4.03e-03)$-$ & 2.33e-01(6.05e-02)$-$ & \hlbest{2.32e-02(9.10e-04)} \\
 & 10 & 10 & 6.16e-01(4.29e-03)$-$ & \hlsecond{1.84e-02(9.86e-04)}$-$ & 3.71e-02(2.44e-03)$-$ & 5.21e-02(6.94e-03)$-$ & \hlbest{1.26e-02(1.62e-04)} \\
\midrule
\multirow{3}{*}{DF6} & 5 & 10 & 7.15e-01(1.38e+00)$-$ & \hlsecond{6.91e-01(8.83e-02)}$-$ & 1.80e+00(4.30e-01)$-$ & 2.18e+00(3.20e-01)$-$ & \hlbest{1.62e-01(2.01e-01)} \\
 & 10 & 5 & 4.35e+00(7.91e-01)$-$ & \hlsecond{1.27e+00(2.33e-01)}$-$ & 6.08e+00(7.90e-01)$-$ & 6.24e+00(7.43e-01)$-$ & \hlbest{4.26e-01(2.27e-01)} \\
 & 10 & 10 & 3.43e+00(5.64e-01)$-$ & \hlsecond{9.50e-01(4.05e-02)}$-$ & 5.44e+00(1.34e+00)$-$ & 3.00e+00(2.72e-01)$-$ & \hlbest{3.62e-01(6.43e-02)} \\
\midrule
\multirow{3}{*}{DF7} & 5 & 10 & 1.52e+00(5.14e-02)$-$ & 3.25e-01(7.91e-02)$-$ & 1.15e-01(3.38e-02)$-$ & \hlbest{6.76e-02(1.70e-02)}$+$ & \hlsecond{9.01e-02(1.83e-02)} \\
 & 10 & 5 & 6.05e-01(4.53e-03)$-$ & 2.12e-01(8.24e-04)$-$ & 2.56e-01(9.75e-03)$-$ & \hlsecond{1.19e-01(3.46e-02)}$-$ & \hlbest{7.20e-02(2.22e-02)} \\
 & 10 & 10 & 5.95e-01(2.50e-03)$-$ & 2.13e-01(6.36e-03)$-$ & 8.83e-02(7.41e-03)$-$ & \hlbest{4.43e-02(2.45e-02)}$+$ & \hlsecond{5.98e-02(8.51e-03)} \\
\midrule
\multirow{3}{*}{DF8} & 5 & 10 & 5.29e-02(1.83e-02)$-$ & 6.22e-02(4.60e-03)$-$ & 4.33e-02(3.32e-03)$-$ & \hlsecond{3.49e-02(2.34e-03)}$-$ & \hlbest{1.60e-02(5.88e-04)} \\
 & 10 & 5 & 5.44e-01(6.32e-03)$-$ & \hlsecond{6.40e-02(9.37e-04)}$-$ & 7.57e-02(4.61e-04)$-$ & 7.12e-02(4.65e-03)$-$ & \hlbest{1.79e-02(1.49e-04)} \\
 & 10 & 10 & 5.34e-01(6.21e-03)$-$ & 5.86e-02(9.69e-04)$-$ & 4.52e-02(2.81e-03)$-$ & \hlsecond{4.20e-02(3.63e-03)}$-$ & \hlbest{1.56e-02(7.03e-04)} \\
\midrule
\multirow{3}{*}{DF9} & 5 & 10 & \hlsecond{1.01e-01(6.42e-02)}$-$ & 3.01e-01(5.22e-04)$-$ & 3.72e-01(1.04e-01)$-$ & 3.38e-01(5.27e-02)$-$ & \hlbest{6.81e-02(2.35e-03)} \\
 & 10 & 5 & 8.81e-01(5.87e-02)$-$ & \hlsecond{3.25e-01(6.65e-03)}$-$ & 6.65e-01(5.78e-02)$-$ & 5.64e-01(1.02e-01)$-$ & \hlbest{8.00e-02(9.30e-03)} \\
 & 10 & 10 & 6.70e-01(3.10e-02)$-$ & \hlsecond{2.02e-01(9.69e-04)}$-$ & 2.77e-01(8.25e-02)$-$ & 2.39e-01(4.92e-02)$-$ & \hlbest{4.28e-02(5.25e-03)} \\
\midrule
\multirow{3}{*}{DF10} & 5 & 10 & 3.28e-01(5.29e-02)$-$ & \hlsecond{1.90e-01(4.46e-04)}$-$ & 1.92e-01(1.04e-02)$-$ & 2.01e-01(1.25e-02)$-$ & \hlbest{1.53e-02(6.86e-04)} \\
 & 10 & 5 & 3.77e-01(4.44e-03)$-$ & \hlsecond{1.95e-01(7.00e-04)}$-$ & 2.45e-01(9.24e-03)$-$ & 2.21e-01(1.32e-02)$-$ & \hlbest{2.55e-02(2.31e-03)} \\
 & 10 & 10 & 3.93e-01(3.42e-03)$-$ & 2.06e-01(3.44e-04)$-$ & 2.10e-01(4.80e-02)$-$ & \hlsecond{2.05e-01(1.34e-02)}$-$ & \hlbest{1.22e-02(7.26e-04)} \\
\midrule
\multirow{3}{*}{DF11} & 5 & 10 & 3.40e-01(7.89e-02)$-$ & 1.26e-01(3.79e-04)$-$ & 1.18e+00(3.14e-03)$-$ & \hlsecond{1.19e-01(6.71e-03)}$-$ & \hlbest{9.00e-03(2.11e-04)} \\
 & 10 & 5 & 8.34e-01(1.50e-03)$-$ & \hlsecond{1.33e-01(7.56e-04)}$-$ & 1.25e+00(1.89e-03)$-$ & 1.36e-01(4.83e-03)$-$ & \hlbest{1.21e-02(2.82e-04)} \\
 & 10 & 10 & 8.26e-01(1.63e-03)$-$ & 1.25e-01(6.56e-04)$-$ & 1.06e+00(3.90e-03)$-$ & \hlsecond{1.07e-01(3.92e-03)}$-$ & \hlbest{7.45e-03(2.93e-04)} \\
\midrule
\multirow{3}{*}{DF12} & 5 & 10 & 4.46e-01(5.60e-02)$-$ & 2.80e-01(6.35e-04)$-$ & 4.44e-01(3.67e-02)$-$ & \hlbest{5.68e-02(7.78e-03)}$\approx$ & \hlbest{5.68e-02(7.22e-03)} \\
 & 10 & 5 & 5.64e-01(4.69e-02)$-$ & 1.94e-01(8.24e-04)$-$ & 2.61e-01(1.03e-02)$-$ & \hlsecond{5.94e-02(8.90e-03)}$-$ & \hlbest{5.11e-02(5.68e-03)} \\
 & 10 & 10 & 4.68e-01(5.47e-02)$-$ & 1.77e-01(1.54e-03)$-$ & 4.73e-01(3.11e-02)$-$ & \hlsecond{3.05e-02(6.35e-03)}$\approx$  & \hlbest{2.89e-02(2.68e-03)}\\
\midrule
\multirow{3}{*}{DF13} & 5 & 10 & 2.47e-01(1.12e-02)$-$ & 1.95e-01(2.76e-03)$-$ & \hlsecond{1.85e-01(6.46e-03)}$-$ & 2.72e-01(8.13e-03)$-$ & \hlbest{1.14e-01(4.74e-03)} \\
 & 10 & 5 & 6.48e-01(7.15e-03)$-$ & \hlsecond{2.19e-01(6.97e-04)}$-$ & 2.49e-01(2.43e-02)$-$ & 3.46e-01(1.81e-02)$-$ & \hlbest{1.25e-01(5.71e-03)} \\
 & 10 & 10 & 6.21e-01(1.04e-02)$-$ & 2.02e-01(8.87e-03)$-$ & \hlsecond{1.71e-01(4.48e-03)}$-$ & 2.56e-01(4.92e-03)$-$ & \hlbest{1.14e-01(6.58e-03)} \\
\midrule
\multirow{3}{*}{DF14} & 5 & 10 & 8.89e-02(2.02e-03)$-$ & \hlsecond{7.17e-02(4.03e-04)}$-$ & 1.10e-01(4.18e-03)$-$ & 8.34e-02(2.59e-03)$-$ & \hlbest{6.34e-02(1.60e-04)} \\
 & 10 & 5 & 1.68e-01(6.05e-03)$-$ & \hlsecond{8.21e-02(7.21e-03)}$-$ & 2.09e-01(4.19e-02)$-$ & 1.49e-01(1.66e-02)$-$ & \hlbest{6.83e-02(1.38e-03)} \\
 & 10 & 10 & 1.53e-01(3.38e-03)$-$ & \hlsecond{6.90e-02(9.81e-04)}$-$ & 1.60e-01(4.50e-03)$-$ & 7.36e-02(2.09e-03)$-$ & \hlbest{6.13e-02(1.68e-04)} \\
\midrule
\multirow{3}{*}{FDA1} & 5 & 10 & 4.90e-01(7.10e-03)$-$ & 3.98e-01(3.88e-04)$-$ & 3.86e-01(1.48e-02)$-$ & \hlsecond{6.99e-02(1.08e-02)}$-$  & \hlbest{1.64e-02(8.05e-03)} \\
 & 10 & 5 & 5.34e-01(4.71e-03)$-$ & 3.30e-01(6.65e-03)$-$ & 2.58e-01(3.59e-02)$-$ & \hlsecond{1.49e-01(3.17e-02)}$-$ & \hlbest{2.58e-02(3.02e-03)} \\
 & 10 & 10 & 5.08e-01(5.54e-03)$-$ & 1.48e-01(7.47e-04)$-$ & 2.19e-01(3.24e-02)$-$ & \hlsecond{3.95e-02(2.77e-03)}$-$  & \hlbest{1.22e-02(2.25e-03)} \\
\midrule
\multirow{3}{*}{FDA2} & 5 & 10 & 1.24e+00(7.21e-03)$-$ & 1.96e-01(7.96e-02)$-$ & 2.05e-01(2.20e-03)$-$ & \hlsecond{2.78e-02(1.80e-03)}$-$ & \hlbest{2.45e-02(3.24e-03)} \\
 & 10 & 5 & 1.91e-01(1.43e-03)$-$ & 2.16e-01(6.92e-04)$-$ & 1.21e+00(3.55e-03)$-$ & \hlsecond{7.86e-02(1.12e-02)}$-$  & \hlbest{1.15e-02(5.87e-03)}\\
 & 10 & 10 & 1.15e+00(2.10e-03)$-$ & 9.98e-01(9.42e-04)$-$ & 1.04e+00(2.78e-03)$-$ & \hlsecond{2.38e-02(1.18e-03)}$\approx$  & \hlbest{2.34e-02(5.26e-03)}\\
\midrule
\multirow{3}{*}{FDA3} & 5 & 10 & 1.47e-01(2.12e-03)$-$ & 1.54e-01(1.24e-02)$-$ & 2.29e-01(2.99e-02)$-$ & \hlsecond{1.23e-01(1.18e-02)}$-$ & \hlbest{5.43e-02(4.79e-04)} \\
 & 10 & 5 & 1.71e-01(3.12e-03)$-$ & \hlsecond{1.23e-01(3.80e-04)}$-$ & 2.09e-01(2.33e-02)$-$ & 2.61e-01(3.35e-02)$-$ & \hlbest{5.94e-02(1.30e-03)} \\
 & 10 & 10 & \hlsecond{7.53e-02(4.41e-03)}$-$ & 2.46e-01(3.42e-04)$-$ & 4.39e-01(4.14e-02)$-$ & 8.50e-02(9.16e-03)$-$ & \hlbest{5.35e-02(6.42e-04)} \\
\midrule
\multirow{3}{*}{FDA4} & 5 & 10 & 5.34e-01(1.04e-01)$-$ & 7.86e-01(5.68e-03)$-$ & 1.85e+00(6.89e-03)$-$ & \hlsecond{1.03e-01(4.95e-03)}$-$ & \hlbest{7.32e-02(1.25e-03)} \\
 & 10 & 5 & 5.84e-01(2.74e-01)$-$ & 6.48e-01(7.49e-04)$-$ & 2.49e-01(3.75e-02)$-$ & \hlsecond{1.76e-01(1.36e-02)}$-$ & \hlbest{8.58e-02(1.03e-03)} \\
 & 10 & 10 & 4.38e-01(5.39e-02)$-$ & 5.59e-01(3.09e-04)$-$ & 1.75e-01(5.02e-03)$-$ & \hlsecond{9.45e-02(2.41e-03)}$-$ & \hlbest{7.26e-02(7.87e-04)} \\
\midrule
\multirow{3}{*}{FDA5} & 5 & 10 & \hlbest{5.89e-02(2.40e-03)}$+$ & 8.24e-02(5.42e-04)$+$ & \hlsecond{7.67e-02(4.68e-03)}$+$ & 3.33e-01(2.87e-02)$-$ & 1.14e-01(8.36e-03) \\
 & 10 & 5 & 3.51e-01(2.08e-03)$-$ & 2.81e-01(7.33e-04)$-$ & \hlsecond{2.06e-01(1.87e-02)}$-$ & 4.64e-01(3.83e-02)$-$ & \hlbest{9.33e-02(9.55e-03)} \\
 & 10 & 10 & 4.48e-01(1.92e-03)$-$ & 4.46e-01(2.31e-04)$-$ & \hlsecond{1.16e-01(5.60e-03)}$\approx$ & 3.34e-01(2.86e-02)$-$ & \hlbest{1.16e-01(8.10e-03)} \\
\midrule
\midrule
\multicolumn{3}{l|}{\textbf{Average Rank } \hspace{2em}} & 4.30 & 2.95 & 3.63 & \hlsecond{2.82} & \hlbest{1.29} \\

\bottomrule
\end{tabular}
}
\end{table*}

\begin{table*}[htbp]
\centering
\caption{The MHV values and standard deviation in parentheses obtained by SOTAs and Ours.}
\label{tab:hv_comparison}
\definecolor{bestbg}{gray}{0.75} 
\definecolor{secondbg}{gray}{0.85} 

\newcommand{\hlbest}[1]{\cellcolor{bestbg}#1}
\newcommand{\hlsecond}[1]{\cellcolor{secondbg}#1}

\resizebox{\textwidth}{!}{
\begin{tabular}{c|c|c|c|c|c|c|c}
\toprule
Problem & $n_t$ & $\tau_t$ & STT-MOEA/D & DIP-DMOEA & VARE & SIKT-DMOEA & Ours \\
\midrule
    \multirow{3}{*}{FDA1} & 5 & 10 & 1.07e-01(4.61e-03)$-$ & 5.31e-01(1.68e-03)$-$ & 6.01e-01(1.57e-03)$-$ & \hlsecond{6.17e-01(1.49e-02)}$-$ & \hlbest{6.27e-01(6.56e-03)} \\
      & 10 & 5 & 7.13e-02(3.59e-03)$-$ & 6.51e-01(4.69e-02)$-$ & \hlsecond{6.56e-01(6.34e-03)}$-$ & 5.08e-01(4.21e-02)$-$ & \hlbest{6.74e-01(3.47e-03)} \\
      & 10 & 10 & 1.02e-01(3.55e-03)$-$ & \hlsecond{7.19e-01(4.46e-03)}$+$ & \hlbest{8.07e-01(5.19e-04)}$+$ & 6.60e-01(3.99e-03)$\approx$ & 6.75e-01(4.19e-03) \\
    \midrule
    \multirow{3}{*}{FDA2} & 5 & 10 & 1.74e-03(3.06e-03)$-$ & 5.54e-01(3.50e-03)$-$ & \hlsecond{6.84e-01(1.08e-02)}$-$ & 5.64e-01(1.69e-03)$-$ & \hlbest{2.00e+00(1.14e-02)} \\
      & 10 & 5 & 1.33e-02(1.79e-03)$-$ & \hlsecond{7.11e-01(1.46e-02)}$-$ & 6.07e-01(2.29e-04)$-$ & 5.57e-01(1.35e-02)$-$ & \hlbest{1.82e+00(2.01e-02)} \\
      & 10 & 10 & 2.63e-01(1.68e-03)$-$ & \hlsecond{7.90e-01(4.57e-03)}$-$ & 5.85e-01(5.56e-04)$-$ & 5.65e-01(1.71e-03)$-$ & \hlbest{1.96e+00(2.85e-02)} \\
    \midrule
    \multirow{3}{*}{FDA3} & 5 & 10 & 2.96e-01(1.16e-03)$-$ & 5.33e-01(4.75e-04)$-$ & \hlsecond{6.30e-01(8.86e-03)}$-$ & 4.59e-01(1.00e-02)$-$ & \hlbest{6.72e-01(1.01e-03)} \\
      & 10 & 5 & 4.39e-01(2.24e-03)$-$ & 3.39e-01(7.72e-04)$-$ & \hlsecond{5.71e-01(8.31e-03)}$-$ & 3.38e-01(2.23e-02)$-$ & \hlbest{6.59e-01(2.78e-03)} \\
      & 10 & 10 & \hlsecond{6.50e-01(7.14e-03)}$\approx$ & 6.37e-01(5.32e-04)$-$ & 5.82e-01(2.93e-04)$-$ & 5.06e-01(8.91e-03)$-$ & \hlbest{6.72e-01(1.04e-03)} \\
    \midrule 
    \multirow{3}{*}{FDA4} & 5 & 10 & 1.70e-01(8.38e-03)$-$ & 3.36e-01(4.65e-04)$-$ & 3.33e-01(8.18e-03)$-$ & \hlsecond{4.58e-01(7.47e-03)}$\approx$ & \hlbest{4.68e-01(2.19e-03)} \\
      & 10 & 5 & 1.11e-01(3.86e-02)$-$ & 3.05e-01(1.36e-03)$-$ & 2.26e-01(1.40e-03)$-$ & \hlsecond{3.34e-01(1.56e-02)}$-$ & \hlbest{3.42e-01(3.32e-03)} \\
      & 10 & 10 & 1.70e-01(3.00e-02)$-$ & 4.12e-01(2.38e-04)$-$ & 4.51e-01(1.57e-04)$-$ & \hlsecond{4.77e-01(4.99e-03)}$-$ & \hlbest{4.93e-01(8.06e-04)} \\
    \midrule
    \multirow{3}{*}{FDA5} & 5 & 10 & \hlbest{4.90e-01(3.94e-02)}$\approx$ & 4.40e-01(2.27e-04)$-$ & 4.18e-01(1.37e-04)$-$ & \hlsecond{4.51e-01(1.37e-02)}$\approx$ & 4.59e-01(3.86e-04) \\
      & 10 & 5 & 7.12e-02(1.50e-02)$-$ & 2.42e-01(5.28e-04)$-$ & \hlsecond{3.95e-01(1.16e-03)}$-$ & 3.33e-01(1.47e-02)$-$ & \hlbest{4.00e-01(8.76e-04)} \\
      & 10 & 10 & 1.20e-01(1.71e-02)$-$ & 3.54e-01(4.65e-03)$-$ & \hlsecond{4.62e-01(5.53e-04)}$-$ & 4.58e-01(8.49e-03)$-$ & \hlbest{4.98e-01(2.09e-04)} \\
\midrule
\midrule

\multicolumn{3}{l|}{\textbf{Average Rank} \hspace{1em}} & 4.40 & 3.27 & \hlsecond{2.80} & 3.33 & \hlbest{1.20} \\
\bottomrule
\end{tabular}
}
\end{table*}

\subsection{Dynamic Adaptability Analysis}
{To gain a deeper understanding of the adaptability of the proposed framework across dynamic environments, this section provides a detailed analysis of its evolutionary behavior.} Specifically, we examine the algorithm's tracking stability over sequential environmental changes, visualize the mechanisms underlying its zero-shot population generation, and investigate its generalization boundaries when encountering out-of-distribution topological dynamics.

\subsubsection{Evolutionary Stability Tracking}
\begin{figure}[htbp]
    \centering
    \includegraphics[width=0.49\linewidth]{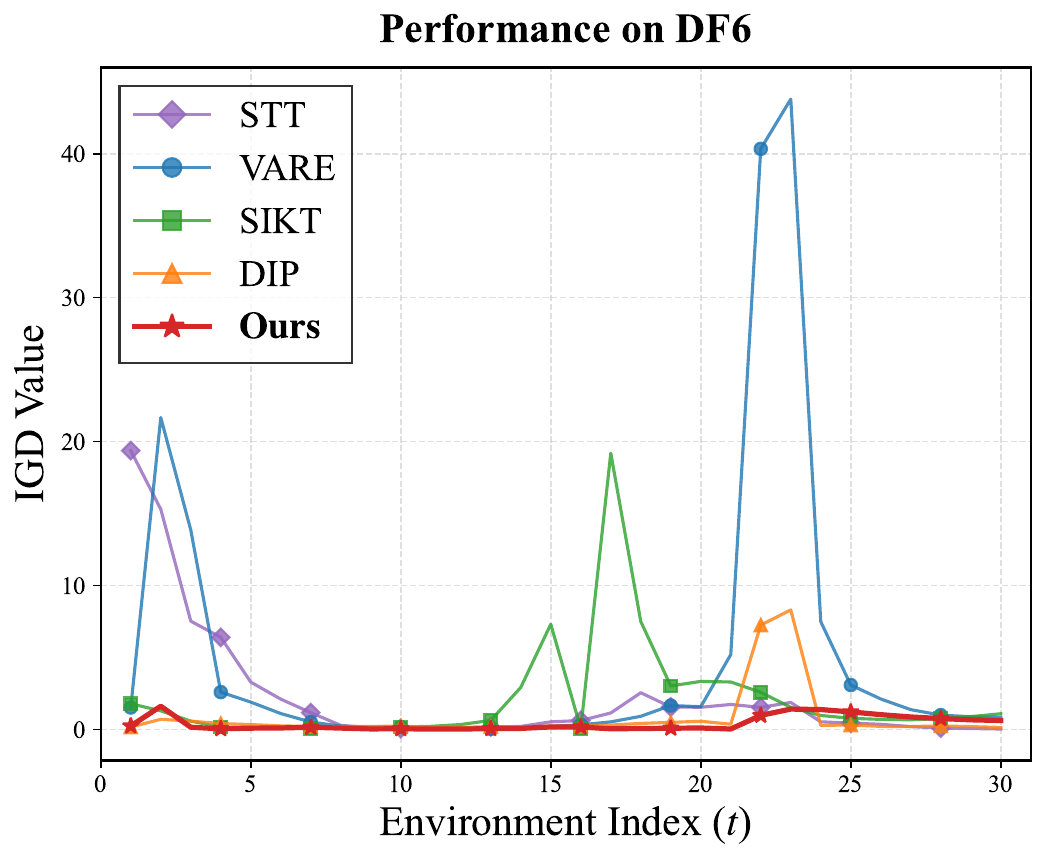}
    \hfill
    \includegraphics[width=0.49\linewidth]{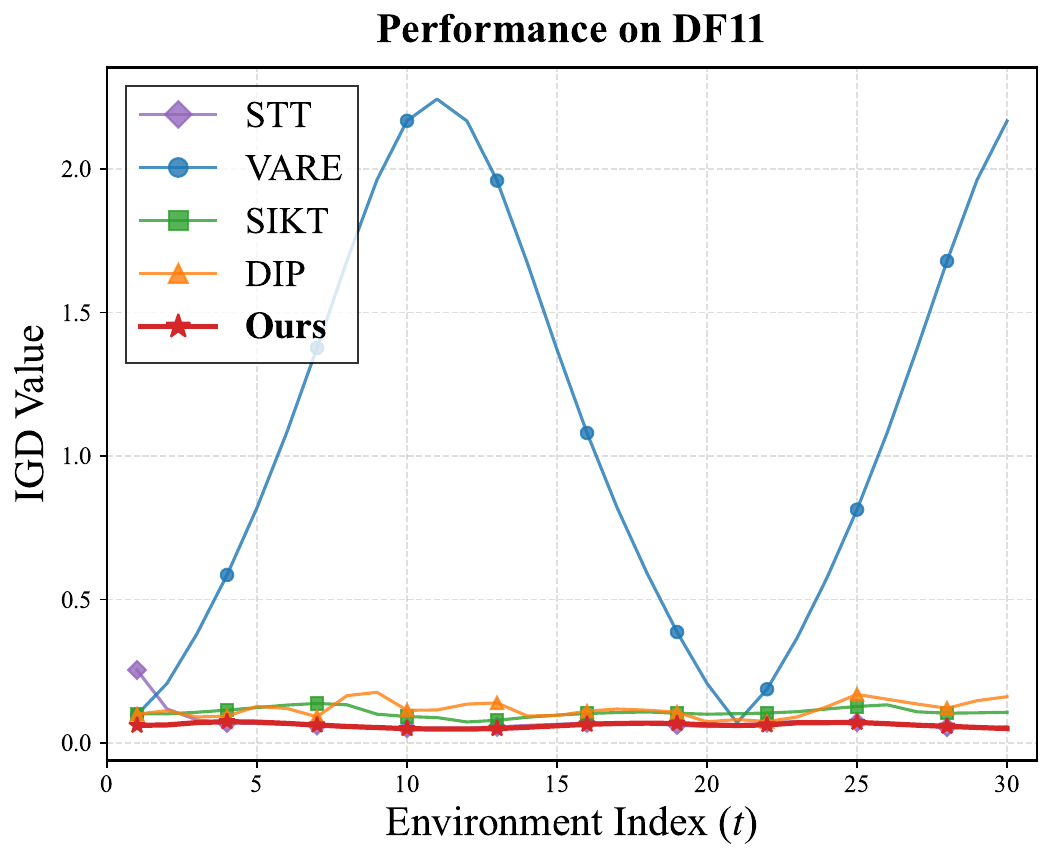}
    \caption{The IGD variation curves over environments.}
    \label{fig:igd_curve}
\end{figure}

Fig.~\ref{fig:igd_curve} illustrates the trajectory of the final IGD values obtained at the end of each environment on representative problems. This metric serves as an indicator of the ability of the algorithm to converge to the new Pareto Front within a limited evolutionary window. For a comprehensive comparison, the complete IGD tracking curves across the entire DF test suite are provided in Appendix A-C.

As observed in Fig.~\ref{fig:igd_curve}, baseline methods, such as the prediction-based VARE, exhibit noticeable fluctuations across different environments. The variations in their IGD curves suggest that these methods experience varying degrees of convergence difficulty. This behavior indicates that while traditional predictive models are effective for moderate changes, they may encounter challenges when facing more abrupt topological shifts, occasionally resulting in decreased solution quality during specific periods.

In comparison, the proposed method maintains a relatively stable IGD trajectory throughout the sequential changes. The model demonstrates consistent convergence behavior across the tested environments. It suggests that the decoupled frequency encoding and structure learning mechanisms assist the model in adapting to diverse dynamic patterns, facilitating steady convergence across various environmental mutation severities.

\subsubsection{Visualization of Zero-Shot Generation}
\begin{figure}[t] 
    \centering
    \subfloat[ \label{fig:df1_tracking_a}]{
        \includegraphics[width=0.48\linewidth]{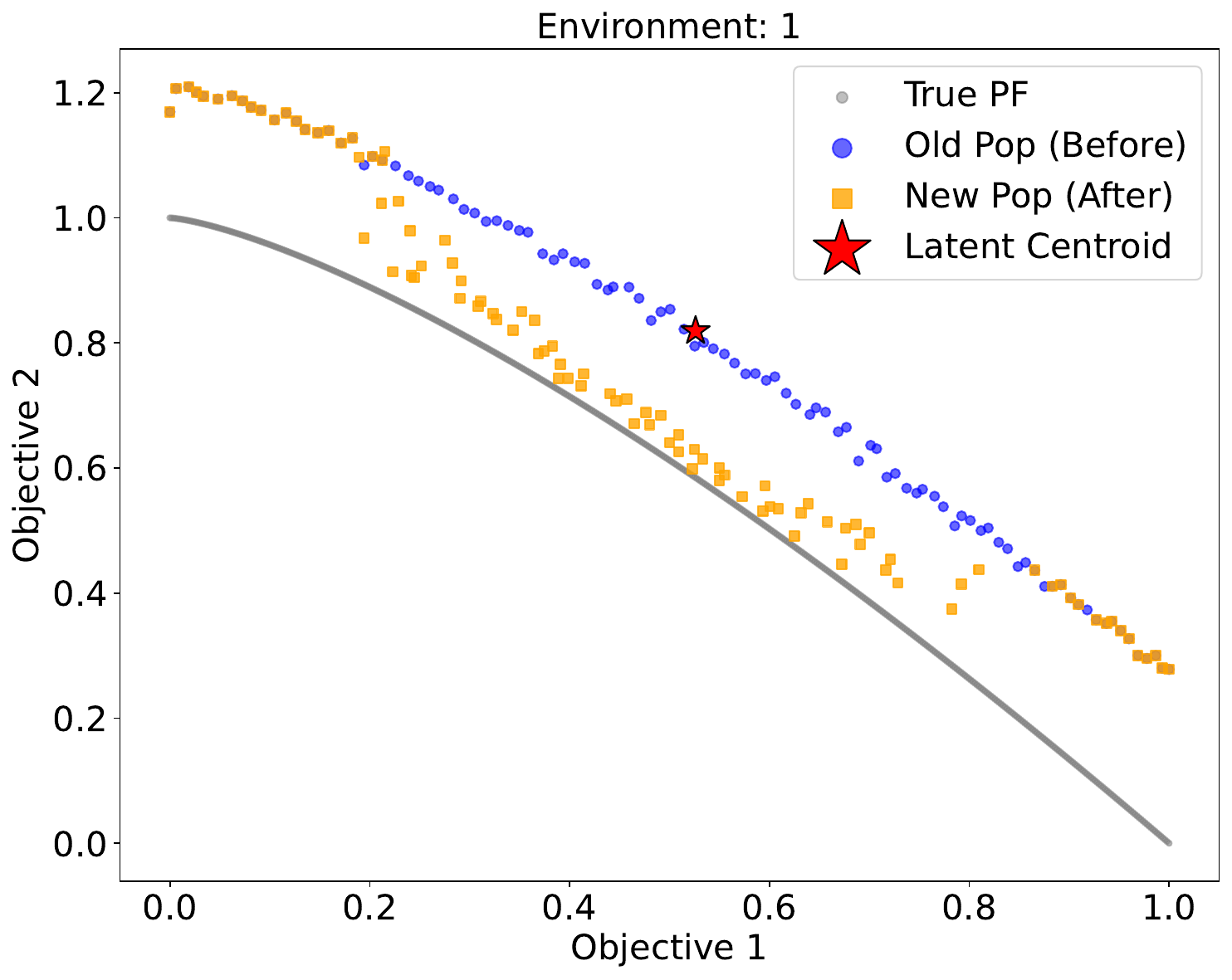}
    }
    \subfloat[\label{fig:df1_tracking_b}]{
        \includegraphics[width=0.48\linewidth]{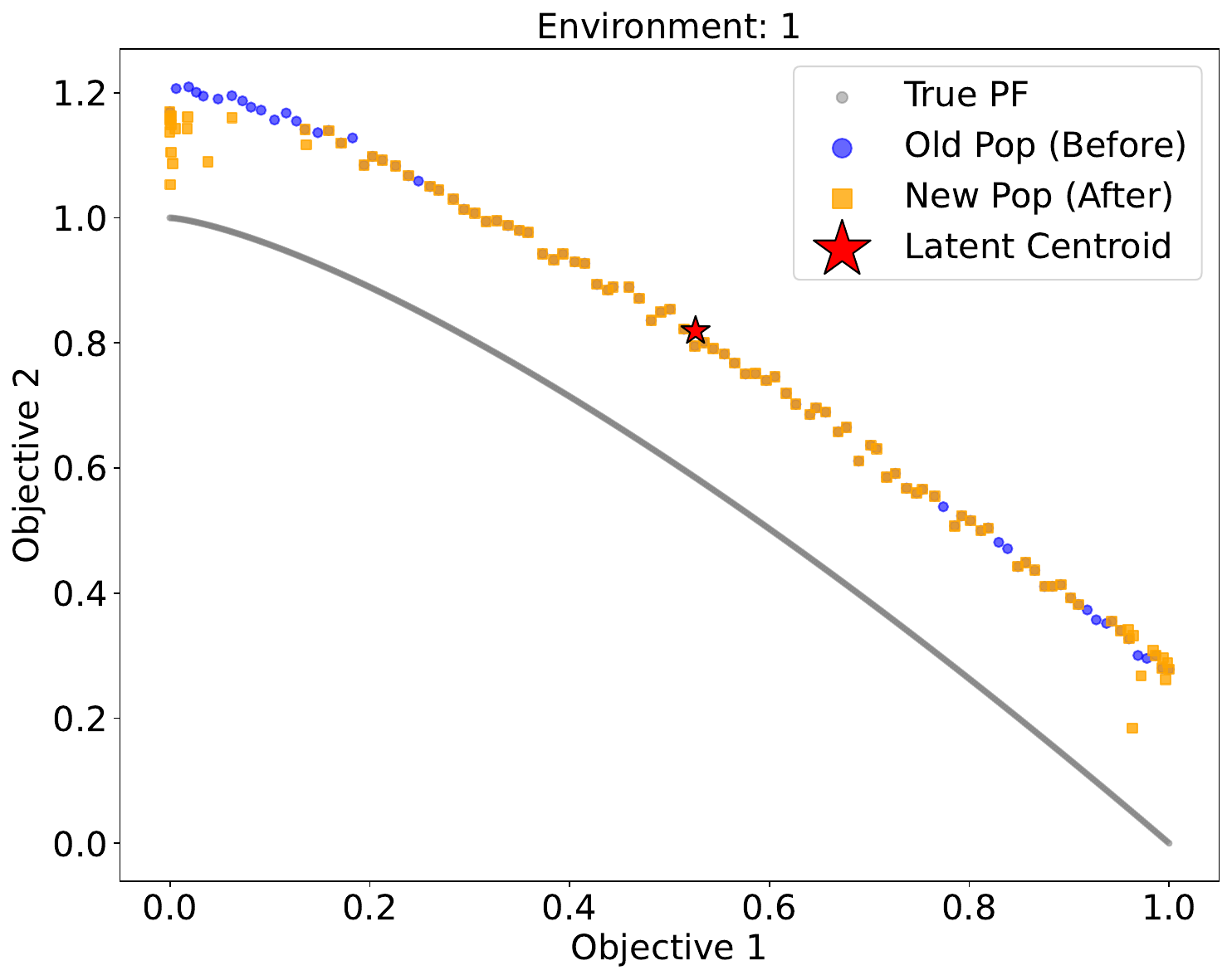}
    }
    \caption{Zero-shot solution generation upon specific environmental change on the DF1 problem. (a) Proposed latent centroid perturbation. (b) Perturbation on all previous solutions.}
    \label{fig:df1_tracking}
\end{figure}

To demonstrate the model's capability to handle data sparsity and adapt to novel environments, Fig.~\ref{fig:df1_tracking} visualizes the distribution of solutions in the objective space upon specific environmental change on the DF1 problem. For a comprehensive view, the complete tracking results across consecutive environments are detailed in Appendix A-B. In these plots, the grey dots delineate the true PF of the new environment. The blue dots represent the lagging population inherited from the previous environment (the old solutions evaluated under the new environmental parameters). The red star marks the evaluation result of the unperturbed latent centroid ($\bar{\mathbf{z}}_{joint}$), while the orange dots denote the final initial population generated after the surrogate-assisted screening.

As shown in Fig.~\ref{fig:df1_tracking}(a), upon an environmental switch, the inherited population (blue dots) exhibits noticeable displacement, often remaining in previous local optima or positioned away from the new PF due to distribution shifts, which quantitatively corresponds to a high initial IGD of 0.1550. Conversely, the proposed model captures the macroscopic evolutionary trend. As observed in the figure, the reconstructed latent centroid (red star) lands in proximity to the center of the true PF, indicating that the model has captured the physical primitives of the new environment. Guided by this structural anchor, the generated candidate population (orange dots) avoids the localized regions covered by the blue dots and is distributed along the true PF (grey dots), substantially reducing the IGD to 0.0488. This suggests that continuous latent space modeling facilitates zero-shot generation: by sampling around the centroid within the learned manifold, DB-GEN provides topologically consistent initial seeds. To further validate this centroid-based design choice, Fig.~\ref{fig:df1_tracking}(b) illustrates an alternative strategy where perturbations are instead applied to the latent variables of all individual solutions from the previous generation. As shown, this all-point perturbation strategy yields a significantly less accurate initial population, only managing to decrease the IGD from 0.1550 to 0.1106. Because the old population is already displaced relative to the new environment, applying noise to their individual latent representations directly propagates this positional error. The resulting candidates tend to scatter around sub-optimal regions and heavily inherit the historical inertia of the old distribution.

\subsection{Data Scalability and Generalization Analysis}
\label{sec:data_scale}
To evaluate the scalability of the proposed framework, we first investigate the impact of the historical dataset size on optimization performance. The expansion of the data scale is achieved by increasing the number of historical problems and performing multiple instantiations under various environmental severity parameters ($n_t$), as detailed in Appendix A-A. Fig.~\ref{fig:data_scale} visualizes the mean IGD values and the average rankings obtained when the number of historical solutions provided for training varies from $4 \times 10^7$ to $1.2 \times 10^8$. 

The statistical significance of the data scale expansion is first evaluated via a non-parametric Friedman test across the matched test instances. The results ($p = 2.48 \times 10^{-18}$) provide strong evidence that the volume of historical data fundamentally impacts the optimization performance, exhibiting a clear monotonic improvement trend in both metrics as the data volume increases. To further investigate the local gains, Wilcoxon signed-rank tests were performed for pairwise comparisons. Crucially, the maximal scale ($1.2 \times 10^8$) significantly outperforms all other configurations ($p < 0.05$ in all cases), including its nearest neighbor ($1.1 \times 10^8$, $p = 0.020$). While the marginal performance gain temporarily plateaus between the scales of $8 \times 10^7$ and $1 \times 10^8$ ($p = 0.966$), suggesting a localized saturation point in manifold coverage for the current benchmark complexity, the overall trend strongly reinforces the scalability of our framework. This indicates that increasing the diversity and density of historical priors allows the model to capture a broader range of dynamic patterns, constructing a more robust and accurate latent manifold for solution generation. Detailed pairwise comparison results and precise $p$-values are provided in Appendix A-F.

\begin{figure}[htbp]
    \centering
    \includegraphics[width=0.95\linewidth]{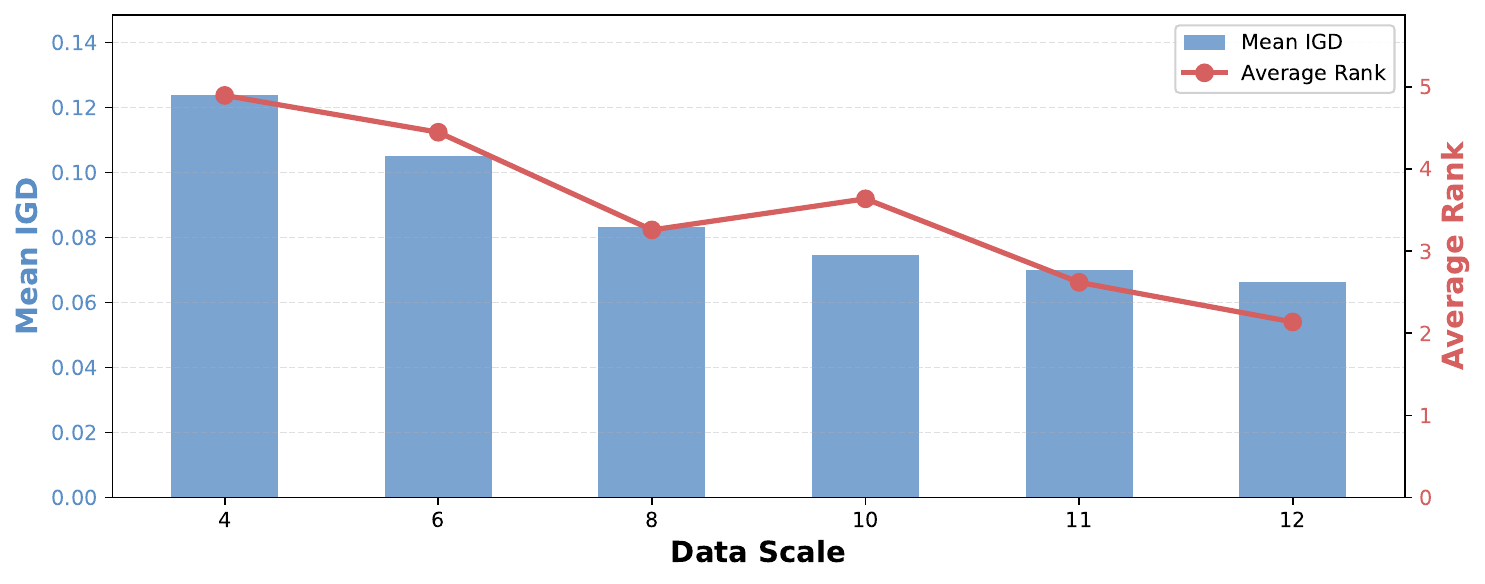}
    \caption{Effect of historical data scale on the mean IGD and the average rank.}
    \label{fig:data_scale}
\end{figure}

\begin{figure}[htbp]
    \centering
    \includegraphics[width=0.95\linewidth]{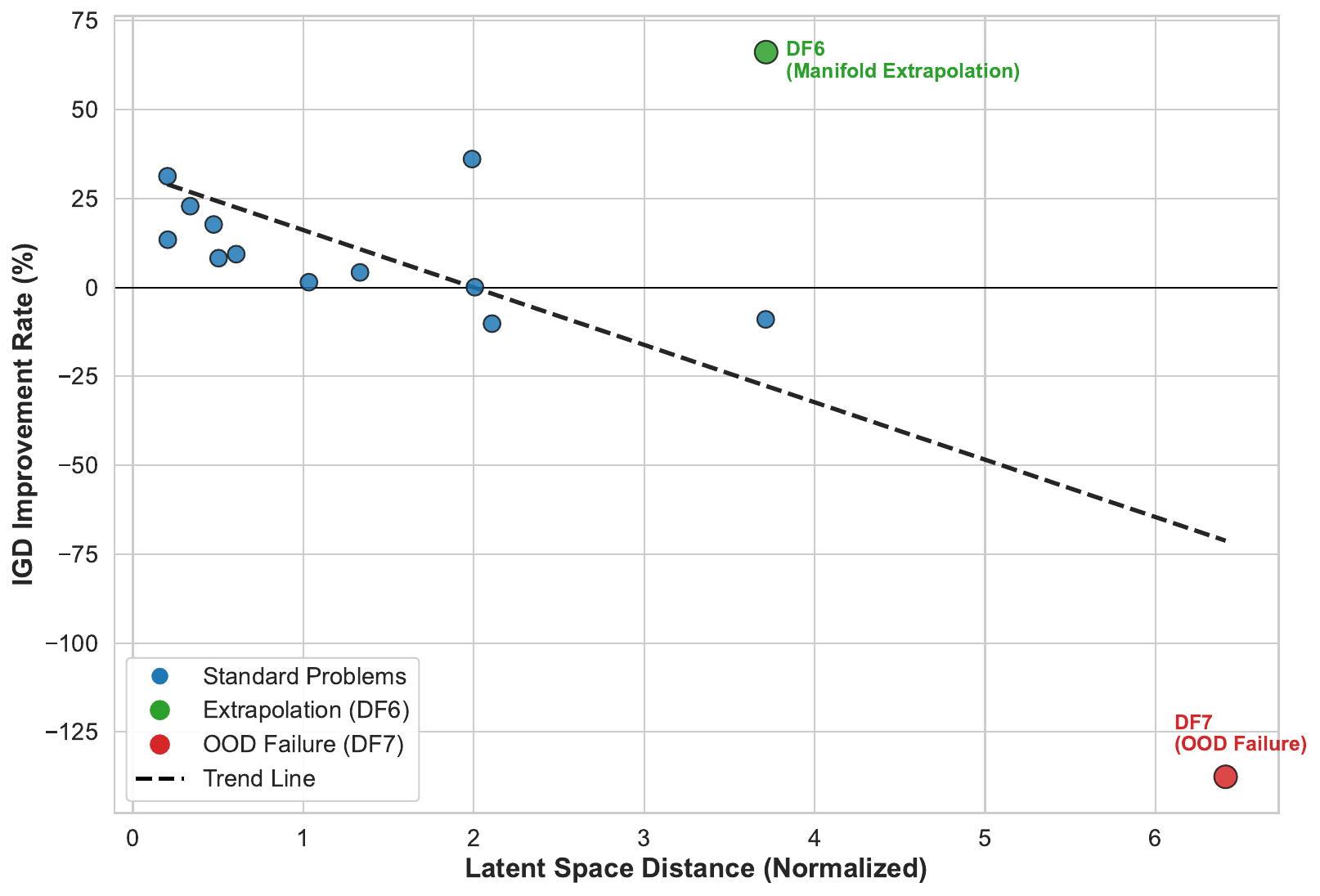}
    \caption{Correlation between the normalized latent space distance of new environments and the average IGD improvement rate.}
    \label{fig:correlation_analysis}
\end{figure}

To further investigate the generalization limits and understand the underlying mechanism of this scalability, we analyze the relationship between distributional shift and performance gain, utilizing the model trained on $8 \times 10^7$ historical solutions. The shift is measured by the \textit{normalized latent space distance} from the new environment to the nearest historical prior, while the gain is quantified by the IGD improvement rate of the generated population over the unperturbed inherited population. As shown in Fig.~\ref{fig:correlation_analysis}, a strong negative Pearson correlation ($r = -0.64, p < 0.001$) is observed. This correlation suggests that the latent distance serves as a confidence indicator: the performance improvement generally decays as the target domain deviates further from the historical data distribution. 

However, specific test instances reveal capabilities beyond proximity-based generation. DF6 presents an exception, achieving a high IGD improvement ($\approx 66.19\%$) despite a large latent distance. This indicates that DB-GEN is capable of \textit{manifold extrapolation}. Rather than merely memorizing historical instances, the model adapts to distant novel environments by combinatorially generalizing its learned basis vectors, provided the new dynamics remain within the subspace spanned by the dictionary (detailed in Section~\ref{sec:anly_basis} and Appendix A-D). Conversely, the performance on DF7 illustrates the boundary of the proposed framework. Characterized by out-of-distribution (OOD) topological dynamics, DF7 exhibits both a large latent distance and a negative improvement rate. This limitation occurs because the dynamics of DF7 introduce variations orthogonal to the manifold spanned by the historical basis vectors. It confirms that while DB-GEN performs interpolation and combinatorial extrapolation, generating solutions for environments that require entirely unobserved physical primitives remains a limitation.

\subsection{Ablation Study}
To verify the independent contributions and necessity of the core components proposed in our framework, we conducted a comprehensive ablation study across 57 environmental change instances. All variants are trained on $8 \times 10^7$ historical solutions. As previously demonstrated in Section \ref{sec:data_scale}, this intermediate scale reaches a localized performance saturation, serving as a highly representative and computationally viable baseline to independently evaluate each component. We constructed five variants by removing or replacing specific modules from the full model, as detailed in Table \ref{tab:ablation_results}.

\begin{table}[htbp]
\centering
\caption{Ablation study results on the core mechanisms.}
\label{tab:ablation_results}
\resizebox{\columnwidth}{!}{
\begin{tabular}{l|c|c|c}
\toprule
\textbf{Variant} & \textbf{MIGD} & \textbf{Degradation ($\Delta$)} & \bm{$+ / - / \approx$} \\
\midrule
\textbf{Full Model} & \textbf{0.0832} & - & - \\
\midrule
w/o High-Low Freq  & 0.0983 & +18.1\% & 4 / 53 / 0 \\
w/o Basis Learning  & 0.1205 & +44.8\% & 14 / 43 / 0 \\
w/o Solution VAE    & 0.1028 & +23.6\% & 23 / 34 / 0 \\
w/o Triplet Loss    & 0.1048 & +26.0\% & 15 / 42 / 0 \\
w/o Classify Loss   & 0.0934 & +12.3\% & 17 / 40 / 0 \\
\bottomrule
\end{tabular}
}
\end{table}

Table \ref{tab:ablation_results} confirms each module's positive contribution. The $+/-/\approx$ indicates instances where the variant performs significantly better, worse, or similarly to the Full Model. We first observe that removing the High-Low Frequency Decoupling module (w/o High-Low Freq.) results in an average IGD increase from 0.0832 to 0.0983. The full model outperforms this variant in 53 out of 57 test instances ($4/53/0$). This statistical difference confirms that treating all environmental changes uniformly hinders predictive ability. By explicitly decoupling high-frequency local jitter from low-frequency global trends, this mechanism reduces the complexity of representing environmental changes, thereby facilitating the learning of underlying evolutionary dynamics.

The most significant performance degradation occurs when the Basis Vector Learning module is removed (w/o Basis Learning). Without it, the mean IGD increases sharply to 0.1205 (+44.8\%), and the variant is defeated by the full model in 43 instances. In this baseline, the interpretable sparse combination of physical primitives is replaced by a black-box MLP mapping. The resulting poor tracking performance validates our hypothesis: simply mapping observations to a latent space frequently induces negative transfer. The proposed basis vector mechanism acts as a robust physical prior, enabling the model to achieve reliable zero-shot generalization in unseen environments.

Furthermore, we investigated the impact of the generative manifold and auxiliary loss functions. Removing the Solution VAE (w/o Solution VAE) increases the IGD to 0.1028 and incurs 34 losses against the full model, confirming the necessity of performing generative search within a well-structured latent manifold rather than directly in the complex decision space. Omitting the Triplet Loss (w/o Triplet Loss) leads to a similar performance decline (IGD: 0.1048, $15/42/0$), indicating that contrastive learning is vital for aligning the latent topology with the true problem geometry. Discarding the Classify Loss (w/o Classify Loss) also negatively impacts performance ($17/40/0$). These results robustly suggest that both topological constraints and auxiliary tasks are essential for guiding the model to learn meaningful and discriminative representations.

In summary, while the Basis Vector Learning module serves as the most critical backbone for robust generalization, the Frequency Decoupling, Solution VAE, and Auxiliary Tasks further refine the prediction accuracy and topological stability. The full model achieves optimal performance across the vast majority of environments, demonstrating that these architectural components are mutually complementary and indispensable.
\subsection{Analysis of Basis Vectors}
\label{sec:anly_basis}
\subsubsection{Visualization of Latent Representation}
\label{sec:tsne_visualization}
\begin{figure}[htbp]
    \centering
    \includegraphics[width=0.98\linewidth]{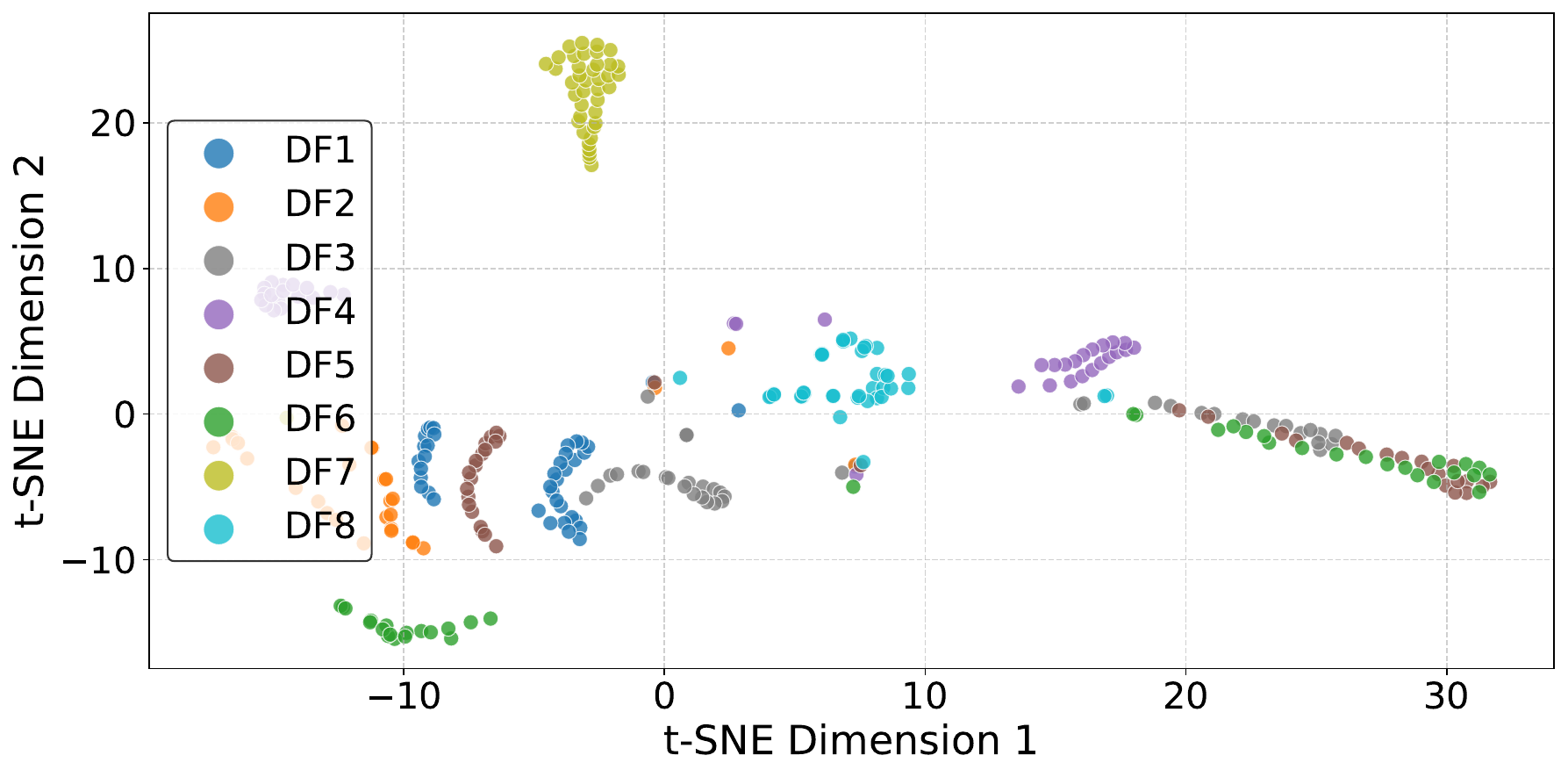}
    \caption{The t-SNE visualization of the environment-level structural latent representations ($\mathbf{z}_{struct}$) generated by the pre-trained model on unseen test problems.}
    \label{fig:tsne}
\end{figure}

To investigate the environmental representations learned by our framework, we project the high-dimensional structural latent representations (denoted as $\mathbf{z}_{struct}$) into a 2D space using t-SNE. Specifically, the macroscopic latent coordinate of each environment is obtained by feeding a large volume of sampled data from that environment into the model, generating their corresponding $\mathbf{z}_{struct}$ codes, and subsequently averaging them. For this visualization, the environmental severity parameter is specifically set to $n_t = 20$. This high-resolution configuration is deliberately chosen to meticulously capture the fine-grained transition dynamics and to reveal the exact correspondence between environmental changes and latent coordinates, particularly for problems with inherent periodic variations. The data points in Fig.~\ref{fig:tsne} represent these aggregated latent locations for \textit{unseen} benchmark test problems during the online inference phase. Each point corresponds to a specific environmental state, with different colors denoting distinct test problems. 

Observing the intra-problem dynamics, the latent representations of consecutive environments form continuous, smooth trajectories rather than randomly scattered clusters. This indicates that the model successfully tracks the temporal continuity of environmental changes. Furthermore, several trajectories exhibit U-shaped folding and cyclic patterns. Such structures objectively reflect the periodic dynamics, such as sinusoidal parameter variations, inherent in the definitions of the benchmark suite. A striking spatial bifurcation can be explicitly observed in the trajectory of the DF6 problem, whose environmental states are mapped into two distant, isolated sub-regions. This separation physically corresponds to the distinct evolutionary phases of the problem. During the expansion phase, where the underlying parameters drive the problem structure to deform continuously in a specific direction, the latent coordinates cluster tightly in one region. Conversely, during the contraction phase, which encapsulates reverse structural transformations and opposite spatial offsets, the representations are mapped to the other distant region. This proves the model's capability to effectively discern distinct and opposing geometric deformations occurring within the same problem.

Extending the observation to inter-problem relationships, the trajectories of most test problems occupy distinct sub-regions in the latent space. This global separability suggests that the learned basis vectors and their combinations possess sufficient expressiveness to differentiate various modes of macroscopic topological mutations across unobserved environments. However, this spatial distribution is driven by geometric similarity rather than strict mathematical formulations. A notable example is the distinct intertwining phenomenon observed in the lower-right region of Fig.~\ref{fig:tsne}, specifically between the trajectories of the DF5 and DF6 problems. Although DF5 relies on a static linear base and DF6 features dynamic curvature scaling, the dynamic curvature of DF6 periodically flattens during its evolutionary cycle, causing its macroscopic geometry to intersect directly with the linear base of DF5. Because the topology-aware triplet loss prioritizes global geometric homology over minor differences in their localized sinusoidal ripples, it groups these macroscopically similar states into adjacent sub-manifolds. A detailed mathematical derivation of this topological intersection is provided in Appendix A-D.

Overall, the t-SNE visualization confirms that the organization of the latent space is primarily governed by topological semantics. This geometric structuring provides a verifiable context for understanding how the underlying basis vectors combine to construct diverse environmental manifolds.

\subsubsection{Semantic Correlation Analysis}
To further demonstrate the interpretability of the proposed method and verify whether the learned dictionary vectors capture meaningful physical primitives, we conduct an in-depth case study on the DF6 problem. As previously observed in the generalization analysis (Fig.~\ref{fig:correlation_analysis}), DF6 exhibits a relatively large distance from historical priors in the latent space, yet our model still achieves highly competitive optimization performance on it. 

\begin{figure}[htbp]
    \centering
    \includegraphics[width=0.7\linewidth]{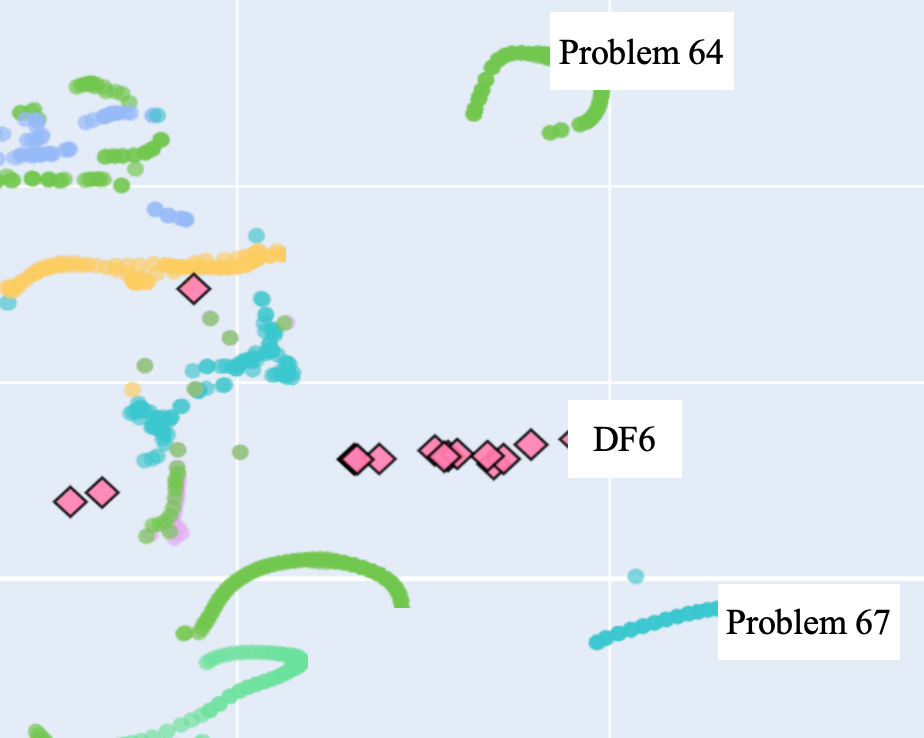} 
    \caption{The t-SNE visualization of DF6 and Problem 64/67. }
    \label{fig:latent_train_test}
\end{figure}

To understand the mechanism behind this successful out-of-distribution generalization, we visualize both the test and training environments in a unified latent space, as shown in Fig.~\ref{fig:latent_train_test}. Interestingly, although DF6 is isolated from the majority of historical problems, its trajectory is positioned exactly between two specific training instances: Problem 64 and Problem 67. Further investigation into their mathematical definitions reveals a distinct physical disentanglement. DF6 is a complex dynamic environment characterized by simultaneous non-linear curvature changes in the Pareto Front (PF) and positional shifts in the Pareto Set (PS). Conversely, Problem 64 exclusively shares a similar PF curvature variation pattern with DF6 but maintains a static PS; meanwhile, Problem 67 exclusively shares a similar PS positional shift pattern with DF6 but maintains a static PF geometry. The detailed mathematical formulations and dynamic similarities of these three problems are provided in Appendix A-D. 

This geometric interpolation in the latent space suggests that the model effectively reconstructs the unseen DF6 by recombining the decoupled physical properties (curvature and shift) learned from Problem 64 and Problem 67. By recording the activation coefficients of the learned basis vectors across continuous environments, we computed the Pearson correlation $r$ between the activation sequences of DF6 and those of the auxiliary problems. This quantitative analysis demonstrates a clear functional disentanglement among the learned latent vectors. Specifically, curvature vectors, such as dimensions 9 and 20, act as shape neurons that encode topological deformations like convexity or concavity while ignoring spatial translations. For instance, dimension 9 exhibits a high activation correlation with the curvature-varying Problem 64 ($r=0.955$) and remains uncorrelated with the shift-varying Problem 67 ($r=-0.080$). Conversely, shift vectors function as displacement neurons controlling the macroscopic movement of solutions across the decision space. Dimension 11, for example, shows a strong correlation with the positional shift dynamics of Problem 67 ($r=0.832$) but is independent of curvature changes ($r=-0.034$). Finally, global intensity vectors, such as dimension 26, maintain high correlations of $r \approx 0.902$ with both auxiliary problems, likely encoding the global temporal factor or the overall intensity of the environmental transition.

\subsubsection{Functional Verification via Masking}
\begin{figure}[t]
    \centering
    \includegraphics[width=0.75\linewidth]{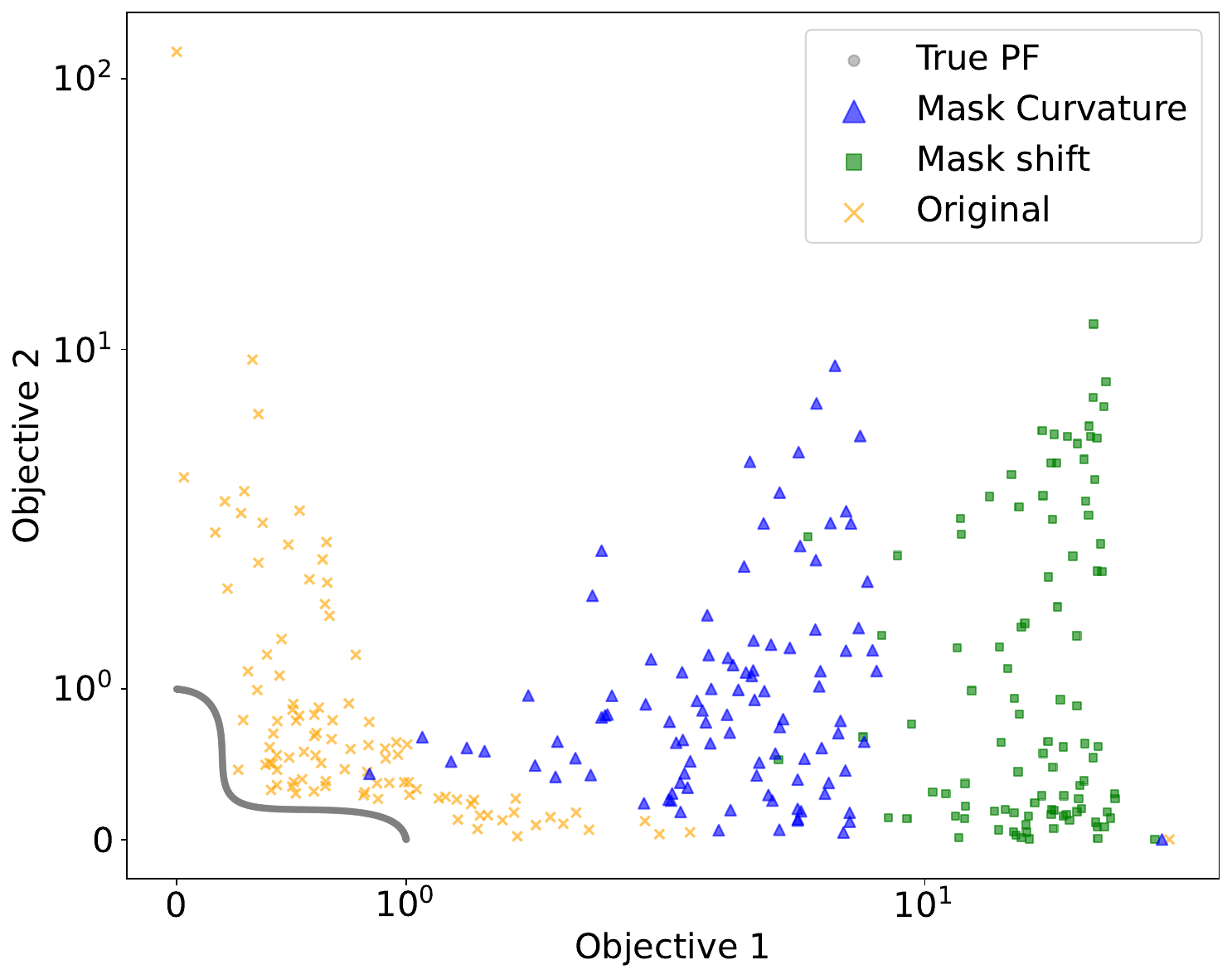}
    \caption{Visualization of the generated solutions under different masking strategies across environments on the DF6 problem.}
    \label{fig:masking_exp}
\end{figure}

Building upon the statistical correlations discovered above, we conduct a masking experiment to explicitly verify the functional roles of these disentangled basis vectors. By manually zeroing specific dimensions of the activation coefficients during inference, we observe the resulting geometric impacts on the generated solutions. Fig.~\ref{fig:masking_exp} visualizes the objective space distribution of the generated solutions under different masking strategies on the DF6 problem.

The full model accurately tracks the true PF, maintaining both correct global positioning and non-linear curvature. When the curvature vectors (9, 20) are masked, the generated solutions remain in the general vicinity of the true PF but exhibit distinct geometric distortion, failing to reconstruct the correct topological shape. This observation confirms their specific function in handling shape deformations. Conversely, masking the shift vectors (11, 16) displaces the generated solutions far from the true PF and causes a complete collapse of their geometric shapes. In highly non-linear problems, local manifold curvature is tightly coupled with global spatial coordinates. Losing the positional anchor leads to severe non-linear distortion during the decoding mapping. This validates that the shift vectors not only govern precise spatial localization but also provide the necessary structural foundation for stable manifold reconstruction.

\subsection{Parameter Sensitivity Analysis}
To investigate the robustness of the proposed DB-GEN framework and understand its key hyperparameters, we conducted sensitivity analyses on the number of basis vectors $K$, the Gaussian perturbation radius $\sigma$, and the candidate pool size $N_{cand}$. The results, evaluated using the mean IGD across dynamic environments, are summarized in Table~\ref{tab:parameter_sensitivity}.

\begin{table}[htbp]
\centering
\caption{Parameter sensitivity analysis result.}
\label{tab:parameter_sensitivity}
\resizebox{\columnwidth}{!}{
\begin{tabular}{l|cccccc}
\toprule
\textbf{Parameter} & \multicolumn{6}{c}{\textbf{Tested Values and Mean IGD}} \\
\midrule
\multirow{2}{*}{Basis Vectors ($K$)} 
& 32 & 64 & 128 & \textbf{256} & 384 & 512 \\
& 0.1048 & 0.0975 & 0.1046 & \textbf{0.0832} & 0.0854 & 0.0872 \\
\midrule
\multirow{2}{*}{Perturbation Radius ($\sigma$)} 
& 0.05 & 0.10 & 0.15 & \textbf{0.20} & 0.25 & 0.30 \\
& 0.1040 & 0.0923 & 0.0970 & \textbf{0.0832} & 0.0886 & 0.0871 \\
\midrule
\multirow{2}{*}{Candidate Pool ($N_{cand}$)} 
& 50,000 & 100,000 & 150,000 & \textbf{200,000} & 250,000 & 300,000 \\
& 0.1140 & 0.0910 & 0.0920 & \textbf{0.0832} & 0.0870 & 0.0840 \\
\bottomrule
\end{tabular}
}
\end{table}

The parameter $K$ determines the capacity of the learned dictionary. As observed in Table~\ref{tab:parameter_sensitivity}, performance exhibits a U-shaped trend, achieving the optimal mean IGD at $K=256$. A smaller dictionary causes underfitting, as the limited basis vectors are insufficient to express complex environmental dynamics. Conversely, an excessively large dictionary introduces semantic redundancy and risks overfitting to environmental noise, hindering the sparse reconstruction process.

The perturbation radius $\sigma$ directly controls the exploration-exploitation balance within the latent manifold. The best performance is recorded at $\sigma=0.20$. A smaller radius leads to over-exploitation around the latent centroid, degrading the diversity of the generated population. On the other hand, an excessively large radius pushes the sampling process into low-probability-density regions of the latent space, where the decoder mapping becomes highly non-linear and unreliable, ultimately yielding sub-optimal solutions with poor convergence.

Finally, the candidate pool size $N_{cand}$ dictates the sampling density for the surrogate-assisted screening. Performance improves significantly as the pool size increases up to 200,000, as a denser pool ensures that the Tchebycheff decomposition can precisely match high-quality candidates for diverse weight vectors. Beyond this threshold, the performance plateaus. This indicates that the latent space has been adequately covered, and further expanding the pool size yields diminishing returns while unnecessarily increasing the computational burden.

\subsection{Performance on Real-World Scenarios}
To evaluate the practical applicability of the proposed framework, we extend the experiments to two real-world dynamic formulations: DRA and DPP, whose detailed mathematical definitions and environmental settings are provided in Appendix B. Table~\ref{tab:real_world} summarizes the mean IGD results of all algorithms on these two problems.

The DRA problem models the trade-off between resource cost and system return under spatio-temporal environmental shifts. As shown in Table~\ref{tab:real_world}, DB-GEN achieves the remarkably lowest IGD of 1.35E-02. Compared to the second-best baseline, DIP-DMOEA (3.63E-02), DB-GEN yields a substantial quantitative improvement of approximately 62.8\%. In practical resource allocation, this significant reduction in IGD translates to a much more accurate approximation of the optimal dispatching frontier. This indicates that upon an environmental change, DB-GEN can rapidly restore the balance between cost and quality, mitigating the excessive transient resource waste typically caused by lagging population convergence.

The DPP problem models the physical trade-off between operation time ($f_1 \propto 1/v$) and energy consumption ($f_2 \propto v^2$). In this scenario, the vector auto-regression model VARE achieves the best performance, though DB-GEN follows closely with a marginal IGD difference. Because DPP dynamics are driven by smooth, auto-correlated periodic functions, explicit numerical extrapolation holds an inherent predictive advantage over continuous manifold sampling. Nevertheless, DB-GEN remains highly competitive and significantly outperforms other baselines. Practically, this tracking accuracy ensures the reliable generation of valid velocity profiles to minimize dynamic energy while satisfying strict temporal constraints.

\begin{table}[htbp]
\centering
\caption{MIGD results on real-world dynamic problems.}
\label{tab:real_world}
\resizebox{\columnwidth}{!}{
\begin{tabular}{l|ccccc}
\toprule
\textbf{Problem} & \textbf{Ours} & \textbf{STT-MOEA/D} & \textbf{DIP-DMOEA} & \textbf{SIKT-DMOEA} & \textbf{VARE} \\
\midrule
DRA & \textbf{1.35E-02} & 9.18E-02 & \underline{3.63E-02} & 1.22E-01 & 5.74E-02 \\
DPP & \underline{7.19E-02}$\approx$ & 1.58E-01 & 2.90E-01 & 1.77E-01 & \textbf{7.07E-02} \\
\bottomrule
\end{tabular}
}
\end{table}

\subsection{Run Time Analysis}
\begin{figure}[htbp]
    \centering
    \includegraphics[width=\columnwidth]{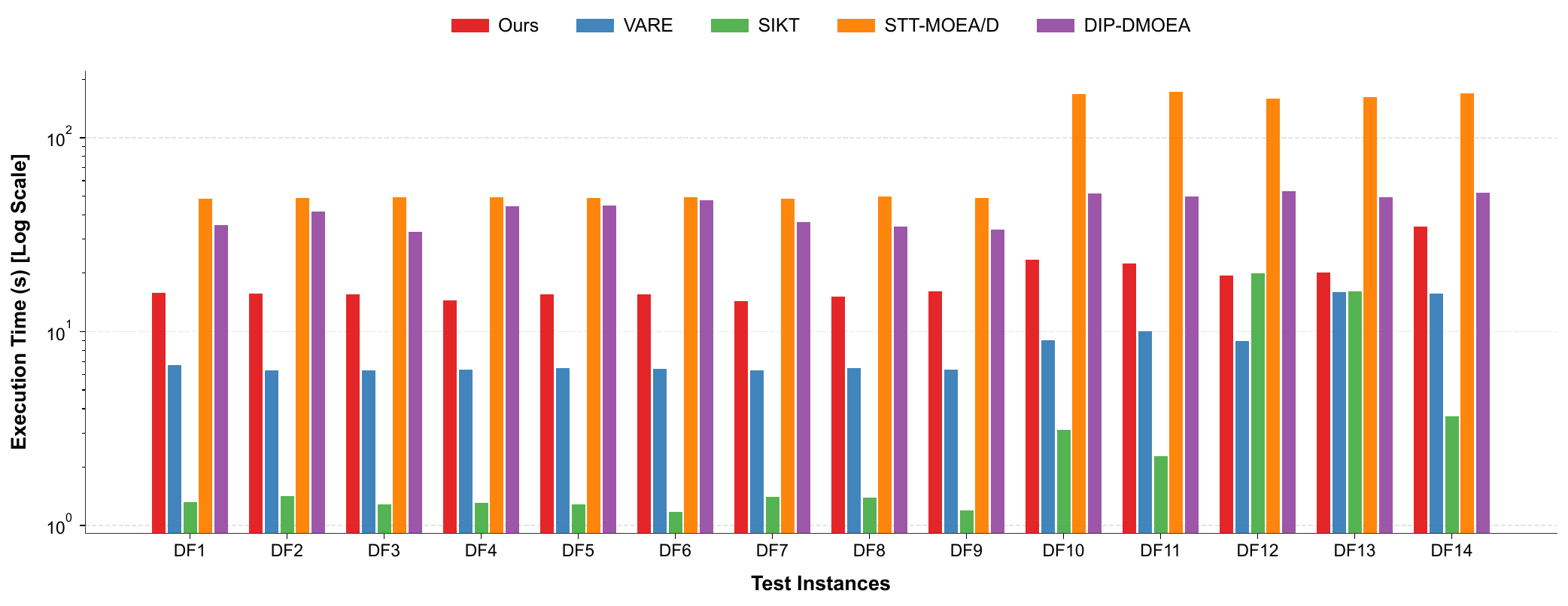}
    \caption{Comparison of execution time (in seconds) across the DF test suite when $\tau_t=10$ and $n_t = 10$.}
    \label{fig:run_time}
\end{figure}

To evaluate the computational efficiency of the proposed framework, we compare the execution time of DB-GEN with the baseline algorithms across the DF test suite. The quantitative execution times are visualized in Fig.~\ref{fig:run_time}.

Overall, DB-GEN exhibits a moderate computational cost, consistently completing the entire environmental tracking process within 15 to 35 seconds. Compared to lightweight prediction methods such as SIKT and VARE, DB-GEN requires slightly more execution time. This is primarily because SIKT relies on a simplified clustering-based mapping, and VARE employs a computationally inexpensive linear vector autoregression. However, this marginal increase in execution time is a worthwhile trade-off. The generative sampling in the continuous latent space and the surrogate-assisted screening in DB-GEN yield tracking accuracy and topological robustness, mitigating the performance limitations observed in SIKT and VARE under highly non-linear dynamics.

Furthermore, compared to advanced structural and learning-based prediction methods, DB-GEN demonstrates a clear efficiency advantage. Both STT-MOEA/D and DIP-DMOEA require higher computational costs, frequently exceeding 100 seconds per environment change. This is particularly evident on the 3-objective instances (i.e., DF10-DF14), where their execution times scale up noticeably. Specifically, STT-MOEA/D involves the intensive computational complexity of spatial-temporal topological tensor decomposition, while DIP-DMOEA requires online backpropagation training for its neural network upon every environmental change. In contrast, DB-GEN avoids online training by performing zero-shot generative inference directly within the pre-trained latent manifold. Consequently, DB-GEN achieves a favorable balance between execution efficiency and optimization performance.

\section{Conclusion}
In this paper, we proposed a decoupled basis-vector-driven generative framework (DB-GEN) for dynamic multi-objective optimization. Rather than relying on task-specific memory inheritance or simple autoregression, DB-GEN achieves robust cross-problem knowledge transfer through manifold generation and structural recombination, thereby mitigating negative transfer and overcoming the cold-start problem.

Specifically, the framework employs the discrete wavelet transform to decouple global trends from local noise. Subsequently, a sparse dictionary learning mechanism extracts universal topological basis vectors from a large-scale offline dataset, constructing a latent manifold instead of rigidly tracking current historical trajectories. Guided by these extracted primitives, DB-GEN performs cross-problem zero-shot generation in the continuous latent space for completely unseen environments. A surrogate-assisted Tchebycheff screening strategy is then utilized to sample a high-quality initial population from the generated candidates. Experimental results and ablation studies across various benchmarks indicate that DB-GEN improves tracking accuracy and convergence stability compared to existing baselines.

Despite its advantages, the current framework has certain limitations. The dictionary capacity and sampling radius are configured as static hyperparameters, which limits adaptability to varying dynamic severities. Additionally, the current evaluation primarily focuses on unconstrained environments with low-dimensional spaces. Therefore, future work will focus on adaptive mechanisms to dynamically adjust the manifold structure and sampling strategies. Furthermore, extending DB-GEN to handle high-dimensional and constrained dynamic multi-objective optimization problems represents an important direction for further investigation\cite{li2025causal}. For instance, integrating space discretization or dimensionality reduction techniques \cite{qian2025provable} into our continuous latent generation process could be a promising avenue to overcome the curse of dimensionality.


\bibliographystyle{IEEEtran}   
\bibliography{main}   

\end{document}


\title{APPENDIX}
\maketitle

\appendices
\section{Detail of the Experiment}

\subsection{Parameter Configurations and Implementation Details}
\label{app:parameters}

To ensure the reproducibility of the proposed framework, Table~\ref{tab:params} summarizes the complete parameter configurations utilized in our experiments. This includes the environment settings for the benchmark problems, the scale of the pre-training data, the architectural hyperparameters of the proposed DB-GEN framework, and the specific control parameters for the MOEA/D static optimizer. 

Furthermore, all algorithmic implementations, including the offline latent space pre-training and the online evolutionary tracking, were developed in Python using the PyTorch deep learning framework. The experiments were deployed and executed on a workstation running the Ubuntu 22.04.4 LTS operating system. The hardware environment is equipped with an AMD EPYC 7742 64-Core CPU, 314 GiB of RAM, and NVIDIA GeForce RTX 4090 (24GB) GPU to accelerate the tensor computations and neural network inferences.

\begin{table}[htbp]
\centering
\caption{Parameter Configurations and Implementation Environments}
\label{tab:params}
\resizebox{\columnwidth}{!}{
\begin{tabular}{llc}
\toprule
\textbf{Category} & \textbf{Parameter} & \textbf{Value} \\
\midrule
\multirow{5}{*}{DMOP Benchmarks} 
& Number of Objectives ($M$) & 2, 3 \\
& Decision Variables Dimension ($D$) & 10 \\
& Change Frequency \& Severity ($\tau_t, n_t$) & (10,10), (5,10), (10,5) \\
& Number of Environments & 30 \\
& Total Generations per Run & $30 \times \tau_t$ \\
\midrule
\multirow{6}{*}{DB-GEN Architecture} 
& Latent Space Dimension ($d_z$) & 64 \\
& Dictionary Size ($K$) & 256 \\
& Historical Window Size ($W$) & 3 \\
& Perturbation Radius ($\sigma_r$) & 0.20 \\
& Candidate Pool Size ($N_{cand}$) & 200,000 \\
& Loss Penalty ($\lambda_1 - \lambda_3$) & 0.5, 0.1, 0.01 \\
\midrule
\multirow{5}{*}{Static Optimizer (MOEA/D)} 
& Population Size ($N$) & 100 ($M=2$), 150 ($M=3$) \\
& Neighborhood Size ($T$) & 20 \\
& Neighborhood Selection Prob. ($p$) & 0.8 \\
& DE Scaling Factor ($F$) & 0.5 \\
& DE Crossover Rate ($CR$) & 0.5 \\
\midrule
\multirow{4}{*}{Implementation Details} 
& Programming Language & Python 3.10 \\
& Deep Learning Framework & PyTorch (CUDA 12.2) \\
& CPU \& RAM & AMD EPYC 7742 (64-Core), 314 GiB \\
& GPU Acceleration & NVIDIA RTX 4090 (24GB) \\
\bottomrule
\end{tabular}
}
\end{table}

During the training phase, to encompass various environmental change characteristics and evaluate the proposed method, we utilize a collection of standard test problem suites commonly applied in dynamic multi-objective optimization (DMOP). Specifically, the base training problems are selected from the following classic series: F and ZF \cite{zhou2013population}, HE \cite{helbig2011archive, helbig2014benchmarks}, JY \cite{jiang2016evolutionary}, UDF \cite{biswas2014evolutionary}, DIMP \cite{koo2010predictive}, GTA \cite{gee2016benchmark}, SDP \cite{jiang2019scalable}, SJY \cite{jiang2014framework}, DCP \cite{zhang2011artificial}, and DSW \cite{mehnen2000evolutionary}. Additionally, problems such as the T, DMZDT, and CF series, as well as the ZJZ test problems reviewed in the comprehensive survey \cite{liu2020survey}, are also included.

To construct the offline pre-training dataset, we execute a decomposition-based static optimizer (i.e., MOEA/D) across these historical base problems under diverse environmental change frequencies and severities ($\tau_t, n_t$). During the evolutionary process, periodic random initialization and mutation mechanisms are employed to inject diversity and prevent premature convergence. For each generation, the system records comprehensive evolutionary state features, including the decision variables, objective values, Tchebycheff scalarizing values, and the temporal context of historical objective fluctuations. This highly parallelized collection process yields a high-fidelity trajectory dataset encapsulating diverse dynamic patterns.

To ensure a rigorous evaluation of the zero-shot cross-problem generalization capabilities and to strictly prevent data leakage, we performed a decontamination of the pre-training dataset. Any historical problems sharing identical or highly isomorphic mathematical definitions with our target test suites (i.e., the FDA and DF series) were explicitly excluded. For instance, several instances in the dMOP and DMZDT series share highly similar baseline formulations and Pareto Front dynamic mappings with the DF test suite. To avoid the model merely memorizing these superficial similarities rather than learning transferable structural primitives, all such overlapping benchmarks were removed from the training pool.

\subsection{Detailed Visualization of Zero-Shot Generation on DF1}
\label{app:visualization}

To comprehensively illustrate the generative capability of the proposed DB-GEN framework and validate the design choice of the latent sampling strategy, this section provides the complete visualizations of the zero-shot generation process across all 30 consecutive environments on the DF1 problem.

\begin{figure*}[htbp] 
    \centering
    \includegraphics[width=0.19\linewidth]{figures/DF1_mean_search/Environment_1_visualization.pdf}\hfill
    \includegraphics[width=0.19\linewidth]{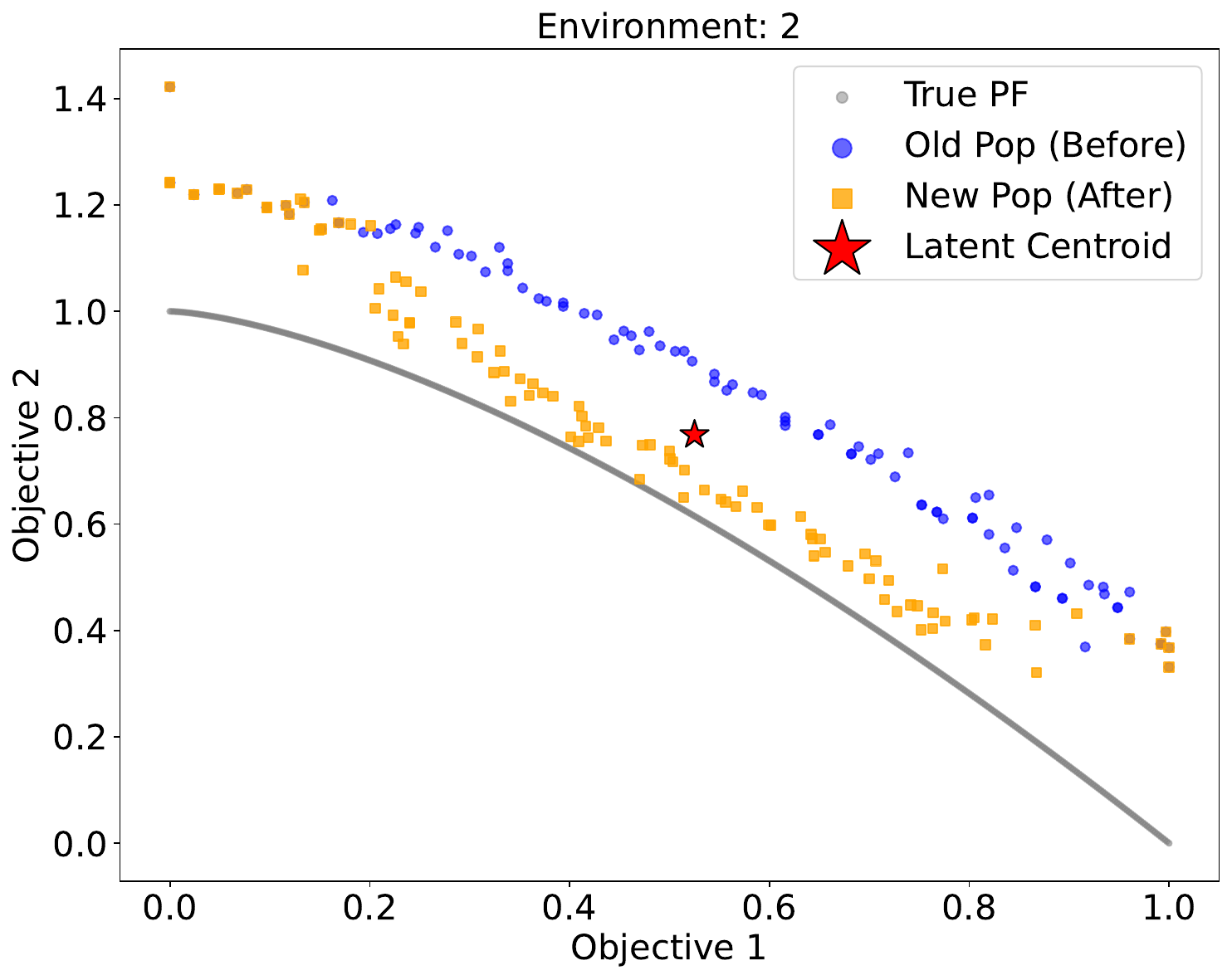}\hfill
    \includegraphics[width=0.19\linewidth]{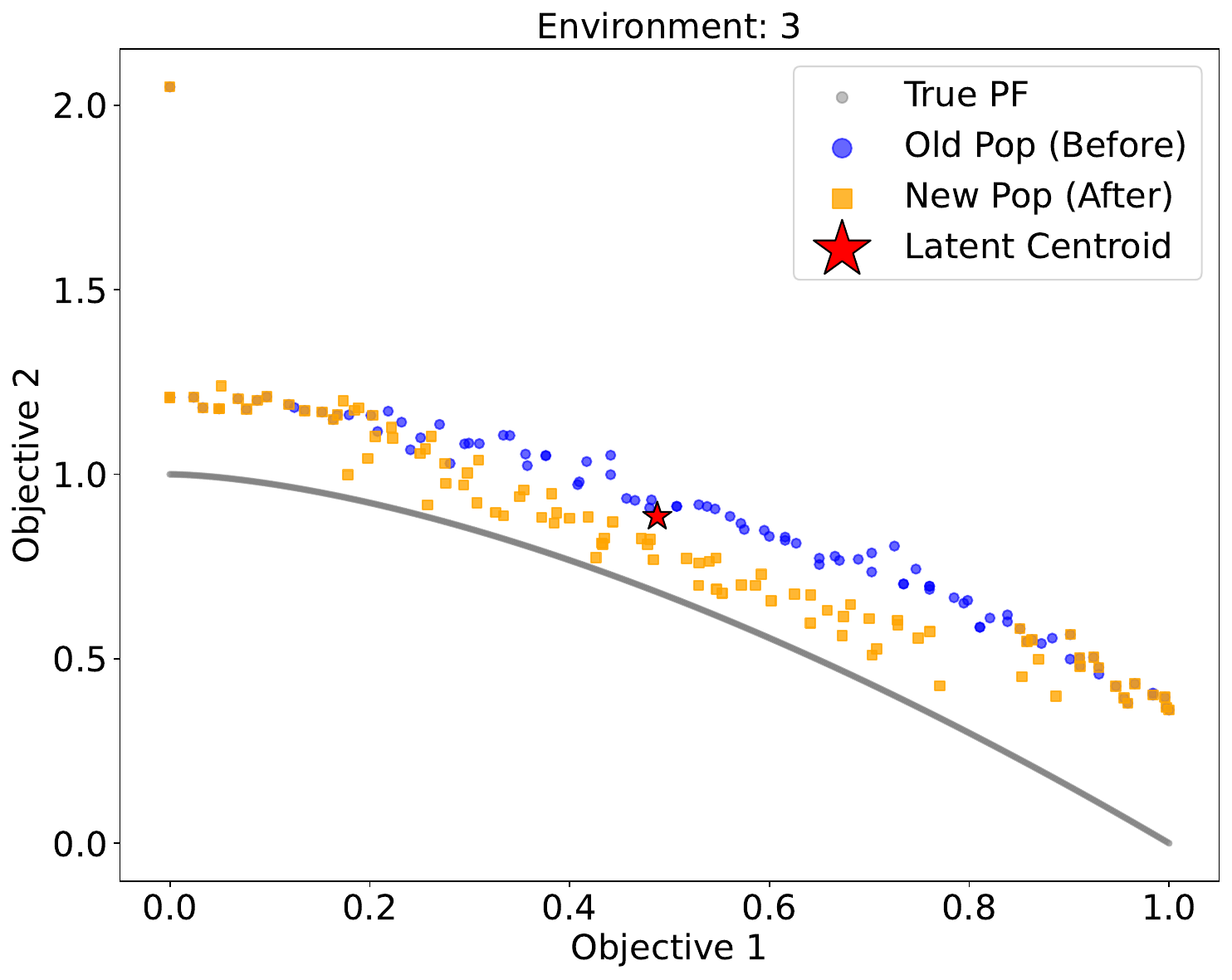}\hfill
    \includegraphics[width=0.19\linewidth]{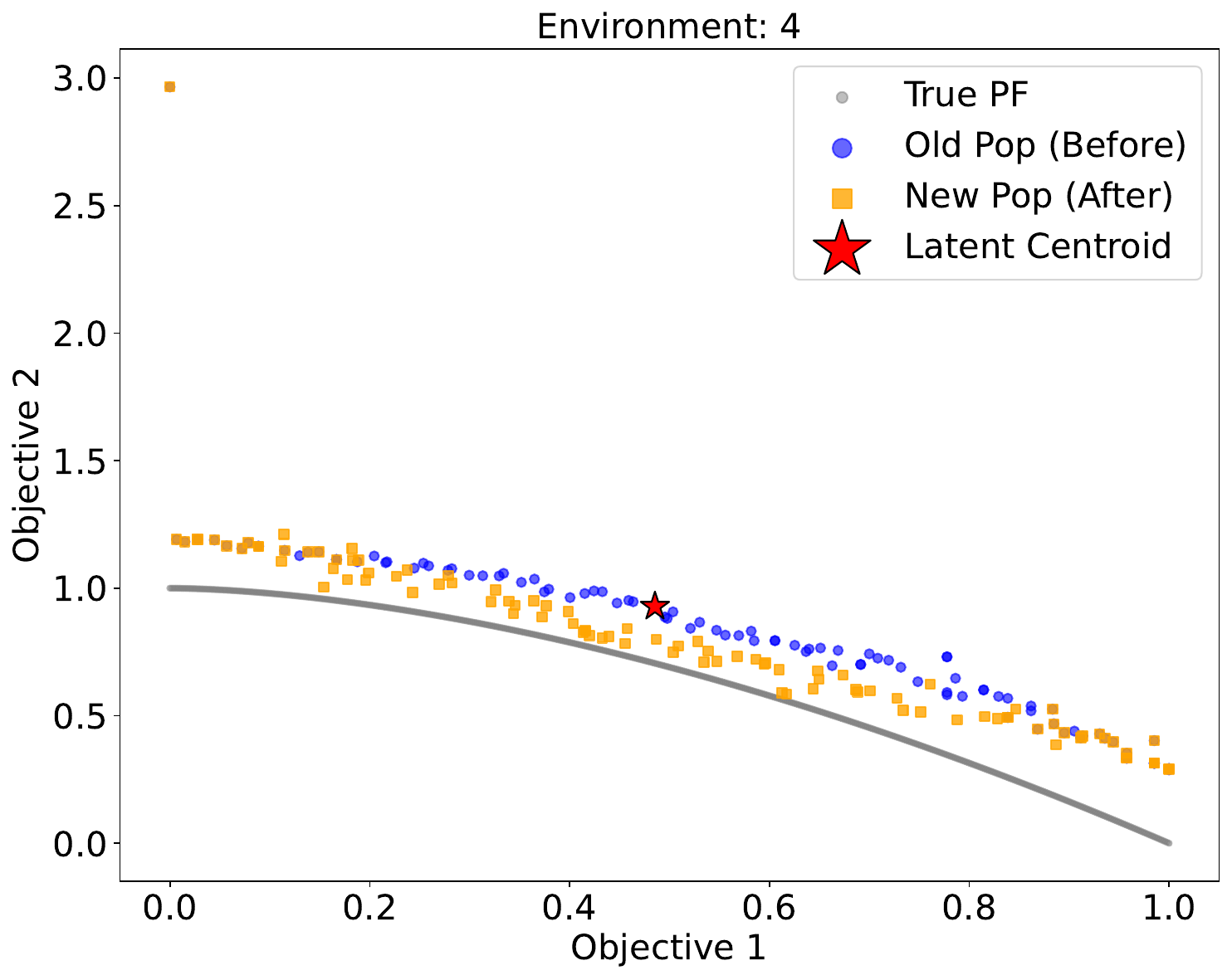}\hfill
    \includegraphics[width=0.19\linewidth]{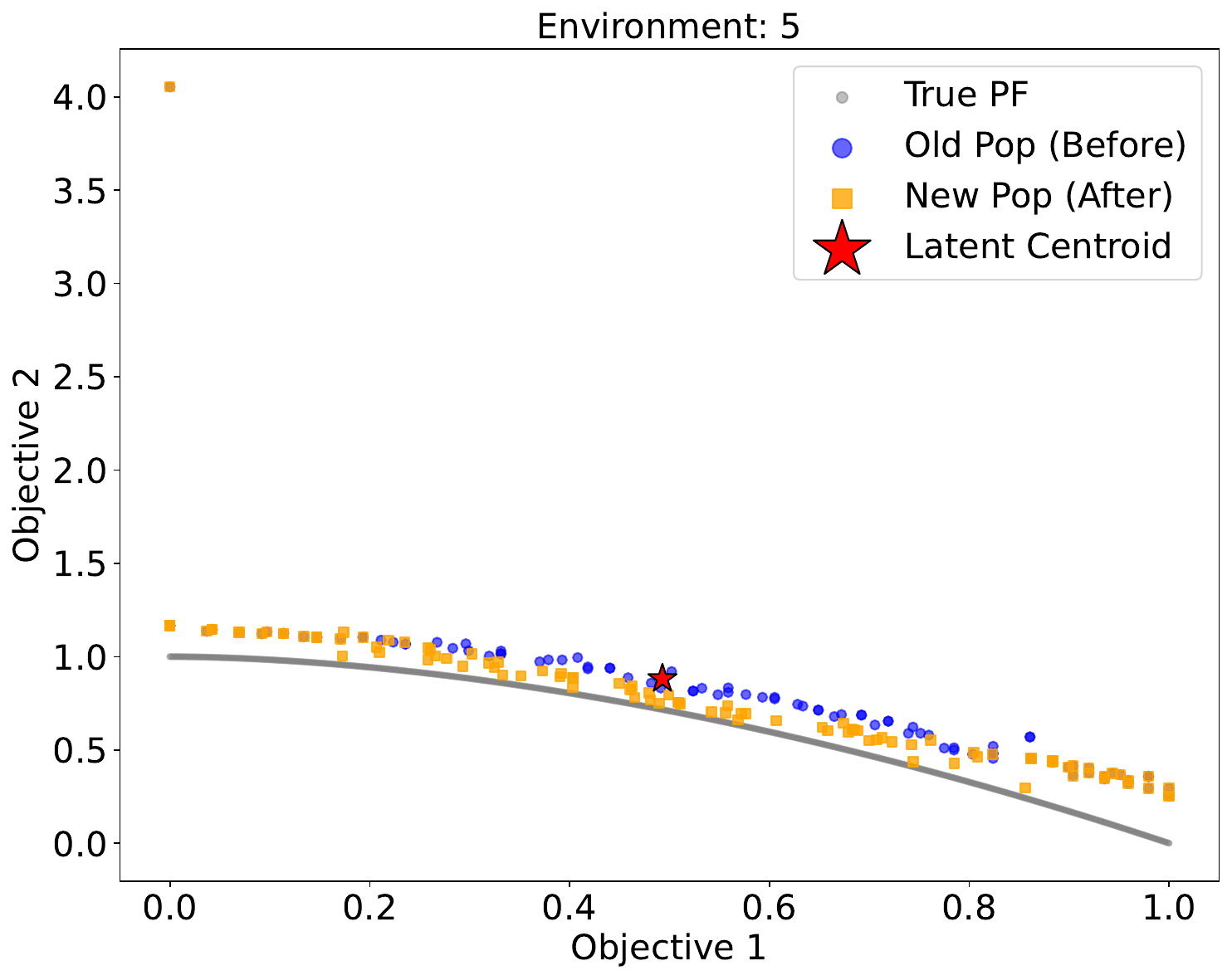}
    \\[4pt] 
    \includegraphics[width=0.19\linewidth]{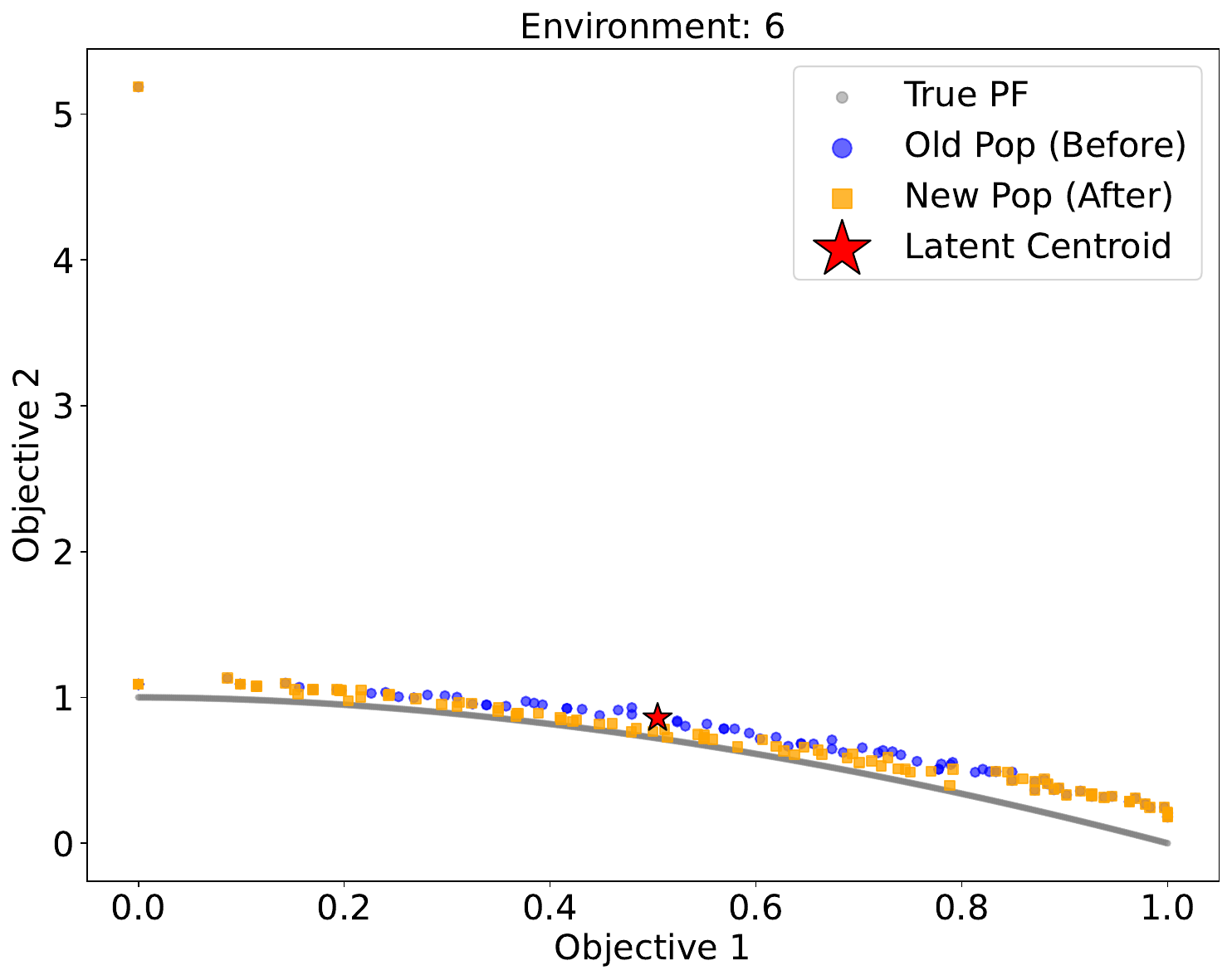}\hfill
    \includegraphics[width=0.19\linewidth]{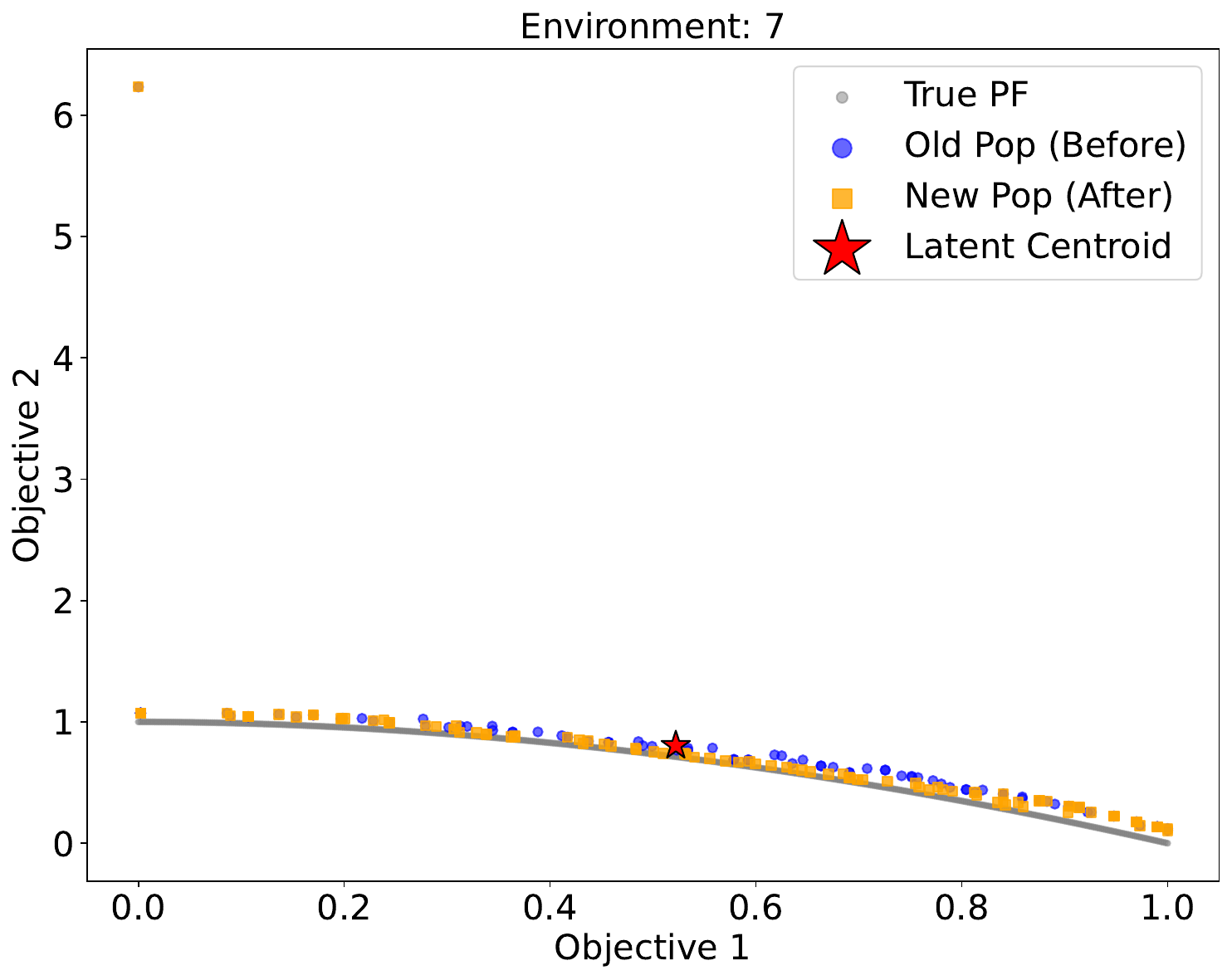}\hfill
    \includegraphics[width=0.19\linewidth]{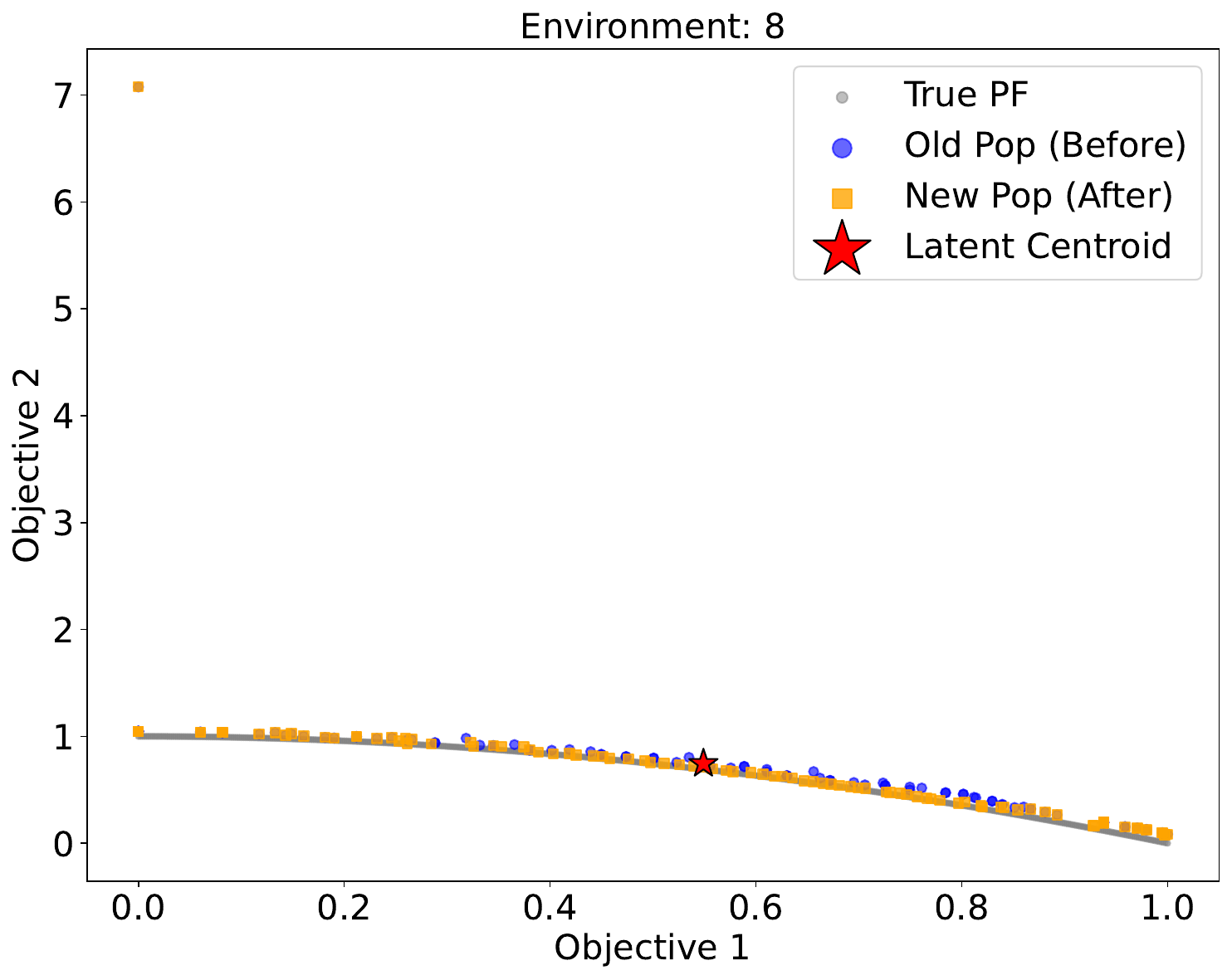}\hfill
    \includegraphics[width=0.19\linewidth]{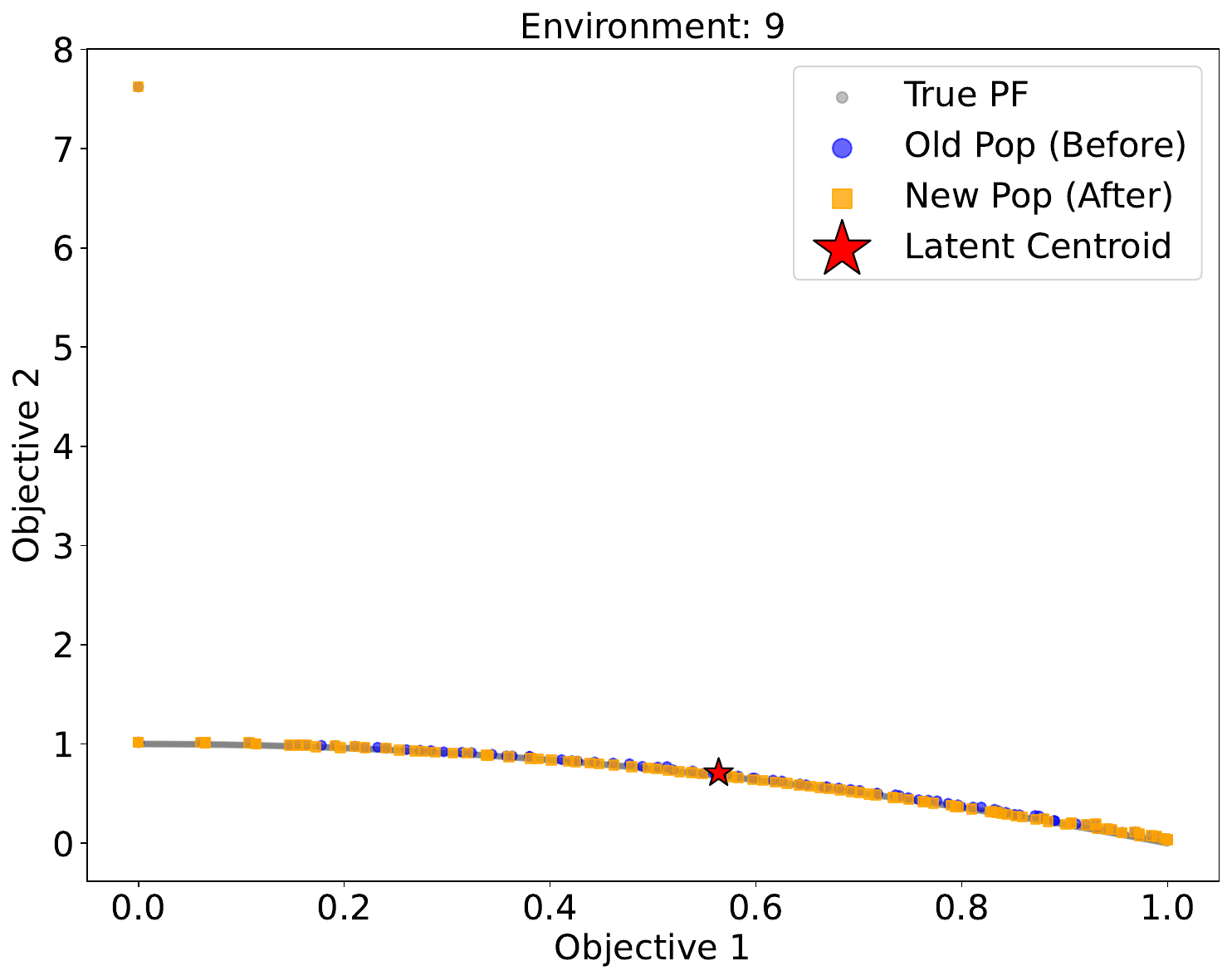}\hfill
    \includegraphics[width=0.19\linewidth]{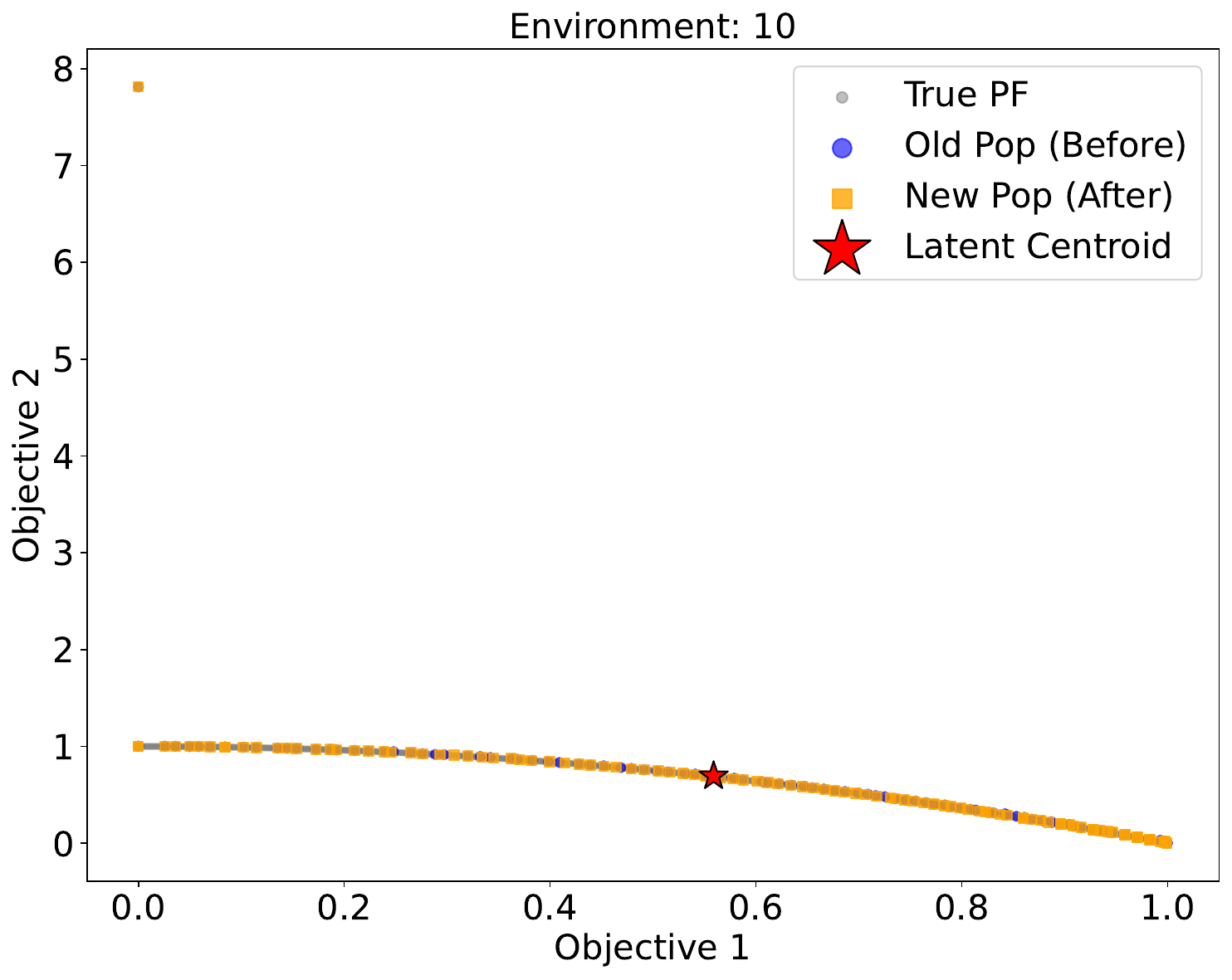}
    \\[4pt]
    \includegraphics[width=0.19\linewidth]{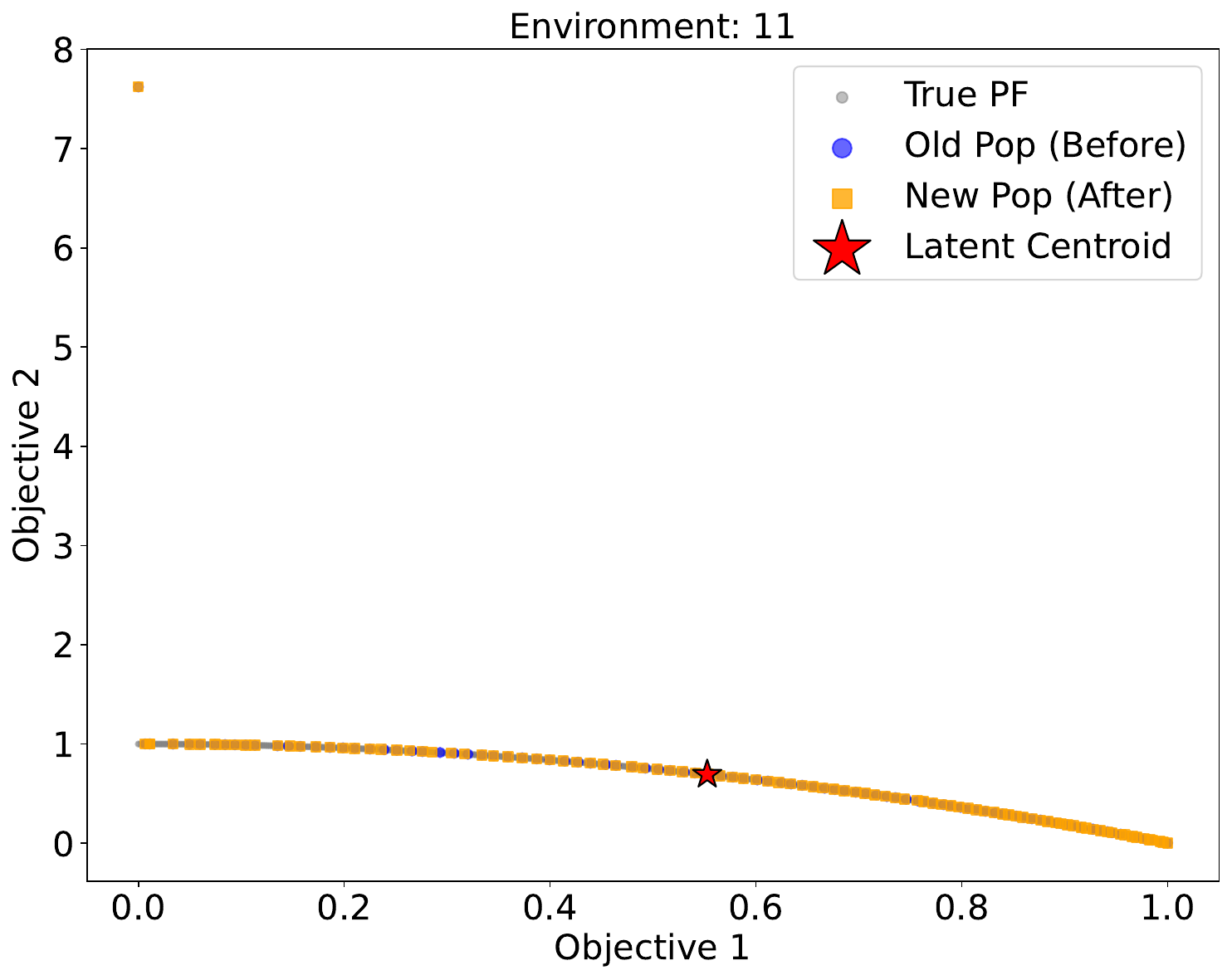}\hfill
    \includegraphics[width=0.19\linewidth]{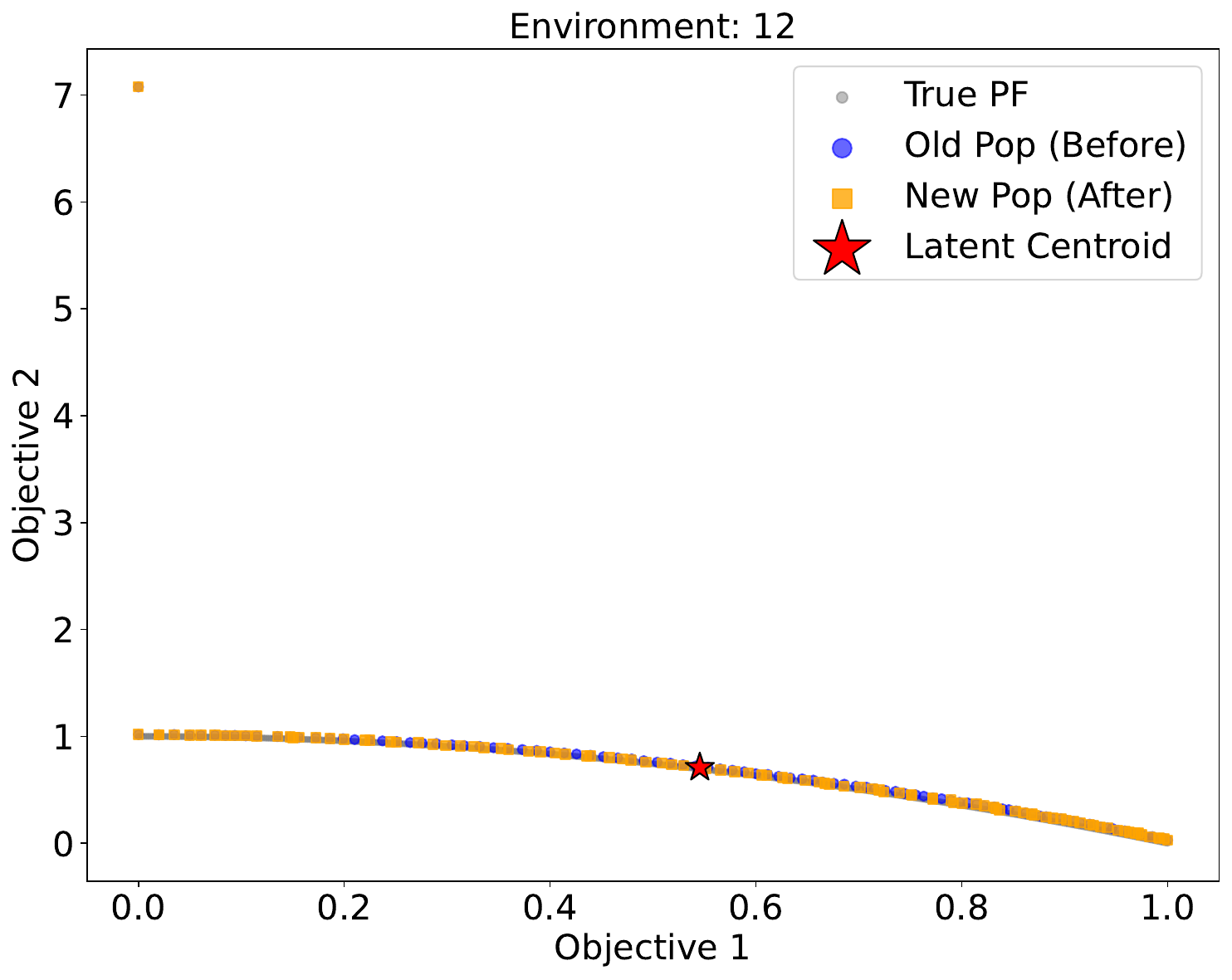}\hfill
    \includegraphics[width=0.19\linewidth]{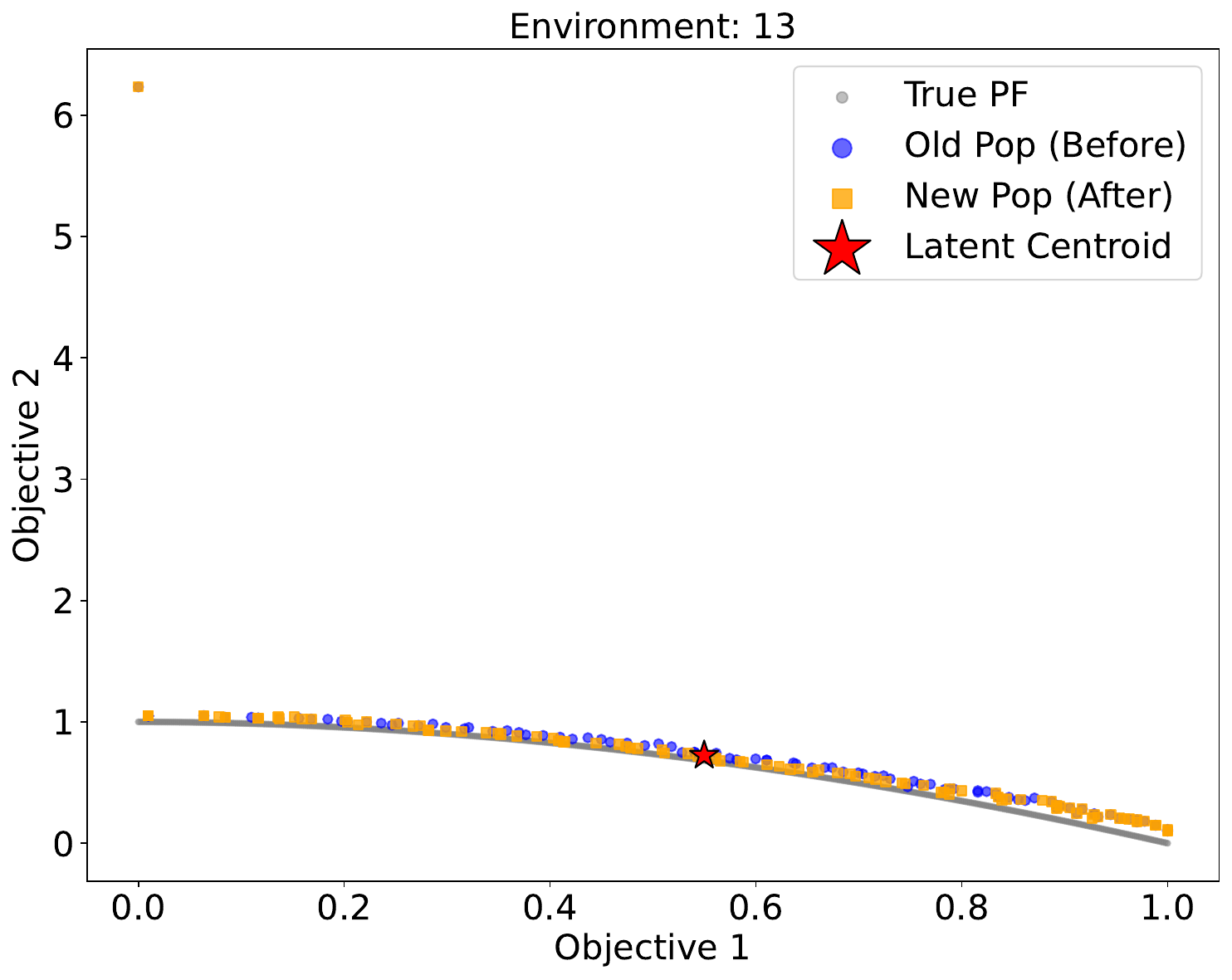}\hfill
    \includegraphics[width=0.19\linewidth]{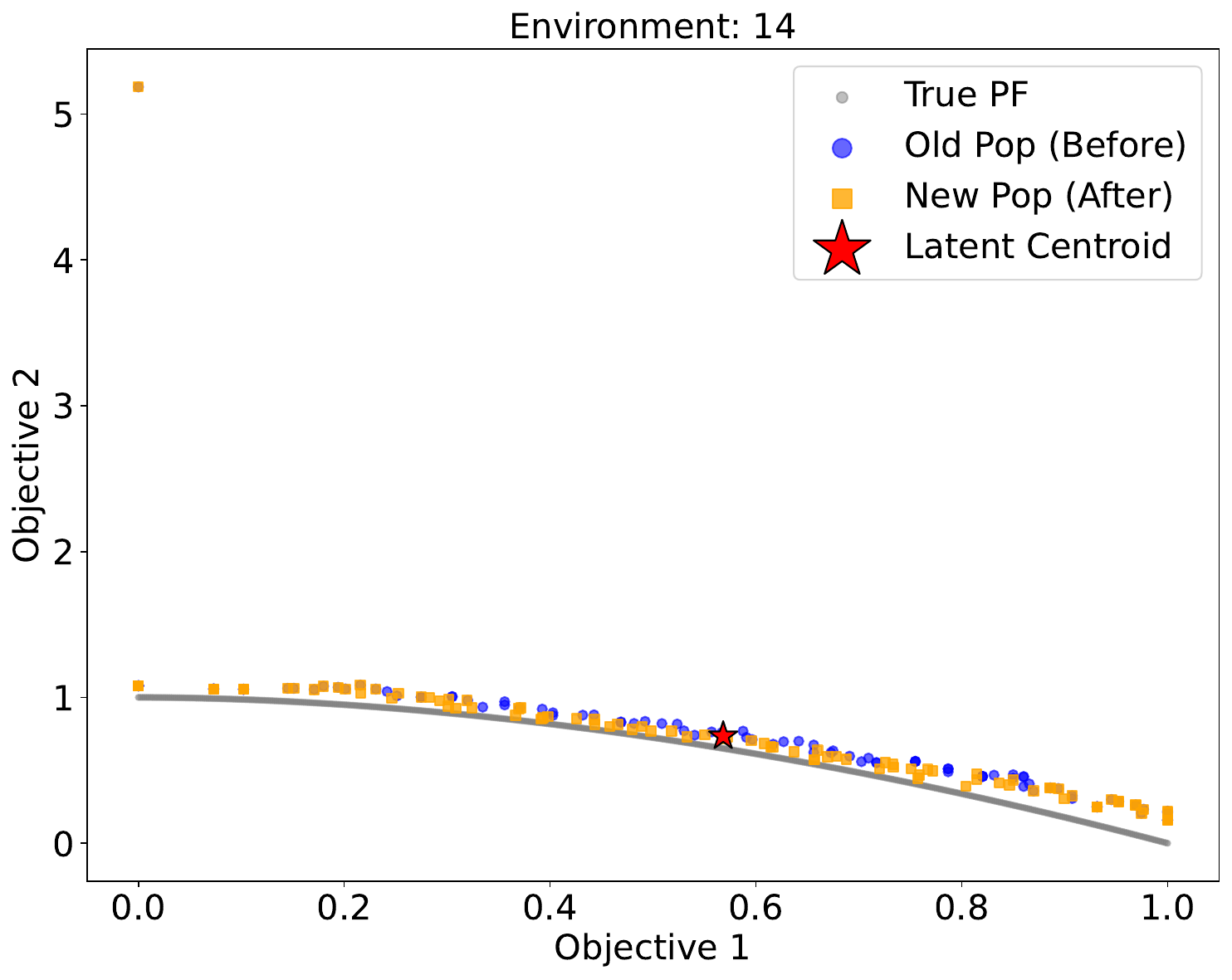}\hfill
    \includegraphics[width=0.19\linewidth]{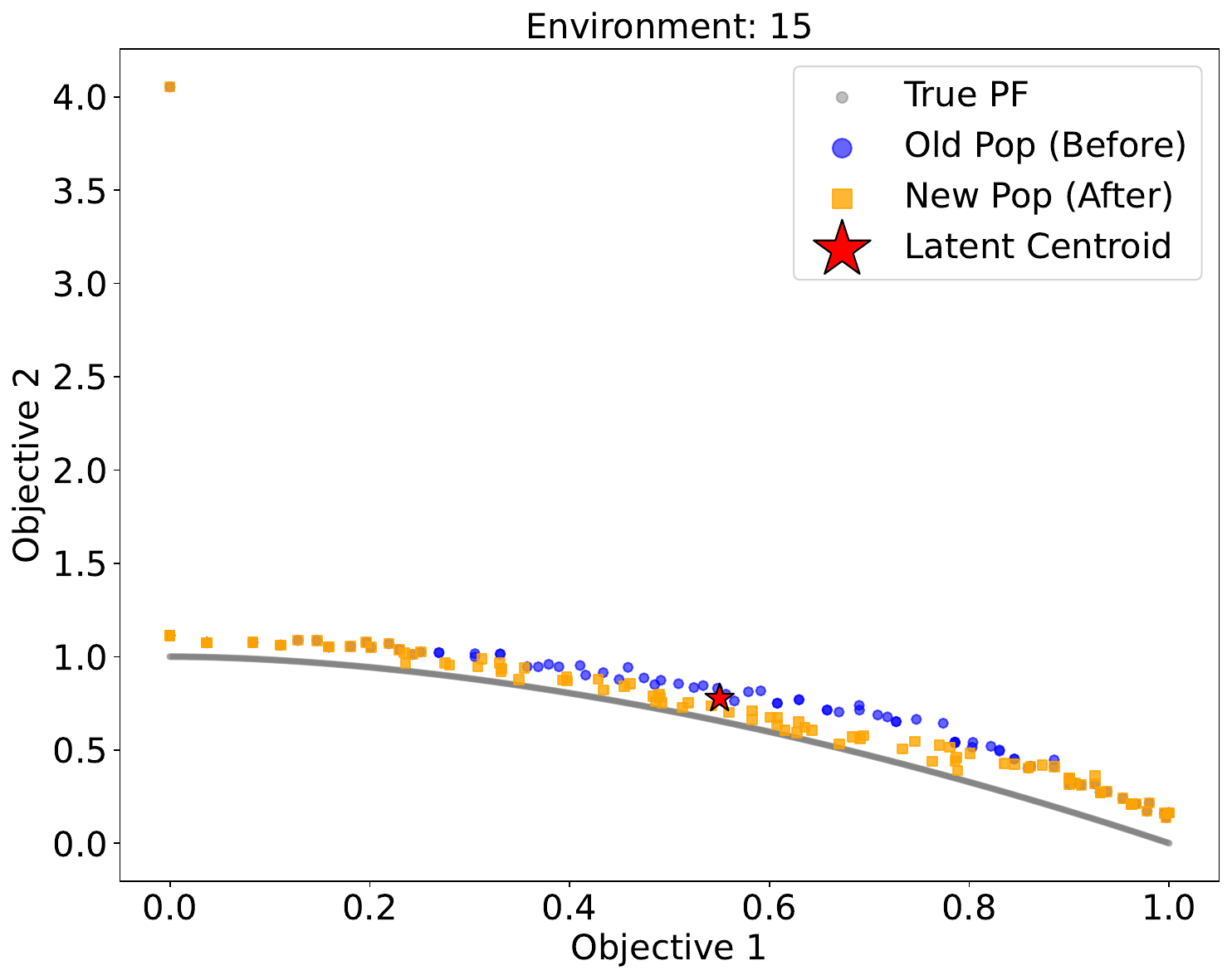}
    \\[4pt]
    \includegraphics[width=0.19\linewidth]{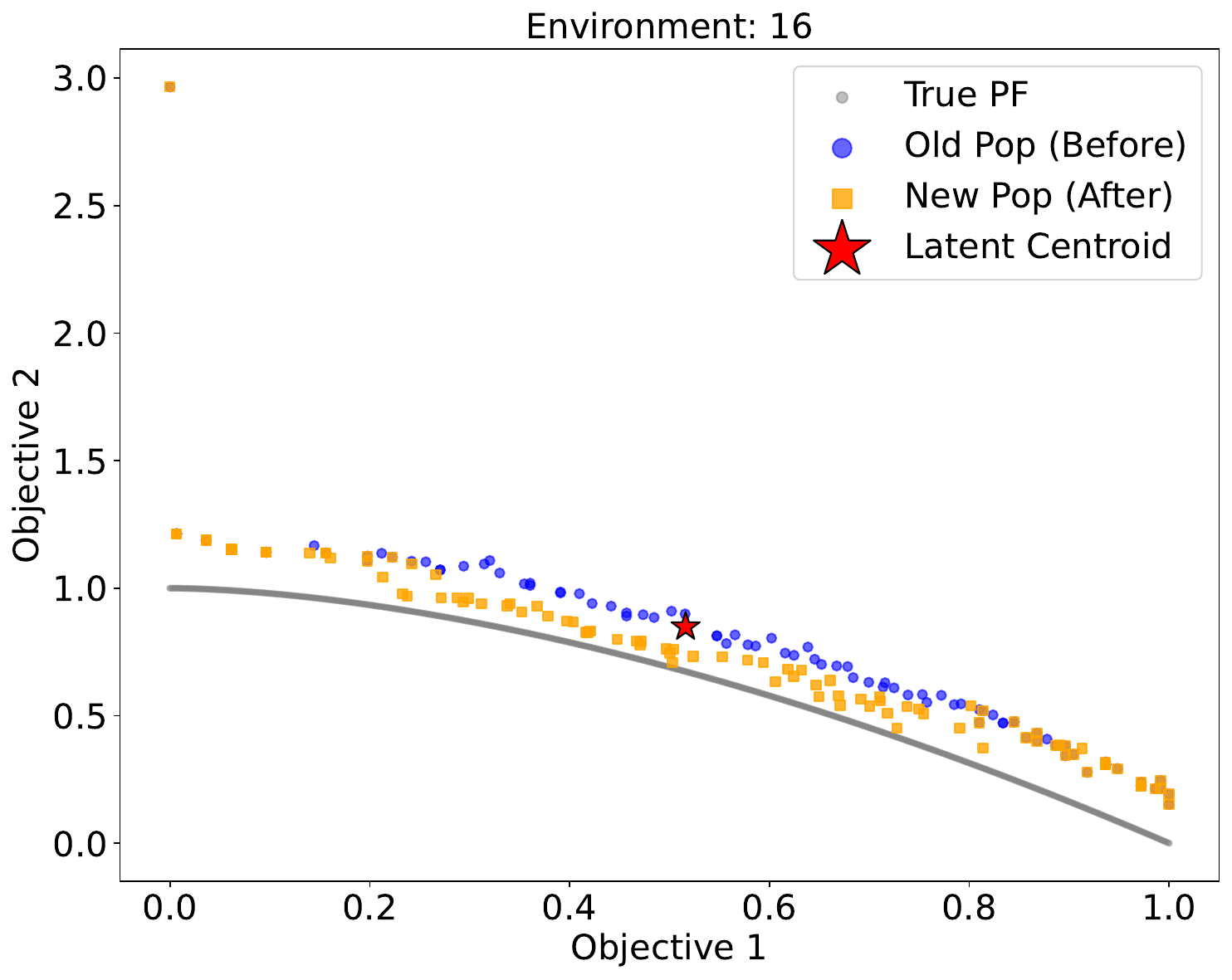}\hfill
    \includegraphics[width=0.19\linewidth]{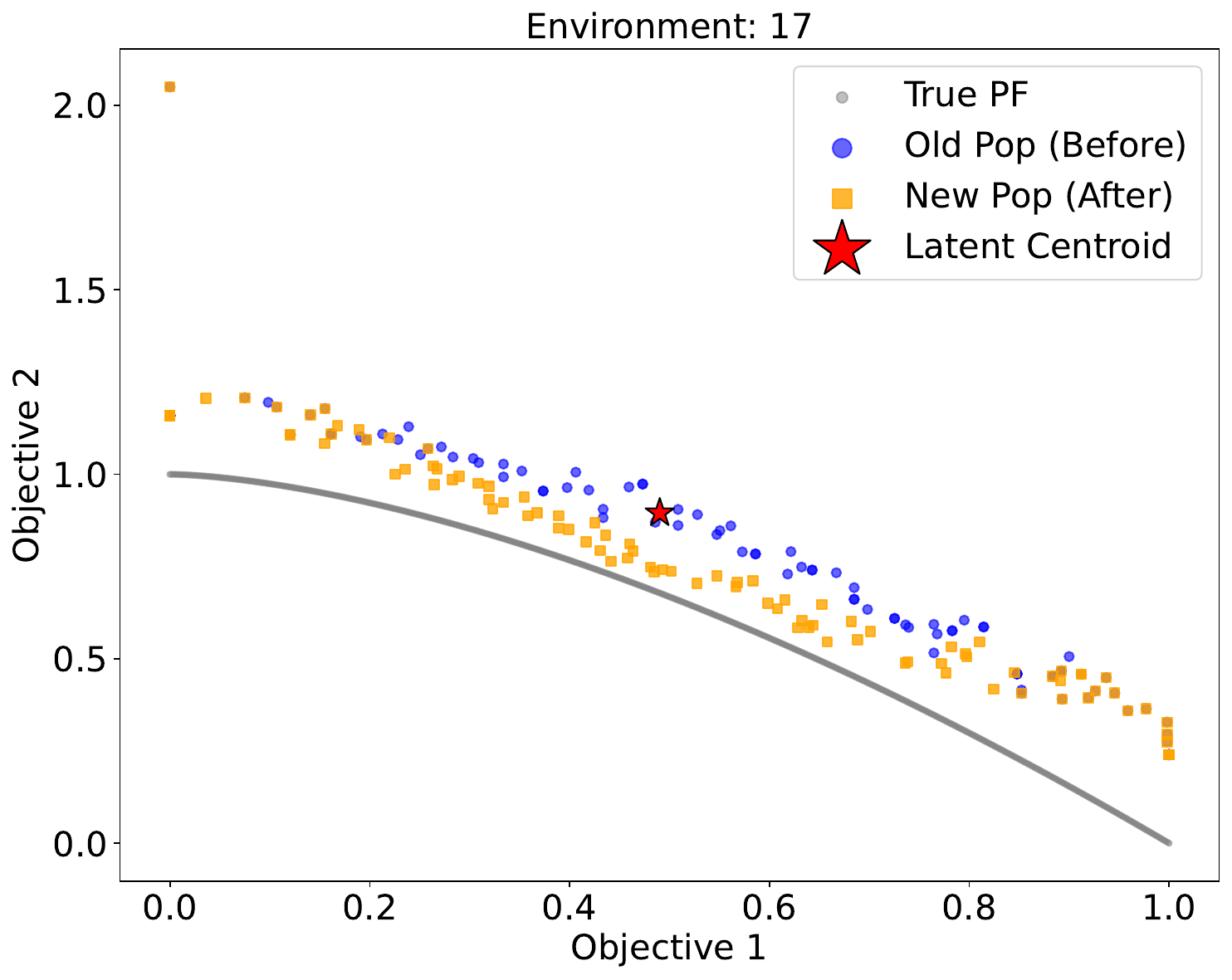}\hfill
    \includegraphics[width=0.19\linewidth]{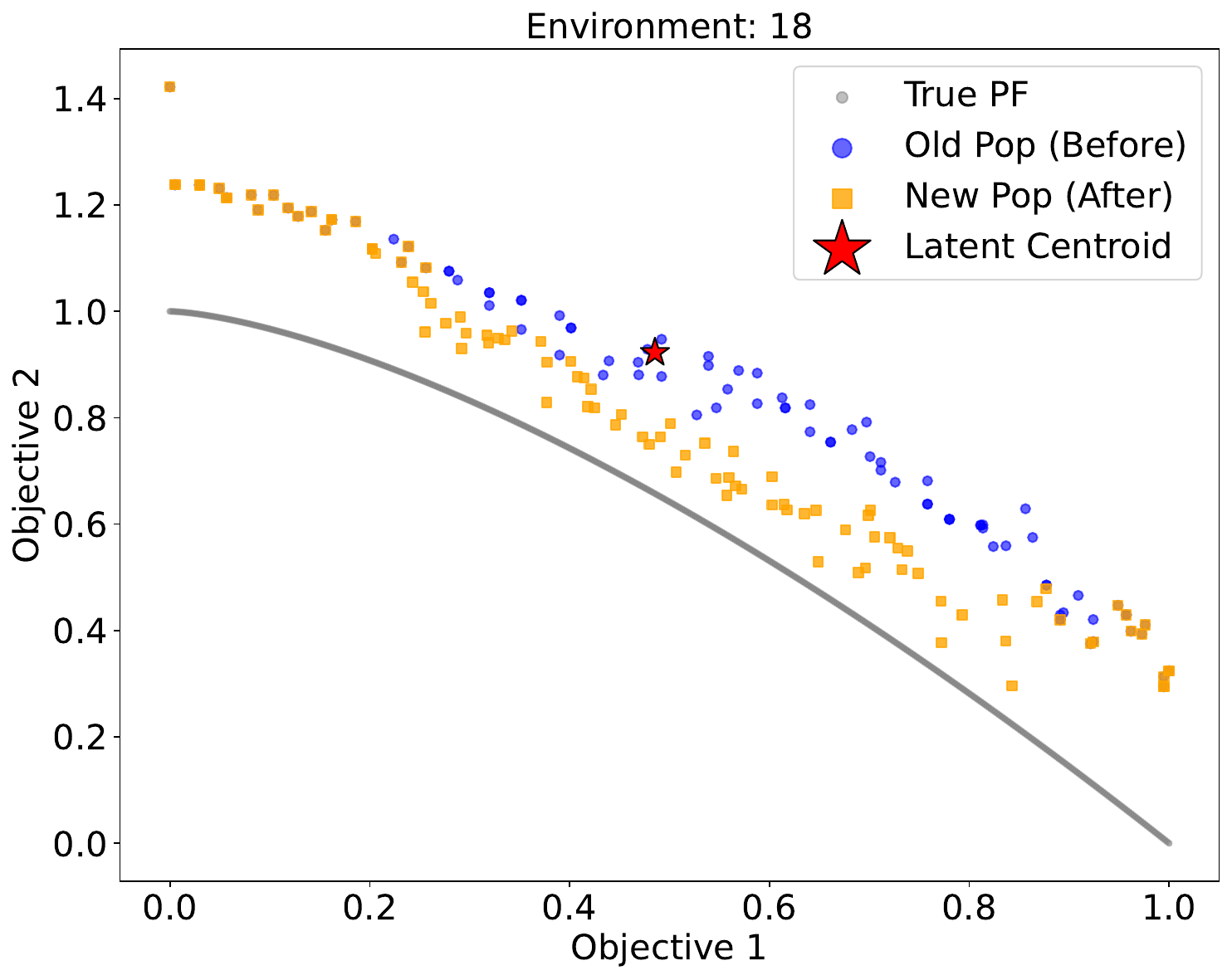}\hfill
    \includegraphics[width=0.19\linewidth]{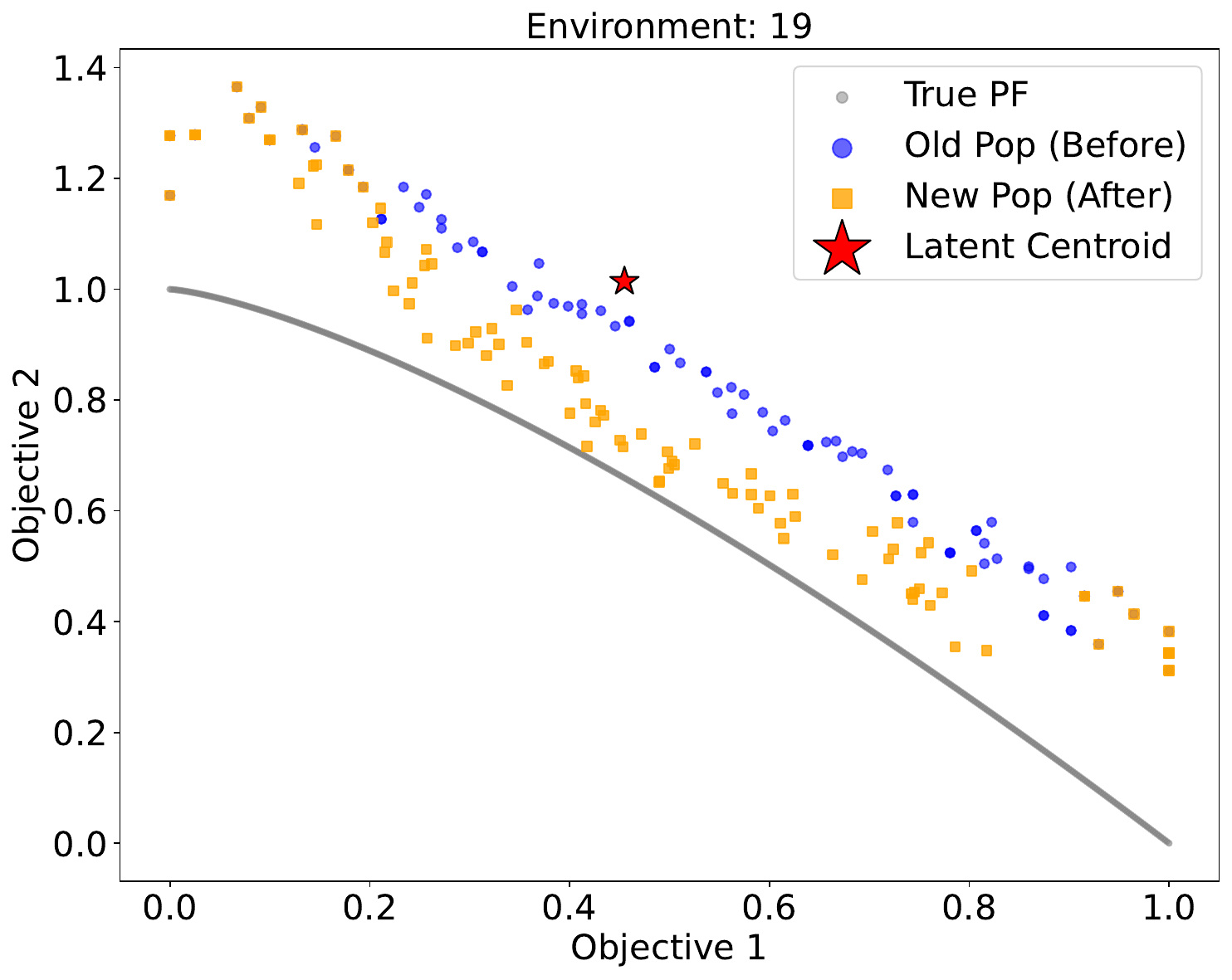}\hfill
    \includegraphics[width=0.19\linewidth]{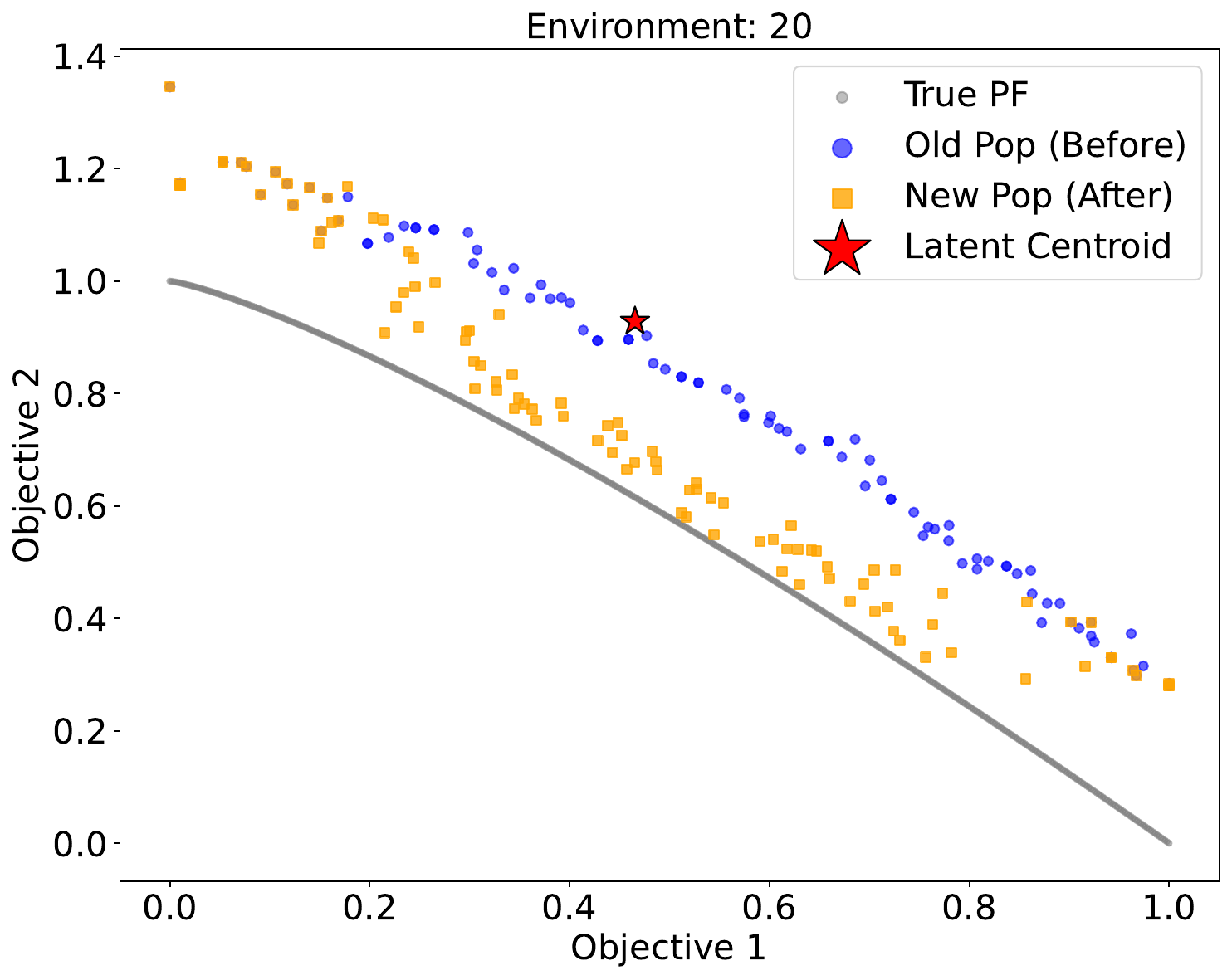}
    \\[4pt]
    \includegraphics[width=0.19\linewidth]{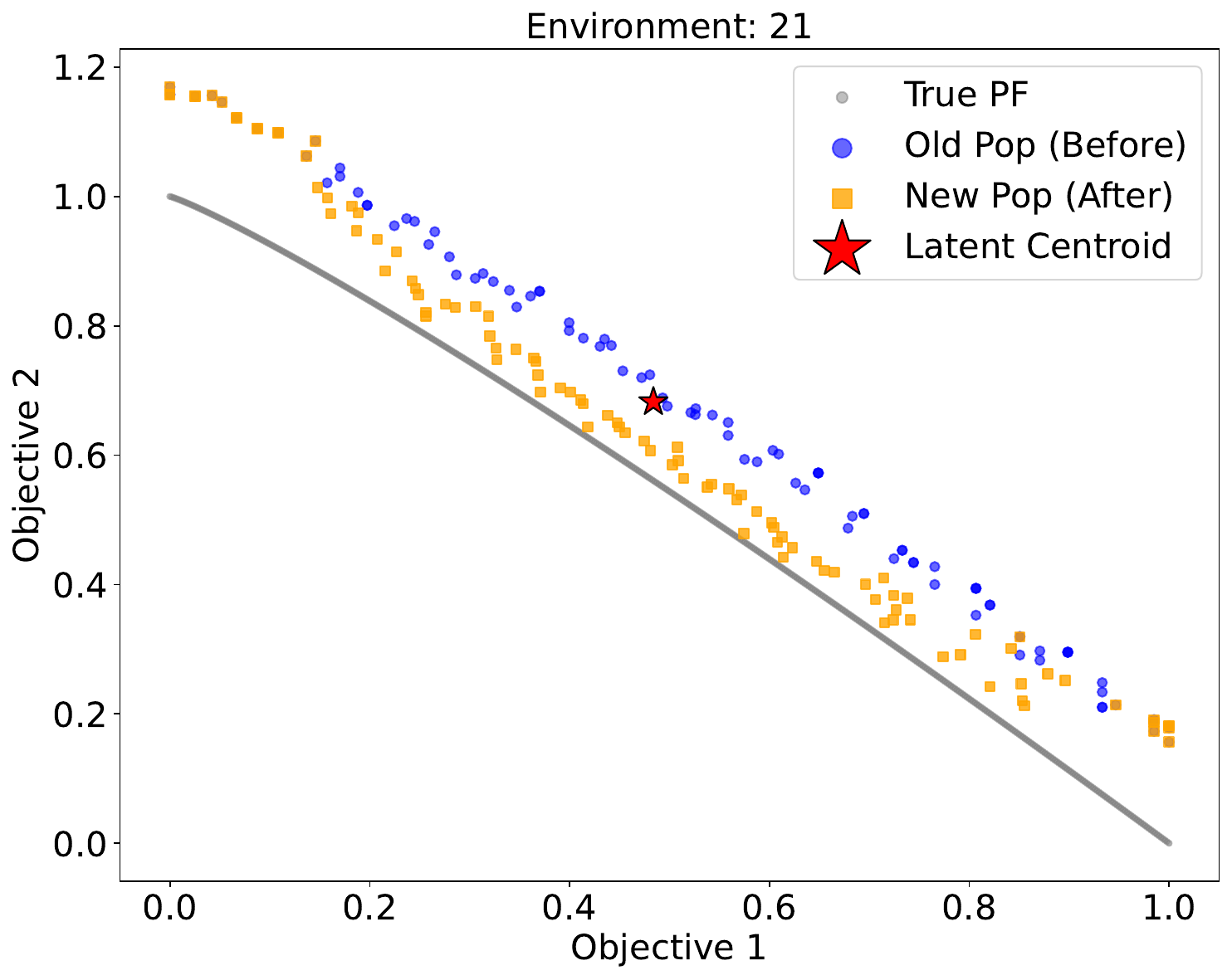}\hfill
    \includegraphics[width=0.19\linewidth]{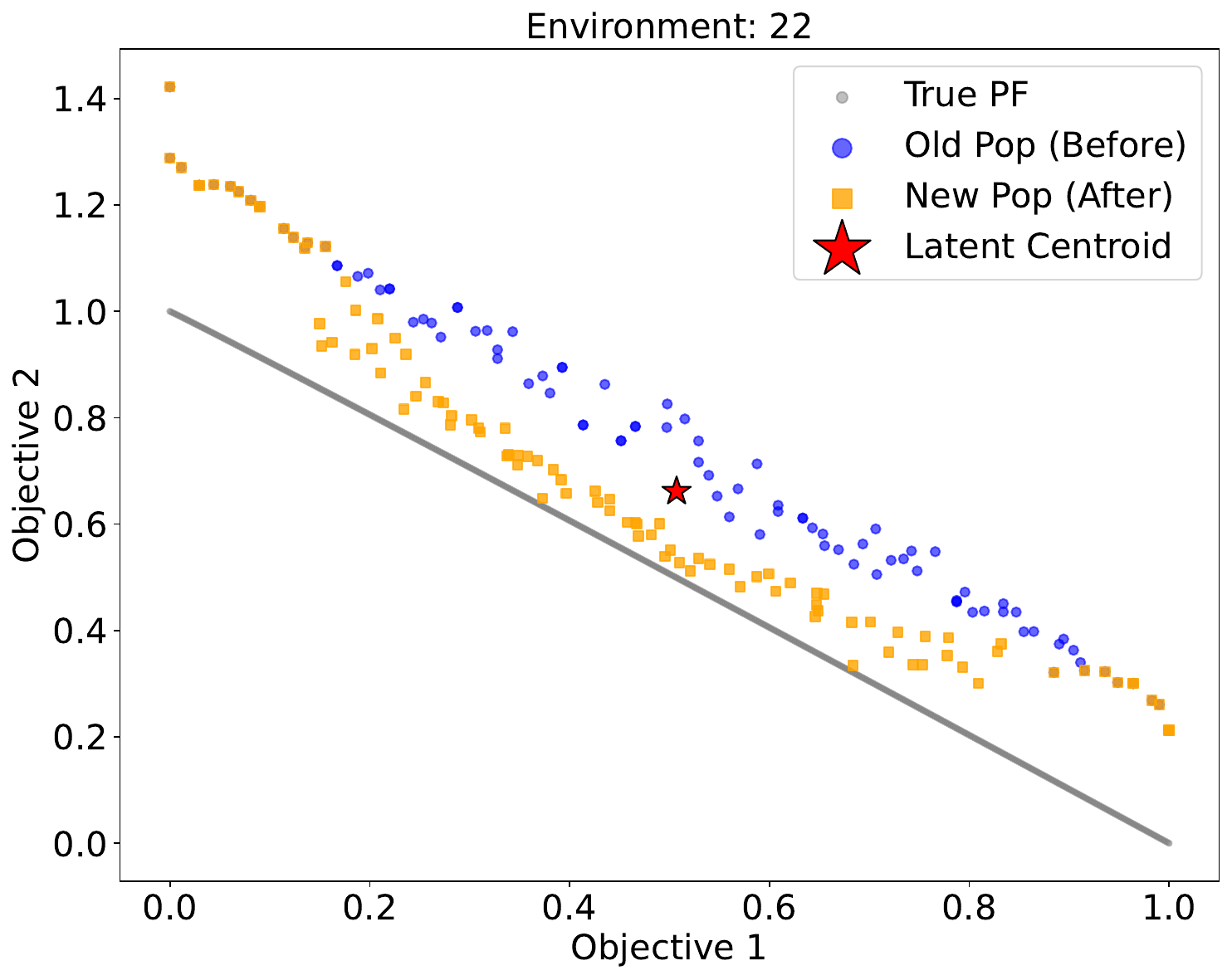}\hfill
    \includegraphics[width=0.19\linewidth]{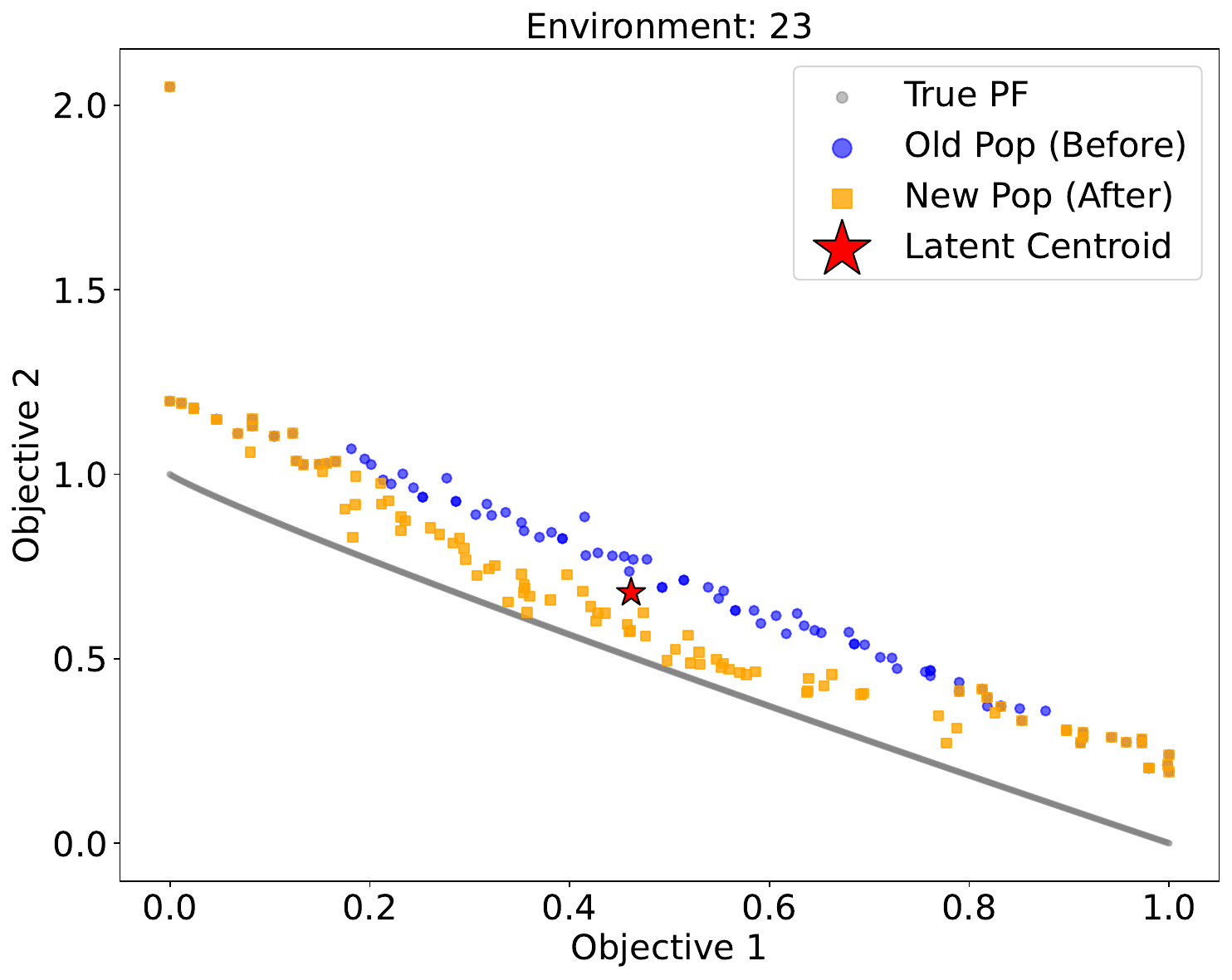}\hfill
    \includegraphics[width=0.19\linewidth]{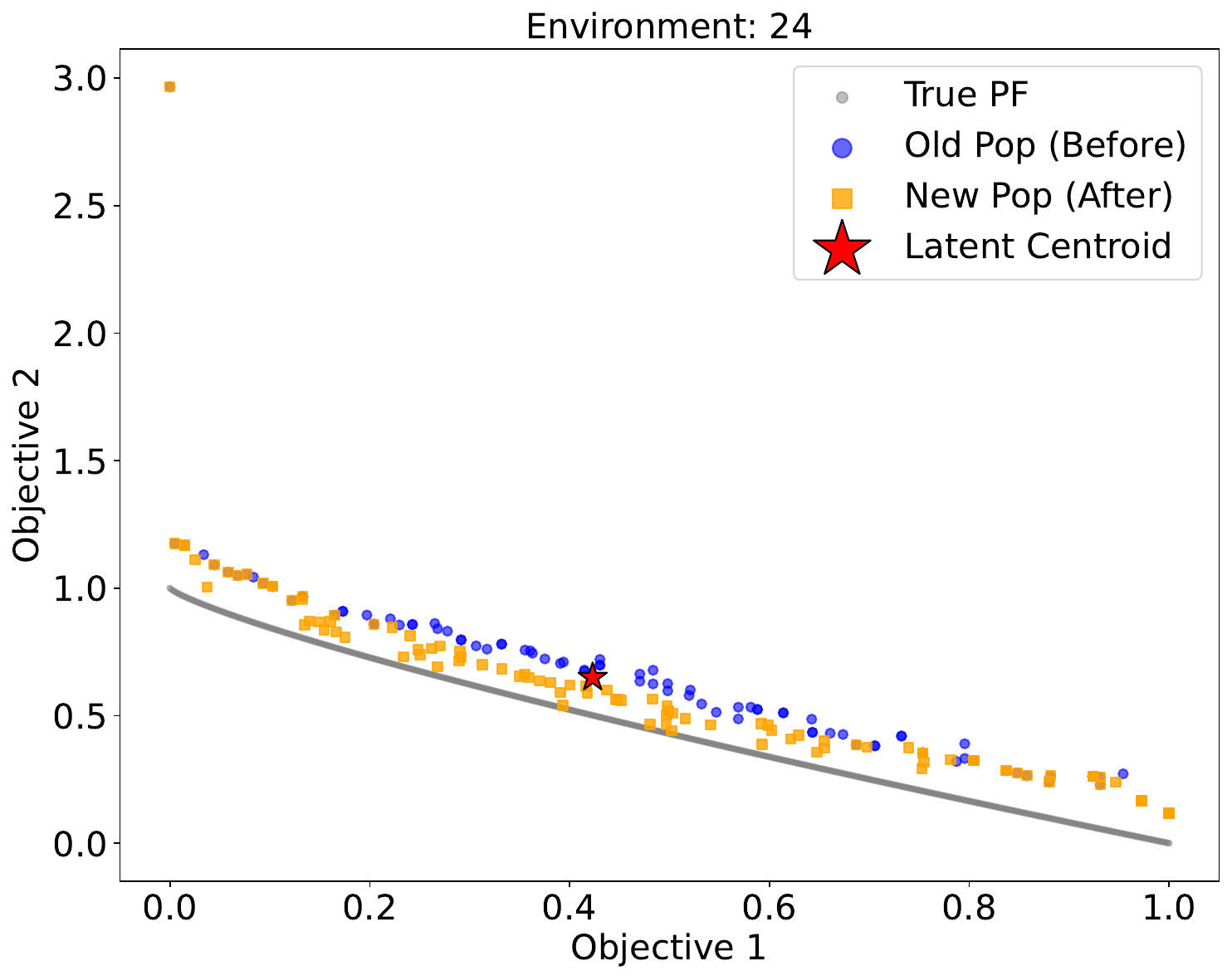}\hfill
    \includegraphics[width=0.19\linewidth]{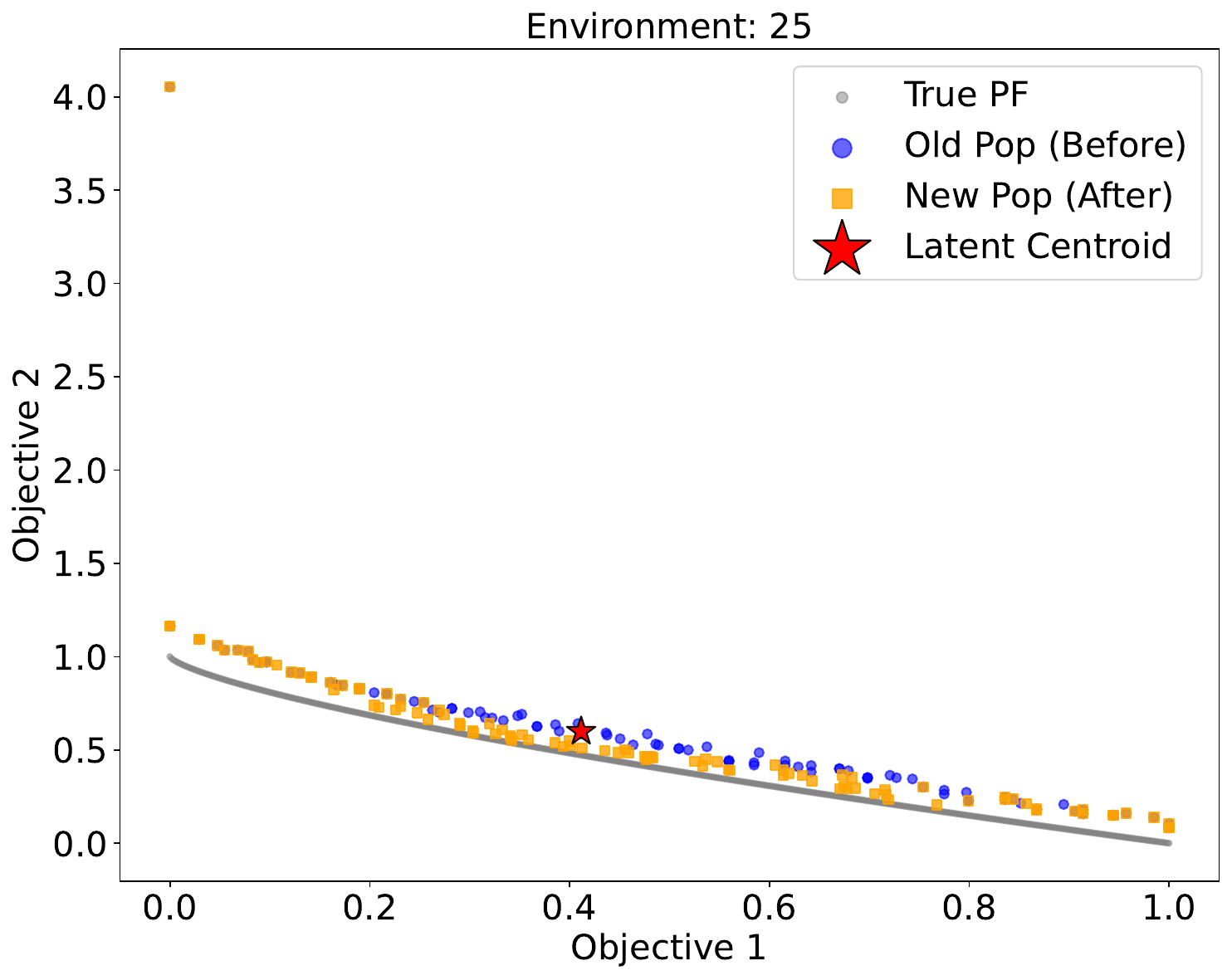}
    \\[4pt]
    \includegraphics[width=0.19\linewidth]{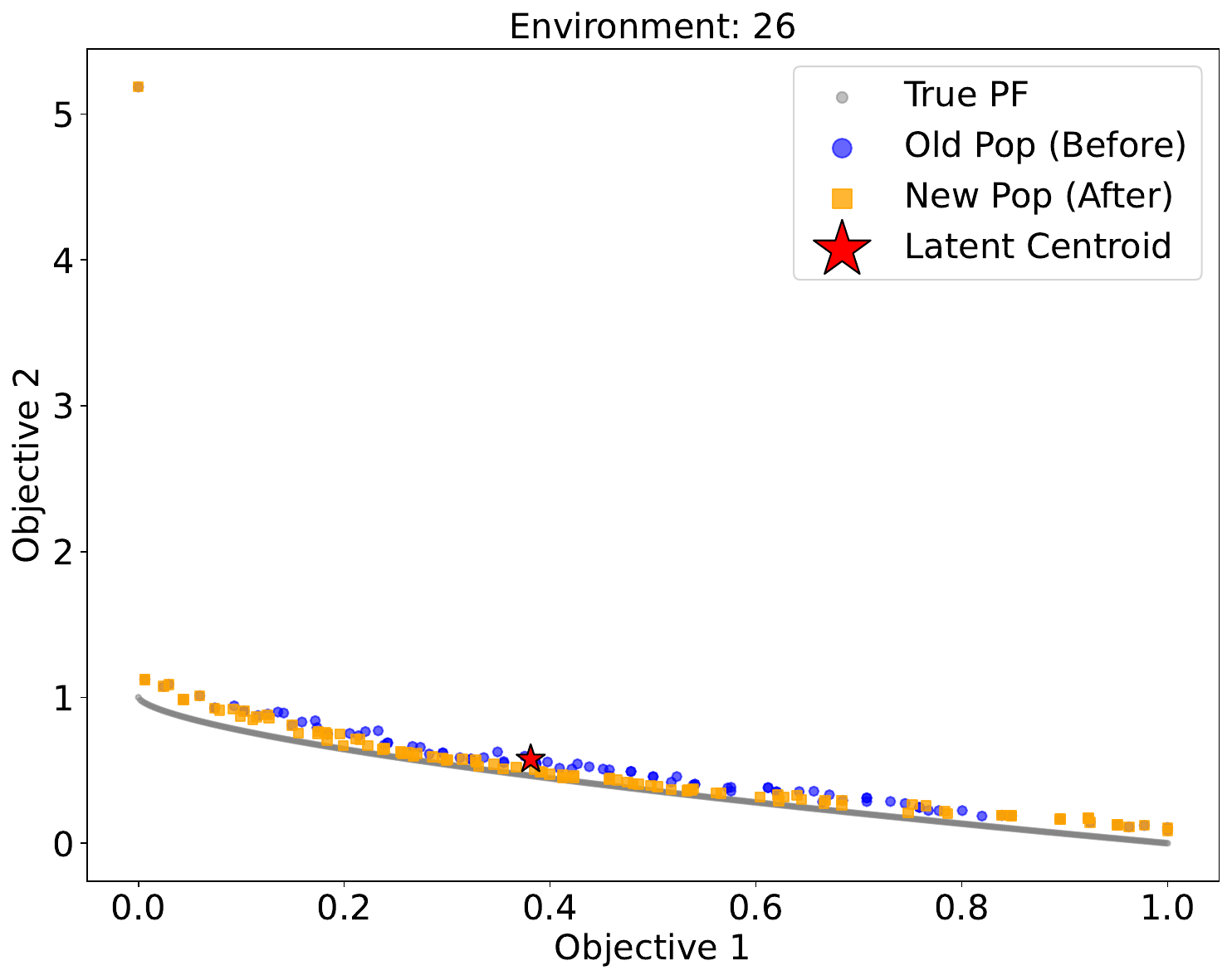}\hfill
    \includegraphics[width=0.19\linewidth]{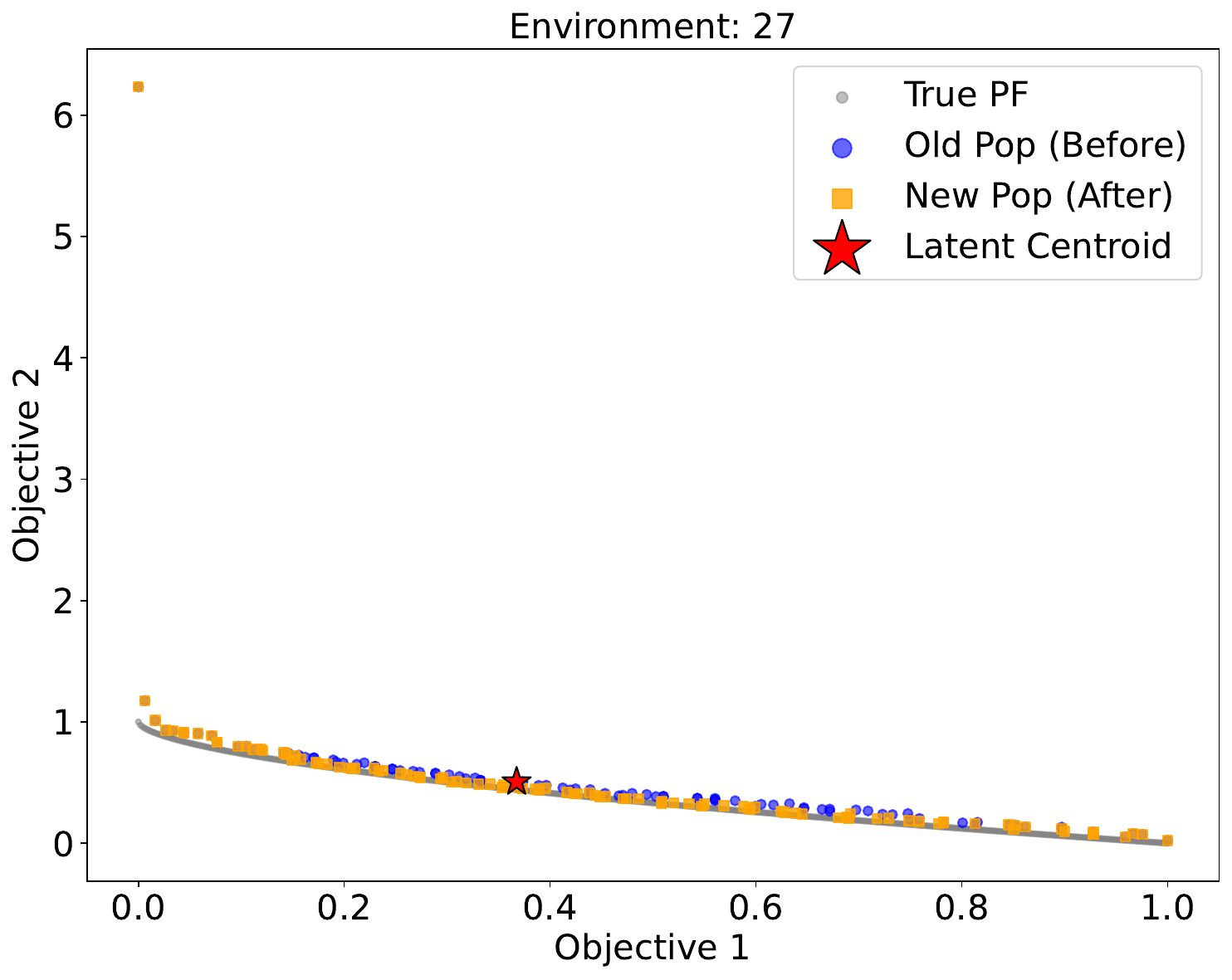}\hfill
    \includegraphics[width=0.19\linewidth]{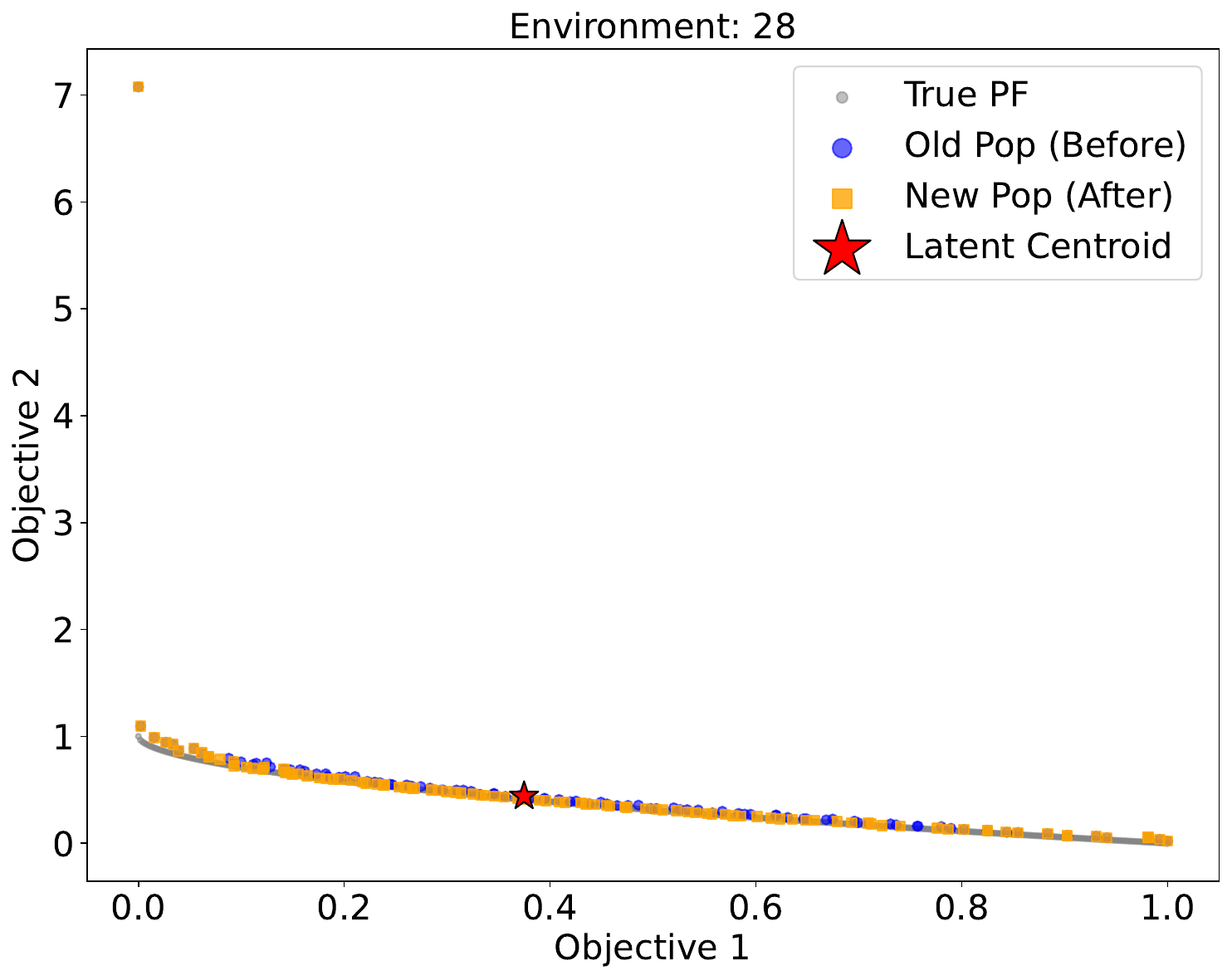}\hfill
    \includegraphics[width=0.19\linewidth]{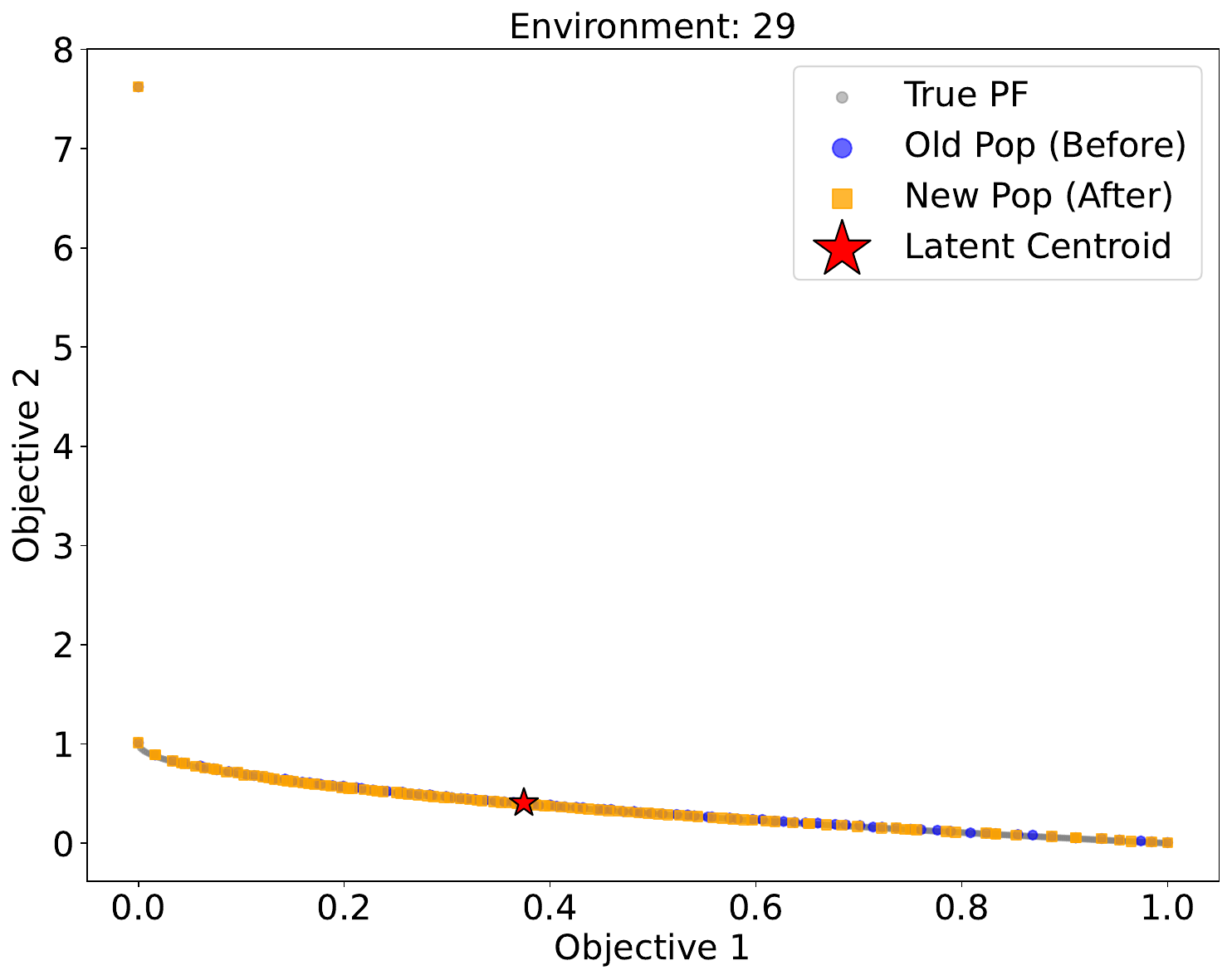}\hfill
    \includegraphics[width=0.19\linewidth]{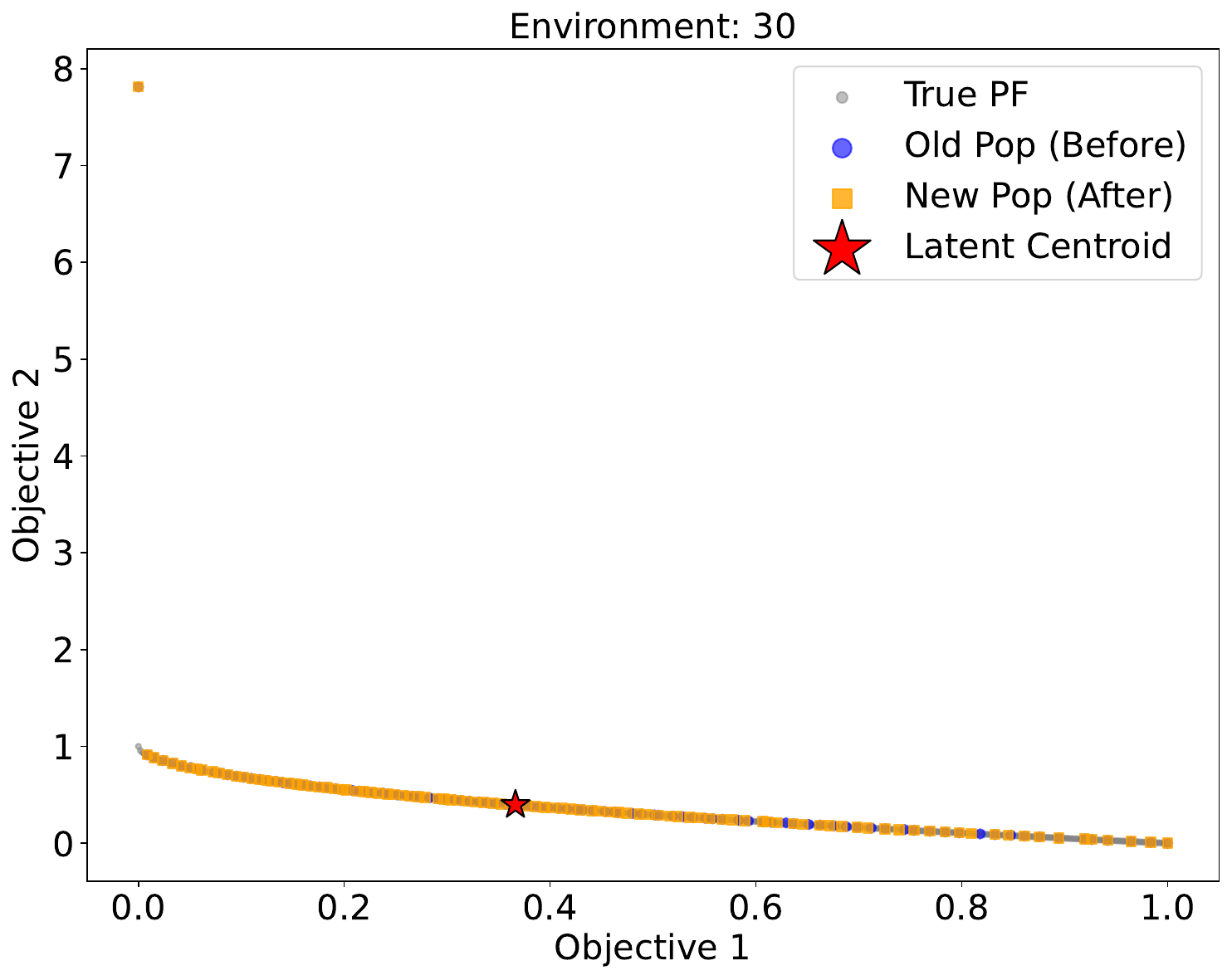}
    \\[2pt] 
    \caption{Complete tracking performance of the proposed DB-GEN utilizing the \textbf{centroid perturbation strategy} across all 30 environments on the DF1 problem. The grey dots delineate the true PF, the blue dots represent the un-evaluated inherited population, the red star marks the latent centroid, and the orange dots denote the generated initial candidates.}
    \label{fig:app_df1_tracking_mean}
\end{figure*}

Fig.~\ref{fig:app_df1_tracking_mean} visualizes the performance of the proposed \textbf{centroid perturbation strategy}. By aggregating the population into a unified joint centroid in the latent space before applying Gaussian perturbation, the model effectively filters out the individual historical biases of the inherited population. As observed across the 30 environments, this strategy consistently generates a topologically correct and well-distributed initial candidate pool (orange dots) that closely approximates the true Pareto Front (grey dots), regardless of the displacement of the old population (blue dots).

\begin{figure*}[htbp] 
    \centering
    \includegraphics[width=0.19\linewidth]{figures/DF1_all_search/Environment_1_visualization.pdf}\hfill
    \includegraphics[width=0.19\linewidth]{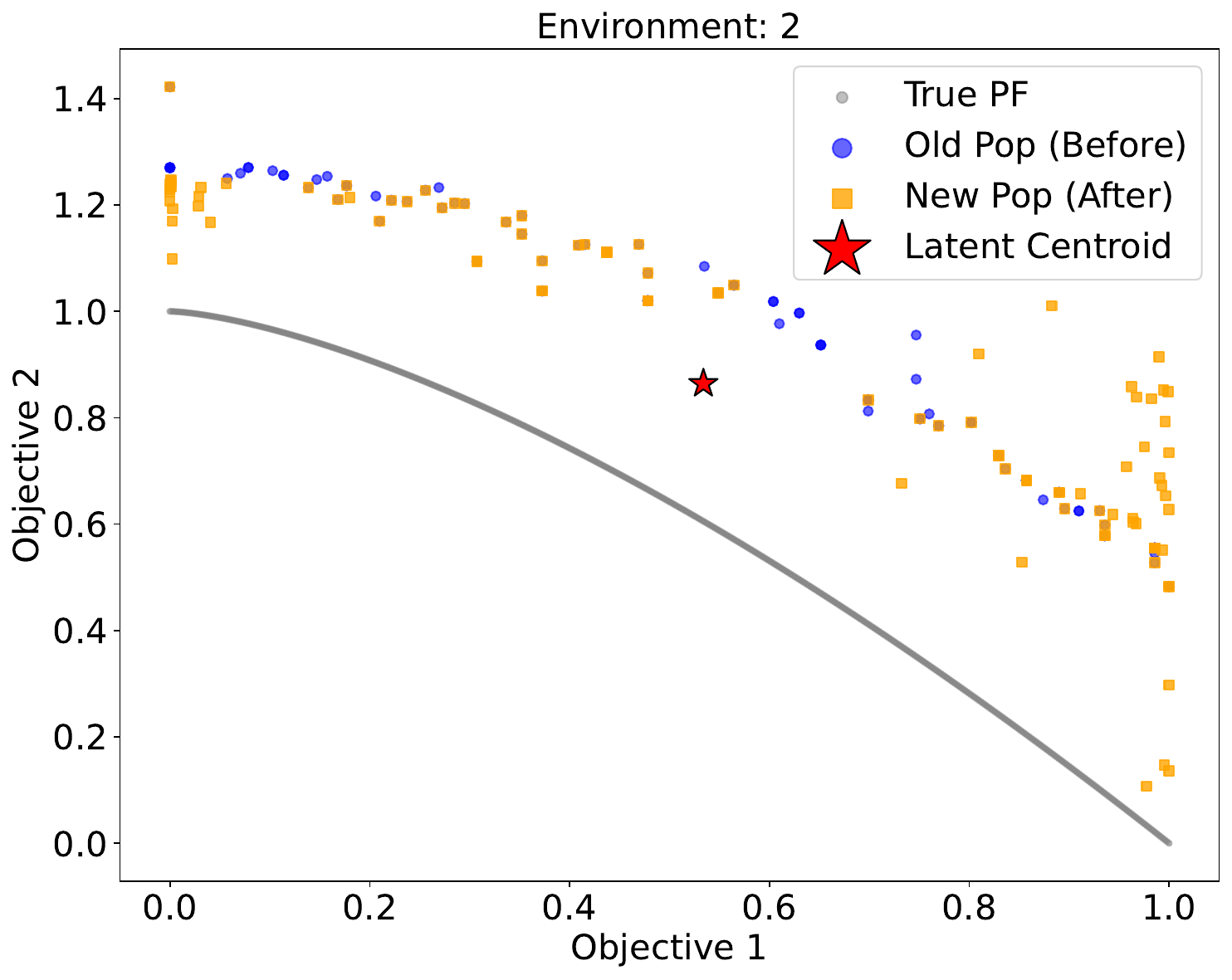}\hfill
    \includegraphics[width=0.19\linewidth]{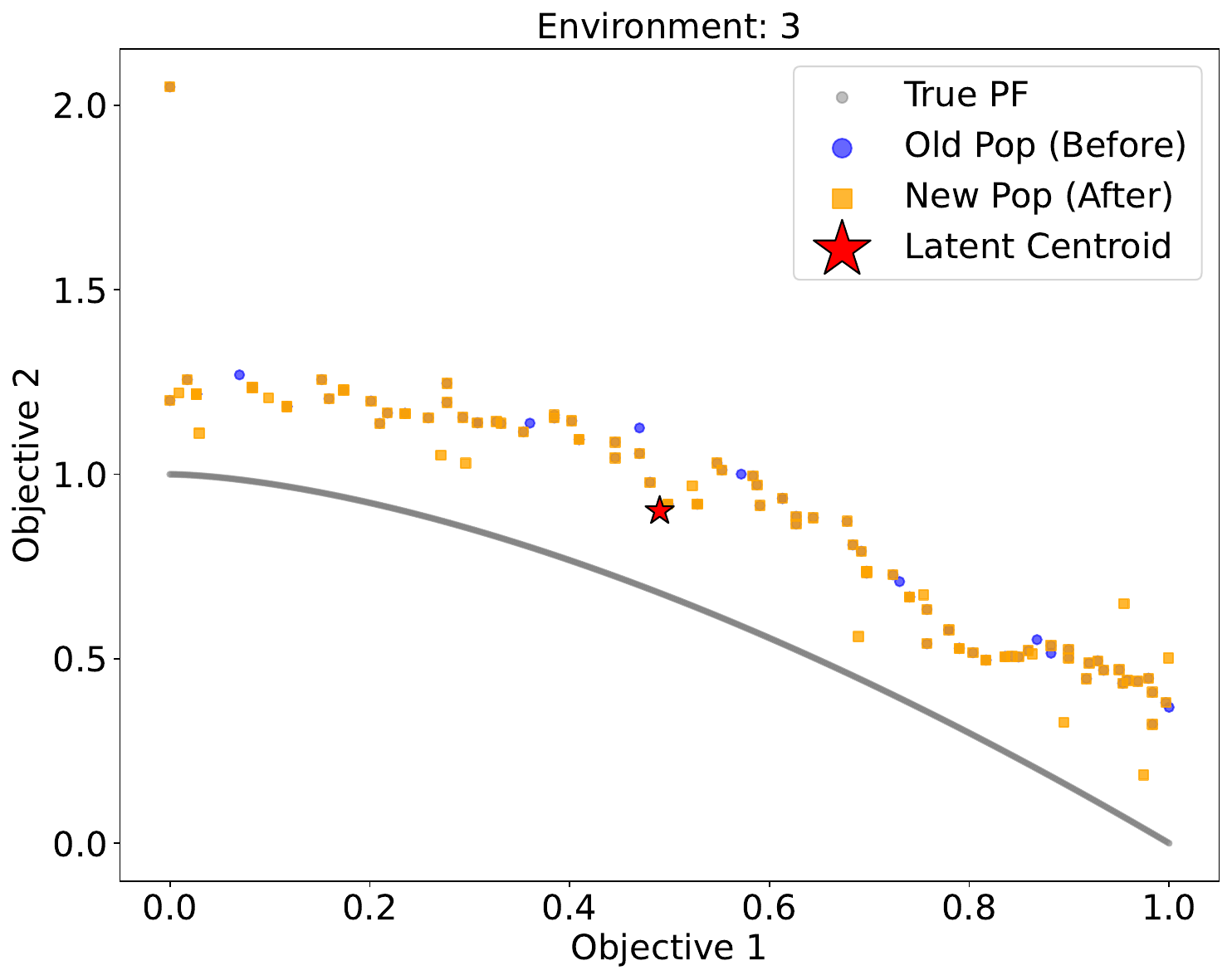}\hfill
    \includegraphics[width=0.19\linewidth]{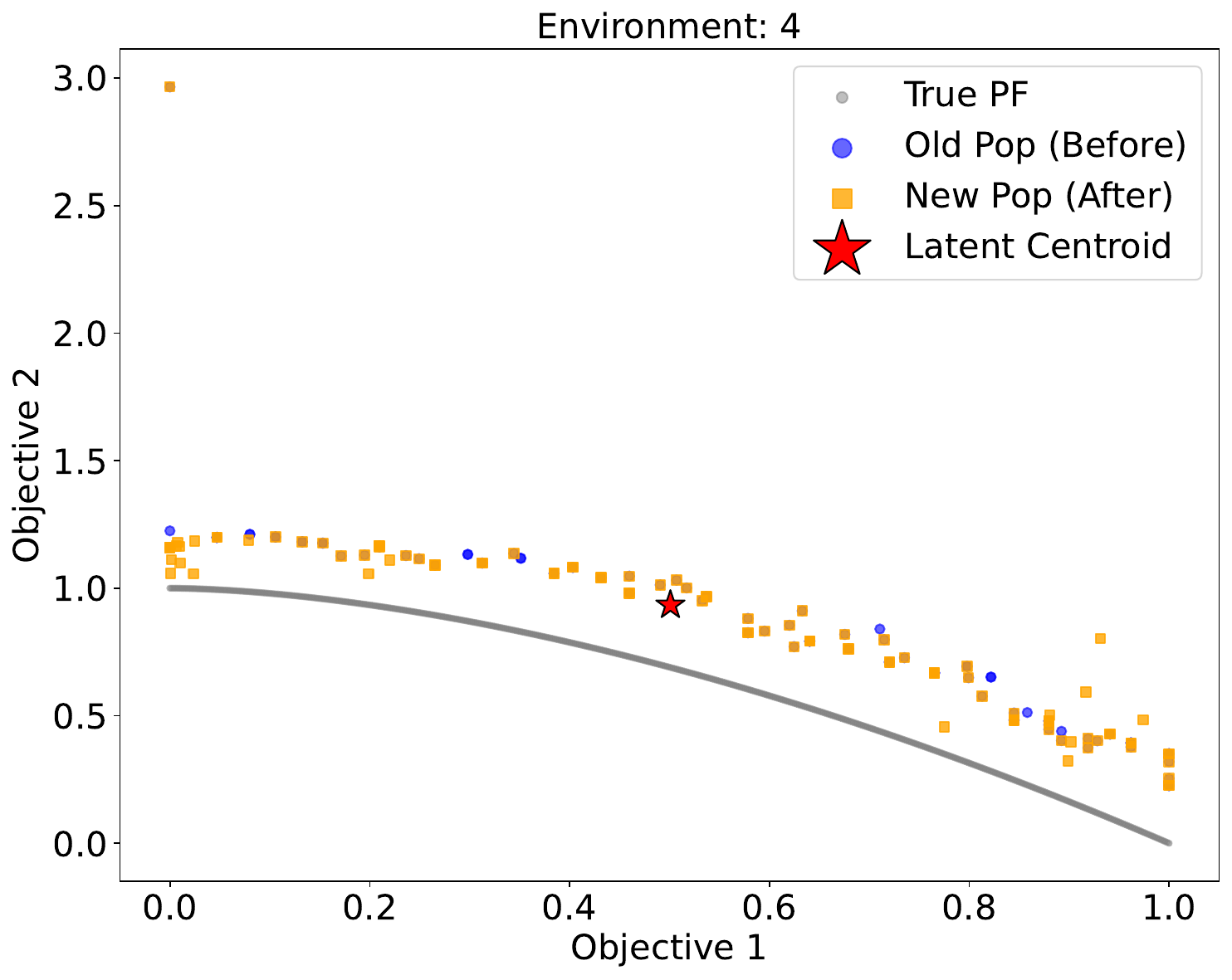}\hfill
    \includegraphics[width=0.19\linewidth]{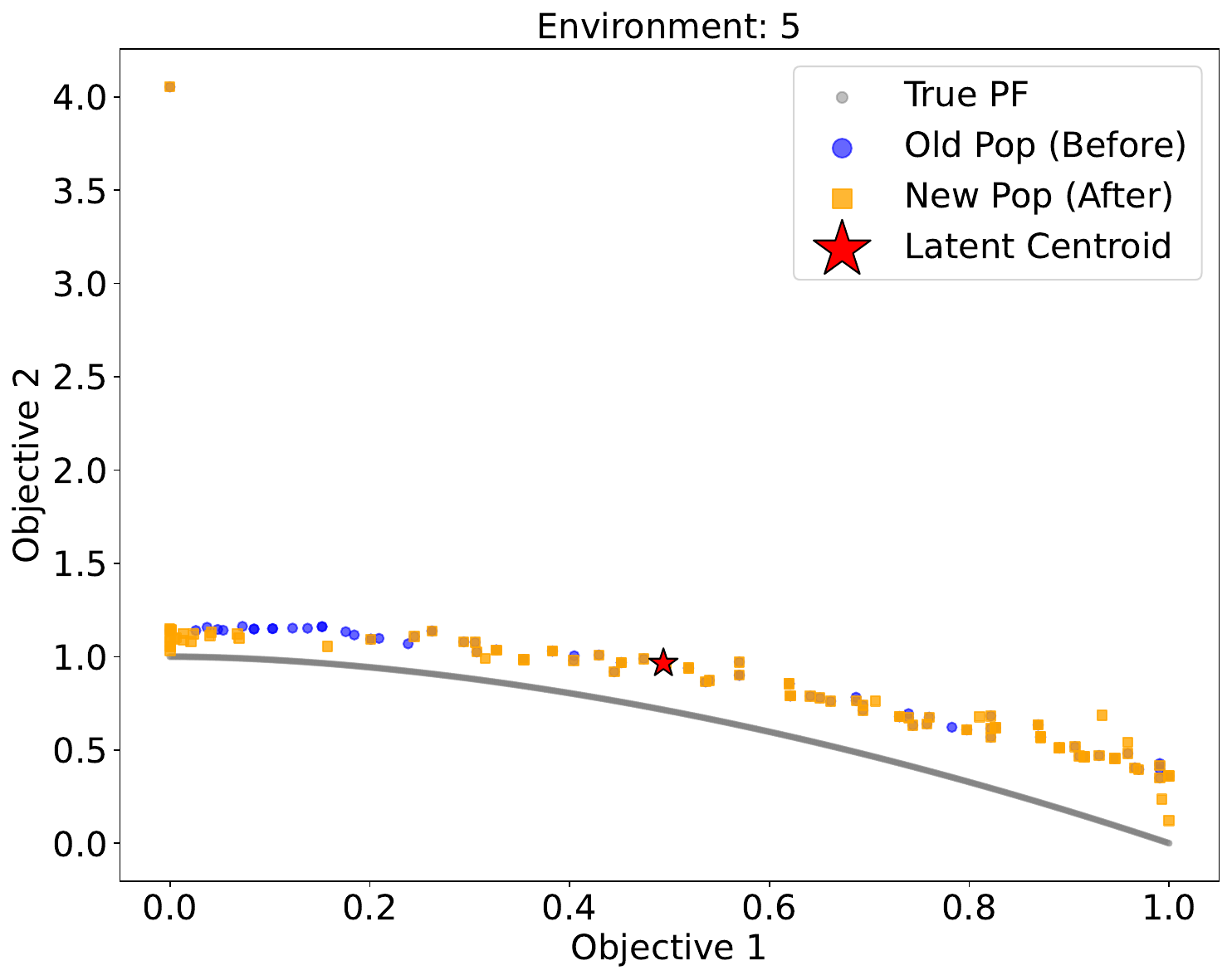}
    \\[4pt] 
    \includegraphics[width=0.19\linewidth]{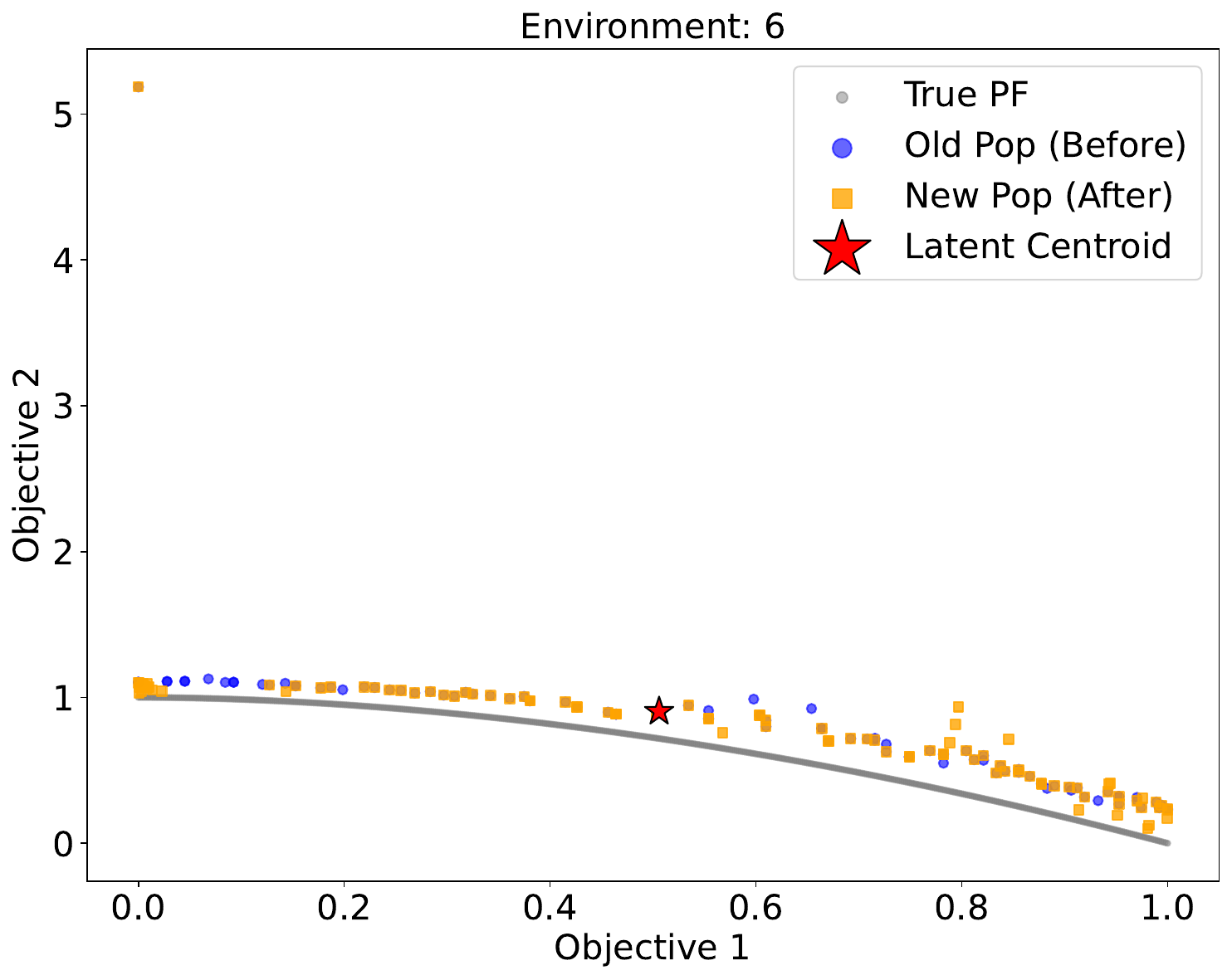}\hfill
    \includegraphics[width=0.19\linewidth]{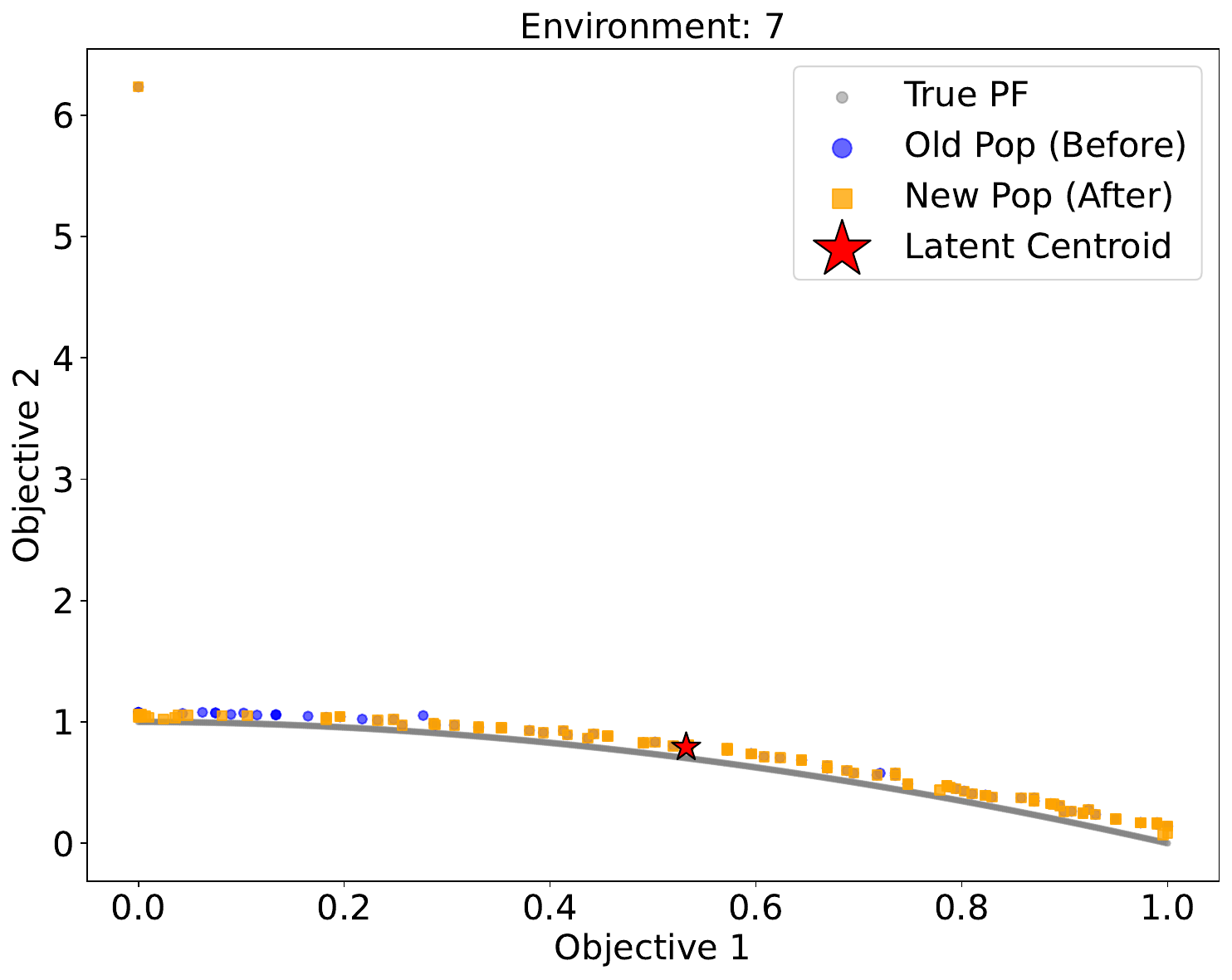}\hfill
    \includegraphics[width=0.19\linewidth]{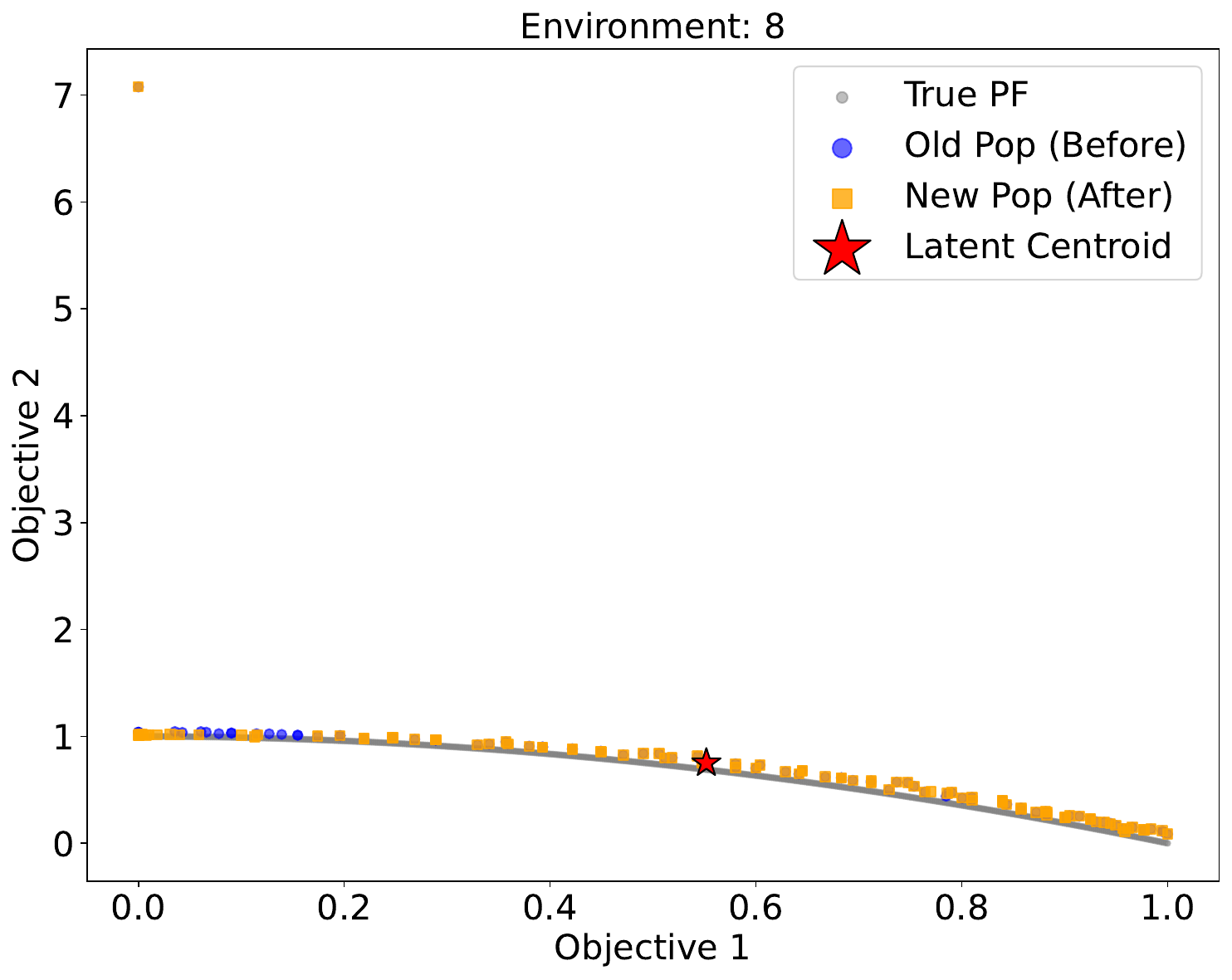}\hfill
    \includegraphics[width=0.19\linewidth]{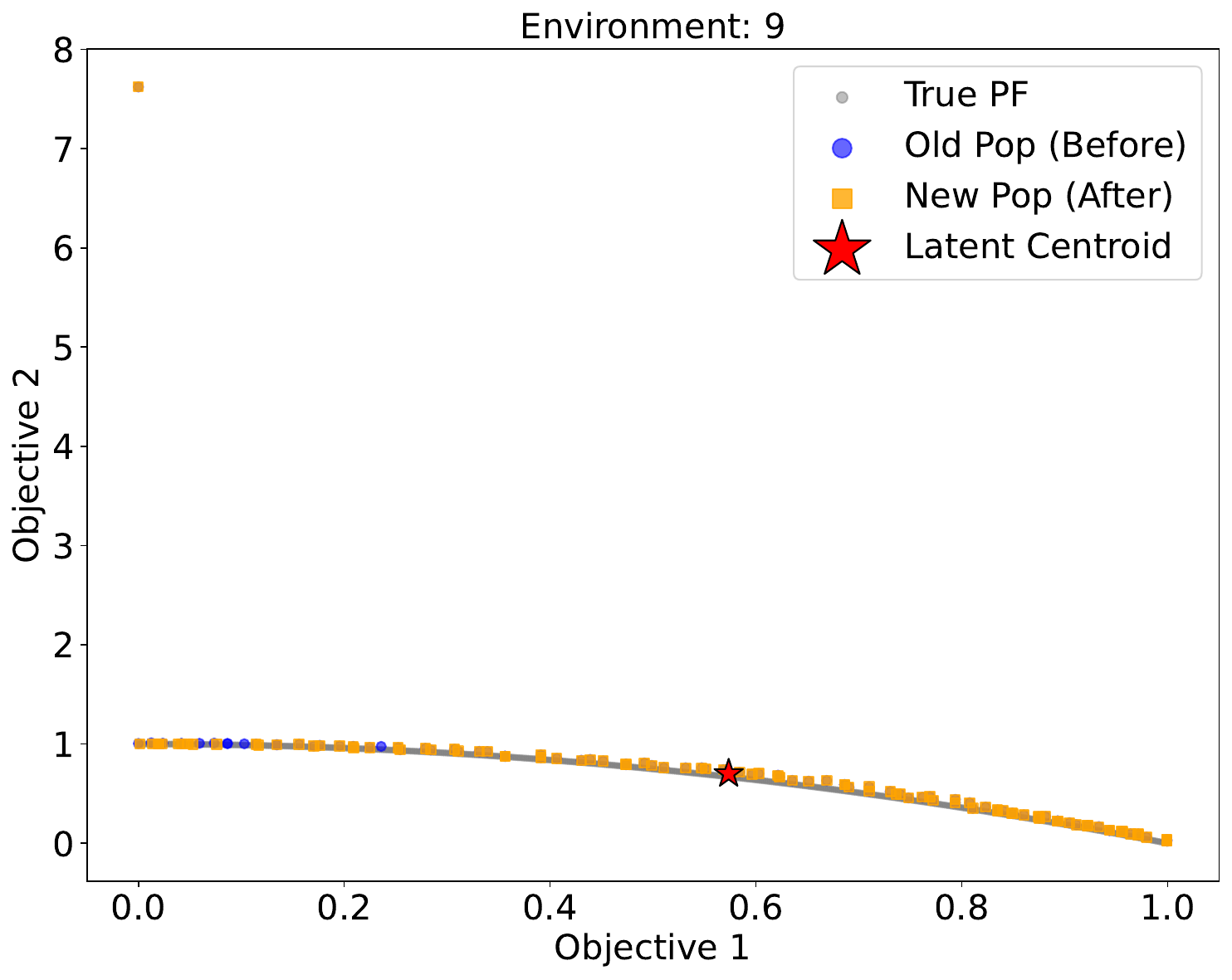}\hfill
    \includegraphics[width=0.19\linewidth]{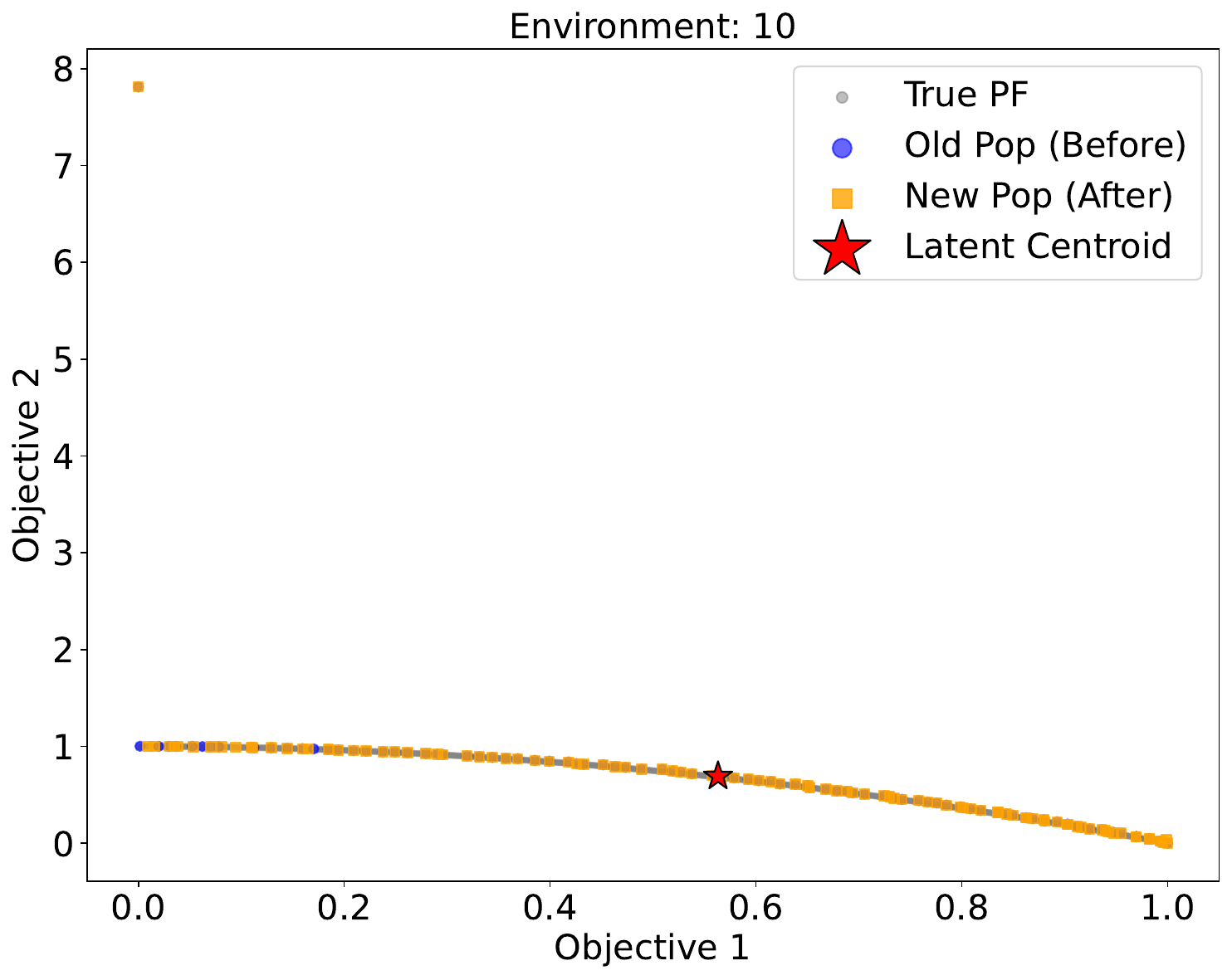}
    \\[4pt]
    \includegraphics[width=0.19\linewidth]{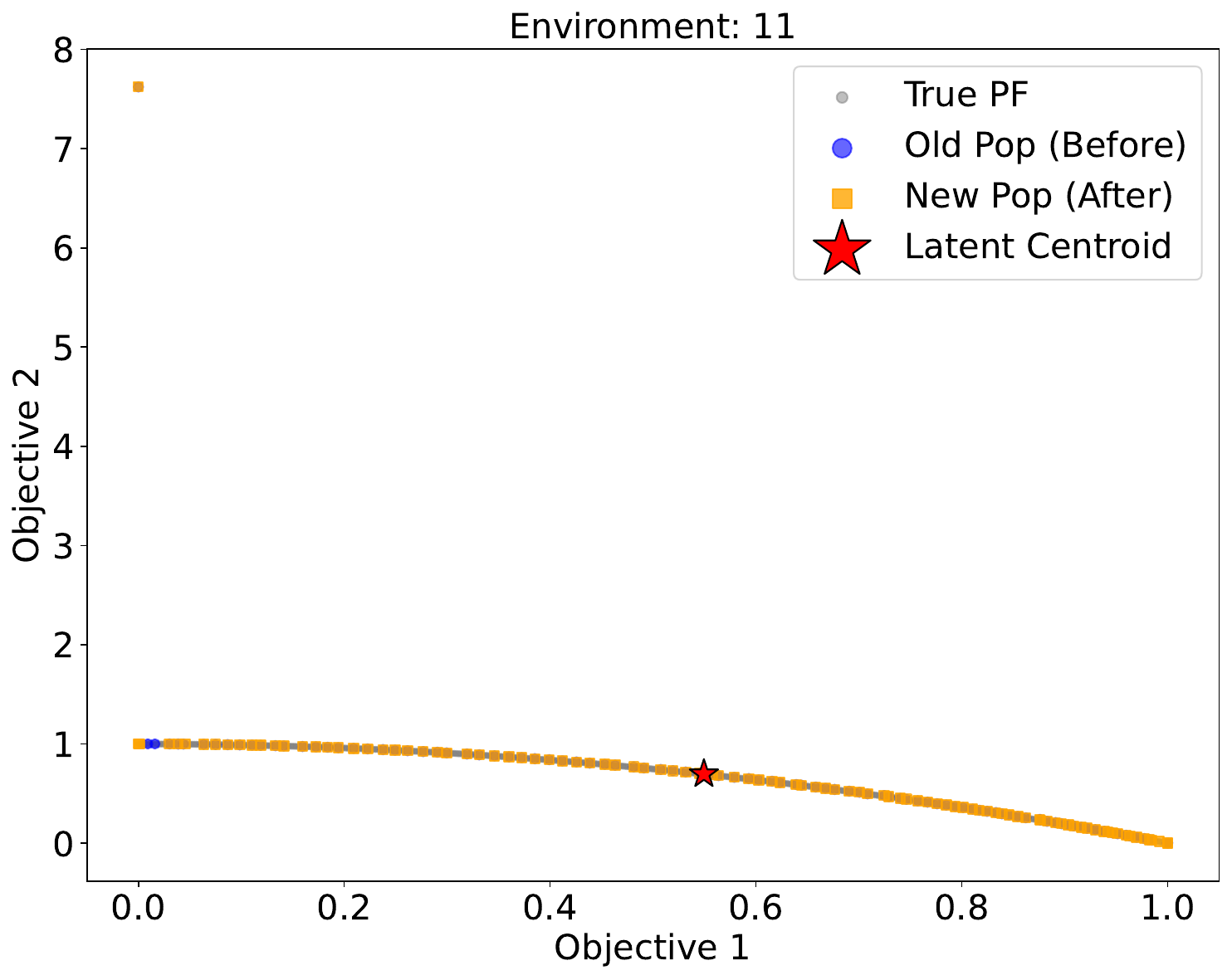}\hfill
    \includegraphics[width=0.19\linewidth]{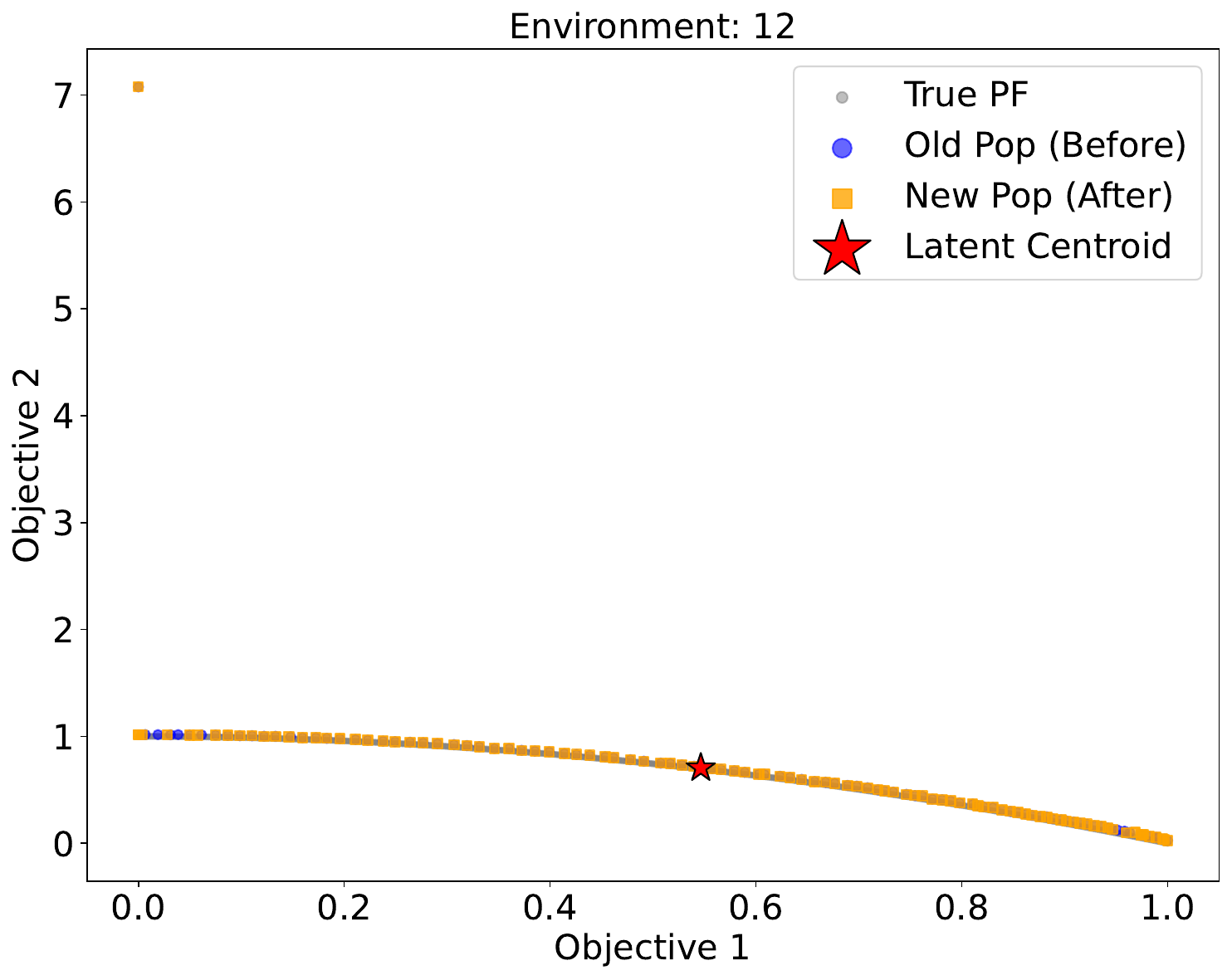}\hfill
    \includegraphics[width=0.19\linewidth]{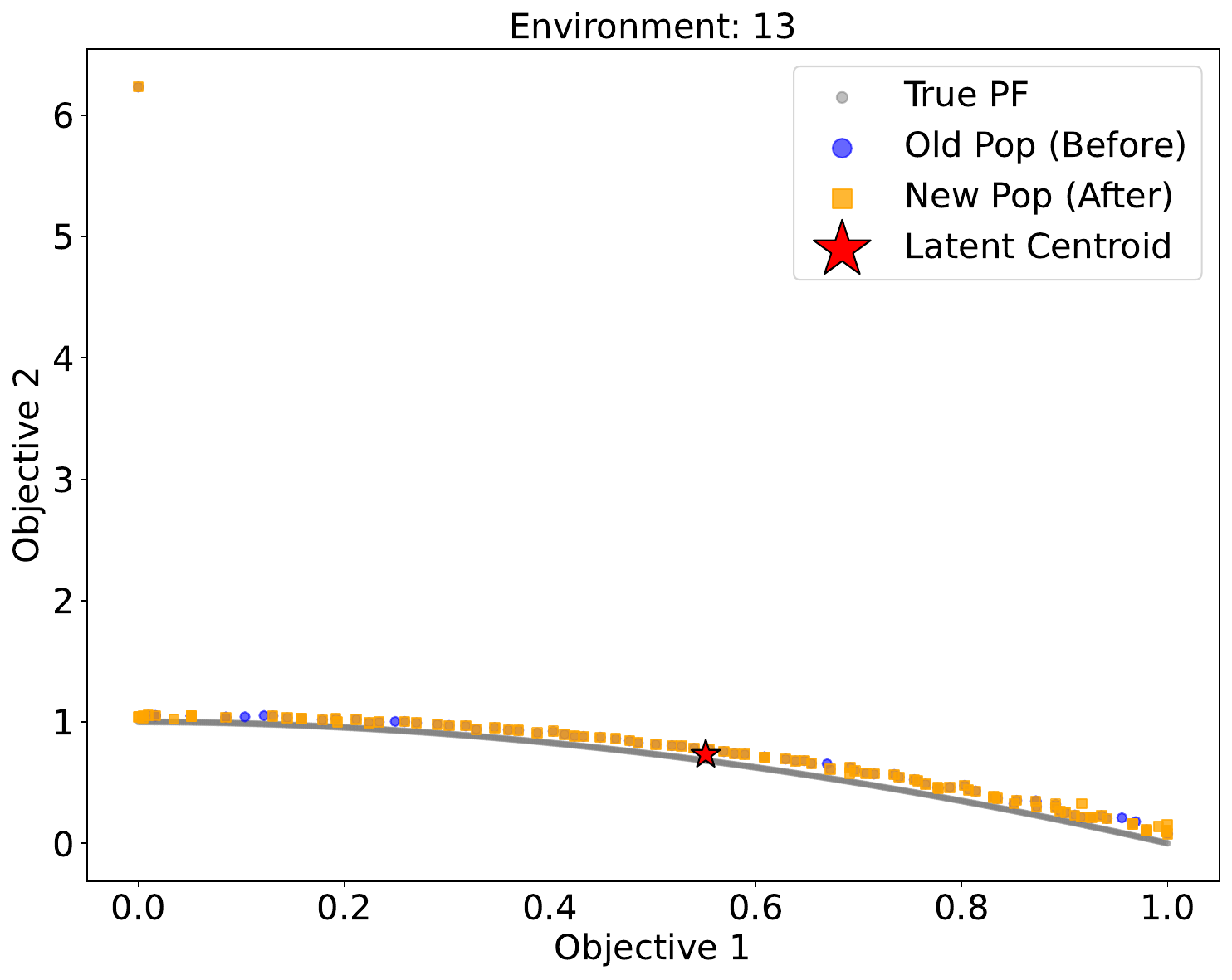}\hfill
    \includegraphics[width=0.19\linewidth]{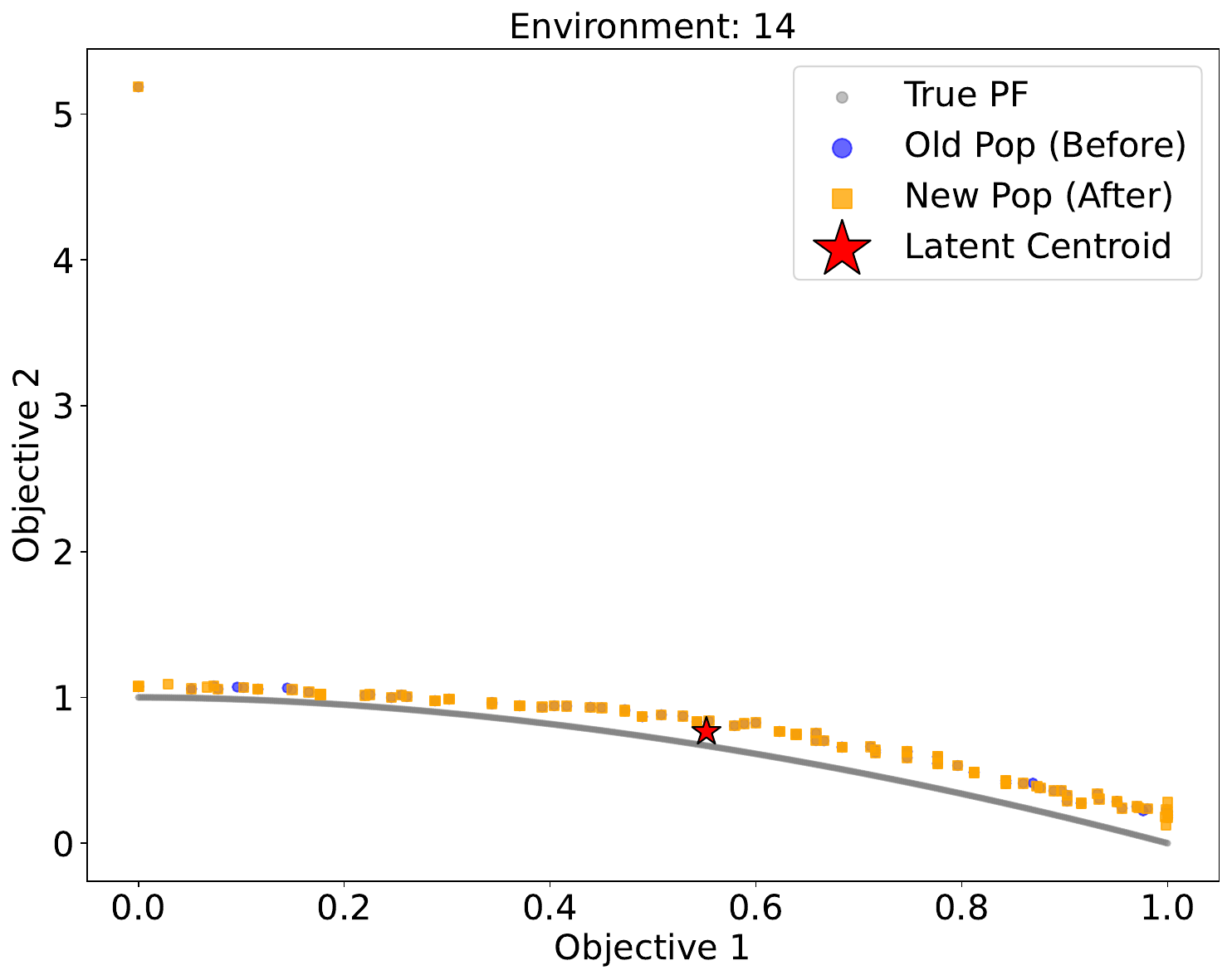}\hfill
    \includegraphics[width=0.19\linewidth]{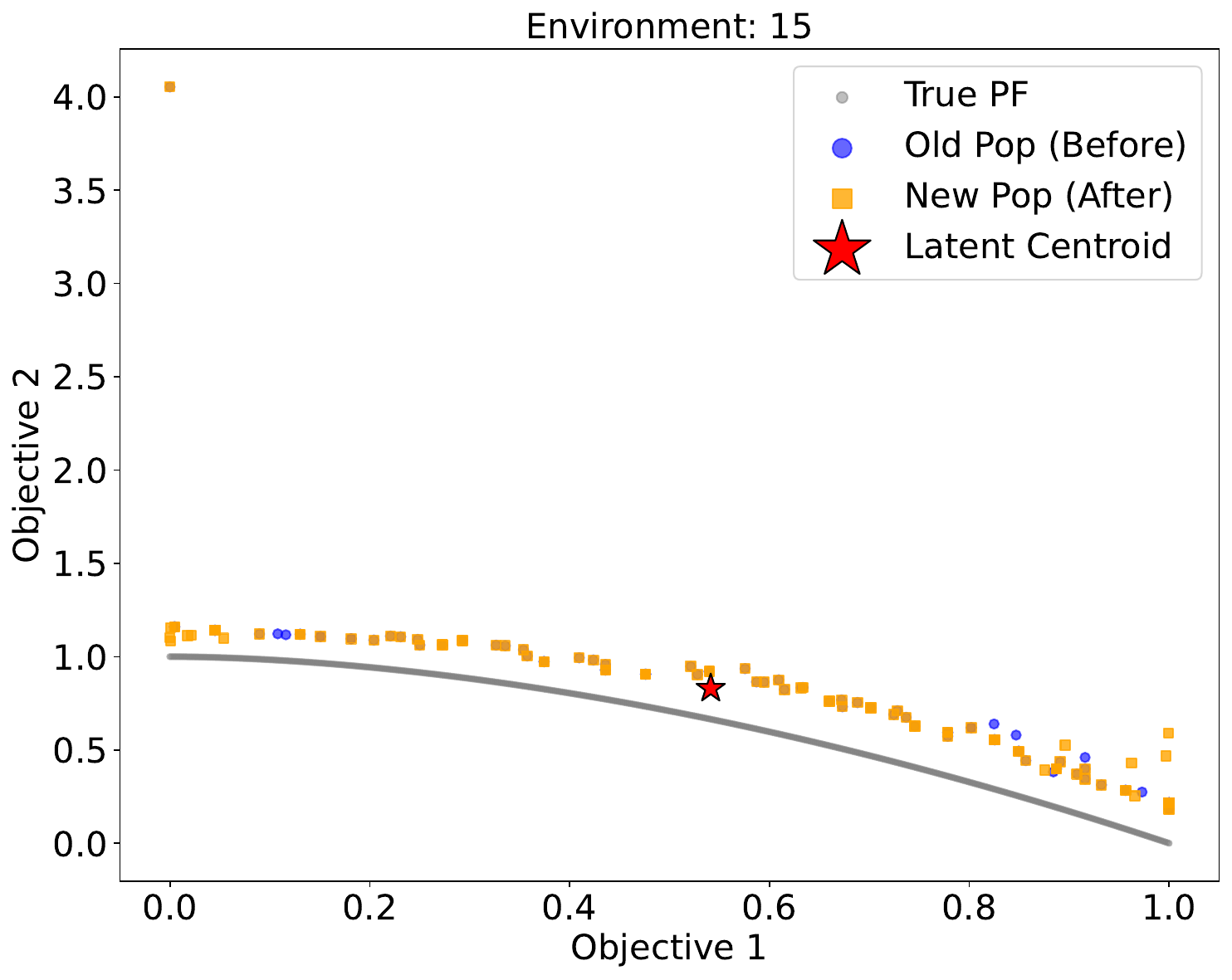}
    \\[4pt]
    \includegraphics[width=0.19\linewidth]{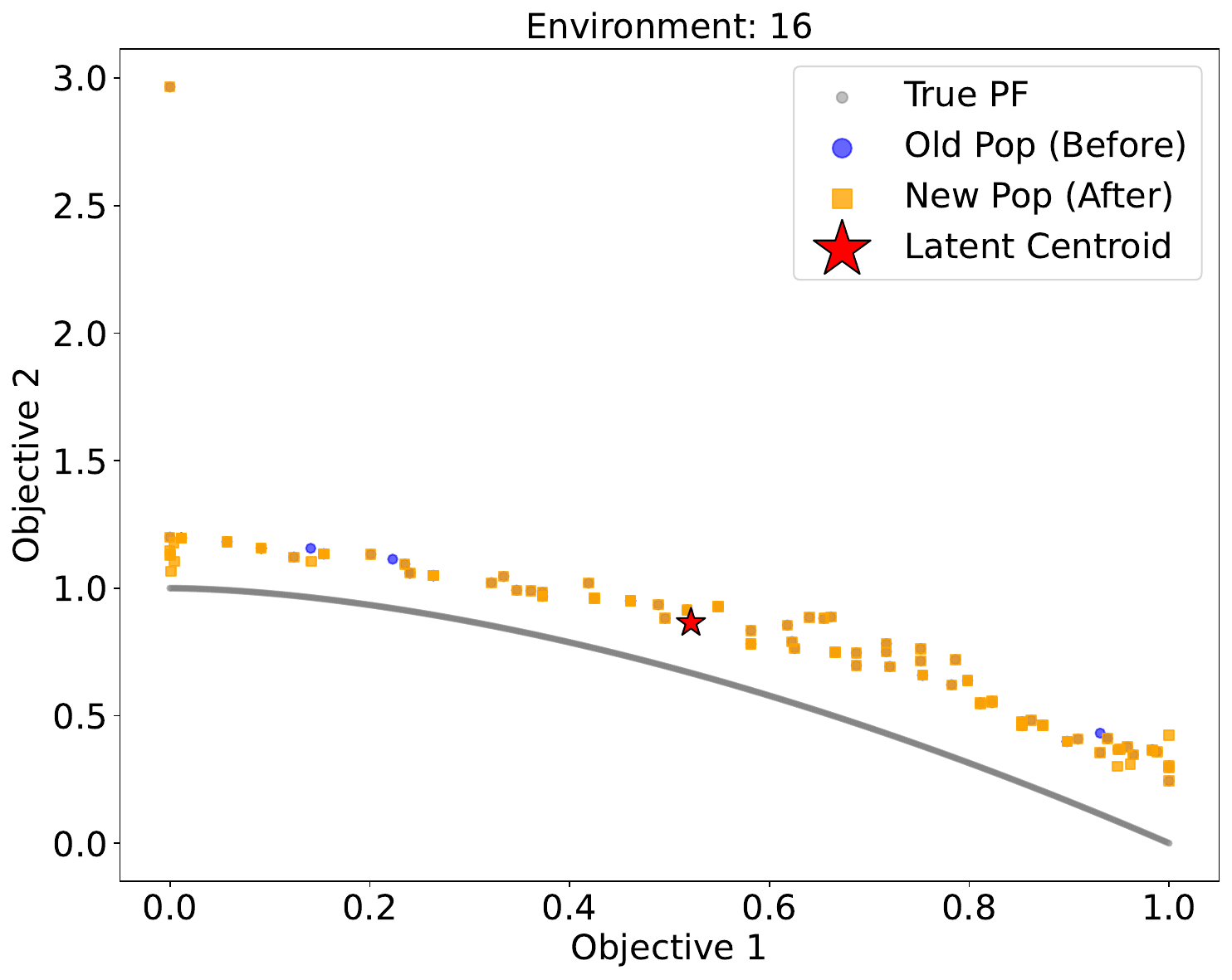}\hfill
    \includegraphics[width=0.19\linewidth]{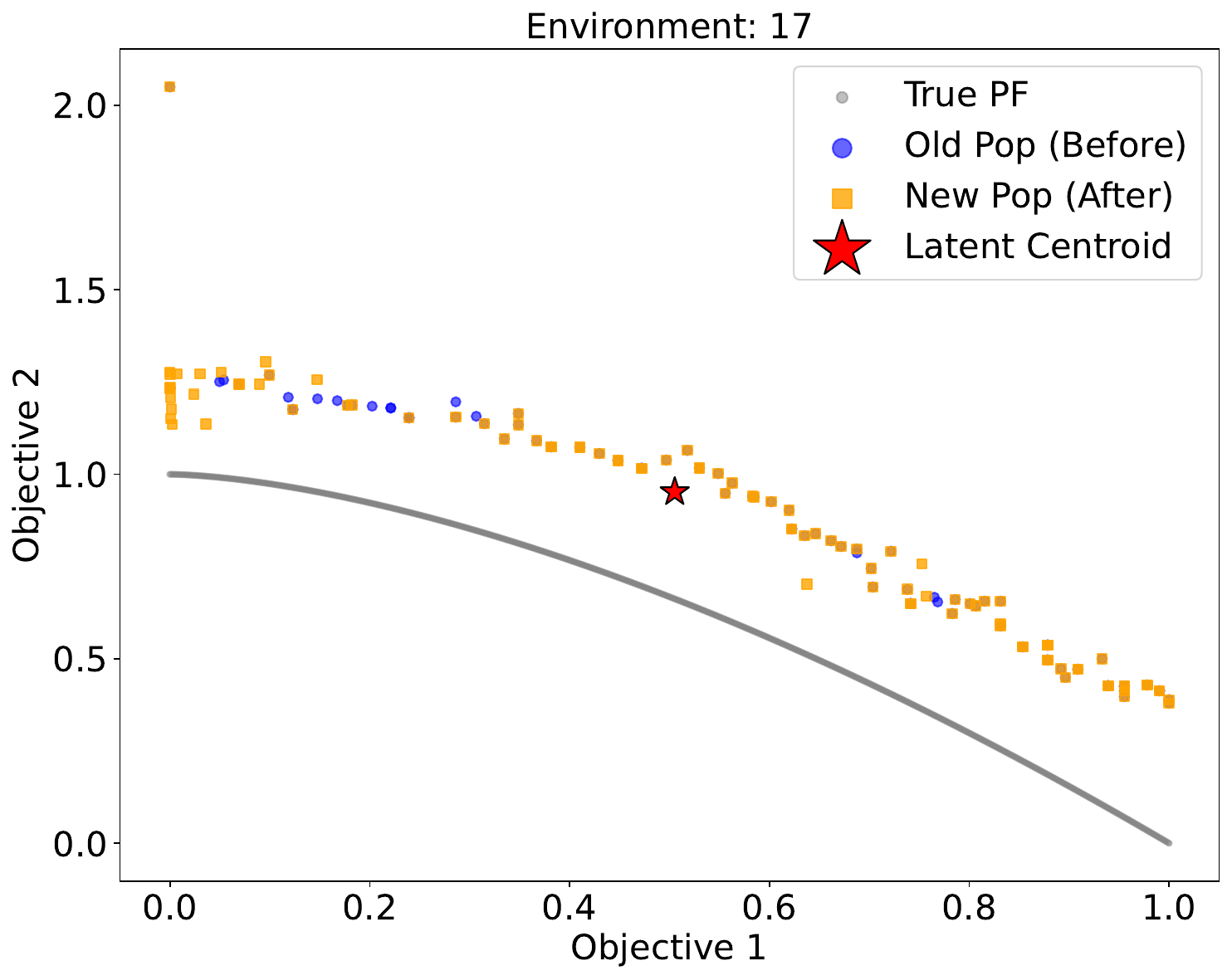}\hfill
    \includegraphics[width=0.19\linewidth]{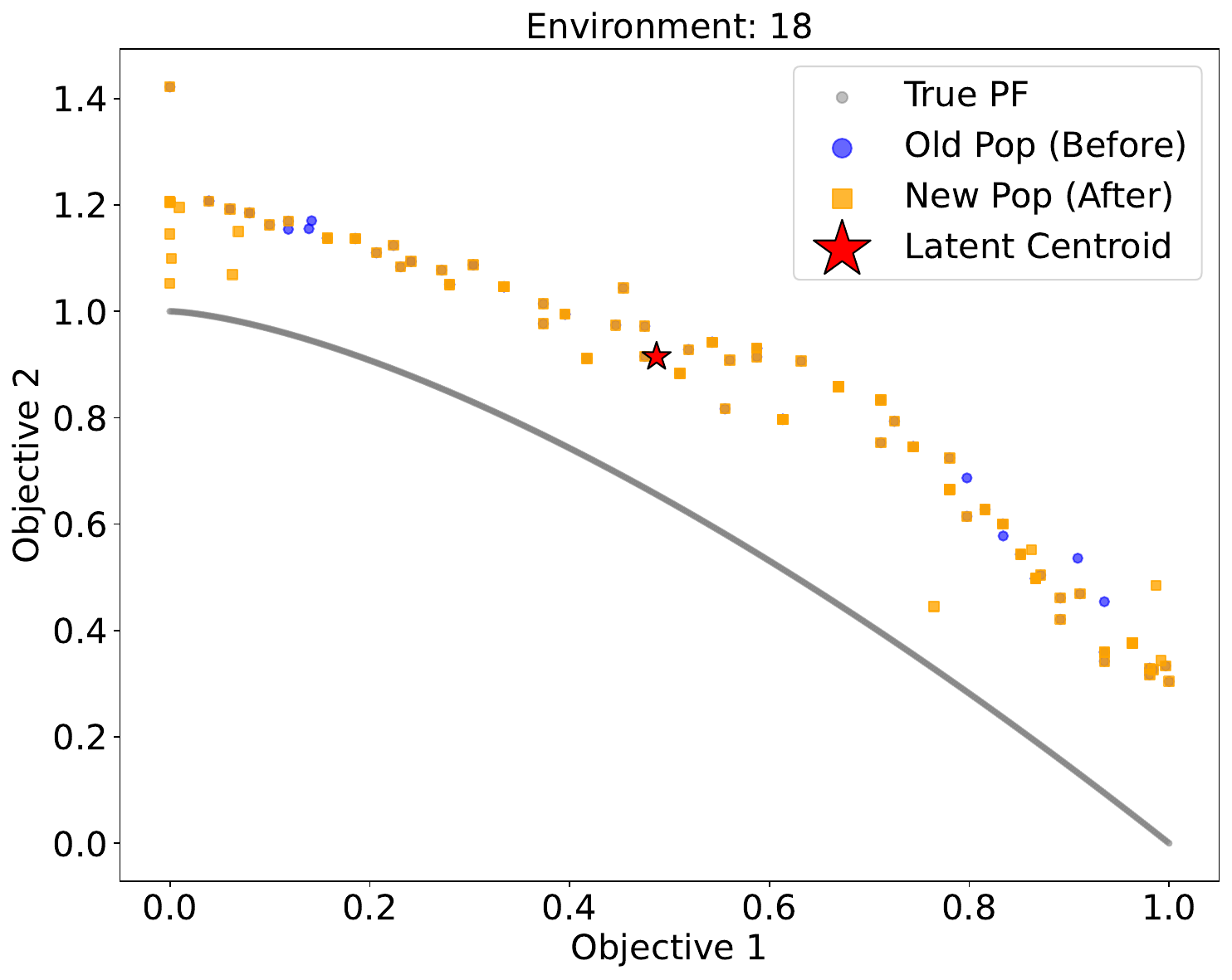}\hfill
    \includegraphics[width=0.19\linewidth]{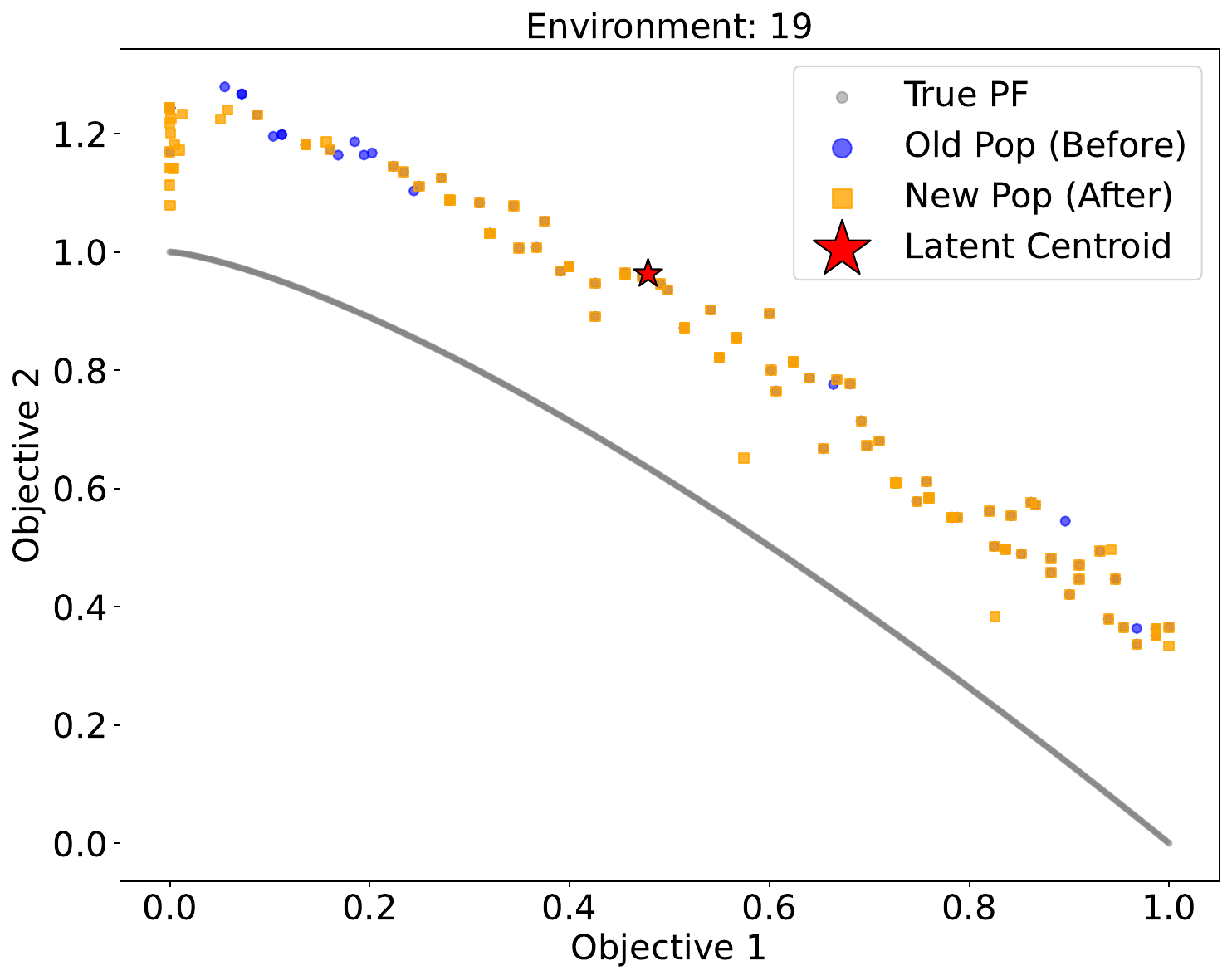}\hfill
    \includegraphics[width=0.19\linewidth]{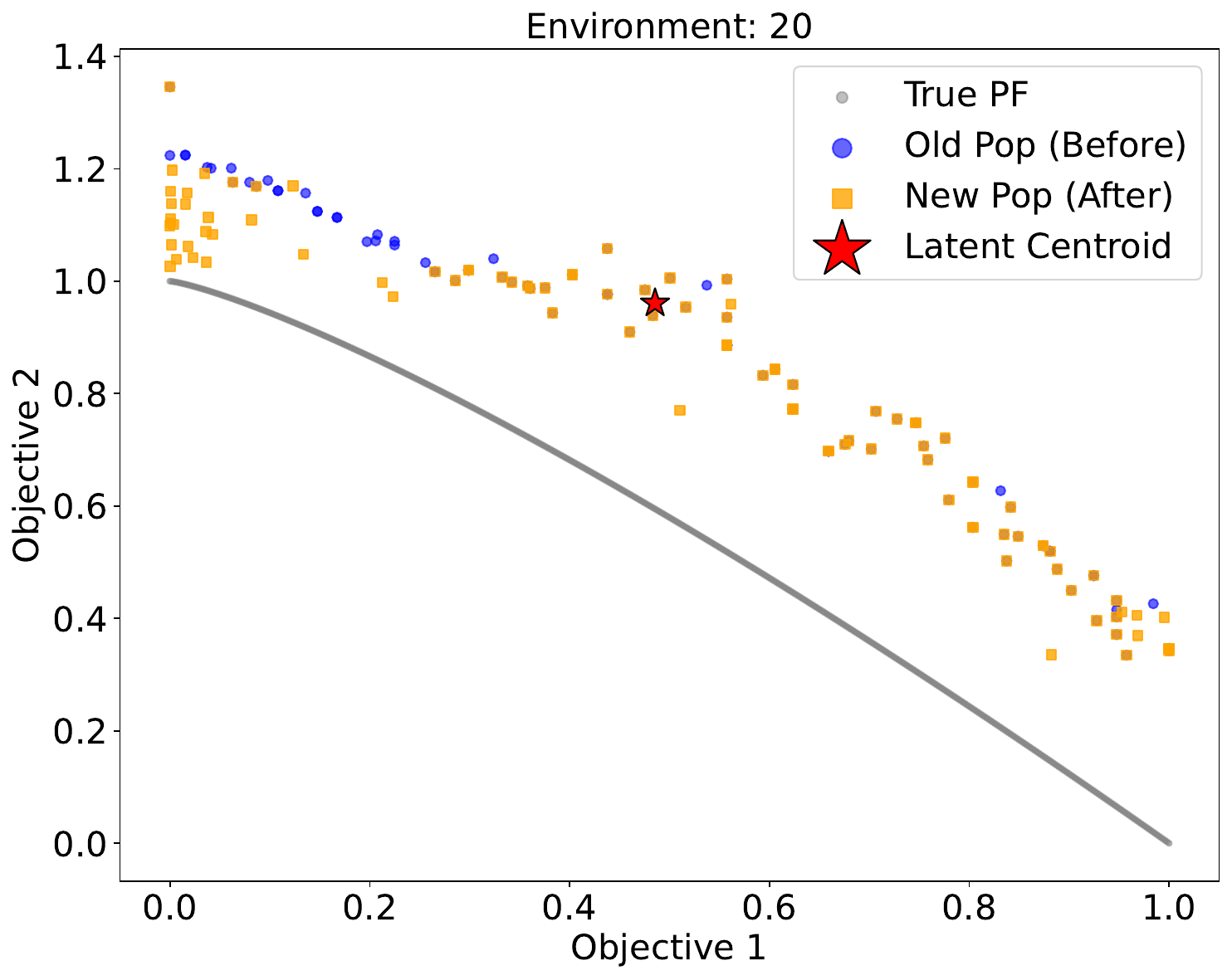}
    \\[4pt]
    \includegraphics[width=0.19\linewidth]{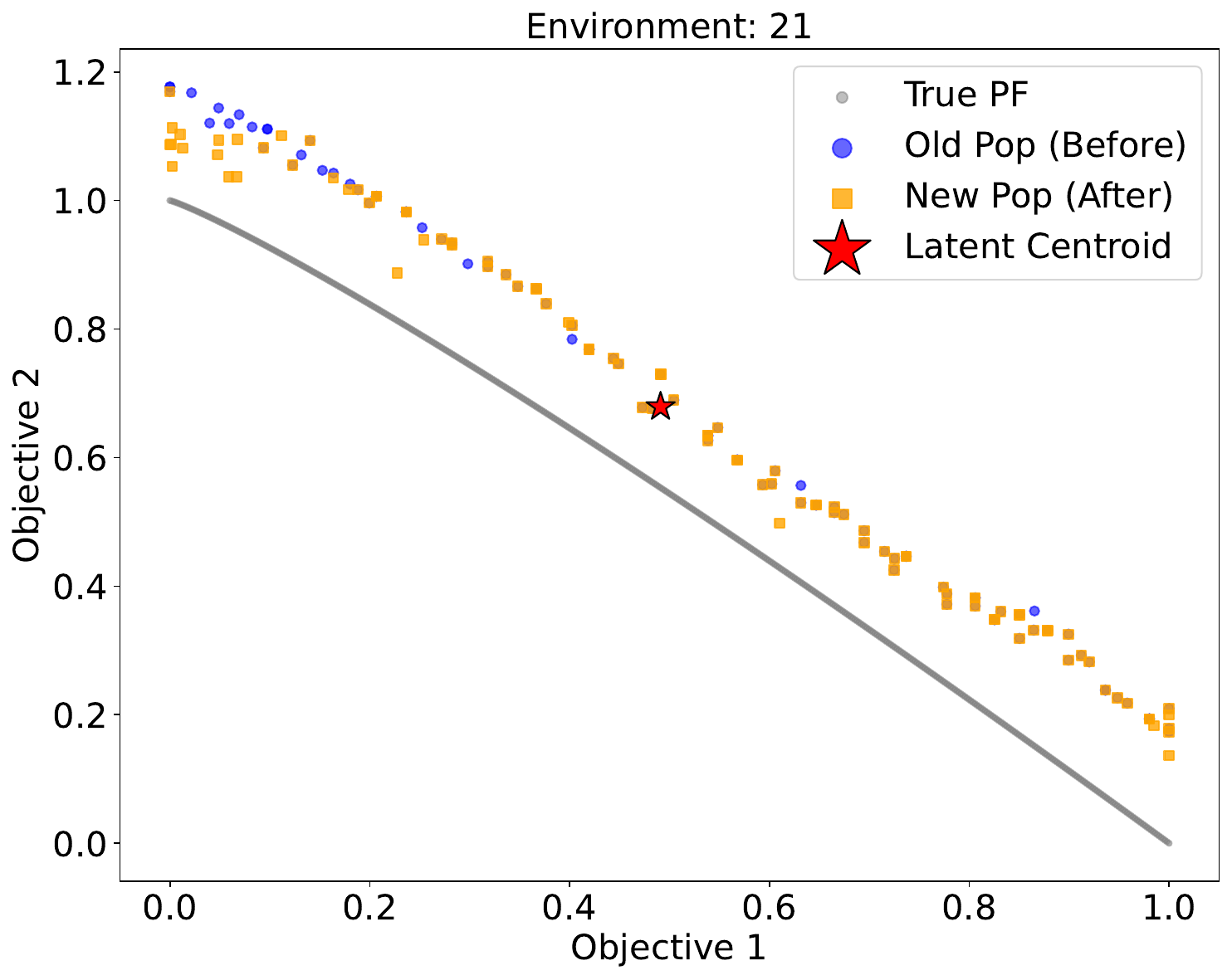}\hfill
    \includegraphics[width=0.19\linewidth]{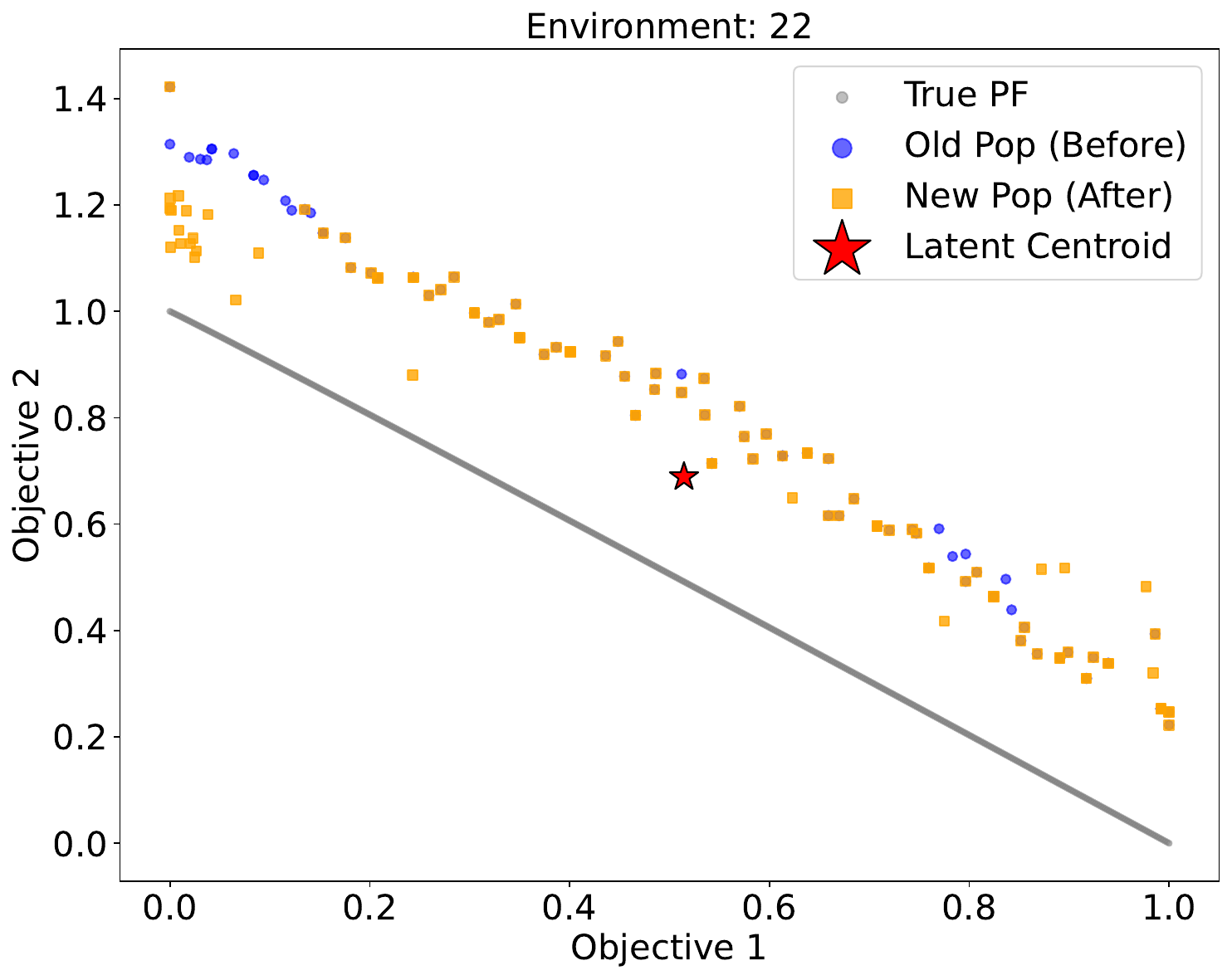}\hfill
    \includegraphics[width=0.19\linewidth]{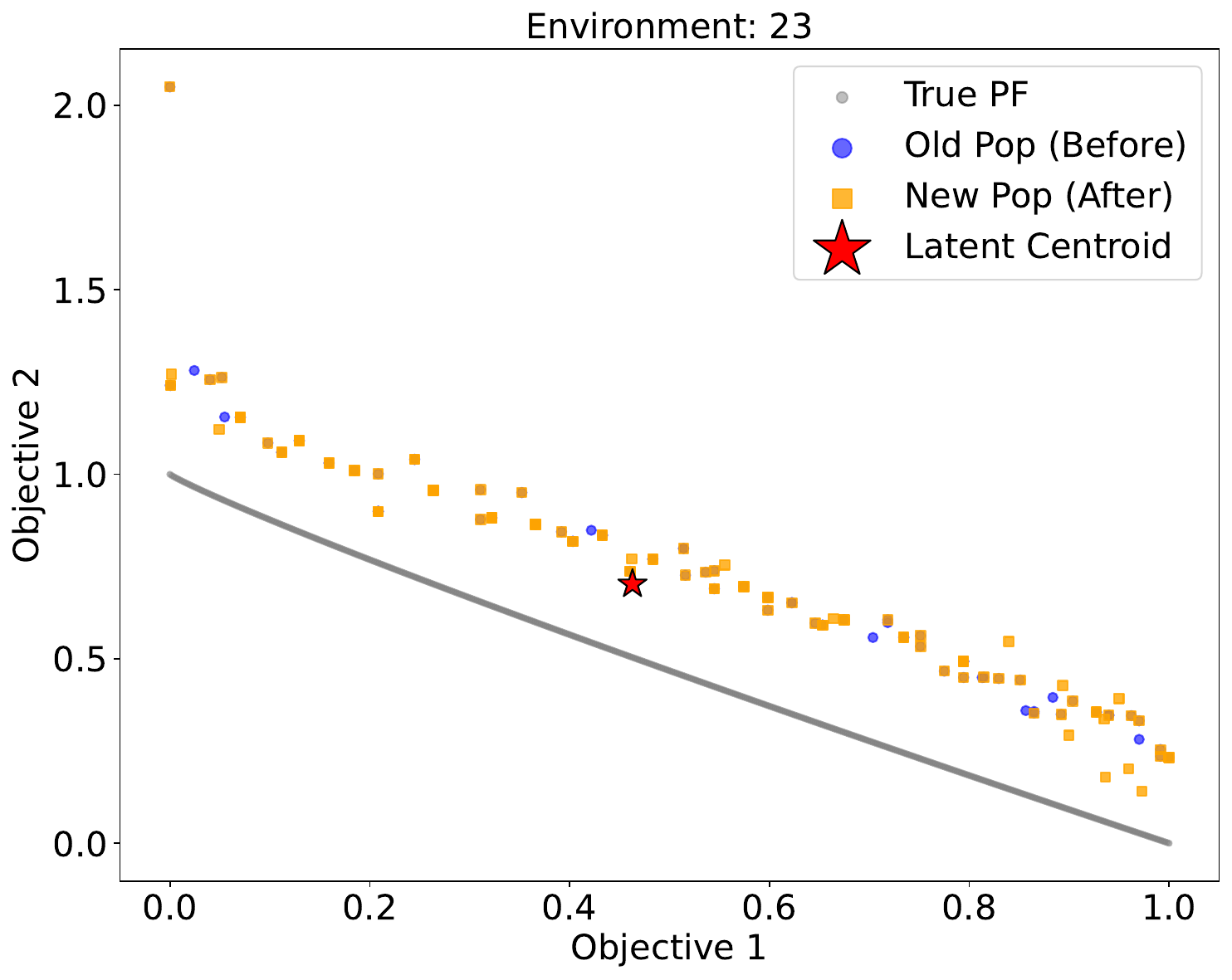}\hfill
    \includegraphics[width=0.19\linewidth]{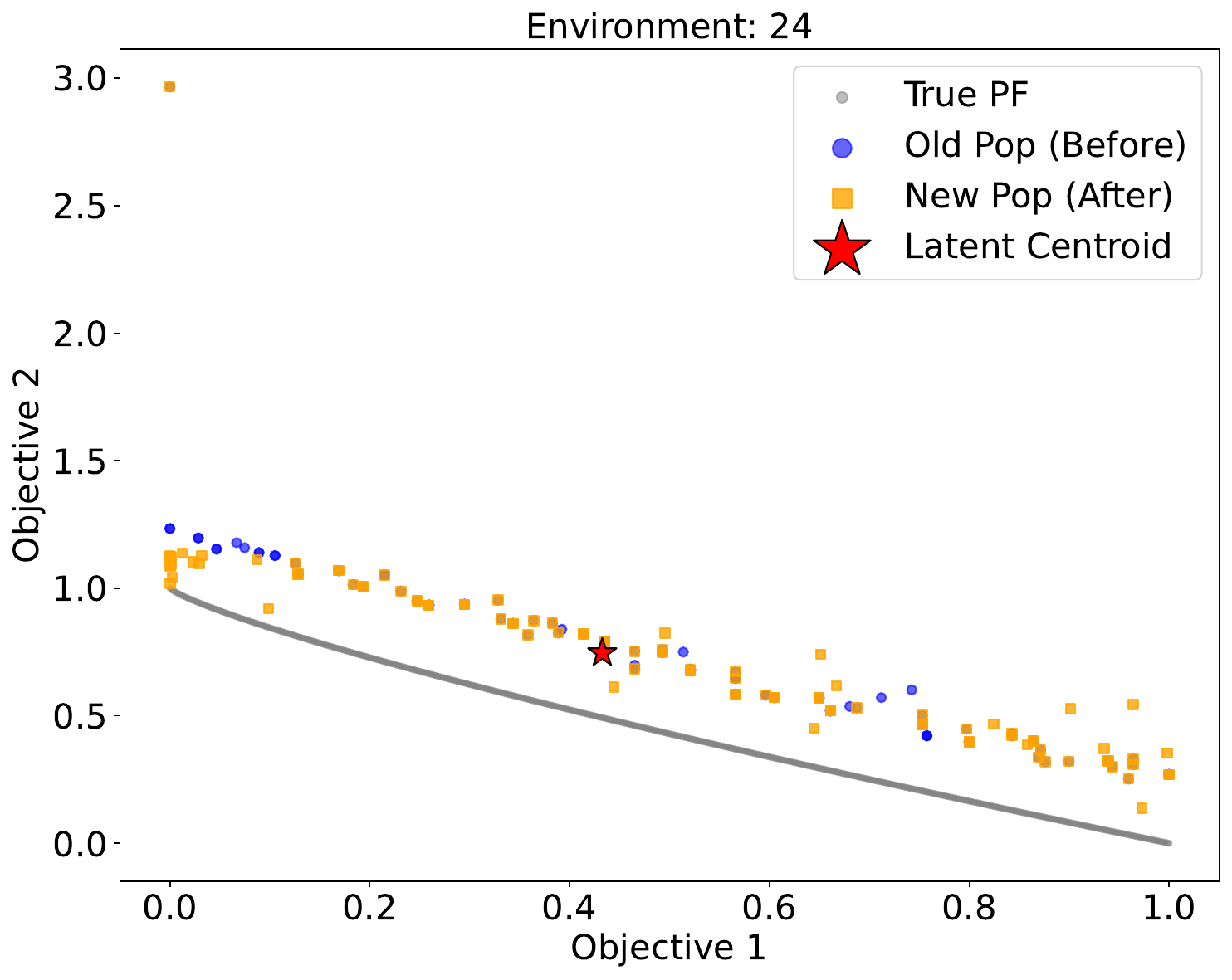}\hfill
    \includegraphics[width=0.19\linewidth]{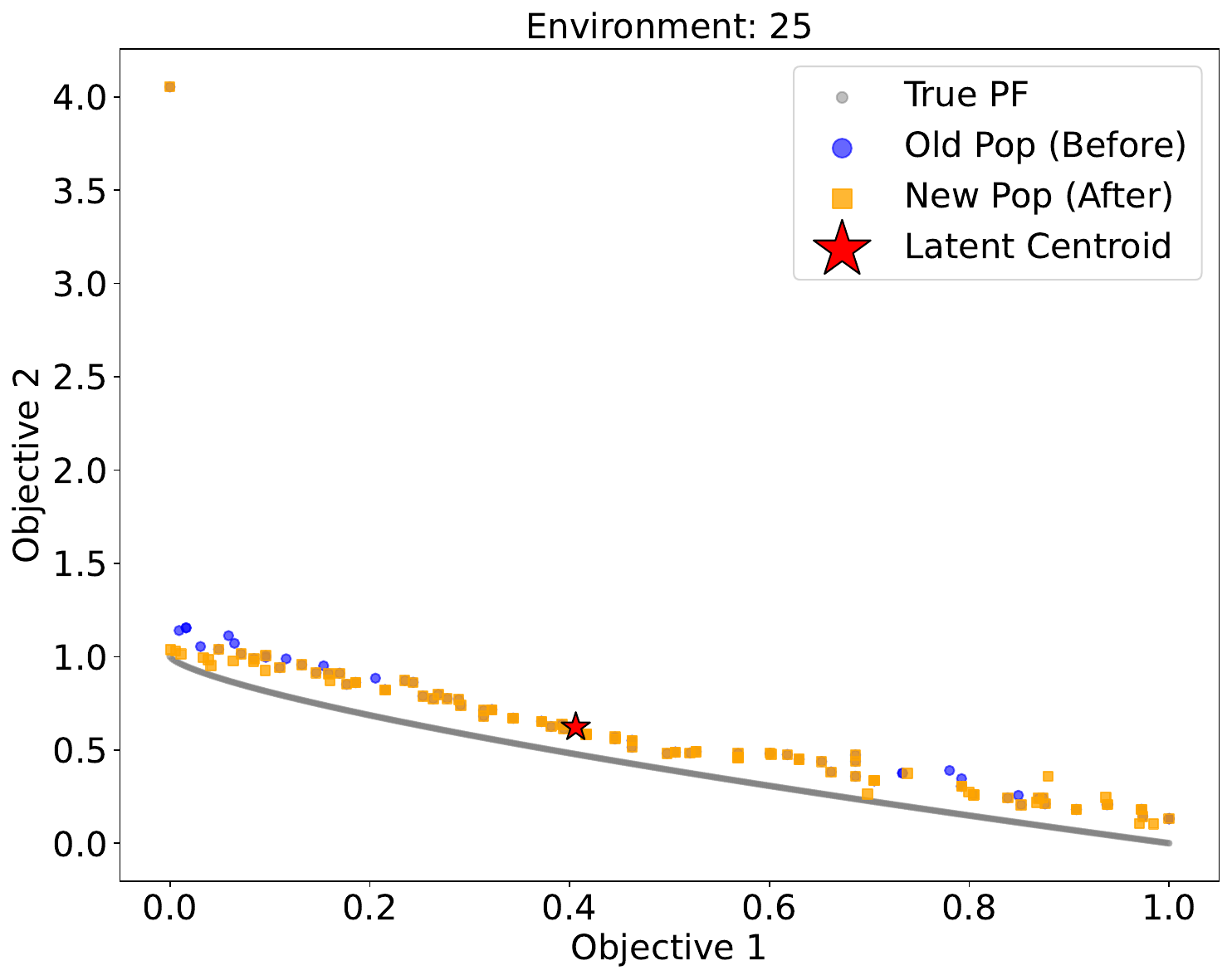}
    \\[4pt]
    \includegraphics[width=0.19\linewidth]{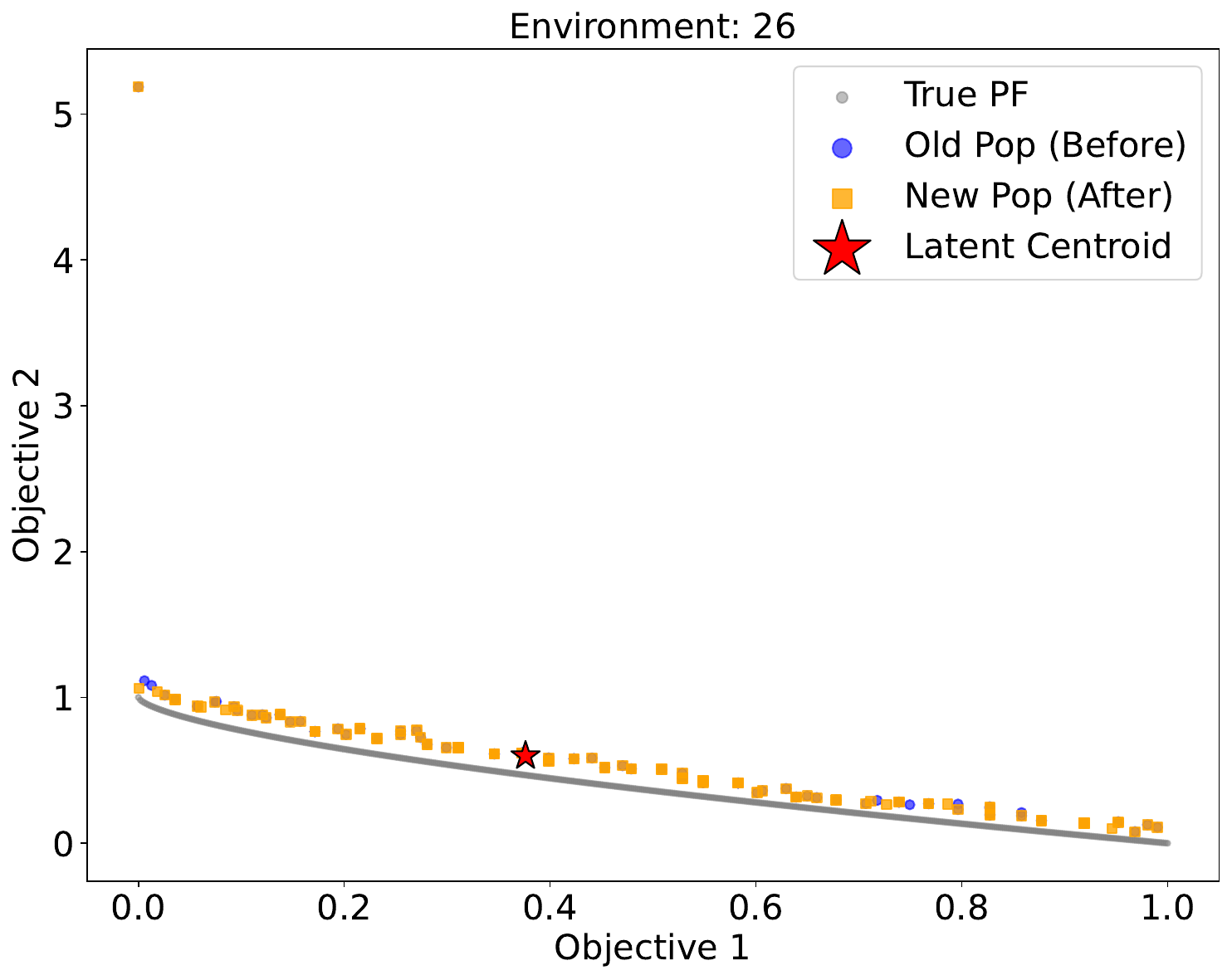}\hfill
    \includegraphics[width=0.19\linewidth]{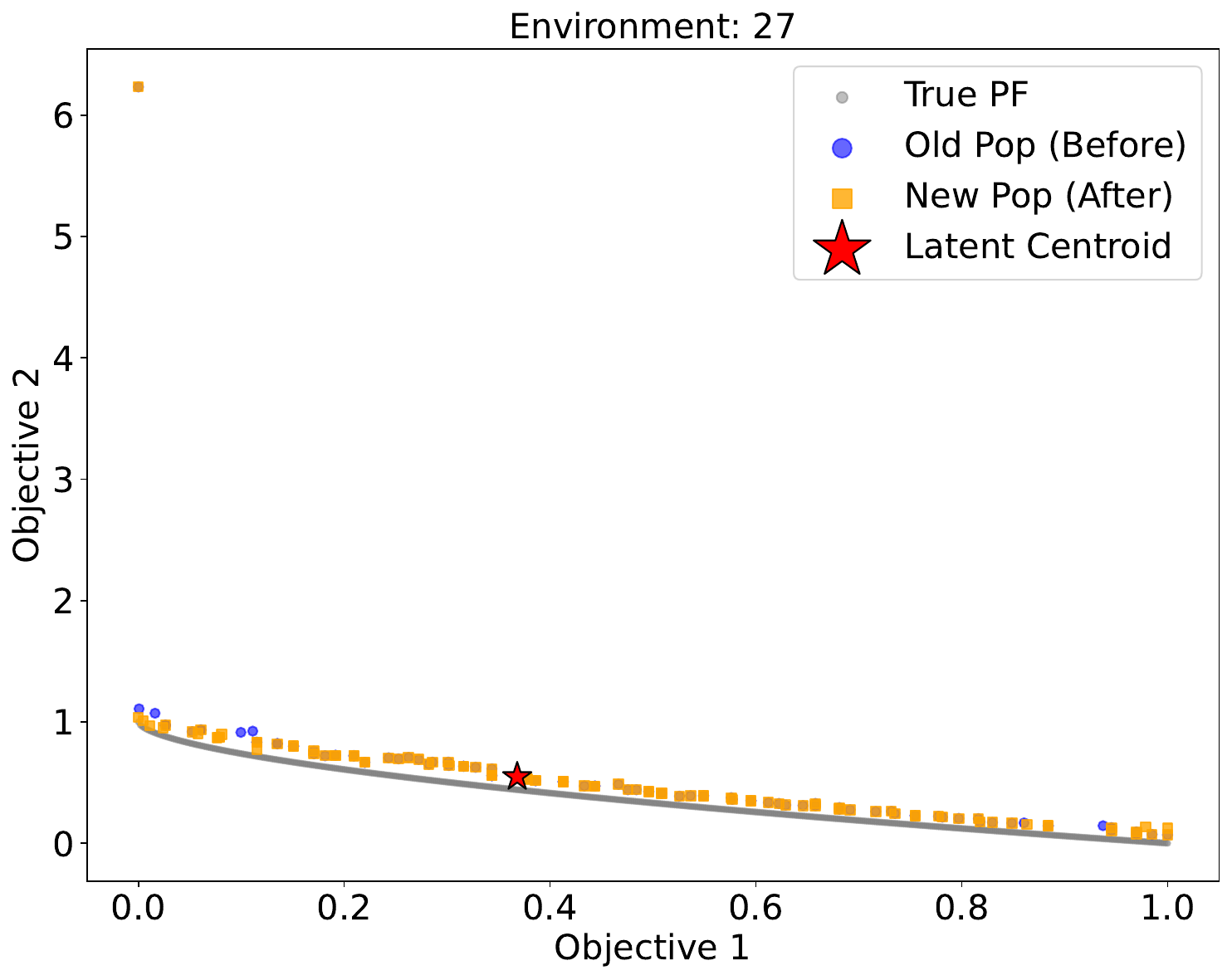}\hfill
    \includegraphics[width=0.19\linewidth]{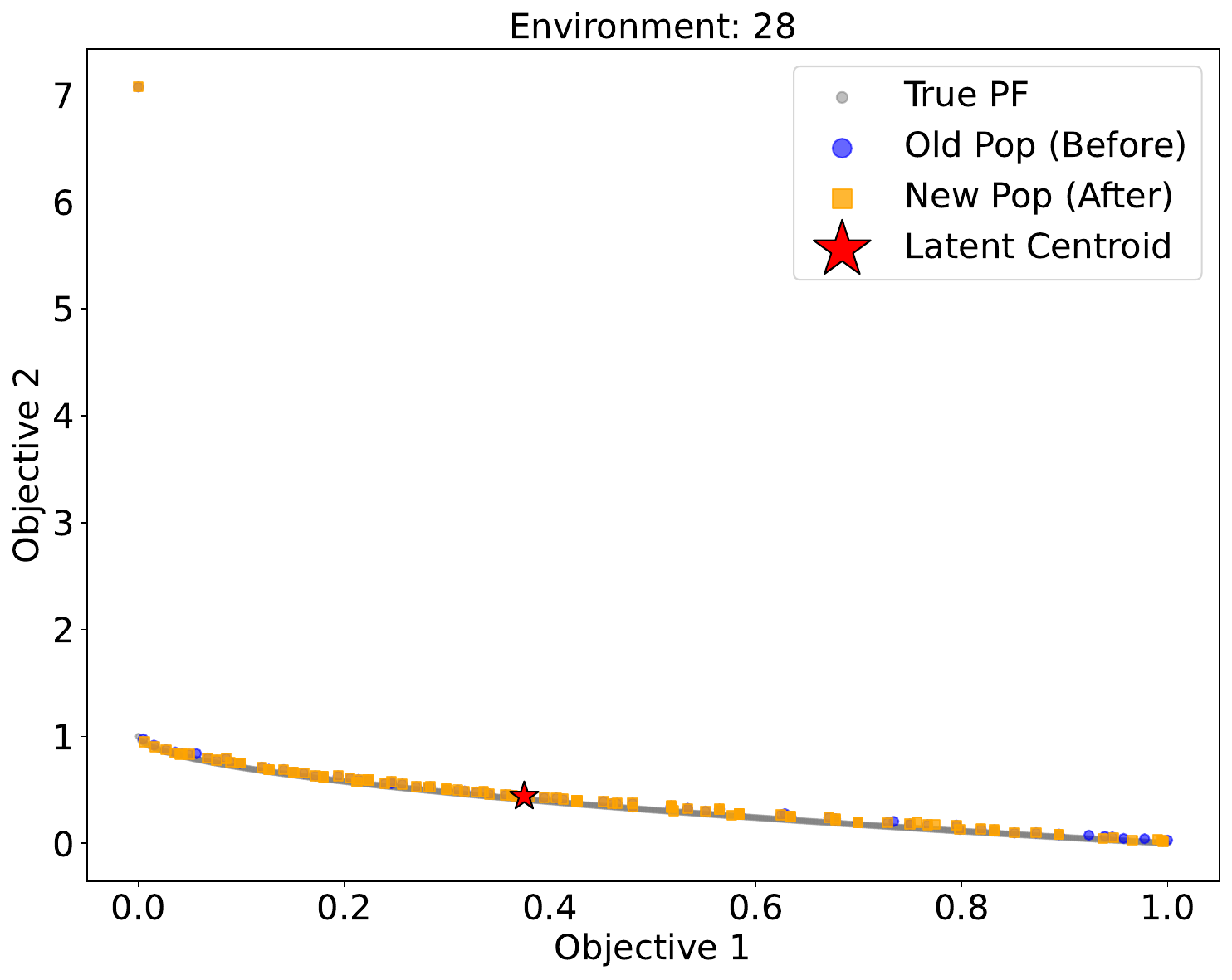}\hfill
    \includegraphics[width=0.19\linewidth]{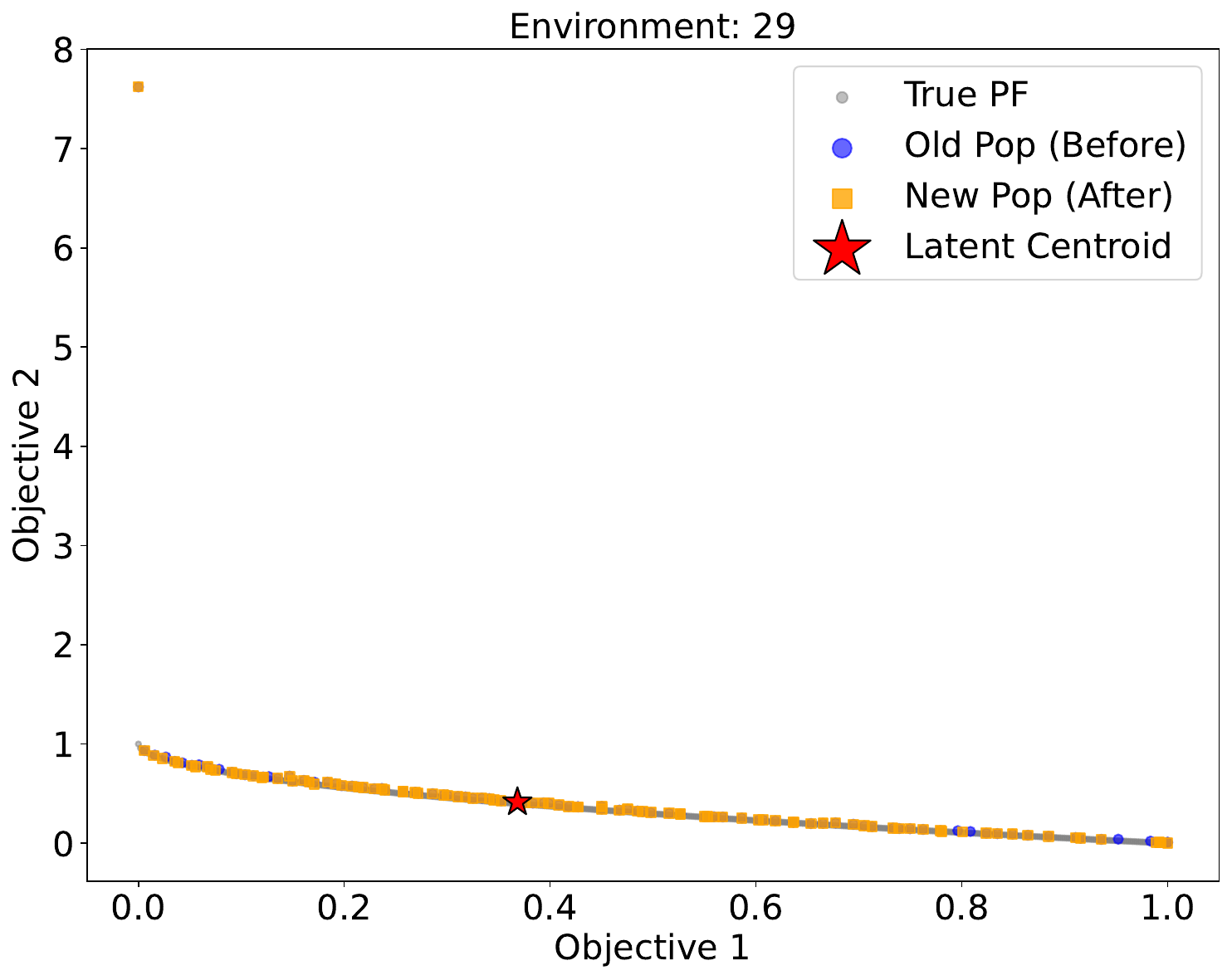}\hfill
    \includegraphics[width=0.19\linewidth]{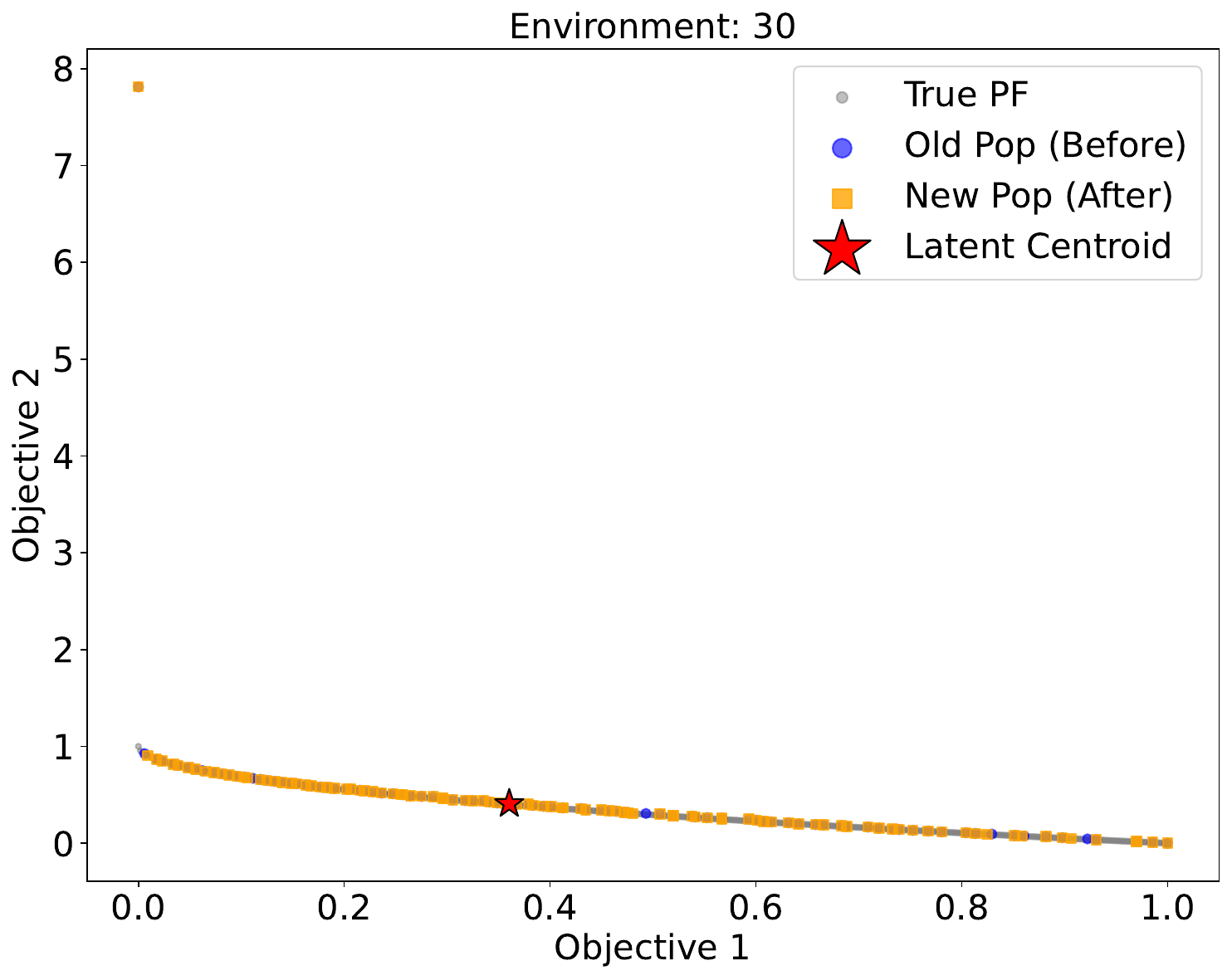}
    \\[2pt] 
    \caption{Complete tracking performance utilizing the alternative \textbf{all-point perturbation strategy} across all 30 environments on the DF1 problem.}
    \label{fig:app_df1_tracking_all}
\end{figure*}

Conversely, Fig.~\ref{fig:app_df1_tracking_all} illustrates the results of the alternative \textbf{all-point perturbation strategy}, where noise is applied directly to the latent representations of all individual historical solutions. Because the inherited population is already misaligned with the new environment, adding individual noise propagates these positional errors. The visualizations demonstrate that the resulting candidates frequently scatter around sub-optimal regions, inheriting the historical inertia of the previous distribution and failing to accurately cover the new Pareto Front.

\subsection{Complete IGD Trajectories for All Test Instances}
\label{app:full_igd}

\begin{figure}[htbp]
    \centering
    \includegraphics[width=0.45\linewidth]{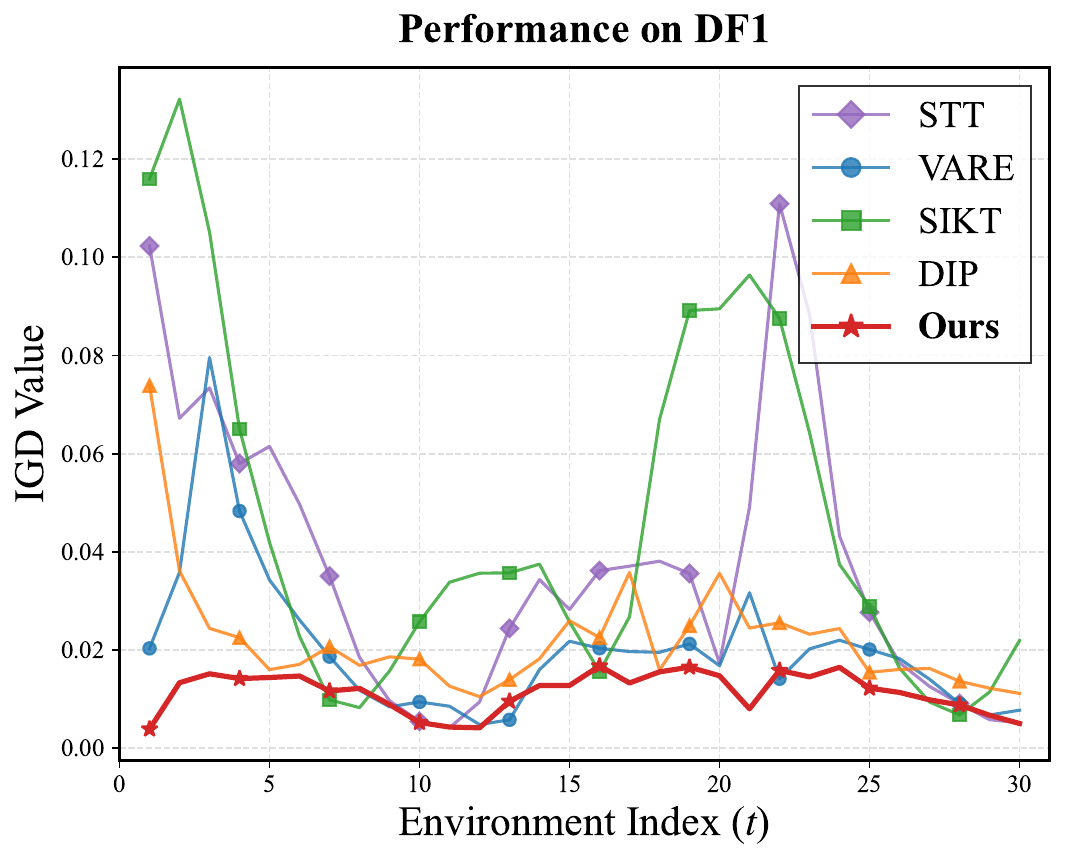}\hfill
    \includegraphics[width=0.45\linewidth]{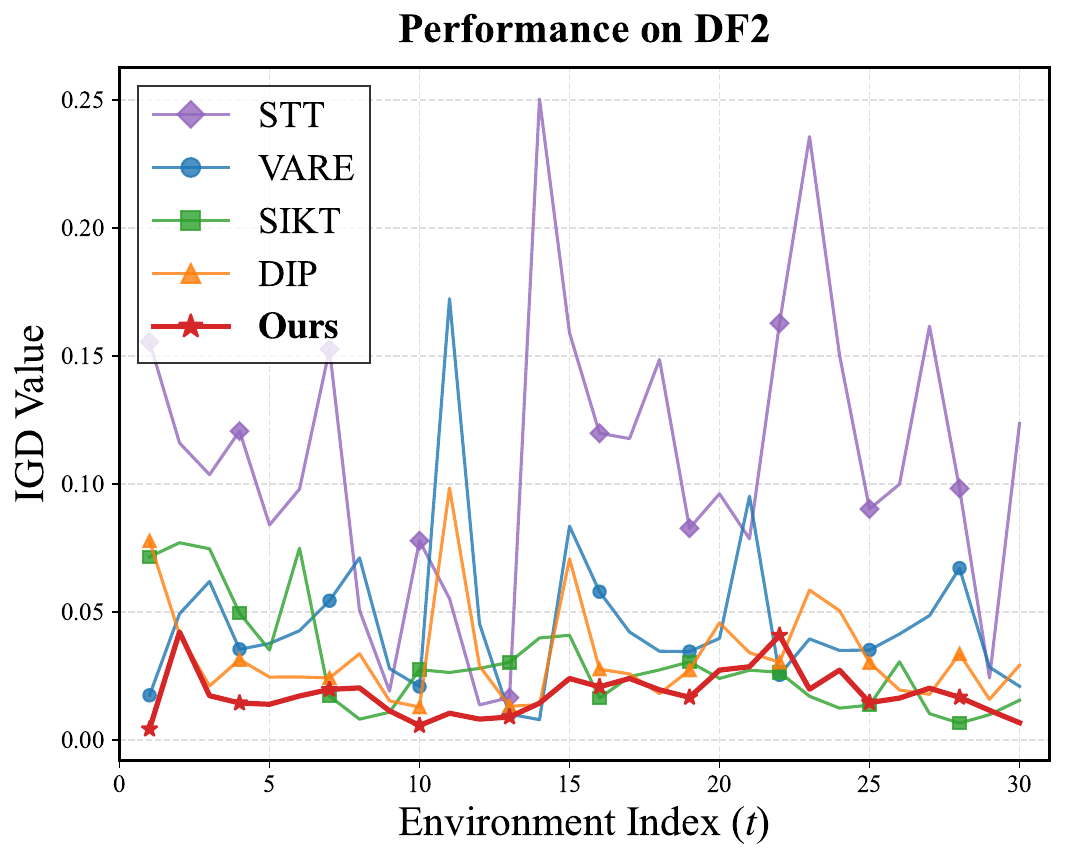}
    \\[4pt] 
    
    \includegraphics[width=0.45\linewidth]{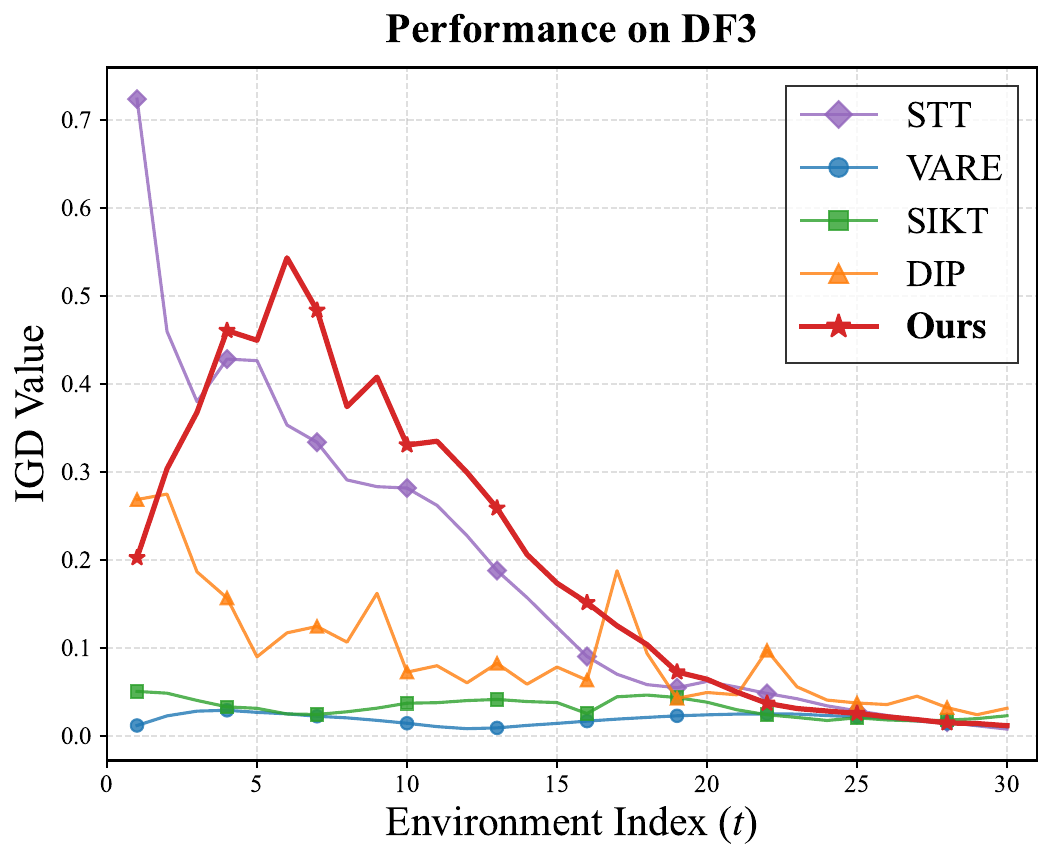}\hfill
    \includegraphics[width=0.45\linewidth]{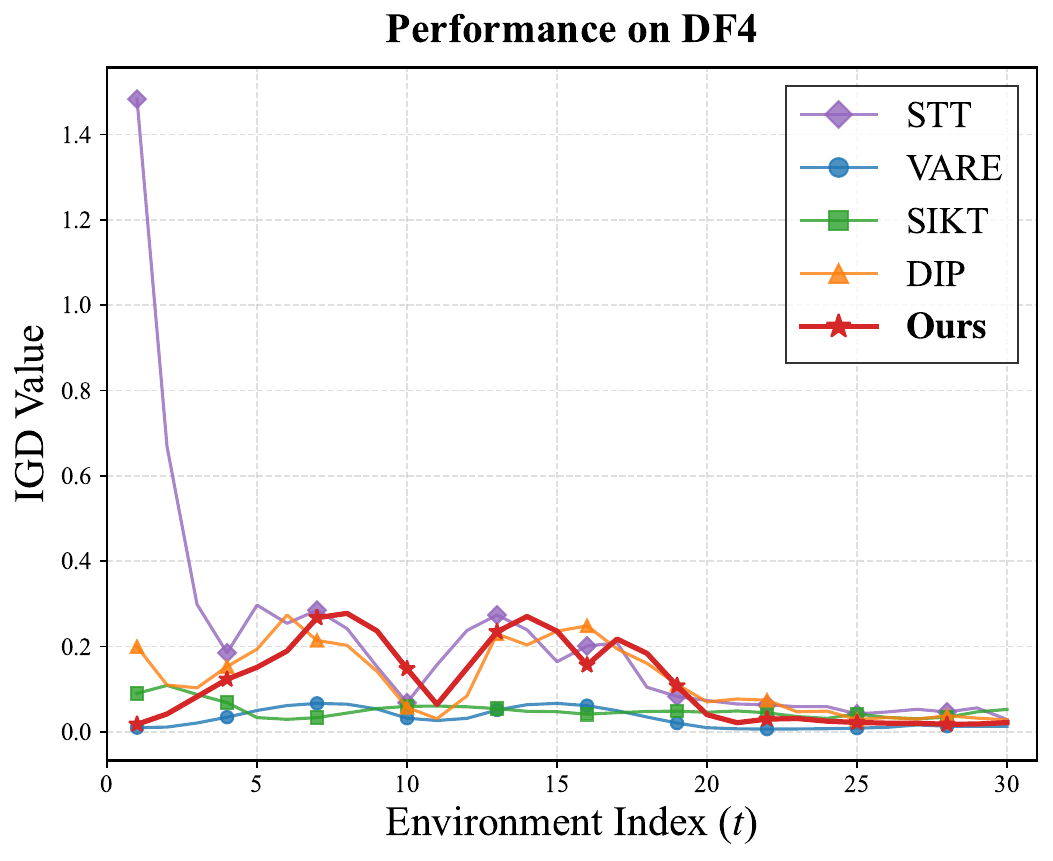}
    \\[4pt] 
    
    \includegraphics[width=0.45\linewidth]{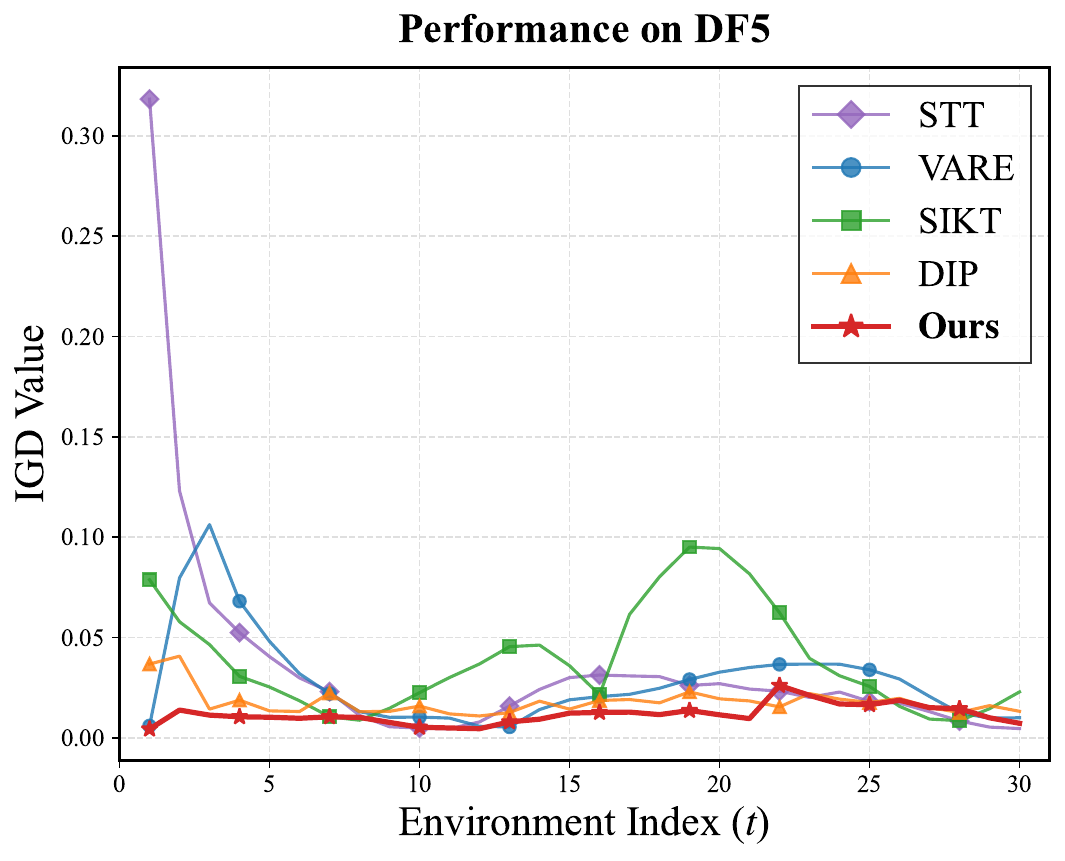}\hfill
    \includegraphics[width=0.45\linewidth]{figures/IGD_curve/DF6_IGD_Curve.pdf}
    \\[4pt] 
    
    \includegraphics[width=0.45\linewidth]{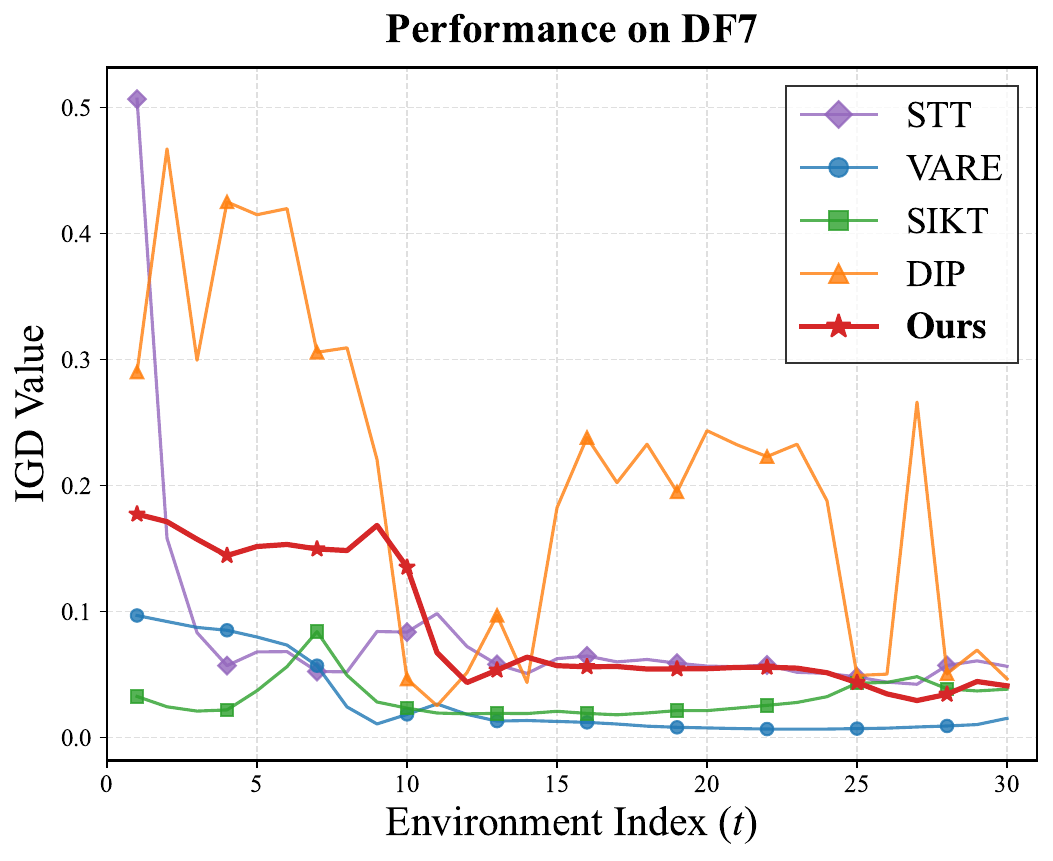}\hfill
    \includegraphics[width=0.45\linewidth]{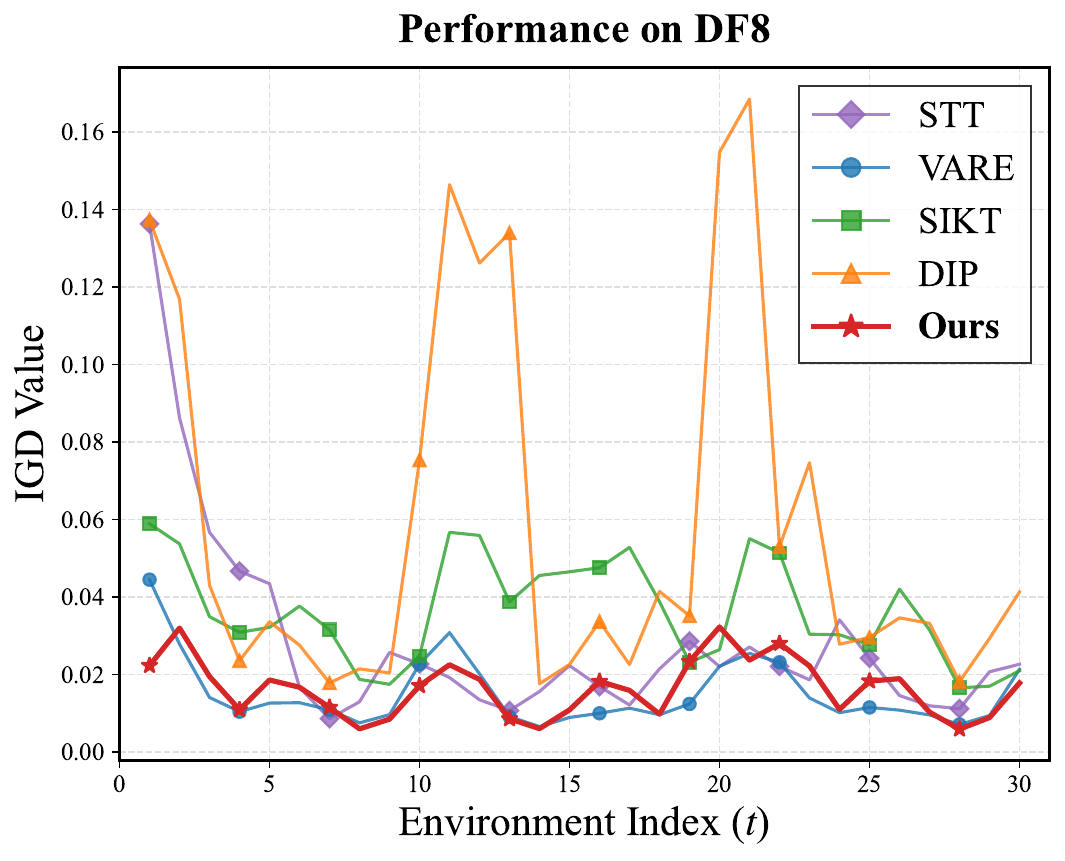}
    \\[4pt] 
    
    \includegraphics[width=0.45\linewidth]{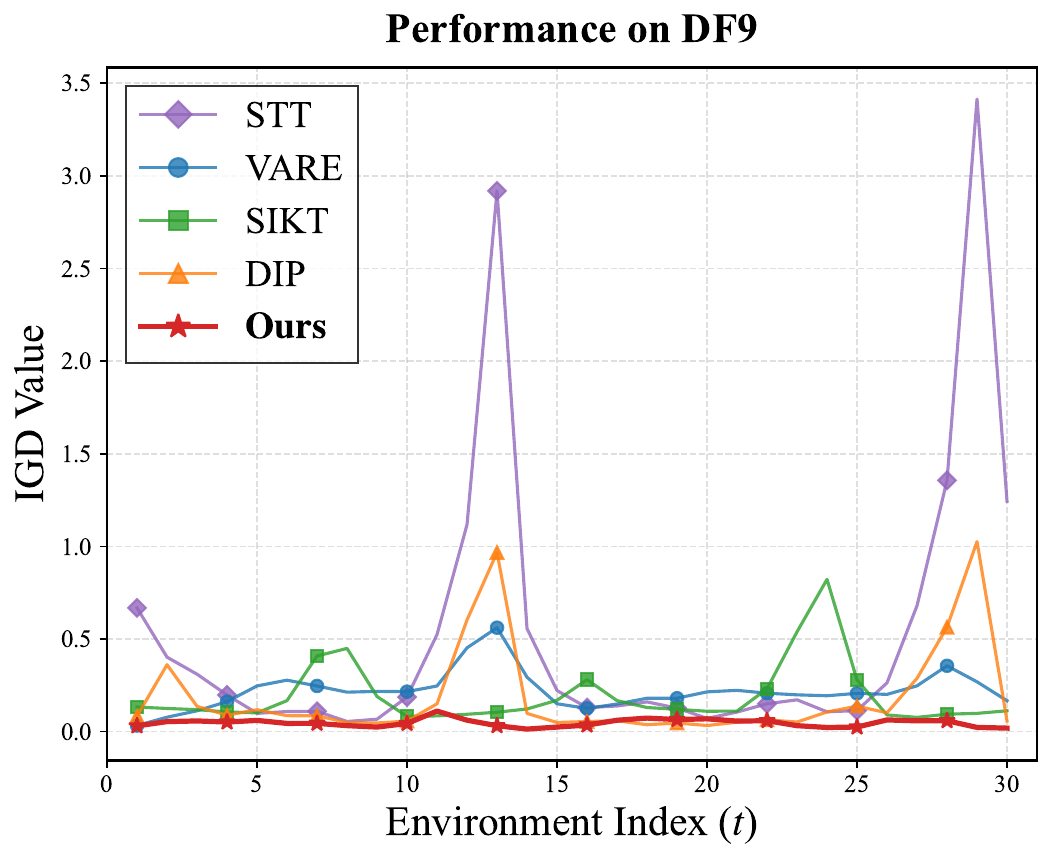}\hfill
    \includegraphics[width=0.45\linewidth]{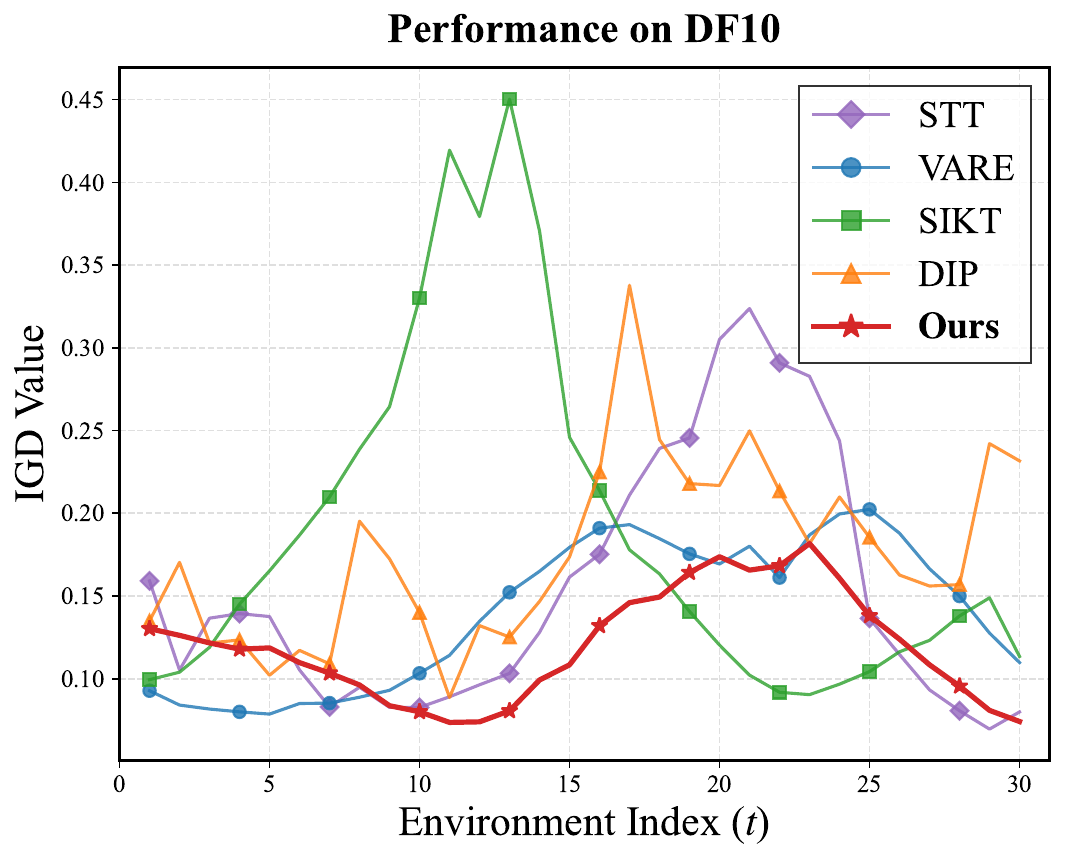}
    \\[4pt] 
    
    \includegraphics[width=0.45\linewidth]{figures/IGD_curve/DF11_IGD_Curve.pdf}\hfill
    \includegraphics[width=0.45\linewidth]{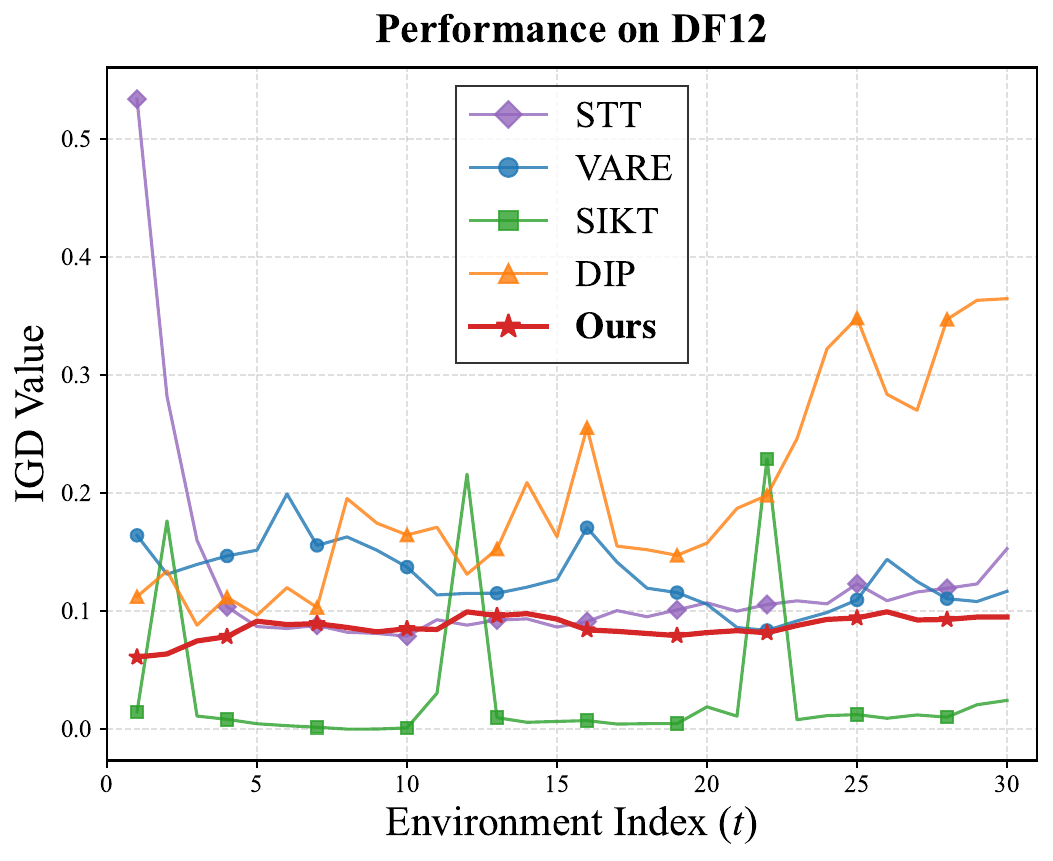}
    \\[4pt] 
    
    \includegraphics[width=0.45\linewidth]{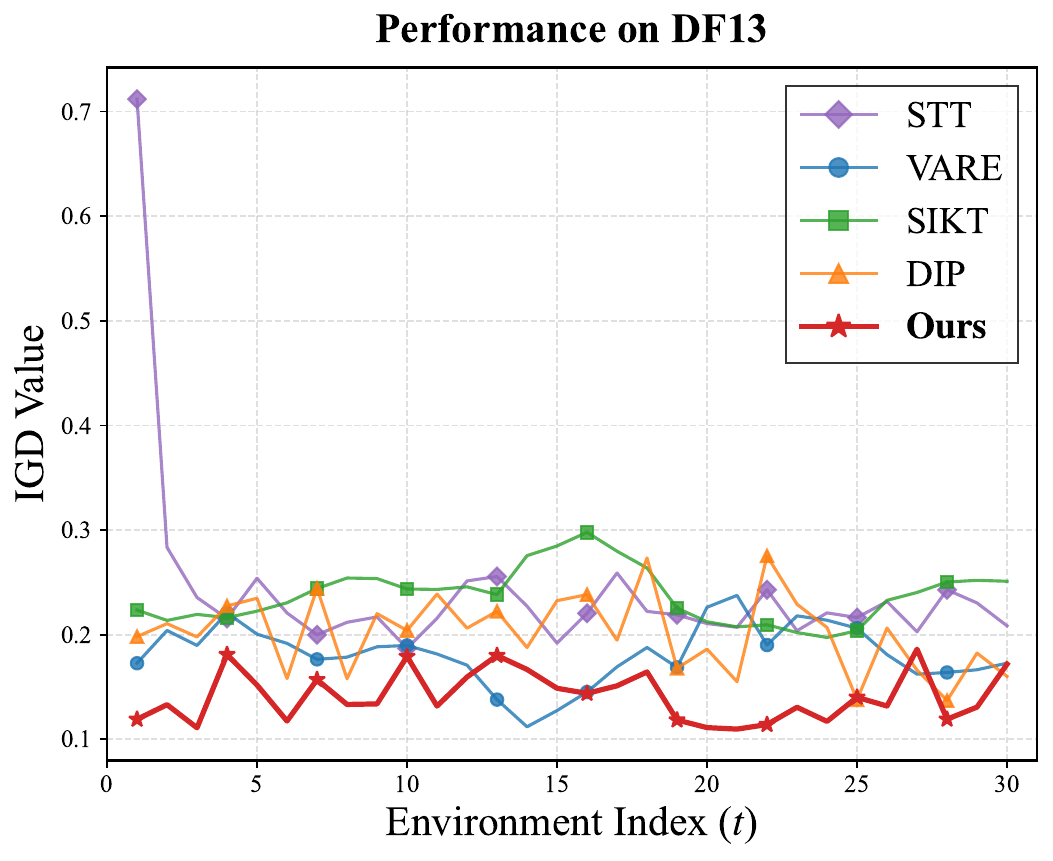}\hfill
    \includegraphics[width=0.45\linewidth]{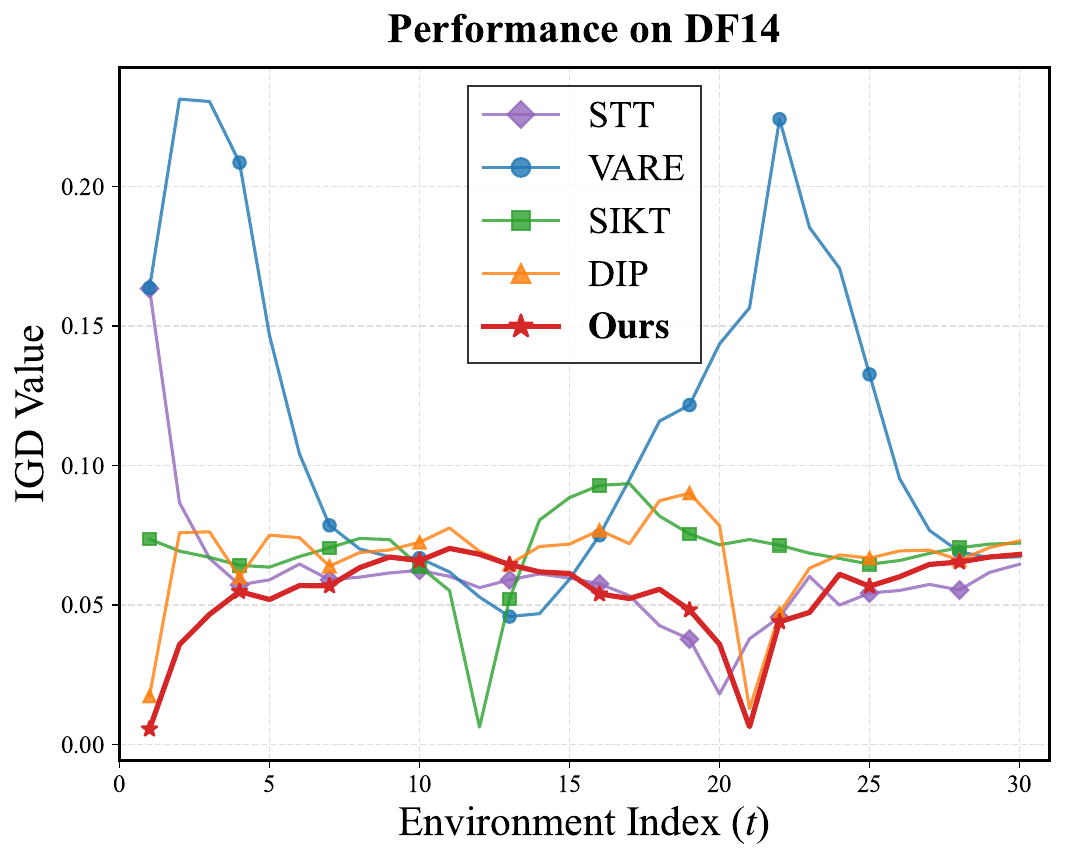}
    
    \caption{Complete IGD variation curves of different algorithms over consecutive environments on all DF test instances.}
    \label{fig:full_igd_curve}
\end{figure}

To provide a comprehensive and transparent view of the evolutionary stability, Fig.~\ref{fig:full_igd_curve} presents the complete IGD trajectories for all 14 test instances (DF1 to DF14) across sequential environmental changes.

As shown in the complete results, DB-GEN generally maintains a stable and low IGD profile across the majority of the benchmark problems, demonstrating its overall robustness to diverse dynamic patterns. However, it is also objectively observable that on certain specific instances (e.g., DF4), the proposed method yields sub-optimal tracking performance or experiences slight fluctuations compared to the best-performing baseline methods. This comprehensive visualization highlights both the general adaptability of the proposed framework and its occasional limitations when encountering specific topological structures.

\subsection{Details of Basis Vector Correlation and Manifold Topology}
\label{apx:basis_vector_details}

\begin{figure}[htbp]
    \centering
    \includegraphics[width=0.65\linewidth]{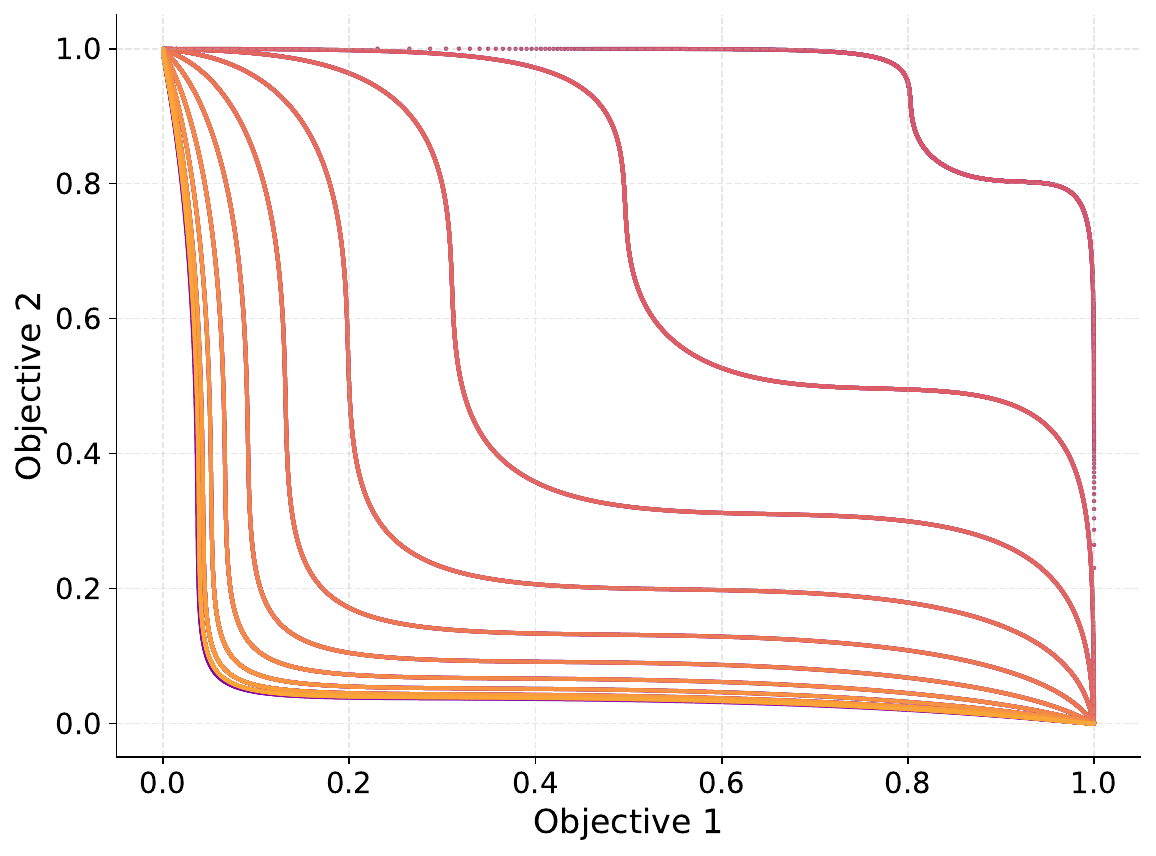}
    \caption{The dynamic Pareto Front (PF) of the DF6 benchmark, illustrating the non-linear curvature changes and localized disconnected regions across different environmental states.}
    \label{fig:df6_pf}
\end{figure}

To quantitatively evaluate the semantic meaning of the learned basis vectors and explain the geometric phenomena observed in the latent space, we introduce DF6 as the primary target for our case studies. As visually illustrated in Fig.~\ref{fig:df6_pf}, DF6 exhibits highly complex dynamics characterized by simultaneous positional shifts in the decision space and non-linear curvature changes in the objective space. 

Let the decision variables be denoted as $\mathbf{x} = (x_1, \dots, x_n)$, where the search space is $x_1 \in [0, 1]$ and $x_i \in [-1, 1]$ for $i = 2, \dots, n$. The DF6 problem is mathematically defined as:
\begin{equation}
    \begin{cases} 
        f_1(\mathbf{x}, t) = g(\mathbf{x}, t)(x_1 + 0.1 \sin(3\pi x_1))^{\alpha_t} \\
        f_2(\mathbf{x}, t) = g(\mathbf{x}, t)(1 - x_1 + 0.1 \sin(3\pi x_1))^{\alpha_t}
    \end{cases}
    \label{eq:df6_def}
\end{equation}
where the distance function is $g(\mathbf{x}, t) = 1 + \sum_{i=2}^{n} (|G(t)|y_i^2 - 10 \cos(2\pi y_i) + 10)$, with the spatial translation term $y_i = x_i - G(t)$. The temporal dynamics are driven by $G(t) = \sin(0.5\pi t)$, which controls both the decision space shift and the objective space curvature exponent $\alpha_t = 0.2 + 2.8|G(t)|$. 

\subsubsection{Semantic Correlation with Auxiliary Problems (Problem 64 and 67)}
To disentangle the coupled dynamics of DF6, we introduce two auxiliary problems that isolate specific dynamic properties. 

\textbf{Problem 64 (Curvature Isolation):} Problem 64 isolates the PF curvature variations without any positional shifts in the PS. The search space is $\mathbf{x} \in [0, 1]^n$. It is defined as:
\begin{equation}
    \begin{cases} 
        f_1(\mathbf{x}, t) = x_1 \\
        f_2(\mathbf{x}, t) = g(\mathbf{x}, t) \left(1.0 - \left(\frac{x_1}{g(\mathbf{x}, t)}\right)^{\alpha_t} \right)
    \end{cases}
\end{equation}
where $g(\mathbf{x}, t) = 1.0 + 9.0 \sum_{i=2}^{n} x_i^2$, and the curvature exponent $\alpha_t = 0.2 + 3.0|G(t)|$. The dynamic exponent $\alpha_t$ shares a highly similar mathematical structure with DF6, causing analogous convex-to-concave geometric deformations, but its PS remains statically anchored at $x_i = 0$ ($i \ge 2$).

\textbf{Problem 67 (Shift Isolation):} Problem 67 isolates the PS positional shifts while maintaining a completely static PF geometry. It is defined as:
\begin{equation}
    \begin{cases} 
        f_1(\mathbf{x}, t) = x_1 \\
        f_2(\mathbf{x}, t) = g(\mathbf{x}, t) \left(1.0 - \sqrt{\frac{x_1}{g(\mathbf{x}, t)}} \right)
    \end{cases}
\end{equation}
where $g(\mathbf{x}, t) = 1.0 + 9.0 \sum_{i=2}^{n} (x_i - s_i(t))^2$. The shift vector $s_i(t)$ alternates between $|G(t)|$ and $-|G(t)|$ across indices, closely mirroring the $y_i = x_i - G(t)$ translation mechanism in DF6. However, its objective space exponent is fixed at $0.5$.

\begin{figure}[htbp]
    \centering
    \includegraphics[width=0.8\linewidth]{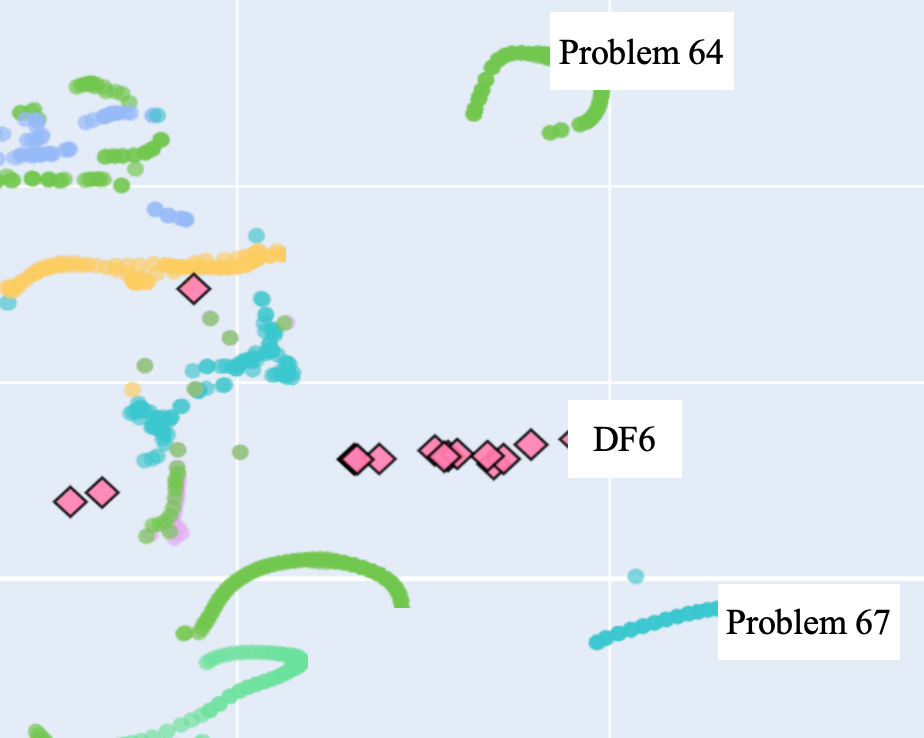}
    \caption{Latent space visualization of DF6, Problem 64, and Problem 67. Although distributed in distant sub-regions, the trajectory of DF6 is spatially positioned between Problem 64 and Problem 67, reflecting its compositional dynamic properties.}
    \label{fig:df_64_67_latent}
\end{figure}

To visually illustrate the structural relationships among these problems, Fig.~\ref{fig:df_64_67_latent} presents their distributions in the latent space. While DF6, Problem 64, and Problem 67 occupy distinct sub-regions, the trajectory of DF6 is situated spatially between Problem 64 and Problem 67. This intermediate positioning in the latent manifold visually corroborates the compositional nature of DF6, proving that the learned latent space captures the underlying mathematical topology and intrinsic structural connections across different dynamic environments.

\subsubsection{Topological Intersection Between DF5 and DF6}
As observed in the latent space t-SNE visualization, the evolutionary trajectories of the DF5 and DF6 problems exhibit a distinct intertwining phenomenon. To mathematically explain this behavior, we compare their respective Pareto Front definitions. 

According to the CEC 2018 benchmark definitions \cite{jiang2018benchmark}, the dynamic PF of DF5 is defined with a \textit{static linear base} and dynamically oscillating high-frequency ripples:
\begin{equation}
    \text{DF5: }
    \begin{cases} 
        f_1(\mathbf{x}, t) = g(\mathbf{x}, t)(x_1 + 0.02 \sin(w_t \pi x_1)) \\
        f_2(\mathbf{x}, t) = g(\mathbf{x}, t)(1 - x_1 + 0.02 \sin(w_t \pi x_1))
    \end{cases}
    \label{eq:df5_pf}
\end{equation}
where the distance function is $g(\mathbf{x}, t) = 1 + \sum_{i=2}^{n}(x_i - G(t))^2$, and the frequency parameter $w_t = \lfloor 10 G(t) \rfloor$. Macroscopically, ignoring the minor $0.02$ amplitude ripple, the PF of DF5 is consistently anchored to a linear geometry ($f_1 + f_2 \approx 1$).

By contrasting Eq.~\ref{eq:df5_pf} with the DF6 definition in Eq.~\ref{eq:df6_def}, it is evident that their local ripples differ mathematically (DF6 uses a static $0.1 \sin(3\pi x_1)$ ripple). However, a macroscopic topological homology occurs when the dynamic curvature of DF6 degenerates into a linear manifold. This condition is satisfied when the exponent of DF6 becomes exactly $1$:
\begin{equation}
    \alpha_t = 0.2 + 2.8|G(t)| = 1
\end{equation}
\begin{equation}
    |G(t)| = |\sin(0.5\pi t)| = \frac{0.8}{2.8} = \frac{2}{7} \approx 0.2857
\end{equation}

This derivation reveals that at specific evolutionary moments within every cycle (when $|\sin(0.5\pi t)| = 2/7$), the non-linear curvature of DF6 dynamically flattens. At these exact steps, the macroscopic PF of DF6 becomes $f_1 + f_2 \approx 1$. Because the Topology-Aware Contrastive Regularization in DB-GEN utilizes the Inverted Generational Distance (IGD) metric, the minor differences in their localized ripple amplitudes ($0.1$ vs $0.02$) contribute negligibly to the overall IGD. 

Consequently, the IGD between the PFs of DF5 and DF6 approaches zero at these temporal nodes. The triplet margin loss directly forces the model to encode these geometrically homologous, yet mathematically distinct, environmental states into adjacent latent regions. This provides a rigorous explanation for their trajectory intersection in the t-SNE projection, confirming that the latent space representation is strictly governed by global manifold geometry.

\subsection{HV}

Table \ref{tab:hv_comparison} presents the complete Mean Hypervolume (MHV) results and their corresponding standard deviations across all 57 dynamic test configurations, comprising the DF1--DF14 and FDA1--FDA5 problem suites. The performance of the proposed method is compared against four state-of-the-art dynamic multi-objective evolutionary algorithms: STT-MOEA/D, DIP-DMOEA, VARE, and SIKT-DMOEA. 

In the table, the best and second-best results for each configuration are highlighted with a dark gray and light gray background, respectively. To ensure statistical rigor, the Wilcoxon rank-sum test at a 5\% significance level is conducted. The symbols ``$+$'', ``$-$'', and ``$\approx$'' indicate that the corresponding comparison algorithm performs significantly better than, significantly worse than, or statistically similar to our proposed method. The bottom row summarizes the overall performance, where the proposed method achieves the best average rank of 1.37 across all test environments.

\begin{table*}[htbp]
\centering
\caption{The MHV values and standard deviation in parentheses obtained by SOTAs and Ours.}
\label{tab:hv_comparison}
\definecolor{bestbg}{gray}{0.75} 
\definecolor{secondbg}{gray}{0.85} 

\newcommand{\hlbest}[1]{\cellcolor{bestbg}#1}
\newcommand{\hlsecond}[1]{\cellcolor{secondbg}#1}

\resizebox{\textwidth}{!}{
\begin{tabular}{c|c|c|c|c|c|c|c}
\toprule
Problem & $n_t$ & $\tau_t$ & STT-MOEA/D & DIP-DMOEA & VARE & SIKT-DMOEA & Ours \\
\midrule
    \multirow{3}{*}{DF1} & 5 & 10 & 9.30e-02(3.84e-03)$-$ & 4.72e-01(3.09e-03)$-$ & \hlsecond{5.81e-01(1.56e-03)}$-$ & 4.55e-01(1.21e-02)$-$ & \hlbest{6.09e-01(2.74e-03)} \\
      & 10 & 5 & 7.91e-02(8.64e-03)$-$ & 4.34e-01(4.66e-04)$-$ & \hlsecond{5.39e-01(4.63e-03)}$-$ & 3.73e-01(1.82e-02)$-$ & \hlbest{5.58e-01(1.74e-03)} \\
      & 10 & 10 & 9.15e-02(8.24e-04)$-$ & \hlsecond{4.95e-01(4.53e-02)}$-$ & 4.43e-01(2.90e-03)$-$ & 4.46e-01(7.95e-03)$-$ & \hlbest{5.15e-01(4.57e-04)} \\
    \midrule
    \multirow{3}{*}{DF2} & 5 & 10 & 2.63e-01(2.99e-03)$-$ & 6.32e-01(7.12e-03)$+$ & \hlsecond{6.55e-01(3.61e-03)}$+$ & \hlbest{6.56e-01(7.68e-03)}$+$ & 6.24e-01(4.49e-03) \\
      & 10 & 5 & 2.45e-01(7.95e-04)$-$ & 5.81e-01(5.35e-04)$-$ & 5.51e-01(1.20e-03)$-$ & \hlsecond{5.84e-01(7.65e-03)}$-$ & \hlbest{6.28e-01(2.90e-03)} \\
      & 10 & 10 & 2.47e-01(6.28e-04)$-$ & 6.71e-01(4.58e-04)$-$ & \hlsecond{6.76e-01(6.94e-03)}$-$ & 6.67e-01(5.75e-03)$-$ & \hlbest{6.81e-01(1.44e-03)} \\
    \midrule
    \multirow{3}{*}{DF3} & 5 & 10 & 8.54e-02(1.80e-03)$-$ & 3.72e-01(1.55e-03)$-$ & \hlsecond{4.41e-01(3.54e-03)}$-$ & 4.33e-01(4.66e-03)$-$ & \hlbest{4.57e-01(3.39e-03)} \\
      & 10 & 5 & 2.28e-01(7.64e-04)$-$ & 3.26e-01(5.42e-04)$-$ & 2.90e-01(2.32e-03)$-$ & \hlsecond{3.84e-01(6.63e-03)}$-$ & \hlbest{4.19e-01(3.44e-03)} \\
      & 10 & 10 & \hlsecond{7.99e-01(8.38e-04)}$+$ & 3.82e-01(7.47e-04)$+$ & \hlbest{9.49e-01(2.23e-03)}$+$ & 4.39e-01(2.47e-03)$+$ & 2.28e-01(7.20e-03) \\
    \midrule
    \multirow{3}{*}{DF4} & 5 & 10 & \hlsecond{4.97e+00(2.49e-03)}$-$ & 6.99e-01(5.41e-04)$-$ & 8.03e-01(5.13e-03)$-$ & 6.78e-01(2.11e-03)$-$ & \hlbest{7.24e+00(7.24e-03)} \\
      & 10 & 5 & \hlsecond{5.19e+00(4.23e-03)}$-$ & 6.73e-01(6.63e-03)$-$ & 7.41e-01(5.74e-03)$-$ & 5.79e-01(4.09e-03)$-$ & \hlbest{7.16e+00(3.75e-03)} \\
      & 10 & 10 & \hlsecond{4.85e+00(4.70e-03)}$-$ & 7.10e-01(2.68e-04)$-$ & 6.23e-01(6.93e-03)$-$ & 6.66e-01(5.30e-03)$-$ & \hlbest{7.24e+00(7.32e-03)} \\
    \midrule
    \multirow{3}{*}{DF5} & 5 & 10 & 1.04e-01(3.20e-03)$-$ & 5.51e-01(4.75e-02)$-$ & \hlbest{6.15e-01(2.68e-03)}$+$ & 4.58e-01(1.73e-02)$-$ & \hlsecond{5.77e-01(6.77e-04)} \\
      & 10 & 5 & 1.03e-01(3.90e-03)$-$ & \hlsecond{5.18e-01(4.32e-02)}$-$ & 4.78e-01(2.98e-03)$-$ & 3.26e-01(3.30e-02)$-$ & \hlbest{5.57e-01(1.29e-03)} \\
      & 10 & 10 & 9.04e-02(3.21e-03)$-$ & \hlsecond{5.54e-01(1.73e-02)}$-$ & 5.37e-01(3.86e-03)$-$ & 4.95e-01(9.06e-03)$-$ & \hlbest{5.78e-01(2.65e-04)} \\
    \midrule
    \multirow{3}{*}{DF6} & 5 & 10 & \hlsecond{7.61e-01(2.26e-03)}$-$ & 2.80e-01(4.54e-03)$-$ & 3.64e-01(2.77e-03)$-$ & 2.15e-01(3.78e-02)$-$ & \hlbest{1.43e+00(1.42e-01)} \\
      & 10 & 5 & \hlsecond{7.84e-01(3.65e-03)}$-$ & 2.36e-01(8.31e-03)$-$ & 2.54e-01(2.88e-03)$-$ & 9.91e-02(2.33e-02)$-$ & \hlbest{1.69e+00(3.92e-02)} \\
      & 10 & 10 & \hlsecond{9.91e-01(2.80e-03)}$-$ & 3.21e-01(3.71e-03)$-$ & 5.70e-01(2.90e-03)$-$ & 2.78e-01(3.99e-02)$-$ & \hlbest{1.82e+00(2.00e-02)} \\
    \midrule
    \multirow{3}{*}{DF7} & 5 & 10 & \hlsecond{5.40e+00(3.27e-03)}$-$ & 3.73e-01(2.61e-03)$-$ & 7.30e-01(2.80e-03)$-$ & 4.31e-01(3.79e-03)$-$ & \hlbest{8.62e+00(1.21e-01)} \\
      & 10 & 5 & \hlsecond{7.75e+00(3.15e-03)}$-$ & 3.91e-01(7.67e-03)$-$ & 5.09e-01(2.84e-03)$-$ & 4.01e-01(8.74e-03)$-$ & \hlbest{1.00e+01(1.47e-01)} \\
      & 10 & 10 & \hlsecond{7.37e+00(3.72e-03)}$-$ & 4.00e-01(3.25e-04)$-$ & 8.12e-01(3.40e-03)$-$ & 4.42e-01(4.01e-03)$-$ & \hlbest{1.03e+01(2.21e-02)} \\
    \midrule
    \multirow{3}{*}{DF8} & 5 & 10 & 3.30e-01(3.19e-03)$-$ & \hlsecond{5.99e-01(5.79e-04)}$-$ & 5.02e-01(3.20e-03)$-$ & 5.93e-01(1.76e-03)$-$ & \hlbest{6.11e-01(5.56e-03)} \\
      & 10 & 5 & 2.83e-01(2.81e-03)$-$ & \hlsecond{5.64e-01(2.44e-04)}$-$ & 4.65e-01(2.66e-03)$-$ & 5.56e-01(2.40e-03)$-$ & \hlbest{5.88e-01(8.83e-03)} \\
      & 10 & 10 & 2.56e-01(3.55e-03)$-$ & \hlsecond{6.01e-01(3.25e-03)}$-$ & 5.23e-01(3.42e-03)$-$ & 6.00e-01(2.21e-03)$-$ & \hlbest{6.39e-01(4.67e-03)} \\
    \midrule
    \multirow{3}{*}{DF9} & 5 & 10 & 4.99e-02(2.68e-03)$-$ & 3.31e-01(3.37e-04)$-$ & \hlsecond{3.42e-01(2.46e-03)}$-$ & 3.09e-01(2.84e-02)$-$ & \hlbest{3.54e-01(4.51e-03)} \\
      & 10 & 5 & 5.61e-02(3.42e-03)$-$ & \hlsecond{2.78e-01(9.64e-04)}$-$ & 2.41e-01(2.78e-03)$-$ & 1.98e-01(3.72e-02)$-$ & \hlbest{3.11e-01(8.20e-03)} \\
      & 10 & 10 & 4.63e-02(2.72e-03)$-$ & 3.79e-01(2.43e-03)$-$ & \hlsecond{4.32e-01(3.20e-03)}$-$ & 4.18e-01(2.71e-02)$\approx$ & \hlbest{4.39e-01(5.47e-03)} \\
    \midrule
    \multirow{3}{*}{DF10} & 5 & 10 & 4.62e-01(4.28e-03)$-$ & \hlbest{6.42e-01(4.57e-04)}$+$ & 5.45e-01(3.31e-03)$-$ & 5.88e-01(8.54e-03)$-$ & \hlsecond{6.36e-01(1.27e-03)} \\
      & 10 & 5 & 3.91e-01(3.14e-03)$-$ & 5.32e-01(5.25e-04)$-$ & \hlsecond{5.65e-01(3.11e-03)}$-$ & 5.22e-01(7.47e-03)$-$ & \hlbest{6.19e-01(3.16e-03)} \\
      & 10 & 10 & 3.72e-01(3.24e-03)$-$ & 5.34e-01(7.45e-04)$-$ & \hlsecond{6.20e-01(2.98e-03)}$-$ & 5.85e-01(4.65e-03)$-$ & \hlbest{6.45e-01(1.26e-03)} \\
    \midrule
    \multirow{3}{*}{DF11} & 5 & 10 & \hlsecond{3.83e-01(2.46e-03)}$-$ & 2.56e-01(7.49e-04)$-$ & 9.38e-02(3.28e-03)$-$ & 2.67e-01(4.66e-03)$-$ & \hlbest{5.18e-01(3.65e-04)} \\
      & 10 & 5 & 5.62e-02(3.18e-03)$-$ & 2.35e-01(5.43e-03)$-$ & \hlbest{6.00e-01(2.73e-03)}$+$ & 2.36e-01(2.54e-03)$-$ & \hlsecond{5.08e-01(6.52e-04)} \\
      & 10 & 10 & 5.56e-02(3.14e-03)$-$ & \hlsecond{2.62e-01(3.69e-04)}$-$ & 9.95e-02(2.77e-03)$-$ & 2.52e-01(3.72e-03)$-$ & \hlbest{5.15e-01(3.88e-04)} \\
    \midrule
    \multirow{3}{*}{DF12} & 5 & 10 & 1.79e-01(3.28e-03)$-$ & \hlsecond{5.46e-01(9.50e-04)}$+$ & \hlbest{1.16e+00(2.56e-03)}$+$ & 6.61e-02(3.65e-03)$-$ & 2.52e-01(1.86e-02) \\
      & 10 & 5 & 1.94e-01(3.11e-03)$-$ & 5.77e-01(4.02e-03)$-$ & \hlbest{1.18e+00(3.24e-02)}$+$ & 3.09e-02(1.41e-03)$-$ & \hlsecond{6.61e-01(2.27e-03)} \\
      & 10 & 10 & 1.82e-01(2.88e-03)$-$ & \hlsecond{6.35e-01(3.37e-03)}$-$ & 1.21e-01(3.63e-03)$-$ & 4.43e-02(2.59e-03)$-$ & \hlbest{6.79e-01(2.12e-04)} \\
    \midrule
    \multirow{3}{*}{DF13} & 5 & 10 & 6.06e-01(2.96e-03)$-$ & 6.30e-01(2.75e-04)$-$ & \hlsecond{6.40e-01(2.77e-03)}$-$ & 4.99e-01(9.88e-03)$-$ & \hlbest{6.52e-01(3.45e-03)} \\
      & 10 & 5 & \hlbest{6.65e-01(3.07e-03)}$+$ & 5.73e-01(3.63e-03)$-$ & 4.96e-01(2.96e-03)$-$ & 4.19e-01(1.47e-02)$-$ & \hlsecond{6.09e-01(1.52e-03)} \\
      & 10 & 10 & \hlsecond{6.41e-01(3.36e-03)}$-$ & 6.10e-01(8.79e-04)$-$ & 6.24e-01(3.33e-03)$-$ & 4.88e-01(9.70e-03)$-$ & \hlbest{6.63e-01(4.26e-03)} \\
    \midrule
    \multirow{3}{*}{DF14} & 5 & 10 & 2.73e-01(3.42e-03)$-$ & 5.82e-01(3.31e-03)$-$ & \hlbest{6.96e-01(2.60e-03)}$+$ & 4.87e-01(5.75e-03)$-$ & \hlsecond{6.90e-01(2.06e-04)} \\
      & 10 & 5 & \hlbest{2.83e+00(3.30e-03)}$+$ & 5.44e-01(4.12e-03)$-$ & 6.67e-01(3.22e-03)$-$ & 4.09e-01(2.02e-02)$-$ & \hlsecond{6.84e-01(2.79e-04)} \\
      & 10 & 10 & \hlbest{2.82e+00(2.89e-03)}$+$ & 5.82e-01(3.21e-04)$-$ & \hlsecond{8.12e-01(3.02e-03)}$+$ & 5.47e-01(4.13e-03)$-$ & 6.91e-01(1.62e-04) \\
    \midrule
    \multirow{3}{*}{FDA1} & 5 & 10 & 1.07e-01(4.61e-03)$-$ & 5.31e-01(1.68e-03)$-$ & 6.01e-01(1.57e-03)$-$ & \hlsecond{6.17e-01(1.49e-02)}$-$ & \hlbest{6.27e-01(6.56e-03)} \\
      & 10 & 5 & 7.13e-02(3.59e-03)$-$ & 6.51e-01(4.69e-02)$-$ & \hlsecond{6.56e-01(6.34e-03)}$-$ & 5.08e-01(4.21e-02)$-$ & \hlbest{6.74e-01(3.47e-03)} \\
      & 10 & 10 & 1.02e-01(3.55e-03)$-$ & \hlsecond{7.19e-01(4.46e-03)}$+$ & \hlbest{8.07e-01(5.19e-04)}$+$ & 6.60e-01(3.99e-03)$\approx$ & 6.75e-01(4.19e-03) \\
    \midrule
    \multirow{3}{*}{FDA2} & 5 & 10 & 1.74e-03(3.06e-03)$-$ & 5.54e-01(3.50e-03)$-$ & \hlsecond{6.84e-01(1.08e-02)}$-$ & 5.64e-01(1.69e-03)$-$ & \hlbest{2.00e+00(1.14e-02)} \\
      & 10 & 5 & 1.33e-02(1.79e-03)$-$ & \hlsecond{7.11e-01(1.46e-02)}$-$ & 6.07e-01(2.29e-04)$-$ & 5.57e-01(1.35e-02)$-$ & \hlbest{1.82e+00(2.01e-02)} \\
      & 10 & 10 & 2.63e-01(1.68e-03)$-$ & \hlsecond{7.90e-01(4.57e-03)}$-$ & 5.85e-01(5.56e-04)$-$ & 5.65e-01(1.71e-03)$-$ & \hlbest{1.96e+00(2.85e-02)} \\
    \midrule
    \multirow{3}{*}{FDA3} & 5 & 10 & 2.96e-01(1.16e-03)$-$ & 5.33e-01(4.75e-04)$-$ & \hlsecond{6.30e-01(8.86e-03)}$-$ & 4.59e-01(1.00e-02)$-$ & \hlbest{6.72e-01(1.01e-03)} \\
      & 10 & 5 & 4.39e-01(2.24e-03)$-$ & 3.39e-01(7.72e-04)$-$ & \hlsecond{5.71e-01(8.31e-03)}$-$ & 3.38e-01(2.23e-02)$-$ & \hlbest{6.59e-01(2.78e-03)} \\
      & 10 & 10 & \hlsecond{6.50e-01(7.14e-03)}$\approx$ & 6.37e-01(5.32e-04)$-$ & 5.82e-01(2.93e-04)$-$ & 5.06e-01(8.91e-03)$-$ & \hlbest{6.72e-01(1.04e-03)} \\
    \midrule 
    \multirow{3}{*}{FDA4} & 5 & 10 & 1.70e-01(8.38e-03)$-$ & 3.36e-01(4.65e-04)$-$ & 3.33e-01(8.18e-03)$-$ & \hlsecond{4.58e-01(7.47e-03)}$\approx$ & \hlbest{4.68e-01(2.19e-03)} \\
      & 10 & 5 & 1.11e-01(3.86e-02)$-$ & 3.05e-01(1.36e-03)$-$ & 2.26e-01(1.40e-03)$-$ & \hlsecond{3.34e-01(1.56e-02)}$-$ & \hlbest{3.42e-01(3.32e-03)} \\
      & 10 & 10 & 1.70e-01(3.00e-02)$-$ & 4.12e-01(2.38e-04)$-$ & 4.51e-01(1.57e-04)$-$ & \hlsecond{4.77e-01(4.99e-03)}$-$ & \hlbest{4.93e-01(8.06e-04)} \\
    \midrule
    \multirow{3}{*}{FDA5} & 5 & 10 & \hlbest{4.90e-01(3.94e-02)}$\approx$ & 4.40e-01(2.27e-04)$-$ & 4.18e-01(1.37e-04)$-$ & \hlsecond{4.51e-01(1.37e-02)}$\approx$ & 4.59e-01(3.86e-04) \\
      & 10 & 5 & 7.12e-02(1.50e-02)$-$ & 2.42e-01(5.28e-04)$-$ & \hlsecond{3.95e-01(1.16e-03)}$-$ & 3.33e-01(1.47e-02)$-$ & \hlbest{4.00e-01(8.76e-04)} \\
      & 10 & 10 & 1.20e-01(1.71e-02)$-$ & 3.54e-01(4.65e-03)$-$ & \hlsecond{4.62e-01(5.53e-04)}$-$ & 4.58e-01(8.49e-03)$-$ & \hlbest{4.98e-01(2.09e-04)} \\
\midrule
\midrule
\multicolumn{3}{l|}{\textbf{Average Rank} \hspace{1em}} & 3.91 & 3.21 & \hlsecond{2.79} & 3.72 & \hlbest{1.37} \\
\bottomrule
\end{tabular}
}
\end{table*}

\subsection{Detailed Statistical Analysis of Data Scale Expansion}
\label{sec:appendix_wilcoxon}

To comprehensively validate the scalability of the proposed framework, we performed pairwise Wilcoxon signed-rank tests across all evaluated historical data scales. Since the Friedman test in the main text confirmed a globally significant difference, these post-hoc pairwise comparisons are utilized to identify exact local improvements. 

\begin{figure}[htbp]
    \centering
    \includegraphics[width=0.7\linewidth]{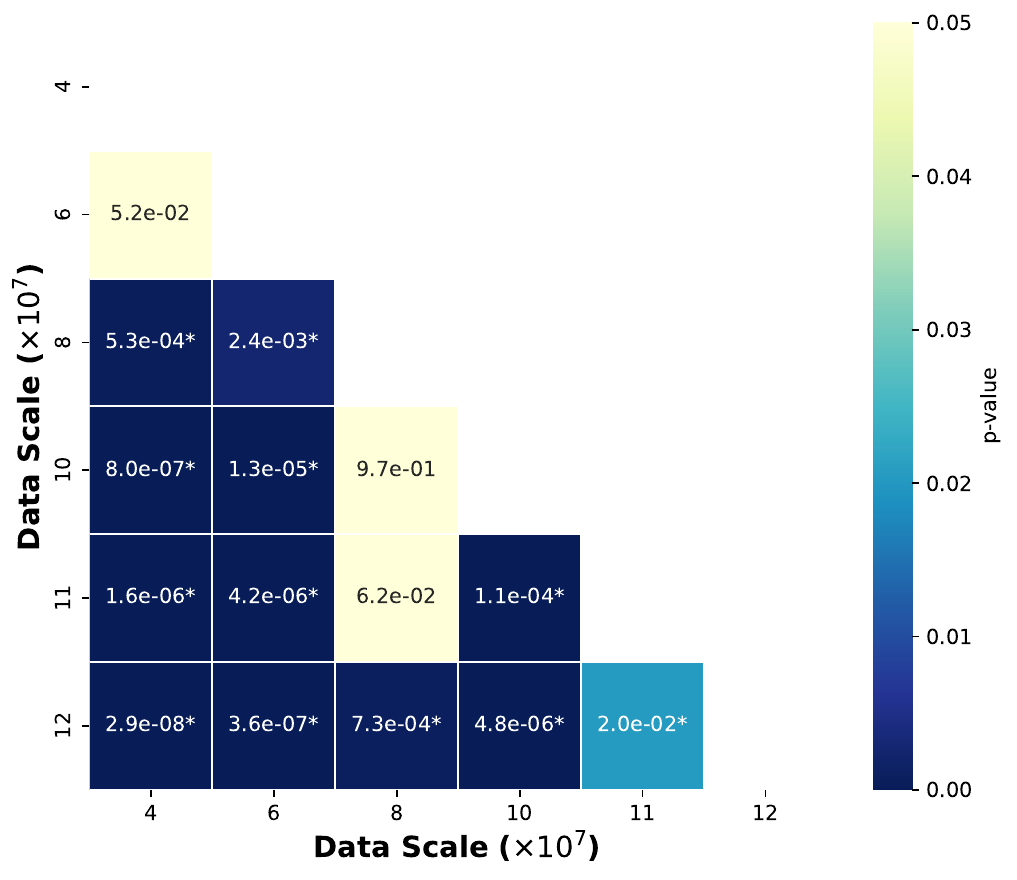}
    \caption{The $p$-value heatmap of the pairwise Wilcoxon signed-rank tests across different historical data scales. Cells annotated with $*$ indicate a statistically significant difference ($p < 0.05$). The scale units are $\times 10^7$.}
    \label{fig:wilcoxon_heatmap}
\end{figure}

Fig.~\ref{fig:wilcoxon_heatmap} visualizes the $p$-value matrix derived from the Wilcoxon tests. In this lower-triangular heatmap, each cell represents the statistical significance of the performance difference between two corresponding data scales. The exact $p$-values are annotated within the cells, where the asterisk ($*$) explicitly denotes statistical significance at the conventional $5\%$ level ($p < 0.05$).

As depicted in Fig.~\ref{fig:wilcoxon_heatmap}, the maximum data scale ($12 \times 10^7$) exhibits highly significant superiority over all other baseline scales (all $p$-values $< 0.05$). Notably, the performance leap from $11 \times 10^7$ to $12 \times 10^7$ remains statistically significant ($p = 2.04 \times 10^{-2}$), indicating that expanding the diverse topological priors continues to benefit the model's generative accuracy at this stage. Additionally, the non-significant cell between the scales of $8 \times 10^7$ and $10 \times 10^7$ ($p = 0.966$) clearly identifies the localized performance plateau mentioned in the main text. Overall, this comprehensive statistical matrix substantiates the robustness and strict monotonicity of the model's scalability.

\section{Real-world Problem Definitions}

To further validate the proposed algorithm in practical scenarios, we introduce two formulated real-world dynamic multi-objective optimization problems. These models are designed with decoupled position and distance variables to explicitly simulate complex topological transformations. Consequently, both the Pareto Set (PS) in the decision space and the Pareto Front (PF) in the objective space undergo continuous and non-linear shifts over time, providing a rigorous testbed for evaluating the manifold generation capability of the algorithm.

\subsection{Dynamic Resource Allocation (DRA)}
The DRA problem models the continuous trade-off between the overall operational cost and the service quality within a dynamically changing computing network. The decision variables $\mathbf{x} = (x_1, \dots, x_D) \in [0, 1]^D$ represent the resource allocation strategy across $D$ heterogeneous nodes. 

To simulate real-world physical constraints, the first variable $x_1$ acts as the primary trade-off controller (e.g., total budget size), while the remaining variables $x_2, \dots, x_D$ represent the convergence accuracy (e.g., load balancing efficiency). The conflicting objectives are formulated as follows:

\begin{itemize}
    \item Objective 1 (Minimize Cost):
    \begin{equation}
        f_1(\mathbf{x}, t) = \bar{c}(t) \cdot x_1 \cdot g(\mathbf{x}, t)
    \end{equation}
    This objective evaluates the normalized operational cost, where $\bar{c}(t)$ denotes the mean cost index across the network at time $t$.
    
    \item Objective 2 (Maximize Quality):
    \begin{equation}
        f_2(\mathbf{x}, t) = \bar{q}(t) - \frac{\bar{q}(t) \sqrt{x_1}}{g(\mathbf{x}, t)}
    \end{equation}
    This objective minimizes the service quality loss. A larger budget $x_1$ reduces the loss marginally, reflecting the physical law of diminishing returns (modeled by the square root). $\bar{q}(t)$ is the mean quality index.
\end{itemize}

The convergence penalty function $g(\mathbf{x}, t)$ models the system efficiency loss and defines the dynamic behavior of the Pareto Set (PS):
\begin{equation}
    g(\mathbf{x}, t) = 1 + 9 \sum_{j=2}^{D} (x_j - G(t))^2, \quad G(t) = |\sin(0.5 \pi t)|
\end{equation}
The optimal state requires $g(\mathbf{x}, t) = 1$, which forces the variables $x_j$ to continuously track the shifting optimal load center $G(t)$. 

The environmental dynamics are characterized by spatial-temporal heterogeneity. The parameters $c_i(t)$ and $q_i(t)$ for the $i$-th node incorporate node-specific base attributes and spatial phase shifts:
\begin{equation}
\begin{aligned}
    c_i(t) &= c_i^{\text{base}} \left(1 + 0.1 \sin(0.5 \pi t + \theta_i)\right) \\
    q_i(t) &= q_i^{\text{base}} \left(1 + 0.1 \cos(0.5 \pi t + \theta_i)\right)
\end{aligned}
\end{equation}
where $c_i^{\text{base}} \in [0.5, 1.5]$, $q_i^{\text{base}} \in [0.5, 2.0]$, and the spatial phase $\theta_i \in [0, \pi]$. The mean indices are computed as $\bar{c}(t) = \frac{1}{D} \sum c_i(t)$ and $\bar{q}(t) = \frac{1}{D} \sum q_i(t)$.

\subsection{Dynamic Path Planning (DPP)}
The DPP problem simulates the velocity planning of an autonomous vehicle navigating through multiple physical segments. It aims to balance the total travel time and energy consumption under varying weather and traffic conditions. 

The decision variable $x_1$ determines the target velocity mapped to a normalized physical interval $v_{\text{norm}} \in [0.2, 1.0]$ via $v_{\text{norm}} = 0.2 + 0.8 x_1$. The remaining variables model the deviation from the optimal trajectory.

\begin{itemize}
    \item Objective 1 (Minimize Travel Time):
    \begin{equation}
        f_1(\mathbf{x}, t) = \frac{\bar{d}(t)}{v_{\text{norm}}} \cdot g(\mathbf{x}, t)
    \end{equation}
    where $\bar{d}(t)$ is the mean dynamic effective distance across all segments, factoring in periodic traffic congestion.
    
    \item Objective 2 (Minimize Energy Consumption):
    \begin{equation}
        f_2(\mathbf{x}, t) = \overline{cd}(t) \cdot v_{\text{norm}}^2 \cdot g(\mathbf{x}, t)
    \end{equation}
    This objective reflects aerodynamic principles where energy consumption is proportional to the square of the velocity. $\overline{cd}(t)$ denotes the spatially averaged coupled drag-distance coefficient.
\end{itemize}

Similar to DRA, the distance function $g(\mathbf{x}, t)$ incorporates the moving optimum $G(t) = |\sin(0.5 \pi t)|$ to simulate the time-varying optimal steering or lane-changing target:
\begin{equation}
    g(\mathbf{x}, t) = 1 + 9 \sum_{j=2}^{D} (x_j - G(t))^2
\end{equation}

The dynamic environmental factors are formulated as:
\begin{equation}
\begin{aligned}
    d_i(t) &= 1 + 0.2 \sin(0.5 \pi t + \theta_i) \\
    c_i(t) &= 1 + 0.2 \cos(0.5 \pi t + \theta_i)
\end{aligned}
\end{equation}
where $d_i(t)$ indicates the congestion length of segment $i$, and $c_i(t)$ represents the varying aerodynamic drag coefficient caused by weather conditions. The aggregate coefficients are derived as $\bar{d}(t) = \frac{1}{D} \sum d_i(t)$ and $\overline{cd}(t) = \frac{1}{D} \sum c_i(t) d_i(t)$.

\bibliographystyle{plain}   
\bibliography{main}   